\documentclass[lettersize,journal]{IEEEtran}%,dvipdfmx
\usepackage{amssymb,amsthm,amsmath}
\usepackage{bm,bbm}
\usepackage[nocompress]{cite}
\usepackage{graphicx}
\usepackage{latexsym}
\usepackage{multirow}
\usepackage{newtxmath}
\usepackage{hyperref}
\usepackage{overpic}
\usepackage{url}
\usepackage{booktabs}
\usepackage{array}
\usepackage{overpic}
\usepackage[table]{xcolor}
\usepackage{float}
\usepackage{pgf}
\newtheorem{definition}{Definition}

\newcommand{\tcr}[1]{{\textcolor{red}{#1}}}
\definecolor{midori}{rgb}{0.0, 0.5, 0.0}
\newcommand{\tcg}[1]{{\textcolor{midori}{#1}}}
\newcommand{\tcb}[1]{{\textcolor{blue}{#1}}}
\newcommand{\tcc}[1]{{\textcolor{cyan}{#1}}}
\definecolor{gold}{rgb}{1.0, 0.843137, 0.0}
\newcommand{\tcy}[1]{{\textcolor{gold}{#1}}}
\newcommand{\tcm}[1]{{\textcolor{magenta}{#1}}}

\definecolor{brown}{rgb}{0.54296875, 0.26953125, 0.07421875}

\newcommand{\prev}{{\rm prev}}
\newcommand{\stri}{{\rm stri}}
\newcommand{\smax}{{\rm sm}}
\newcommand{\tot}{{\rm tot}}
\newcommand{\tra}{{\rm tra}}
\newcommand{\val}{{\rm val}}

\renewcommand{\sl}{{\rm sl}}

\newcommand{\cl}{{\rm cl}}

\newcommand{\cn}{{\tcr{\rm n}}}
\newcommand{\cp}{{\tcg{\rm p}}}
\newcommand{\cs}{{\tcb{\rm s}}}
\newcommand{\an}{{\tcc{\rm n}}}
\newcommand{\ap}{{\tcm{\rm p}}}
\newcommand{\as}{{\tcy{\rm s}}}
\DeclareMathOperator{\argmin}{{arg\,min}}
\DeclareMathOperator{\argmax}{{arg\,max}}

\newcommand{\bg}{{\bm{g}}}

\newcommand{\bp}{{\bm{p}}}
\newcommand{\bP}{{\bm{P}}}
\newcommand{\bq}{{\bm{q}}}
\newcommand{\bQ}{{\bm{Q}}}

\newcommand{\bu}{{\bm{u}}}
\newcommand{\bv}{{\bm{v}}}

\newcommand{\bx}{{\bm{x}}}
\newcommand{\bX}{{\bm{X}}}

\newcommand{\rmH}{{\mathrm{H}}}
\newcommand{\rmI}{{\mathrm{I}}}

\newcommand{\calG}{{\mathcal{G}}}

\DeclareMathOperator{\bbE}{{\mathbb{E}}}
\DeclareMathOperator{\bbI}{{\mathbbm{1}}}
\newcommand{\bbR}{{\mathbb{R}}}
\newcommand{\bbN}{{\mathbb{N}}}

\newcommand{\hyl}[2]{{(\protect\hyperlink{#1}{#2})}}
\newcommand{\hyt}[2]{{\hypertarget{#1}{\rm({#2})}}}
\hypersetup{bookmarksnumbered=true, bookmarksopen=true, colorlinks=true, linkcolor=blue, citecolor=blue,}
\newcolumntype{C}[1]{>{\hfil}m{#1}<{\hfil}}

\newcommand{\vc}[1]{%
  \pgfmathsetmacro{\mixval}{max(0,min(50*(#1+1),100))}%
  \begingroup%
    \pgfmathparse{\mixval < 50 ? 1 : 0}%
    \ifnum\pgfmathresult=1
      \pgfmathsetmacro{\portion}{2*\mixval}%
      \edef\temp{\noexpand\cellcolor{white!\portion!blue!50!white}{\noexpand\textcolor{black}{#1}}}%
    \else%
      \pgfmathsetmacro{\portion}{2*(\mixval-50)}%
      \edef\temp{\noexpand\cellcolor{red!\portion!white!50!white}{\noexpand\textcolor{black}{#1}}}%
    \fi%
    \temp%
  \endgroup%
}
\begin{document}
\title{Unimodality-Promoting Regularized Learning\\for Ordinal Regression}
\author{Ryoya Yamasaki%
}
\markboth{Preprint}
{Shell \MakeLowercase{\textit{et al.}}: Unimodality-Promoting Regularized Learning for Ordinal Regression}
%\IEEEpubid{0000--0000/00\$00.00~\copyright~2026 IEEE}
\maketitle
%========================================%
\begin{abstract}
Ordinal regression, also called ordinal classification, is 
classification of ordinal data, in which the underlying target variable 
is categorical and considered to have a natural ordinal relation.
Previous works have indicated that, in many real-world ordinal data, 
the conditional probability distribution (CPD) of 
the target variable given a value of the explanatory variable 
would be unimodal in a large domain of the explanatory variable 
and close to be unimodal even in a remaining domain.
Therefore, unimodality-promoting regularized learning (UPRL), 
which promotes a predicted CPD closer to be unimodal 
with the aim of decreasing a prediction variance without 
inducing much bias for ordinal data of the unimodality, 
is promising to improve the prediction performance 
especially with small-size training data.
In this study, we show that previous UPRL methods promote 
a predicted CPD to not only become closer to be unimodal but also 
have a larger scale (in other words, be smoother or less-confident).
Therefore, we develop a novel method that 
more strictly reflects the idea of UPRL and evades a scale-related bias, 
and verify through experimental comparison that the unimodality-promotion 
indeed contributes to improve the prediction performance.
Additionally, while our proposed UPRL method could perform better 
for smaller-scale data or with larger-size training data 
compared to a previous UPRL method, 
our analysis explains this experimental observation in terms of 
the presence or absence of an unexpected scale-related bias.
\end{abstract}
%========================================%
\begin{IEEEkeywords}
Ordinal regression, ordinal data, unimodality, scale, 
smoothness, less-confident, regularization
\end{IEEEkeywords}
%========================================%
\section{Introduction}
\label{sec:Introduction}
%==========%
\IEEEPARstart{O}{rdinal} regression (OR), 
also referred to as ordinal classification, is classification for ordinal data, 
where the target variable is categorical but inherently ordered.
In the OR framework, target categories are drawn from an ordinal scale 
that is considered to have a natural ordinal relation.
Such scales are often constructed as graded summaries of objective indicators, 
such as age groups \{`under 10', `10s', `20s', \ldots, `90s', `over 100'\}, 
or as graded evaluations of subjective judgments, such as Likert scale 
\{`strongly agree', `agree', `neutral', `disagree', `strongly disagree'\} \cite{likert1932technique}.
Ordinal data arise in a wide range of applications, including 
face-age estimation \cite{niu2016ordinal,yamasaki2023optimal,yamasaki2024parallel}, 
disease stage estimation \cite{medmnistv2}, information retrieval \cite{liu2009learning}, 
and analysis of rating data \cite{kim2012corporate, yu2006collaborative} and 
questionnaire survey in social research \cite{chen1995response, burkner2019ordinal}.

%==========%
As an example of the OR problem, consider a questionnaire survey 
about a certain policy that requires subjects to respond from 
\{`strongly agree', `agree', `neutral', `disagree', `strongly disagree'\}.
There, subjects with features specific to those who typically 
respond `agree' might respond `strongly agree' or `neutral', 
but would be less likely to respond `disagree' or `strongly disagree'.
Such an underlying data structure can be rephrased as 
the unimodality of the conditional probability distribution (CPD) 
of the target variable given a value of the explanatory variable.
Through numerical experiments with real-world ordinal data, 
the previous work \cite{yamasaki2022unimodal} has verified that the underlying 
CPD of ordinal data would be unimodal at most of the observed values of 
the explanatory variable (high-UR in Definition~\ref{def:unimodal}), 
and \cite{yamasaki2025approximately} has verified that the CPD would be close 
to be unimodal even at values where the CPD is not unimodal strictly 
(low-UD in Definition~\ref{def:unimodal}); 
see Table~\ref{tab:Data-Prop} and further explanation in Section~\ref{sec:Data}.
Many OR users may often judge that the data have a natural ordinal relation 
and decide to treat them within the OR framework, 
with (unconsciously) expecting their unimodality.

% \IEEEpubidadjcol
%==========%
\begin{figure}[!t]
\centering%
\renewcommand{\arraystretch}{1.5}%
\renewcommand{\tabcolsep}{4pt}%
\begin{tabular}{cc}%
\begin{overpic}[width=4cm]{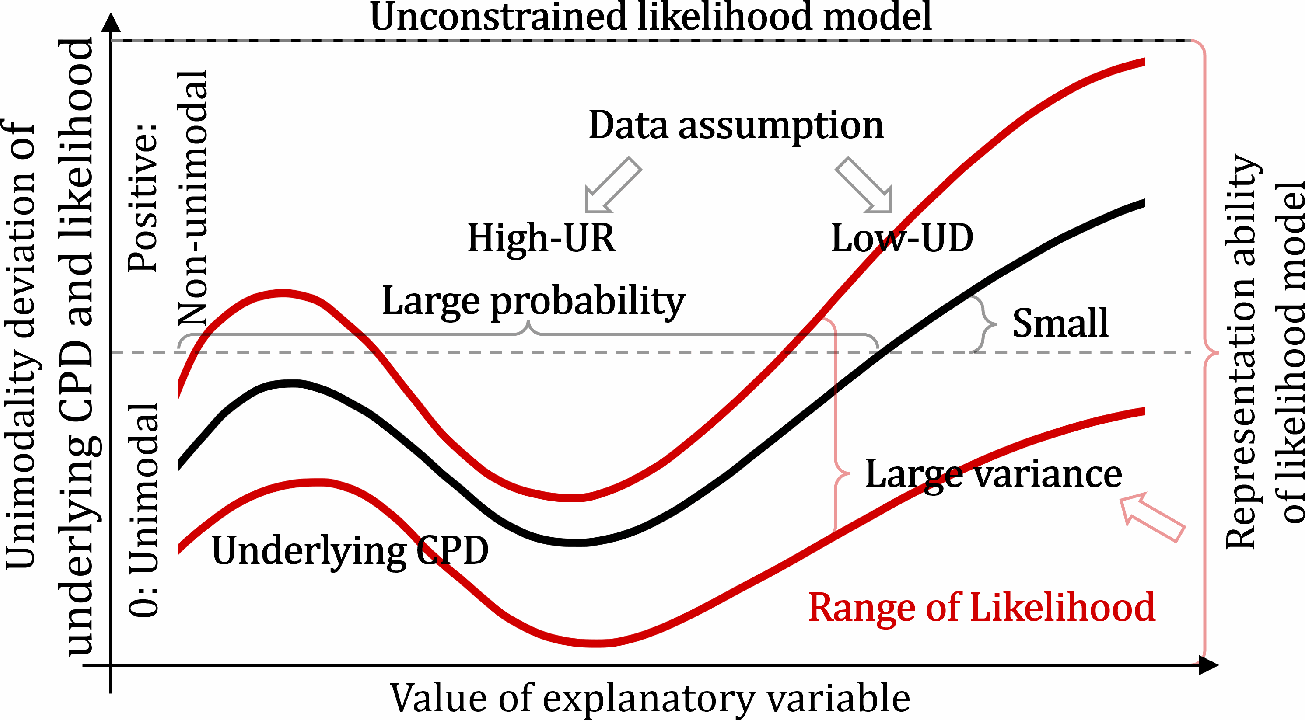}
\put(2,54){{\tiny\rm\hyt{Ideaa}{a}}}\end{overpic}&
\begin{overpic}[width=4cm]{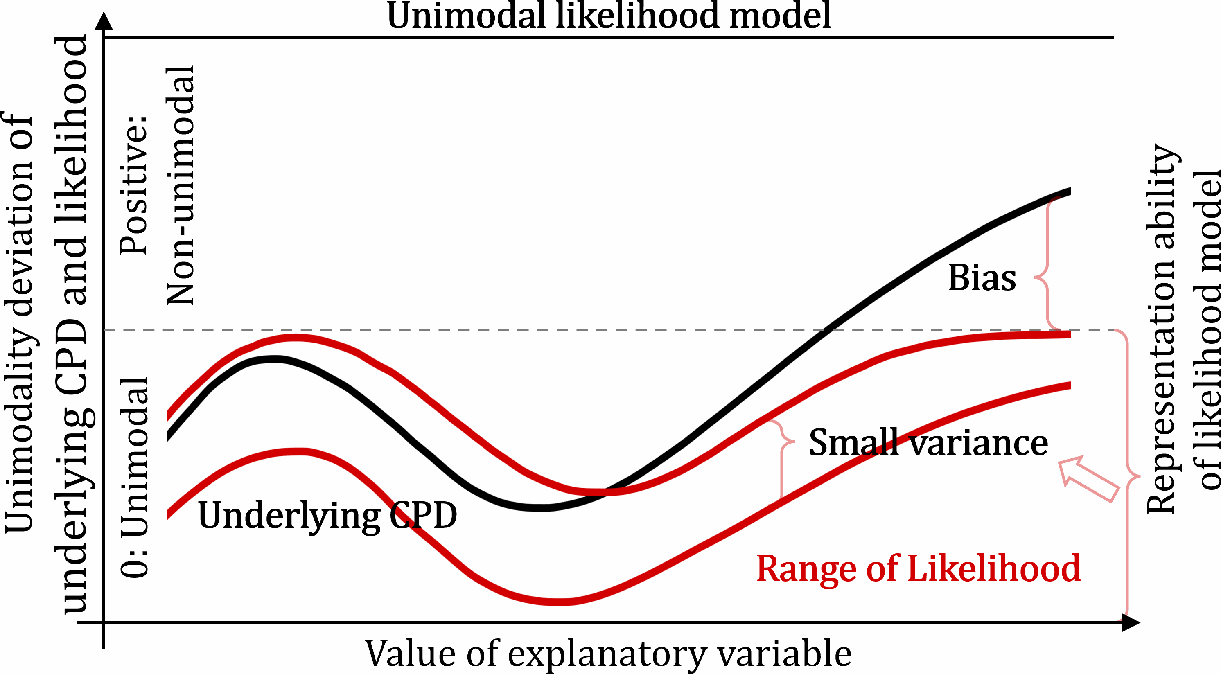}
\put(2,54){{\tiny\rm\hyt{Ideab}{b}}}\end{overpic}
\\
\begin{overpic}[width=4cm]{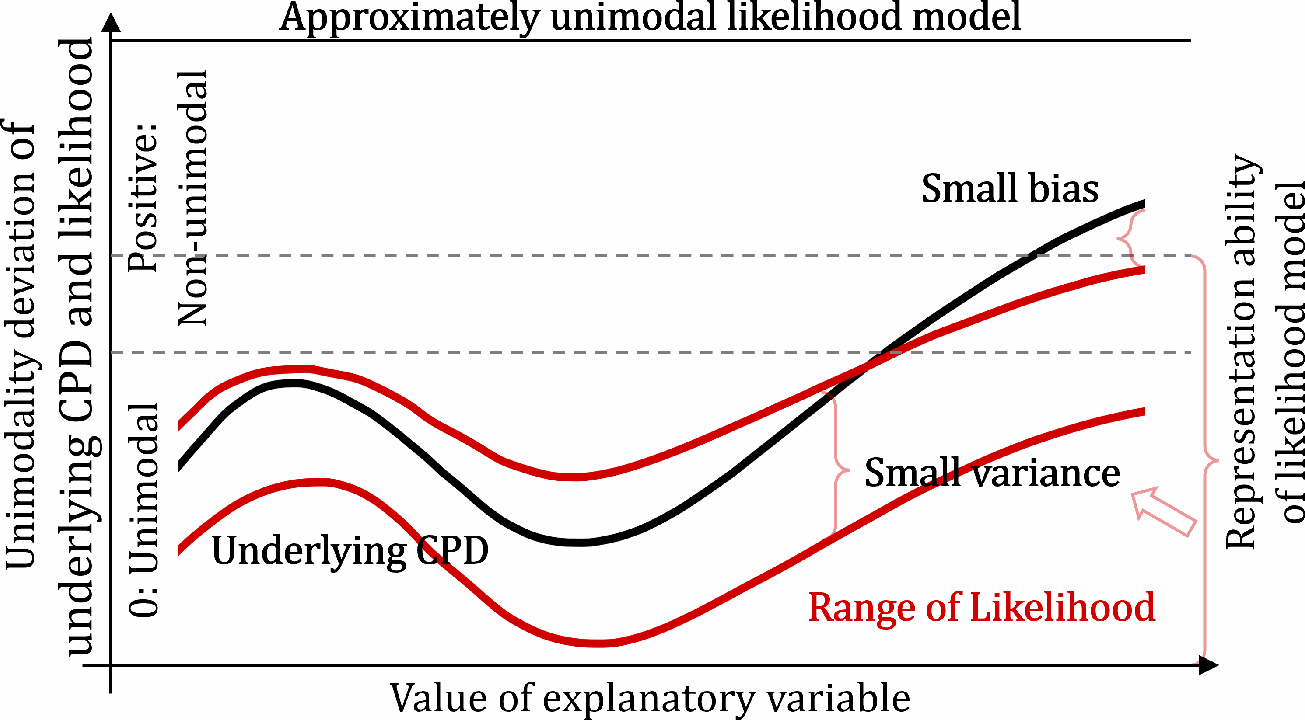}
\put(2,54){{\tiny\rm\hyt{Ideac}{c}}}\end{overpic}&
\begin{overpic}[width=4cm]{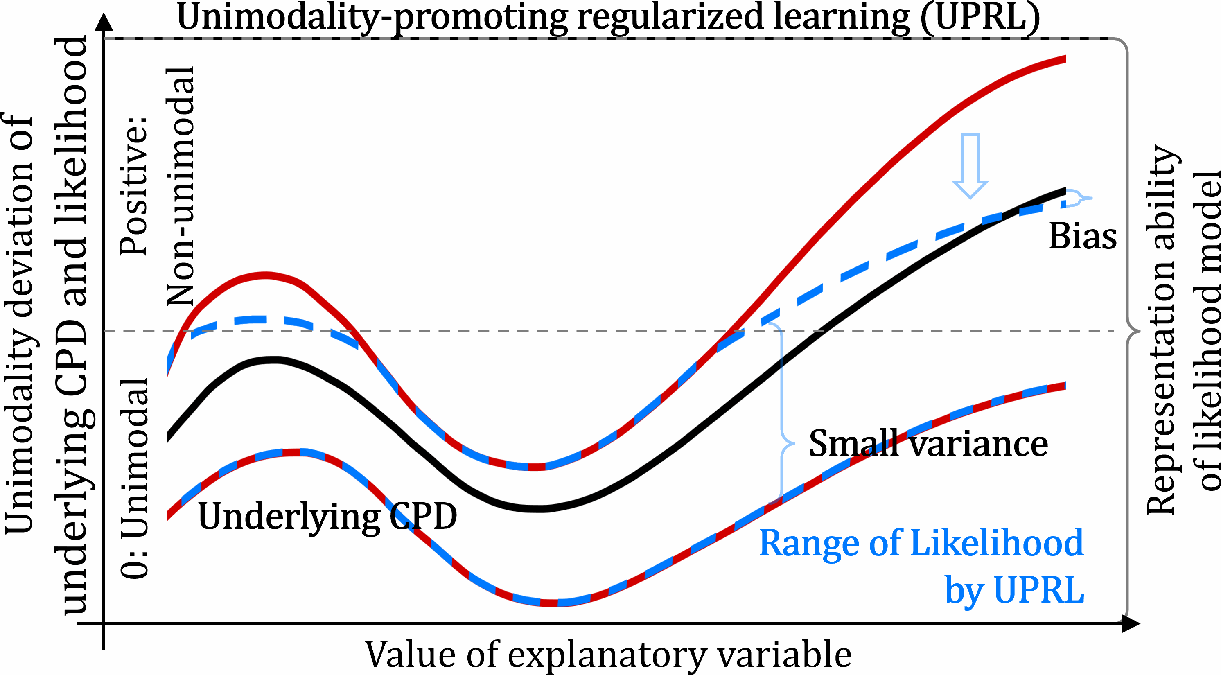}
\put(2,54){{\tiny\rm\hyt{Idead}{d}}}\end{overpic}
\end{tabular}
\caption{%
Illustration of behaviors of OR methods that exploit the unimodality hypothesis:
unimodal likelihood models \hyl{Ideab}{b} can decrease a variance due to 
lower representation ability than unrestricted likelihood models \hyl{Ideaa}{a}, 
but have a bias at a value of the explanatory variable with a non-unimodal CPD; 
approximately unimodal likelihood models \hyl{Ideac}{c} can decrease a bias by enlarging 
the representation ability from the one of unimodal likelihood models \hyl{Ideab}{b}; 
unimodality-promoting regularized learning \hyl{Idead}{d} aims to decrease a variance 
by devising the learning procedure, rather than explicitly restricting 
the representation ability by devising the likelihood model as \hyl{Ideab}{b} and \hyl{Ideac}{c}.}
\label{fig:Idea}
\end{figure}

%==========%
In general, to obtain a prediction model with higher prediction 
performance when the size of available training data is restricted, 
it is promising to design or learn a model with 
leveraging prior domain knowledge on the data.
Therefore, under the unimodality hypothesis for ordinal data, 
previous works \cite{yamasaki2022unimodal,yamasaki2025approximately} 
developed unimodality-driven likelihood models 
(see Figure~\ref{fig:Idea} \hyl{Ideab}{b} and \hyl{Ideac}{c}), 
and \cite{belharbi2020,albuquerque2021ordinal,albuquerque2022quasi,Cardoso2025,kim2025cal} 
developed unimodality-promoting regularized learning (UPRL) methods.
UPRL aims to promote a predicted CPD closer to be unimodal by 
devising a learning procedure, and this study focuses on this approach.
Recall that the prediction error can be roughly 
decomposed into bias- and variance-dependent terms, 
that is, bias-variance trade-off \cite{bishop2006pattern}
(for simplicity, we ignore the optimization error 
\cite{Bottou} in the discussion of this paper); 
enhancing the unimodality-promotion is thought 
to decrease the prediction variance, 
although it may increase the prediction bias in a domain 
of the explanatory variable with a non-unimodal CPD, 
as schematized in Figure~\ref{fig:Idea} \hyl{Idead}{d}.
Thus, the unimodality-promotion by UPRL, if of suitable strength, 
can be expected to improve the prediction performance 
especially when one can access small-size data only 
so that the variance dominates the prediction error.

%==========%
However, in this study, we show that previous UPRL methods 
have a certain degree of the unimodality-promotion, 
but also have strong smoothness-promotion, 
namely, they promote a smoother (in other words, 
less-confident or larger-scale; see Definition~\ref{def:scale}) 
predicted CPD, contrary to the development idea: 
their regularizer can have a non-minimum contribution to 
the expectation value even when the underlying CPD is unimodal
and the predicted CPD takes the same value as the underlying CPD,
and that contribution tends to be smaller for a smoother predicted CPD.
This may lead to an unexpected scale-related bias.
Therefore, we propose a novel UPRL method 
that more strictly reflects the idea of UPRL:
our regularizer has not only a smaller contribution to the expectation 
value when the predicted CPD is closer to be unimodal,
but also a minimum contribution to the expectation 
value when the predicted CPD becomes unimodal.
Because of such design of the regularizer, we expect that our UPRL 
method will promote the predicted CPD closer to be unimodal,
but will not induce a bias into the predicted CPD 
when the underlying CPD is unimodal.

%==========%
The smoothness-promotion by previous UPRL methods may further 
decrease the prediction variance but increase the prediction bias 
especially for data with small-scale underlying CPD 
compared to the unimodality-promotion alone.
Therefore, considering the bias-variance trade-off, 
we expect that the proposed UPRL method may outperform 
a previous UPRL method for smaller-scale data 
especially when the training data size is larger.
In this paper, we will verify this expectation in addition to 
the unimodality- and smoothness-promotion of a previous UPRL method 
and the unimodality-promotion of a proposed UPRL method, through 
numerical experiments with real-world ordinal data.

%==========%
We organized this paper, with the main purposes of 
clarifying whether the unimodality-promotion truly 
contributes to improve the prediction performance and 
how the smoothness-promotion that is a difference between 
proposed strict UPRL methods and previous UPRL methods 
affects the prediction performance, as follows:
Sections~\ref{sec:Data} and \ref{sec:Method} give preparation for the discussion 
of this study, respectively including the formulation of ordinal data 
and unimodality hypothesis, and that of OR tasks and OR methods.
In Section~\ref{sec:UOLM}, we review unimodality-driven likelihood 
models \cite{yamasaki2022unimodal,yamasaki2025approximately} 
as an alternative approach to exploit the unimodality of ordinal data.
In Section~\ref{sec:Previous}, we review previous UPRL methods developed by 
\cite{belharbi2020,albuquerque2021ordinal,albuquerque2022quasi,Cardoso2025,kim2025cal}, and analyze 
those methods to show their unimodality- and smoothness-promotion.
In Section~\ref{sec:Proposed}, we describe a proposed UPRL method, 
and theoretically show that it strengthens the unimodality-promotion 
and weakens the smoothness-promotion.
In Section~\ref{sec:Experiments}, 
we present numerical experiments with real-world ordinal 
data to check the bias tendency of previous and 
proposed UPRL methods and compare their prediction performance.
There, we also experiment with the combination of 
a unimodality-driven likelihood model and UPRL.
Refer also to supplement for experimental comparison with other OR methods, 
including ordinal logistic regression \cite{mccullagh1980regression}.
%and to \url{https://github.com/Anonymized} for used program codes.
Finally, Section~\ref{sec:Conclusion} concludes this paper 
and presents future research directions:
in particular, a crucial weakness of the proposed UPRL method 
is high computational cost as we will verify in supplement, 
but this study prioritizes rigor with respect to the idea of the UPRL 
and leaves resolving this weakness to future research.

%========================================%
\section{Preparation}
\label{sec:Preparation}
%========================================%
\subsection{Ordinal Data and Unimodality Hypothesis}
\label{sec:Data}
%==========%
Suppose that one has data $(\bx_1,y_1),\cdots,(\bx_n,y_n)$ that are assumed to 
be drawn independently from an identical distribution of $(\bX,Y)\in\bbR^d\times[K]$, 
where $n,d,K\in\bbN$ such that $K\ge3$, and where $[K]\coloneq\{1,\ldots,K\}$.
In the OR framework, suppose further that the categorical target variable $Y$ is 
considered to have a natural ordinal relation in the order of $1,2,\ldots,K$ 
as examples described in the head of Section~\ref{sec:Introduction}, 
and such data are called ordinal data (or ordinal categorical data).

\begin{figure}[!t]
\centering%
\renewcommand{\tabcolsep}{0pt}%
\begin{tabular}{C{2.9cm}C{2.9cm}C{2.9cm}}%
\begin{overpic}[height=1.6cm]{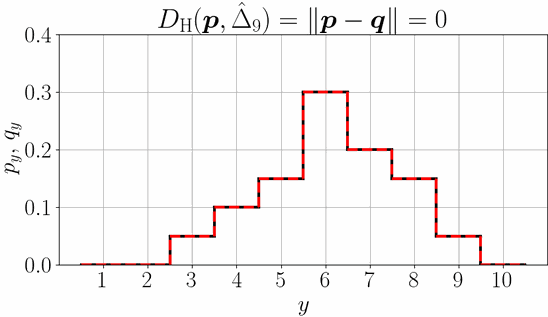}
\put(2,56){{\tiny\rm\hyt{Unima}{a}}}\end{overpic}&
\begin{overpic}[height=1.6cm]{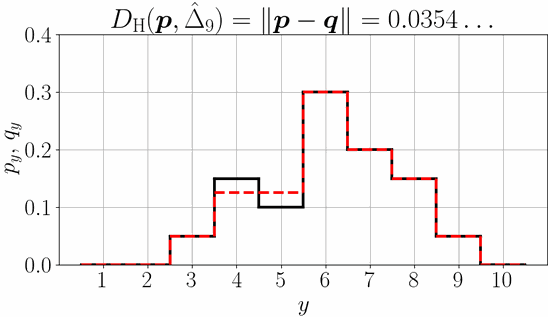}
\put(2,56){{\tiny\rm\hyt{Unimb}{b}}}\end{overpic}&
\begin{overpic}[height=1.6cm]{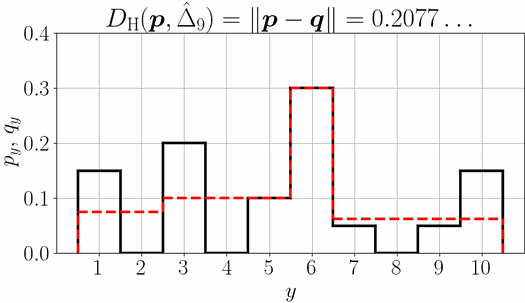}
\put(2,56){{\tiny\rm\hyt{Unimc}{c}}}\end{overpic}
\end{tabular}
\caption{%
Instance of $D_\rmH(\bp,\hat{\Delta}_{K-1})$ (above the figure) with $K=10$ and
$\bp=($0, 0, .05, .1, .15, .3, .2, .15, .05, 0$)^\top$ \hyl{Unima}{a},
$\bp=($0, 0, .05, .15, .1, .3, .2, .15, .05, 0$)^\top$ \hyl{Unimb}{b}, 
or $\bp=($.15, 0, .2, 0, .1, .3, .05, 0, .05, .15$)^\top$ \hyl{Unimc}{c},
where we show $\bp$ (solid black) and $\bq\in\argmin_{\bv\in\hat{\Delta}_{K-1}}\|\bp-\bv\|$ (dashed red).}
\label{fig:Unimodal}
\end{figure}

%==========%
For systematic discussion of OR, it would be important to 
formally define the natural ordinal relation that is likely to be 
common in many ordinal data and that we suppose in the study.
As a characterization of the natural ordinal relation, several previous OR studies 
\cite{da2008unimodal, iannario2011cub, beckham2017unimodal, 
yamasaki2022unimodal, yamasaki2024remarks, yamasaki2025approximately} 
supposed the unimodality hypothesis, stating that many real-world 
ordinal data tend to have a unimodal or close-to-be-unimodal CPD.
This study also adopts this hypothesis, more exactly, supposes that many 
ordinal data are high-UR and low-UD according to the following definition:
\begin{definition}[{high-UR and low-UD data 
\protect\cite{yamasaki2022unimodal,yamasaki2025approximately}}]
\label{def:unimodal}\hfill%
\begin{itemize}
\item%
For a probability mass function (PMF) $\bp=(p_k)_{k\in[K]}\in\Delta_{K-1}$, 
where $\Delta_{K-1}$ is the $(K-1)$-dimensional probability simplex 
$\{(p_k)_{k\in[K]}\in\bbR^K\mid\sum_{k=1}^K p_k=1, p_k\in[0,1]\text{ for all }k\in[K]\}$, 
we call every $M\in\argmax_{k\in[K]} p_k$ a {mode} of $\bp$, 
and say that $\bp$ is {unimodal} if it satisfies 
\begin{align}
\label{eq:unimodality}
	p_1\le\cdots\le p_M\text{~and~}p_M\ge\cdots\ge p_K
\end{align}
with any mode $M$.
Also, we introduce the set $\hat{\Delta}_{K-1}\coloneq\{\bp\in\Delta_{K-1}
\mid\bp\text{ is unimodal.}\}$ of unimodal PMFs.
\item%
We say that the data is {high-UR}, 
if the {unimodality rate (UR)} of the CPD 
$\bP(\bx)\coloneq(\Pr(Y=y|\bX=\bx))_{y\in[K]}$, 
\begin{align}
\label{eq:UR}
	\bbE_{\bx\sim\bX}[\bbI(\bP(\bx)\in\hat{\Delta}_{K-1})],
\end{align}
is high, where $\bbE_{\bx\sim\bX}[\cdot]$ is the expectation over $\bx\sim\bX$, 
and where$\bbI(c)$ is the indicator function that 
values 1 if the condition $c$ is true or 0 otherwise.
\item%
Also, we say that the data is {low-UD}, 
if deviation (which we call {unimodality deviation; UD}) 
$D(\bP(\bx))$ of the CPD $\bP(\bx)$ from the set $\hat{\Delta}_{K-1}$ 
of unimodal PMFs tends to be low at $\bx$ in 
whole the support of the distribution of $\bX$, 
where the UD $D:\Delta_{K-1}\to[0,\infty)$ should satisfy 
\begin{align}
\label{eq:UDrequire}
	D(\bp)=\min_{\bq\in\Delta_{K-1}}D(\bq)
	\text{~if and only if~}\bp\in\hat{\Delta}_{K-1}.
\end{align}
For example, we can measure the UD by $D_\rmH(\cdot,\hat{\Delta}_{K-1})$, 
where $D_\rmH(\bp,S)\coloneq\min_{\bv\in S}\|\bp-\bv\|$ 
with the Euclidean norm $\|\cdot\|$ is the Hausdorff 
distance between $\bp\in\bbR^K$ and $S\subseteq\bbR^K$ 
(namely, $D_\rmH(\bp,\hat{\Delta}_{K-1})$ implies the 
$L_2$-distance from $\bp$ to the nearest unimodal MPF;
see Figure~\ref{fig:Unimodal}).
\end{itemize}
\end{definition}

%==========%

%==========%
Previous works \cite{yamasaki2022unimodal, yamasaki2025approximately} 
verified that the many real-world ordinal data would be high-UR and low-UD, 
through experiments with real-world ordinal datasets.
We here reproduce their experimental verification, 
using 21 real-world ordinal datasets with the total data size 
$n_\tot\ge1000$ used in an OR survey \cite{gutierrez2015ordinal}.
{AB5}, \ldots, {CO5'} (resp.\ {AB10}, \ldots, {CO10'}) are datasets 
generated by discretizing a real-valued target of datasets, 
which are often used to benchmark regression methods, 
by 5 (resp.\,10) different bins with equal proportions.
{CAR}, {ERA}, {LEV}, {SWD}, {WQR} originally have a categorical target, 
and the authors of \cite{gutierrez2015ordinal} judged 
that these data have a natural ordinal relation.%
\footnote{%
One can get real-world ordinal datasets from a researchers'\,site 
(\url{https://www.uco.es/grupos/ayrna/orreview}) of \cite{gutierrez2015ordinal}.}
% or GitHub repository (\url{https://github.com/Anonymized}) for our study.}
Table~\ref{tab:Data-Prop} shows sample-based estimation of 
the UR $\bbE_{\bx\sim\bX}[\bbI(\bP(\bx)\in\hat{\Delta}_{K-1})]$ 
defined in \eqref{eq:UR}, and that of mean Hausdorff distance (MHD) 
\begin{align}
\label{eq:MHD}
	\bbE_{\bx\sim\bX}[D_\rmH(\bP(\bx),\hat{\Delta}_{K-1})]
\end{align}
as an indicator of the UD.
Here, in the estimation of the underlying CPD $\bP(\bx)$, 
we applied multinomial logistic regression (MLR) model 
(with non-regularized learning of $n_\tra=800$ in Section~\ref{sec:Experiments}).
Considering that, for standard classification data with no natural ordinal relation, 
it would be permissible to assign labels $1,\ldots,K$ 
to observations of the target variable in any order, 
we also gave evaluations for the uniform random variable on $\Delta_{K-1}$ 
as a simulation of the CPD of standard classification data.
Comparing the UR and MHD of real-world ordinal data and 
those of uniform random data of the same $K$, one can find that 
many real-world ordinal data enjoy strong tendency of the unimodality.

%==========%
\begin{table}[!t]
\centering%
\renewcommand{\arraystretch}{0.75}%
\renewcommand{\tabcolsep}{4pt}%
\caption{%
Properties of real-world ordinal data: 
`dataset' shows the dataset name used in this paper, 
$n_\tot$ is the total number of used data points, 
$d$ is the dimension of the explanatory variable, 
$K$ is the number of classes of the target variable, 
and `UR' and `MHD' show the mean and standard deviation (STD) 
of 100-trial estimates of UR (see \eqref{eq:UR}) and 
MHD (indicator of UD; see \eqref{eq:MHD}) as `mean$\pm$STD'.
Also, `uniform on $\Delta_{K-1}$' is of 100-trial estimates 
for uniform random data on $\Delta_{K-1}$ 
(instead of the CPD $\bP(\bx)$ in \eqref{eq:UR} and \eqref{eq:MHD}).
The larger UR or the smaller MHD, 
the stronger the tendency for the data to be unimodal.}
\label{tab:Data-Prop}
{\footnotesize\begin{tabular}{ccccccc}\toprule
\multicolumn{2}{c}{dataset}& $n_\tot$ & $d$ & $K$ & UR & MHD \\\midrule
\multirow{21}{*}{\rotatebox{90}{real-world~~~~~~}}
&SWD & 1000 & 10 & 4 & $1.0000\!\pm\!{.0000}$ & $.0000\!\pm\!{.0000}$ \\
&CO5' & 8192 & 21 & 5 & $1.0000\!\pm\!{.0000}$ & $.0000\!\pm\!{.0000}$ \\
&BA5 & 8192 & 8 & 5 & $1.0000\!\pm\!{.0000}$ & $.0000\!\pm\!{.0000}$ \\
&CO5 & 8192 & 12 & 5 & $.9998\!\pm\!{.0007}$ & $.0000\!\pm\!{.0000}$ \\
&BA10 & 8192 & 8 & 10 & $.9971\!\pm\!{.0203}$ & $.0000\!\pm\!{.0000}$ \\
&WQR & 1599 & 11 & 6 & $.9958\!\pm\!{.0212}$ & $.0000\!\pm\!{.0002}$ \\
&LEV & 1000 & 4 & 5 & $.9840\!\pm\!{.0295}$ & $.0000\!\pm\!{.0001}$ \\
&CH5 & 20640 & 8 & 5 & $.9714\!\pm\!{.0568}$ & $.0002\!\pm\!{.0006}$ \\
&CE5' & 22784 & 16 & 5 & $.9510\!\pm\!{.0751}$ & $.0008\!\pm\!{.0012}$ \\
&BA5' & 8192 & 32 & 5 & $.9202\!\pm\!{.1805}$ & $.0008\!\pm\!{.0020}$ \\
&CE5 & 22784 & 8 & 5 & $.9077\!\pm\!{.1095}$ & $.0011\!\pm\!{.0014}$ \\
&AB5 & 4177 & 10 & 5 & $.8925\!\pm\!{.1126}$ & $.0018\!\pm\!{.0021}$ \\
&CO10' & 8192 & 21 & 10 & $.8468\!\pm\!{.1452}$ & $.0004\!\pm\!{.0007}$ \\
&CAR & 1728 & 21 & 4 & $.8210\!\pm\!{.3386}$ & $.0006\!\pm\!{.0015}$ \\
&CO10 & 8192 & 12 & 10 & $.7818\!\pm\!{.1539}$ & $.0009\!\pm\!{.0012}$ \\
&ERA & 1000 & 4 & 9 & $.7367\!\pm\!{.1098}$ & $.0036\!\pm\!{.0029}$ \\
&CE10' & 22784 & 16 & 10 & $.6030\!\pm\!{.2214}$ & $.0032\!\pm\!{.0035}$ \\
&CH10 & 20640 & 8 & 10 & $.5410\!\pm\!{.2510}$ & $.0045\!\pm\!{.0036}$ \\
&CE10 & 22784 & 8 & 10 & $.4660\!\pm\!{.2268}$ & $.0042\!\pm\!{.0035}$ \\
&AB10 & 4177 & 10 & 10 & $.3534\!\pm\!{.1904}$ & $.0082\!\pm\!{.0056}$ \\
&BA10' & 8192 & 32 & 10 & $.2340\!\pm\!{.2398}$ & $.0074\!\pm\!{.0067}$ \\
\midrule
\multirow{5}{*}{\rotatebox{90}{synthesis~~}}
&uniform on $\Delta_3$ & $1000$ & - & 4 & $.3326\!\pm\!{.0135}$ & $.0752\!\pm\!{.0026}$ \\
&\hphantom{unifh}"\hphantom{kon} $\Delta_4$ & " & - & 5 & $.1314\!\pm\!{.0108}$ & $.1000\!\pm\!{.0023}$ \\
&\hphantom{unifh}"\hphantom{kon} $\Delta_5$ & " & - & 6 & $.0443\!\pm\!{.0065}$ & $.1162\!\pm\!{.0023}$ \\
&\hphantom{unifh}"\hphantom{kon} $\Delta_8$ & " & - & 9 & $.0009\!\pm\!{.0010}$ & $.1365\!\pm\!{.0016}$ \\
&\hphantom{unifh}"\hphantom{kon} $\Delta_9$ & " & - & 10 & $.0001\!\pm\!{.0003}$ & $.1385\!\pm\!{.0014}$ \\
\bottomrule\end{tabular}}
\end{table}

%========================================%
\subsection{Ordinal Regression Tasks and Methods}
\label{sec:Method}
%==========%
OR tasks are typically formulated as searching for a classifier 
$f:\bbR^d\to[K]$ that leads to smaller task risk 
$\bbE[\ell(f(\bX),Y)]=\bbE_{\bx\sim\bX}[\sum_{y=1}^K\Pr(Y=y|\bX=\bx)\ell(f(\bx),y)]$ 
for a user-specified task loss $\ell:[K]^2\to[0,\infty)$.
For OR tasks, popular instances of the task loss are $\ell(u,v)=\bbI(u\neq v), |u-v|, (u-v)^2$, 
where the corresponding (empirical) task risk is called mean zero-one error (MZE), 
mean absolute error (MAE), and mean squared error (MSE).
Since the optimal classifier 
$\bar{f}\in\argmin_{f:\bbR^d\to[K]}\bbE[\ell(f(\bX),Y)]$ in this setting satisfies 
$\bar{f}(\bx)\in\argmin_{k\in[K]}\sum_{y=1}^K\Pr(Y=y|\bX=\bx)\ell(k,y)$ a.s., 
many OR methods build their classifier via estimating the conditional probability.
Since good estimation of the conditional probability will 
lead to good performance of an OR method in OR tasks as well, 
we hereafter focus our discussion on the conditional probability estimation task.

%==========%
For the conditional probability estimation task, 
we specify a likelihood model $(Q,\calG)$, which consists of 
a fixed part $Q$ and a set $\calG$ of a learnable part, such that 
we estimate the conditional probability $\Pr(Y=y|\bX=\bx)$ 
as $Q(\bg(\bx),y)$ with a learned model $\bg\in\calG$ 
and typically build a likelihood-based classifier 
$f(\bx)\in\argmin_{k\in[K]}\sum_{y=1}^KQ(\bg(\bx),y)\ell(k,y)$;
for example, MLR model adopts softmax function
$Q_\smax(\bu,y)\coloneq e^{u_y}/\sum_{k=1}^Ke^{u_k}$ as $Q(\bu,y)$ and 
$\calG\subseteq\{\bg:\bbR^d\to\bbR^K\}$ (say, a certain neural network model).
In the learning procedure of a learnable part $\bg$, we typically minimize 
a loss $\phi$ based on a distribution-to-distribution divergence 
between the likelihood $\bQ(\bg(\cdot))\coloneq(Q(\bg(\cdot),y))_{y\in[K]}$ 
and the empirical distribution of the data $(\bx_1,y_1),\ldots,(\bx_n,y_n)$, 
that is, $\min_{\bg\in\calG}\frac{1}{n}\sum_{i=1}^n\phi(\bQ(\bg(\bx_i)),y_i)$.
Here, we suppose that the loss function $\phi$ has the consistency 
\cite{bianco1996robust,yamasaki2023LS}, that is, $\bP(\bx)=\bQ(\bar{\bg}(\bx))$ a.s.\;for $\bx\sim\bX$
with $\bar{\bg}\in\argmin_{\bg\in\calG}\bbE[\phi(\bQ(\bg(\bX)),Y)]$ 
if $\bP\in\{\bQ(\bg(\cdot))\mid\bg\in\calG\}$.
A representative divergence-based loss is the negative log likelihood (NLL) 
$-\frac{1}{n}\sum_{i=1}^n\log Q(\bg(\bx_i),y_i)$ with $\phi(\bu,y)=-\log u_y$.

%==========%
The learning procedure using a consistent loss alone works well when 
the training data size is large enough as it does not yield unnecessary bias, 
but there is possibility for improvement when the training data size is small.
To improve the prediction performance 
when the size of available training data is small, 
it is promising to leverage prior domain knowledge on the data additionally.
Thus, as we will see in following sections, 
OR studies propose devising a likelihood model (in Section~\ref{sec:UOLM}) 
or learning procedure (in Sections~\ref{sec:Previous} and \ref{sec:Proposed}) 
to exploit the unimodality of ordinal data.

%========================================%
\subsection{Related Ordinal Regression Methods: Unimodality-Oriented Likelihood Models}
\label{sec:UOLM}
%==========%
Of the two approaches to exploit the unimodality of ordinal data, 
`devising a likelihood model' and `devising a learning procedure,' 
we mainly discuss the latter and propose a method following that approach in this paper.
In this section, we review methods following the former as a contender to our proposal 
and as a component to be combined with our proposal.

%==========%
Previous works \cite{yamasaki2022unimodal, da2008unimodal, iannario2011cub, beckham2017unimodal} 
developed unimodal likelihood (UL) models $(Q,\calG)$ that satisfy 
$\bQ(\bg(\bx))\subseteq\hat{\Delta}_{K-1}$ for any $\bx\in\bbR^d$ and $\bg\in\calG$.
Especially, a UL model $(Q,\calG)$, which is 
developed by \cite{yamasaki2022unimodal} and consisted of 
\begin{align}
\label{eq:VSLLink}
	Q(\bu,y)=Q_\smax(-\tau(\rho[\bu]),y)
	\text{~for~}y\in[K],\bu\in\bbR^K
\end{align}
can represent arbitrary unimodal CPD with $\calG=\{\bg:\bbR^d\to\bbR^K\}$.
Here, $\rho[\bu]=\acute{\bu}$ is constructed as 
\begin{align}
	\acute{u}_k
	=\begin{cases}
	u_1&\text{for }k=1,\\
	\acute{u}_{k-1}+\rho(u_k)&\text{for }k=2,\ldots,K,
	\end{cases}
\end{align}
so that it satisfies $\acute{u}_1\le\cdots\le\acute{u}_K$, 
with a user-specified non-negative function $\rho$ satisfying 
\begin{align}
\label{eq:RHO}
	\{\rho(u)\mid u\in\bbR\}=[0,\infty)
\end{align}
such as $\rho(u)=u^2$.
Also, $\tau(\rho[\bu])=\tau(\acute{\bu})=\check{\bu}$ is constructed as 
\begin{align}
	\check{u}_k=\tau(\acute{u}_k)\text{ for }k\in[K],
\end{align}
so that $\acute{\bg}$ becomes V-shaped regarding the index, 
with a user-specified V-shaped function $\tau$ satisfying 
\begin{align}
	&\tau(u)\text{~is non-increasing in~}u<0\text{~and non-decreasing in~}u>0
\nonumber\\
\label{eq:TAU}
	&\text{and~}\{\tau(u)\mid u\le0\}=\{\tau(u)\mid u\ge0\}=[\tau(0),\infty)
\end{align}
such as $\tau(u)=u^2$.

%==========%
Experiments by \cite{yamasaki2022unimodal} showed that 
many real-world ordinal data partly have values of the explanatory variable 
where the underlying CPD will not be unimodal 
(as UR in Table~\ref{tab:Data-Prop} is not equal to 1).
UL models have a bias for such a CPD.
The succeeding work \cite{yamasaki2025approximately} verified that 
the underlying CPD of ordinal data would be close to be unimodal 
even at values where the CPD is not unimodal strictly 
(see MHD in Table~\ref{tab:Data-Prop}), 
and, in order to represent such a CPD and to mitigate a bias of UL models, 
it developed a model $(Q,\calG)$ consisted of 
\begin{align}
	&Q((\bu_1,\bu_2), y)
	=(1-r)Q_\smax(-\tau(\rho[\bu_1]),y)+r Q_\smax(\bu_2,y)
\nonumber\\
\label{eq:MAUL}
	&\text{and }\calG\subseteq\{\bg:\bbR^d\to\bbR^{2K}\}
\end{align}
with a user-specified mixture rate parameter $r\in[0,1]$.
This model $(Q,\calG)$ is an instance of what is called an approximately unimodal 
likelihood (AUL) model in \cite{yamasaki2025approximately} and satisfies 
$D_\rmH(\bQ(\bg(\bx)),\hat{\Delta}_{K-1})\le\sqrt{2}r$ for any $\bx\in\bbR^d$ and $\bg\in\calG$; 
refer to \cite[Theorem 3]{yamasaki2025approximately}.

%==========%
Experiments in \cite{yamasaki2022unimodal, yamasaki2025approximately} showed that 
UL models and AUL models provide better prediction performance than unrestricted 
likelihood models such as MLR model when the training data size is small.
These results can be understood by considering the bias-variance trade-off 
\cite{bishop2006pattern}, that many ordinal data are high-UR and low-UD 
(i.e., small bias), and that UL and AUL models have lower 
{representation ability} $\{\bx\mapsto\bQ(\bg(\bx))\mid\bg\in\calG\}$ 
than unrestricted likelihood models (i.e., small variance); 
see also Figure~\ref{fig:Idea} \hyl{Ideaa}{a}--\hyl{Ideac}{c}.

%========================================%
\section{Previous Unimodality-Promoting Regularized Learning Methods}
\label{sec:Previous}
%========================================%
\subsection{Review}
\label{sec:Review}
%==========%
Another promising approach to reflect the unimodality hypothesis 
is to devise the learning procedure.
Such a device is called regularization, and typically addressed 
by changing the objective function used for learning a model:
it adds a regularization term $\lambda\Omega(\bg)$ 
to the conventional objective function 
consisted of a consistent loss function $\phi$, 
where $\lambda>0$ and $\Omega$ are called 
regularization parameter and regularizer, respectively.
In order to reflect the unimodality of ordinal data to the learning result, 
it would be appropriate that the regularizer $\Omega(\bg)$ should take 
a smaller value when the predicted CPD $\bQ(\bg(\bx))$ is closer to be unimodal; 
we call such regularization-based methods, 
which promote a predicted CPD to be close to be unimodal, 
unimodality-promoting regularized learning (UPRL) methods in this paper.

\begin{figure}[!t]
\centering%
\renewcommand{\arraystretch}{1.5}%
\renewcommand{\tabcolsep}{4pt}%
\begin{tabular}{cc}%
\begin{overpic}[width=4cm]{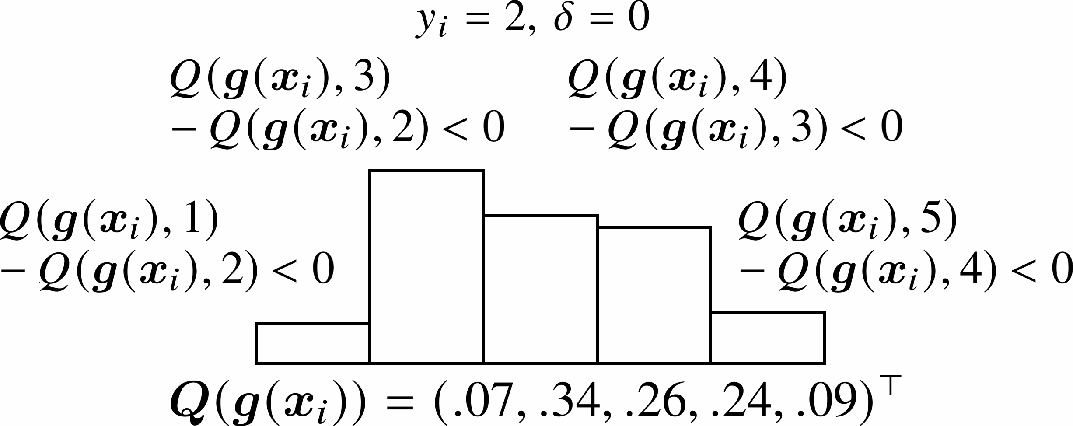}
\put(2,36){{\tiny\rm\hyt{Preva}{a}}}\end{overpic}&
\begin{overpic}[width=4cm]{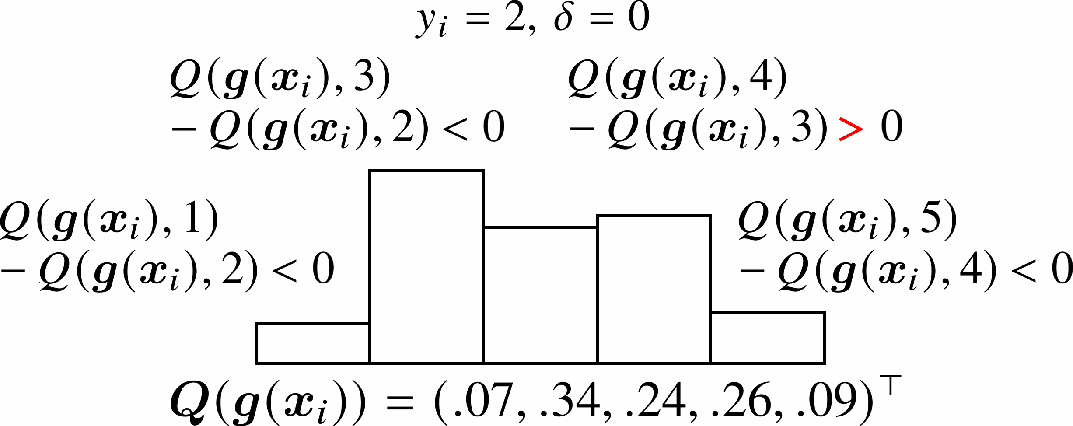}
\put(2,36){{\tiny\rm\hyt{Prevb}{b}}}\end{overpic}
\\%
\begin{overpic}[width=4cm]{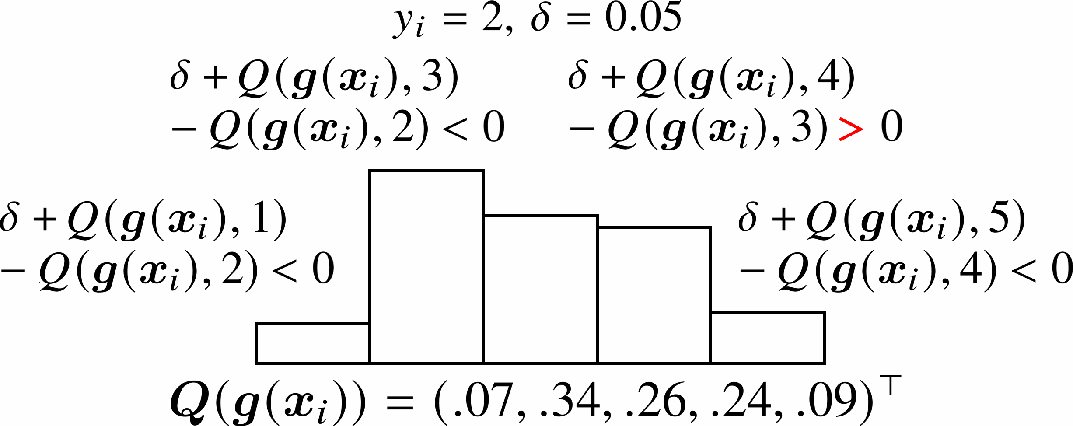}
\put(2,36){{\tiny\rm\hyt{Prevc}{c}}}\end{overpic}&
\begin{overpic}[width=4cm]{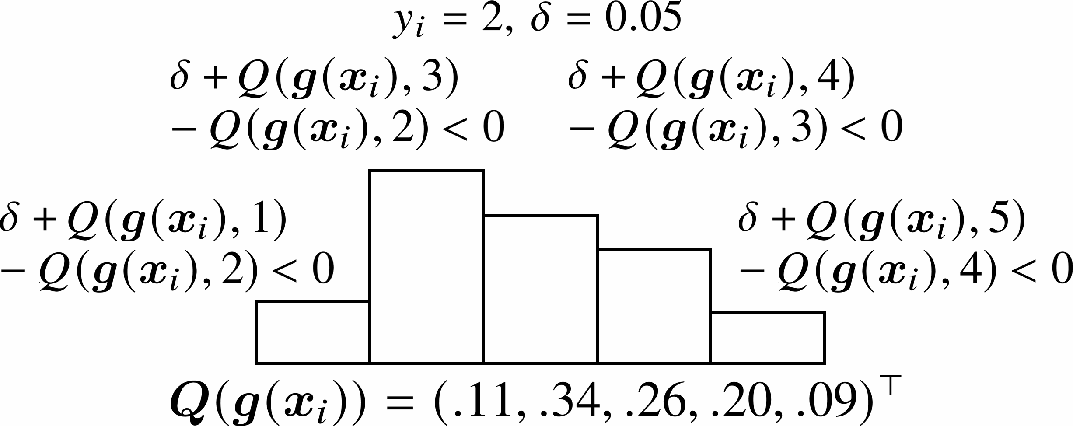}
\put(2,36){{\tiny\rm\hyt{Prevd}{d}}}\end{overpic}
\\%
\multicolumn{2}{c}{%
\begin{overpic}[width=4cm]{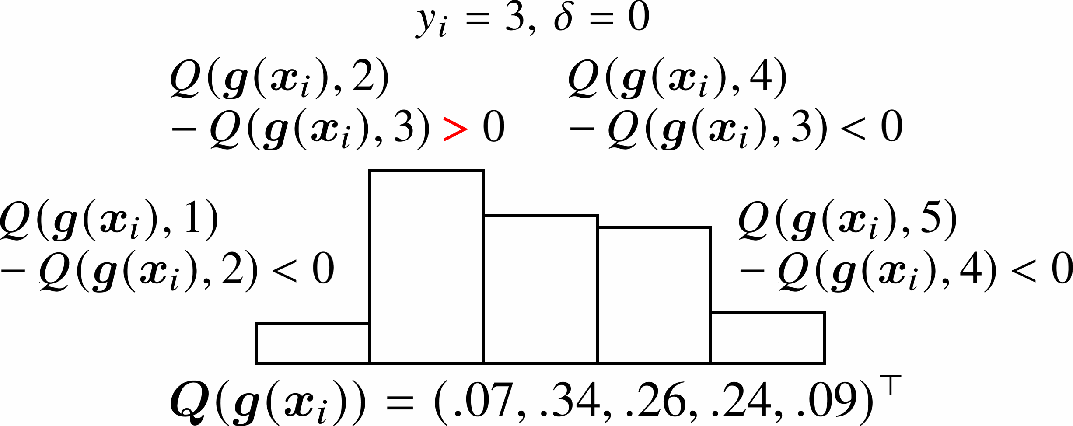}
\put(2,36){{\tiny\rm\hyt{Preve}{e}}}\end{overpic}}
\end{tabular}
\caption{%
Illustration of behaviors of the regularizer of previous UPRL methods:
$i$-th summand of the regularizer \eqref{eq:alb21} becomes minimum 
at a specified $\bQ(\bg(\bx_i))$ in \hyl{Preva}{a} and \hyl{Prevd}{d}
and not in \hyl{Prevb}{b}, \hyl{Prevc}{c}, and \hyl{Preve}{e}.}
\label{fig:Previous}
\end{figure}

%==========%
With the aim of penalizing the non-unimodality of a predicted CPD $\bQ(\bg(\cdot))$, 
the previous works \cite{belharbi2020,albuquerque2021ordinal} proposed the idea of UPRL.
Representative one of their regularizers is 
\begin{align}
	&\Omega_{\prev,\delta}(\bg)
\nonumber\\
	&=\frac{1}{n}\sum_{i=1}^n\sum_{k=1}^{K-1}
	\Bigl(\bbI(k<y_i)[\delta+Q(\bg(\bx_i),k)-Q(\bg(\bx_i),k+1)]_+
\nonumber\\
\label{eq:alb21}
	&+\bbI(k\ge y_i)[\delta+Q(\bg(\bx_i),k+1)-Q(\bg(\bx_i),k)]_+\Bigr)
\end{align}
with the ReLU function $[u]_+\coloneq\max\{0,u\}$
and a user-specified parameter $\delta\ge0$.
This regularizer is based on the fact that 
the unimodality \eqref{eq:unimodality} is 
a condition based on comparison of consecutive probabilities, 
and a graphical understanding that the unimodal 
$\bQ(\bg(\bx_i))$ in Figure~\ref{fig:Previous} \hyl{Preva}{a} is 
promoted over the one in Figure~\ref{fig:Previous} \hyl{Prevb}{b}.
Also, the succeeding works \cite{albuquerque2022quasi,Cardoso2025,kim2025cal} 
developed a similar variant that can be understood as 
a weighted version of the regularizer \eqref{eq:alb21}, 
or a valiant in which ReLU is replaced with another function.

%==========%

%==========%
The authors of these works explained behaviors of these methods by the bias-variance trade-off;
increasing the regularization parameter $\lambda$ increases a bias-dependent term 
of the prediction performance, and decreases a variance-dependent term.
Also, \cite{yamasaki2025approximately} verified that previous UPRL methods 
bring a large improvement especially for small-size training data.

%==========%
As a likelihood model $(Q,\calG)$, 
\cite{belharbi2020,albuquerque2021ordinal,albuquerque2022quasi,Cardoso2025,kim2025cal} 
considered only MLR model, 
and \cite{yamasaki2025approximately} considered 
an AUL model (see Section~\ref{sec:UOLM}) as well;
because an AUL model only controls an upper bound of 
the UD $D_\rmH(\bQ(\bg(\bx)),\hat{\Delta}_{K-1})$, 
but does not have the effect of reducing the UD further 
below that bound, and works differently from UPRL,
combining the previous UPRL method and an AUL model 
could further improve the prediction performance.

%========================================%
\subsection{Analysis}
\label{sec:PrevAnaly}
%==========%
The work \cite{albuquerque2021ordinal} introduced the parameter $\delta>0$ 
into \eqref{eq:alb21} with the purpose for ensuring that the difference between 
consecutive probabilities (say, $|Q(\bg(\bx_i),1)-Q(\bg(\bx_i),2)|$) is at least $\delta$;
for example, they considered that 
$\bQ(\bg(\bx_i))$ in Figure~\ref{fig:Previous} \hyl{Prevd}{d}
is promoted over the one in Figure~\ref{fig:Previous} \hyl{Prevc}{c}.
However, this purpose is clearly not established when 
the number $K$ of classes is large 
(consider that the summation of the components of 
the unimodal$(0,\delta,2\delta,\cdots,(K-1)\delta)^\top$ 
is $\frac{(K-1)K}{2}\delta$ and can exceed 1). 
We thus have to note that the introduction of 
$\delta>0$ is not rationally motivated.

%==========%
Even excluding the issue of $\delta>0$, we consider that 
the previous method \eqref{eq:alb21} was not designed 
in a theoretically adequate manner for the purpose of 
penalizing the non-unimodality of a predicted CPD $\bQ(\bg(\bx))$.
For example, $i$-th summand of the regularizer \eqref{eq:alb21} 
takes a non-minimum value when $y_i\not\in\argmax_{k\in[K]}Q(\bg(\bx_i),k)$ 
(compare Figure~\ref{fig:Previous} \hyl{Prevc}{a} and \hyl{Preve}{e}), 
but such situations are not taken into consideration.
In this section, we present an analysis that demonstrates 
behaviors of the previous method \eqref{eq:alb21} 
more rigorously from a theoretical perspective.

%==========%
The expectation value of the summands of the regularizer \eqref{eq:alb21} conditioned 
on $\bX=\bx$ is $\Pr(\bX=\bx)\omega_{\prev,\delta}(\bP(\bx),\bQ(\bg(\bx)))$ with
\begin{align}
\label{eq:PrevUPRLloss}
	\begin{split}
	&\omega_{\prev,\delta}(\bp,\bq)\coloneq\sum_{y=1}^K p_y \sum_{k=1}^{K-1} 
	\Bigl(\bbI(k<y) [\delta+q_k-q_{k+1}]_+
\\
	&\hphantom{\omega_{\prev,\delta}(\bp,\bq)\coloneq\sum_{y=1}^K p_y}
	+\bbI(k\ge y)[\delta+q_{k+1}-q_k]_+\Bigr),
	\end{split}
\end{align}
which we call expected UPRL loss of the previous UPRL method.
Considering that the data may be distributed such that 
$\bx_1=\bx_2$ and $y_1\neq y_2$ and that UPRL may 
have different effects on different $\bx$'s when it uses 
a likelihood model $\bQ(\bg(\bx))$ of sufficient representation ability, 
examining behaviors of the expected UPRL loss should 
help understand bias tendency of that UPRL method.

%==========%
First, we present a simple demonstration of behaviors of the expected UPRL loss:
when $K=5$ and $\bp=(0,0.2,0.6,0.2,0)^\top$, it holds that
\begin{align}
\label{eq:Exa11}
	\begin{split}
	&\omega_{\prev,0}(\bp,\bq)
\\
	&=\begin{cases}
	0.4&\text{for }\bq=(0,0,1,0,0)^\top,\\
	0.28&\text{for }\bq=(0,0.1,0.8,0.1,0)^\top,\\
	0.16&\text{for }\bq=(0,0.2,0.6,0.2,0)^\top,\\
	0.08&\text{for }\bq=(0.1,0.2,0.4,0.2,0.1)^\top,\\
	0&\text{for }\bq=(0.2,0.2,0.2,0.2,0.2)^\top,
	\end{cases}
	\end{split}
\end{align}
where $\bp$ and $\bq$ are unimodal 
(here, $\bp$ and $\bq$ play the roles of the underlying 
CPD $\bP(\bx)$ and likelihood $\bQ(\bg(\bx))$).
In this demonstration, the expected UPRL loss $\omega_{\prev,0}(\bp,\bq)$ 
is not minimum at $\bq=\bp$ even for unimodal $\bp\neq(0.2,0.2,0.2,0.2,0.2)^\top$.
This indicates that the regularizer \eqref{eq:alb21} 
may induce an unexpected bias other than the unimodality-promotion, 
and raises the question of whether that regularizer indeed has the unimodality-promotion.
Regarding a bias other than the unimodality-promotion, 
with reference to the notion of the scale or equivalently smoothness defined below, 
this demonstration suggests that the regularizer \eqref{eq:alb21} may have 
the smoothness-promotion that enlarges the scale of the predicted CPD, 
because the expected UPRL loss $\omega_{\prev,0}(\bp,\bq)$ 
is smaller for a smoother $\bq$:
\begin{definition}[{Scale or smoothness \protect\cite{yamasaki2022unimodal}}]
\label{def:scale}
For a PMF $\bp=(p_k)_{k\in[K]}\in\Delta_{K-1}$ having a mode $M\in[K]$, 
we call the degree of spread of $\bp$ around $M$ the {scale} of $\bp$, 
and say `having a large scale' as `being {smoother}'.
In this paper, we measure the scale of $\bp$ by $1-p_M$.
\end{definition}

\begin{figure}[!t]
\centering%
\renewcommand{\arraystretch}{1.5}%
\renewcommand{\tabcolsep}{4pt}%
\begin{tabular}{cc}%
\begin{overpic}[width=4cm]{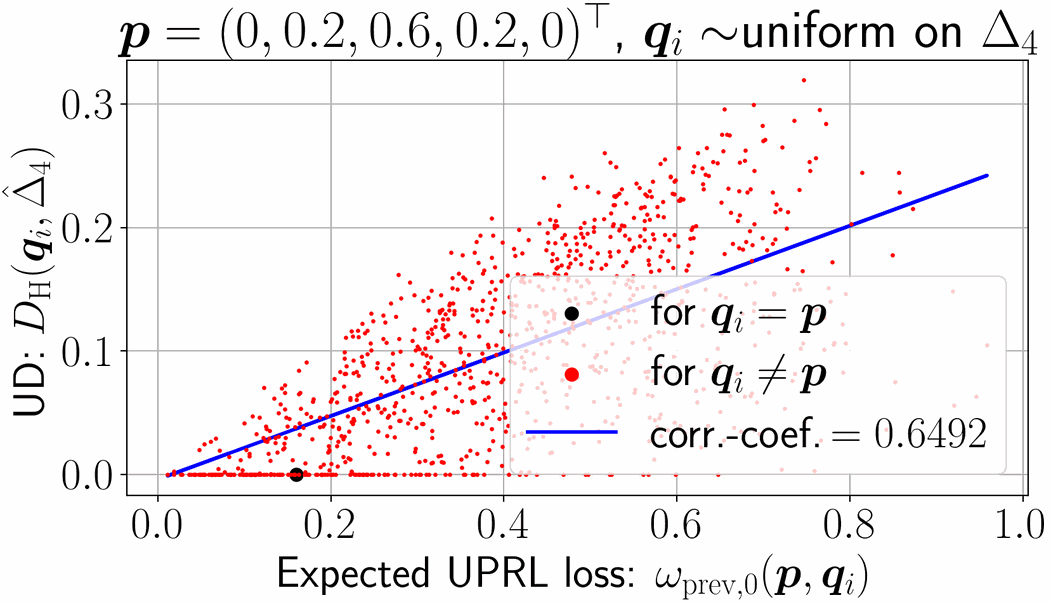}
\put(2,54){{\tiny\rm\hyt{UDau}{au}}}\end{overpic}&
\begin{overpic}[width=4cm]{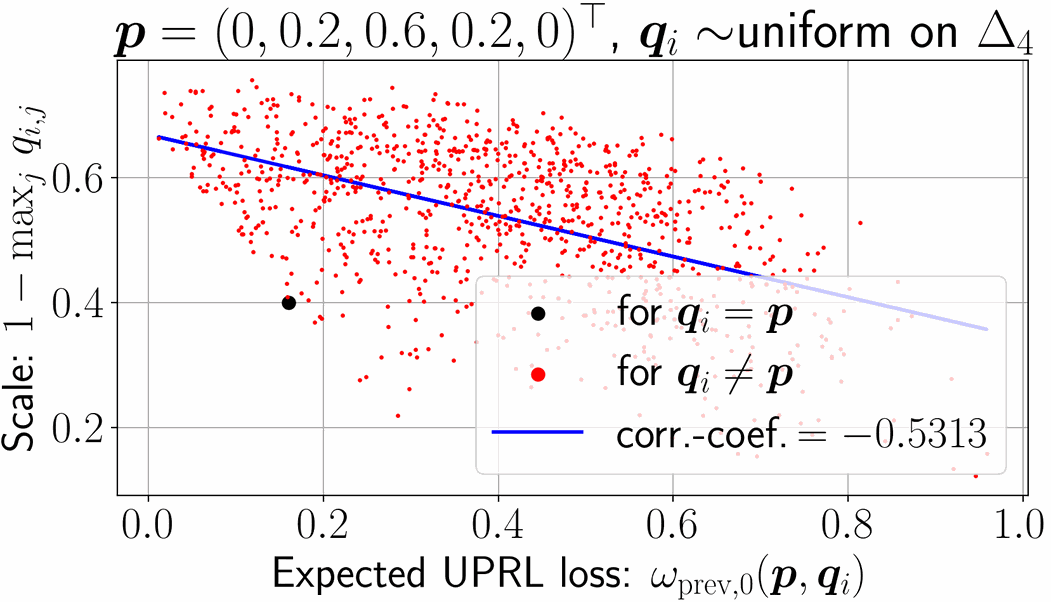}
\put(2,54){{\tiny\rm\hyt{UDas}{as}}}\end{overpic}
\\%
\begin{overpic}[width=4cm]{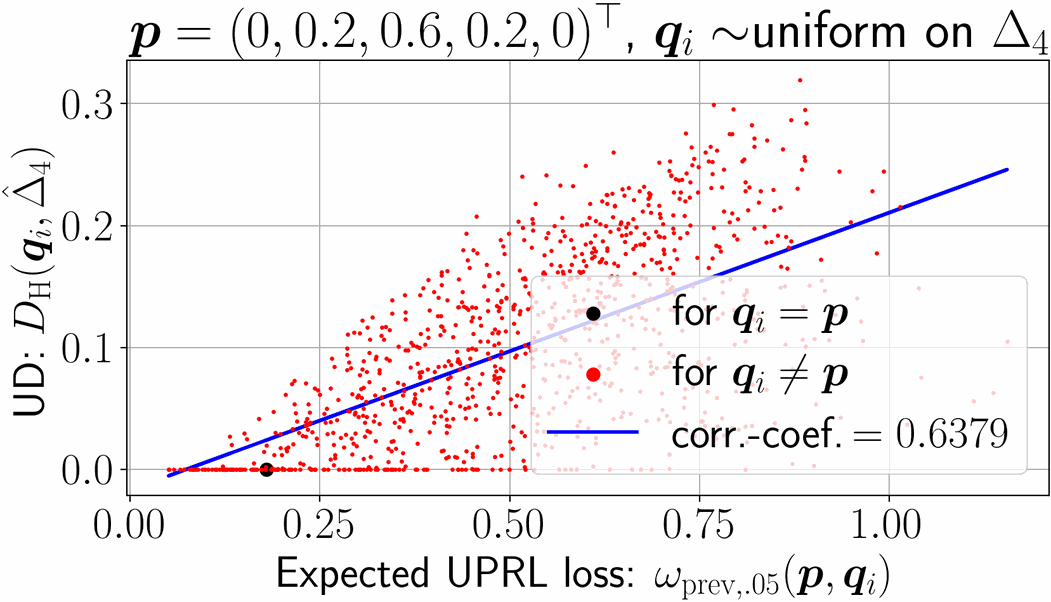}
\put(2,54){{\tiny\rm\hyt{UDbu}{bu}}}\end{overpic}&
\begin{overpic}[width=4cm]{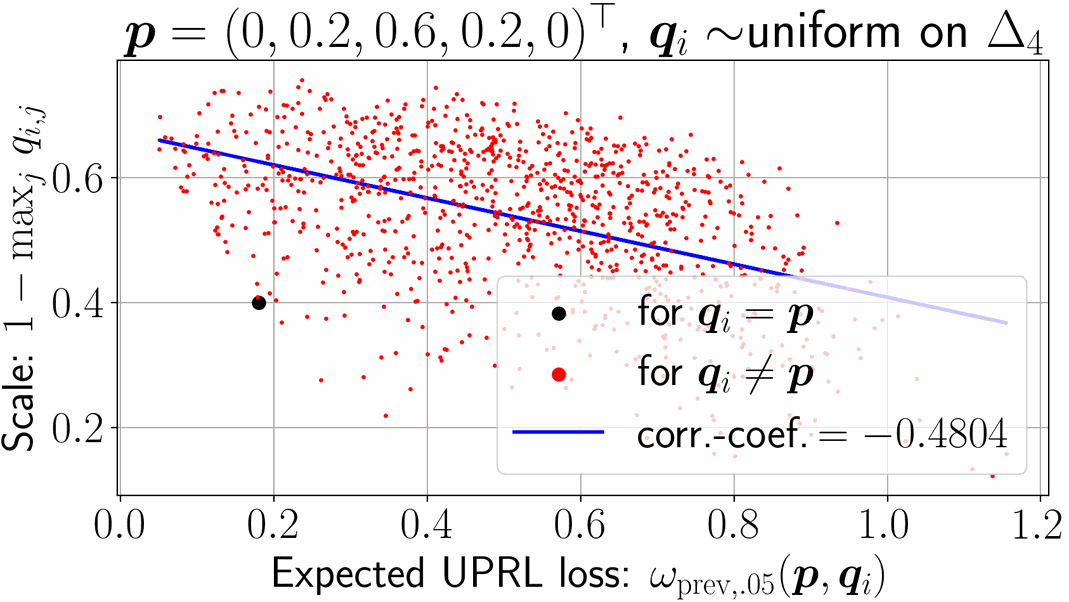}
\put(2,54){{\tiny\rm\hyt{UDbs}{bs}}}\end{overpic}
\\%
\begin{overpic}[width=4cm]{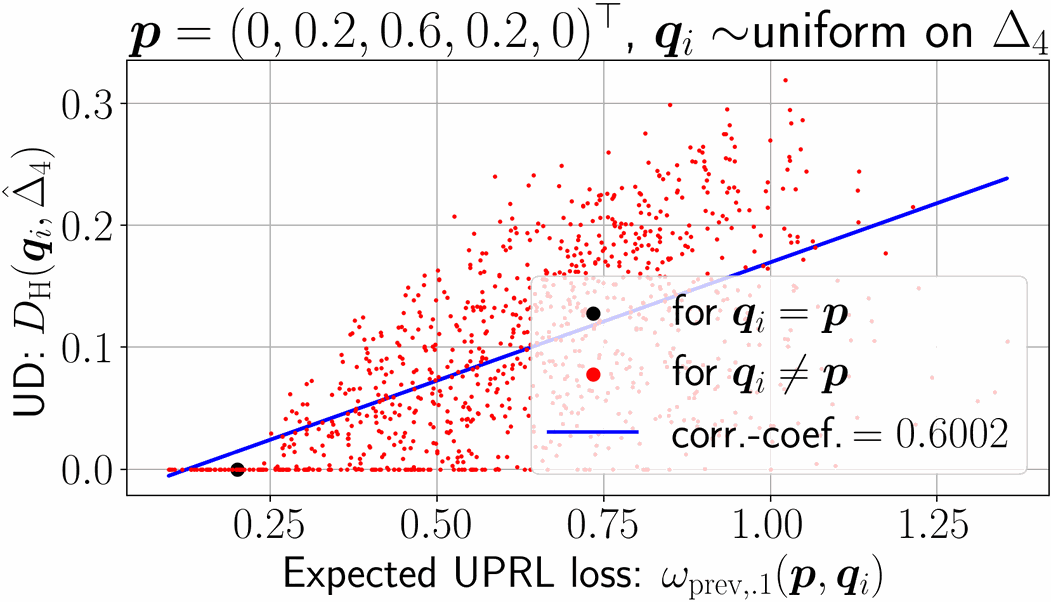}
\put(2,54){{\tiny\rm\hyt{UDcu}{cu}}}\end{overpic}&
\begin{overpic}[width=4cm]{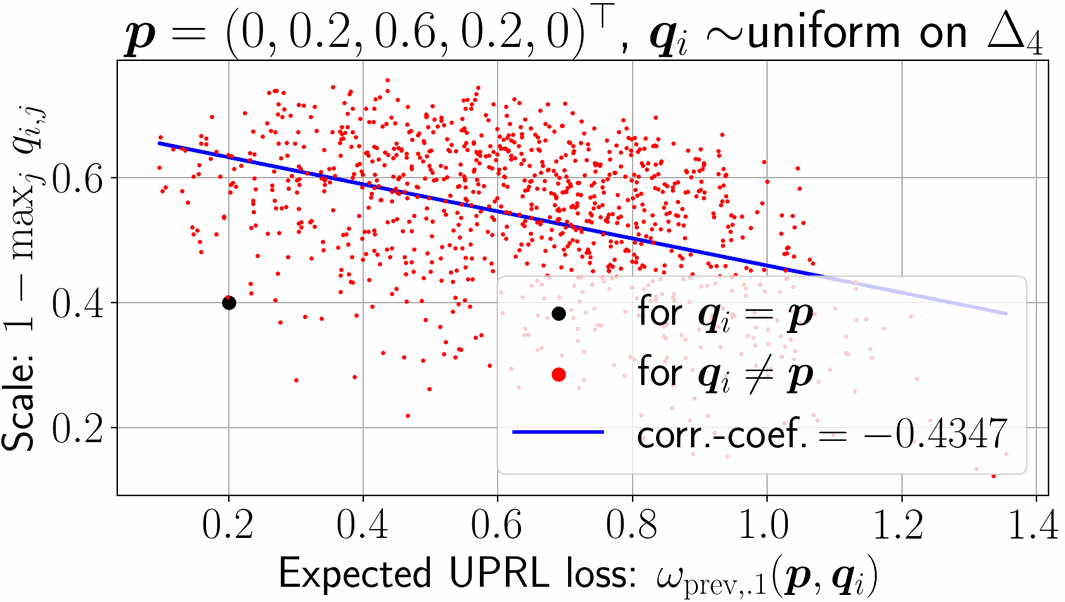}
\put(2,54){{\tiny\rm\hyt{UDcs}{cs}}}\end{overpic}
\\%
\begin{overpic}[width=4cm]{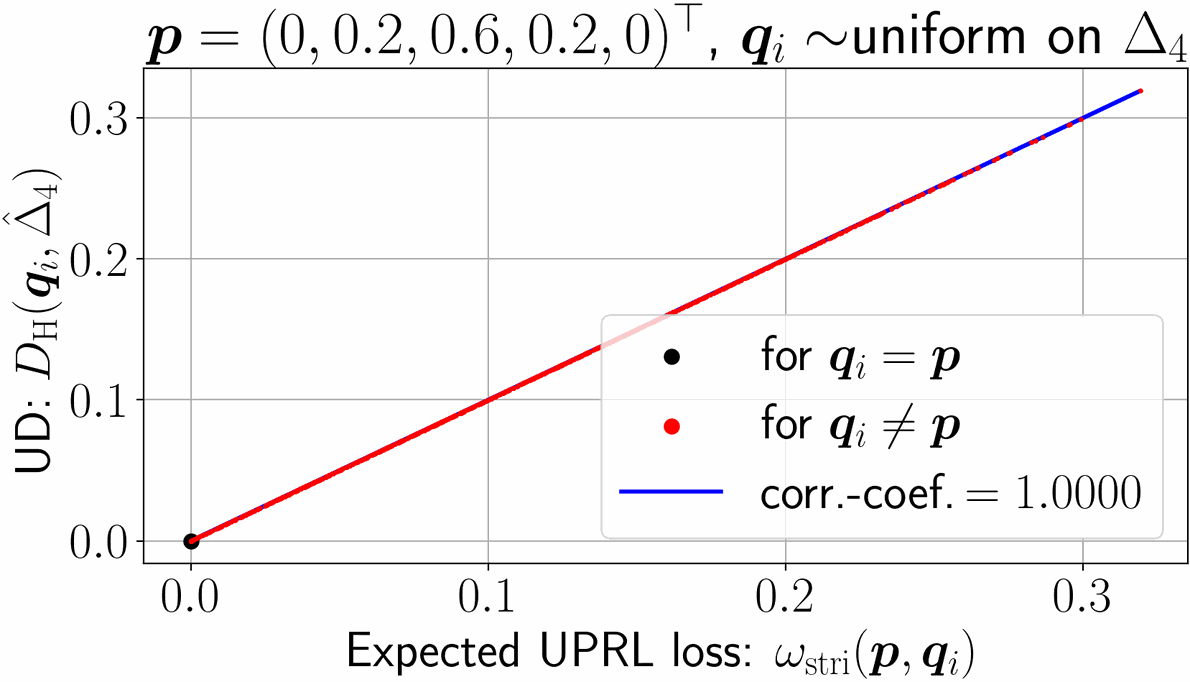}
\put(2,54){{\tiny\rm\hyt{UDdu}{du}}}\end{overpic}&
\begin{overpic}[width=4cm]{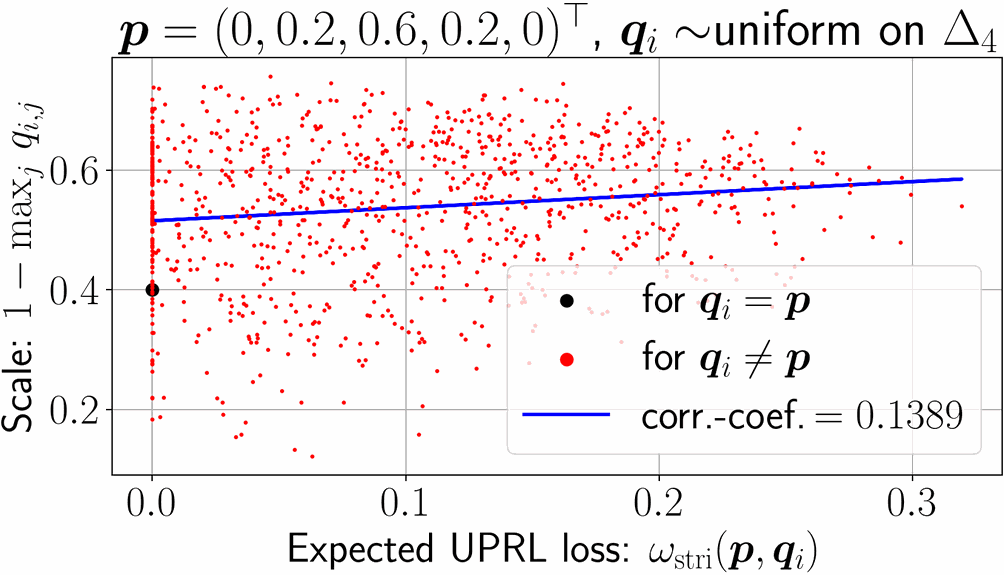}
\put(2,54){{\tiny\rm\hyt{UDds}{ds}}}\end{overpic}
\end{tabular}
\caption{%
Relationship between the expected UPRL loss,
previous UPRL with $\delta=0$ in (a{\textasteriskcentered}), 
$\delta=0.05$ in (b{\textasteriskcentered}), 
or $\delta=0.1$ in (c{\textasteriskcentered}), 
or strict UPRL in (d{\textasteriskcentered}),
and UD $D_\rmH(\bq_i,\hat{\Delta}_4)$ ({\textasteriskcentered}u)
or scale $1-\max_{j\in[5]}q_{i,j}$ ({\textasteriskcentered}s).
Note that the solid blue line is the simple linear regression line.}
\label{fig:UDScale}
\end{figure}

%==========%
In order to verify this question, we performed additional simulation:
Figure~\ref{fig:UDScale} shows the results of simulation with 
$K=5$, $\bp=(0,0.2,0.6,0.2,0)^\top$, and $\bq=\bq_i$, $i\in[10^3]$ 
drawn from the uniform distribution on $\Delta_4$, 
and the supplement shows similar results with different $K$'s and $\bp$'s.
First, the correlation coefficient between 
the expected UPRL loss $(\omega_{\prev,\delta}(\bp,\bq_i))_{i\in[10^3]}$ 
and the UD $(D_\rmH(\bq_i,\hat{\Delta}_4))_{i\in[10^3]}$ is moderately positive.
This indicates that decreasing the value of the regularizer \eqref{eq:alb21} 
leads to decreasing the UD of a predicted CPD $\bQ(\bg(\bx))$.
Thus, the previous UPRL method based on the regularizer \eqref{eq:alb21} 
would achieve the unimodality-promotion to some extent.
Furthermore, it can also be observed that the correlation coefficient between 
the expected UPRL loss $(\omega_{\prev,\delta}(\bp,\bq_i))_{i\in[10^3]}$ and 
the scale $(1-\max_{j\in[5]}q_{i,j}))_{i\in[10^3]}$ is a negative unignorable value.
In other words, we can see that reducing the value of the regularizer 
\eqref{eq:alb21} would have the smoothness-promotion.

%==========%
The smoothness-promotion was not intended in previous UPRL studies 
\cite{belharbi2020,albuquerque2021ordinal,albuquerque2022quasi,Cardoso2025,kim2025cal}.
However, it has been studied in various papers along with 
other keywords such as `less confidence' or `avoiding over-fitting,' 
and has been incorporated into various methods, for example,
ridge regression \cite{hoerl1970ridge}, and label-smoothing 
\cite{szegedy2016rethinking,yamasaki2023LS}.
Like these preceding methods the smoothness-promotion by previous UPRL methods 
would induce the estimation bias but may reduce the estimation variance.
For this reason, although it has been confirmed that 
previous UPRL methods can perform well with small-size 
training data in experiments by \cite{yamasaki2025approximately}, 
it is no longer clear whether that improvement is due to 
the unimodality- or smoothness-promotion or both.

%==========%

%==========%
Note that, by enlarging $\delta$, regularization of previous methods with 
a fixed $\lambda$ seems to weaken the unimodality- and smoothness-promotion;
in Figure~\ref{fig:UDScale} \hyl{UDau}{au}--\hyl{UDcu}{cu} and \hyl{UDas}{as}--\hyl{UDcs}{cs},
the absolute value of the correlation coefficient decreases as $\delta$ increases
(see the supplement for additional verification).
However, such differences of the regularization effect due to differences 
in $\delta$ will be absorbed by changing $\lambda$.
Therefore, we could not clarify the significance of introducing $\delta>0$ 
with respect to the unimodality- and smoothness-promotion.

%==========%
It should be noted that, the analysis thus far has been 
conducted under the simplification that the regularization 
exerts an independent effect on each point $\bx$.
This simplification is reasonable only when using a likelihood 
model $\bQ(\bg(\bx))$ with sufficient representation ability.
When using a restrictive model such as a linear model, 
the bias induced from the design of $\calG$ would
significantly influence the prediction performance.
Also, we additionally note that our analysis assumes the use of 
the strict minimizer of the regularized training error for prediction, 
but there may be some differences from practical settings:
In a case where a learnable part $\bg$ is implemented by a flexible model, 
it is practical to monitor the error with a holdout validation data during the learning 
process and select the model according to behaviors of that validation error
(e.g., early stopping \cite{prechelt2002early,goodfellow2016deep}),
rather than using the strict minimizer of the regularized training error for prediction.
This study presents an argument assuming strict 
minimization of the regularized training error, 
in the hope that it can serve as a suggestion for practical situations.

%========================================%
\section{Proposed Strict Unimodality-Promoting Regularized Learning Method}
\label{sec:Proposed}
%========================================%
\subsection{Proposal}
\label{sec:Proposal}
%==========%
We consider that a regularizer for UPRL should not only
take a smaller expectation value when the predicted CPD 
gets closer to be unimodal for the unimodality-promotion,
but also take a minimum expectation value if the predicted 
CPD is unimodal not to induce an unexpected bias.
Also, the development of a UPRL method with only the unimodality-promotion 
is significant in order to verify the effectiveness of the unimodality-promotion 
and to understand the reasons for the success of 
the previous UPRL methods through comparison with them.
On the basis of these motivations, we propose a novel UPRL method
(which we call strict UPRL method) with the regularizer
\begin{align}
\label{eq:OurUPRL}
	\Omega_\stri(\bg)
	=\frac{1}{n}\sum_{i=1}^n D_\rmH(\bQ(\bg(\bx_i)),\hat{\Delta}_{K-1}).
\end{align}
Here, we adopt a likelihood model $(P,\calG)$ that is not a UL model.
As a side proposal of this study, we propose to combine UPRL methods
with not only unrestricted likelihood models such as MLR model, 
but also AUL models.

%==========%
We can perform the proposed UPRL method with conventional techniques of 
the gradient descent method and quadratic programming:
In the first step, using a current value of $\bg$,
calculate $\bQ(\bg(\bx_i))$ for all $i\in[n]$.
In the second step, using a current value of $\bQ(\bg(\bx_i))$,
calculate $\bq_i\in\argmin_{\bv\in\hat{\Delta}_{K-1}}\|\bQ(\bg(\bx_i))-\bv\|$, 
which is $\bQ(\bg(\bx_i))$ if $\bQ(\bg(\bx_i))$ is unimodal, for all $i\in[n]$.
In the third step, using a gradient information 
$\frac{1}{n}\sum_{i=1}^n\nabla\phi(\bQ(\bg(\bx_i)), y_i)
+\lambda\frac{1}{n}\sum_{i=1}^n
\bbI(\bQ(\bg(\bx_i))\not\in\hat{\Delta}_{K-1})
\nabla\|\bQ(\bg(\bx_i))-\bq_i\|$,
update a value of $\bg$.
We iterate these three steps.

%==========%
The minimization problem in the second step, 
$\bq_i\in\argmin_{\bv\in\hat{\Delta}_{K-1}}\|\bQ(\bg(\bx_i))-\bv\|$, 
can be solved with a technique of quadratic programming:
the minimization problem $\min_{\bv\in\hat{\Delta}_{K-1}}\|\bQ(\bg(\bx_i))-\bv\|$ 
under the assumption that a mode of $\bq_i$ is $M\in[K]$ (subproblem-$M$) reduces to
\begin{align}
	\begin{split}
	&\min_{\bv\in\bbR^K}\|\bQ(\bg(\bx_i))-\bv\|,
\\
	&\text{~\;s.t.~~}
	0\le v_1\le\cdots\le v_M,
	v_M\ge\cdots\ge v_K\ge0,
\\
	&\hphantom{	\text{~~s.t.~~}}
	\text{and }
	\sum_{k=1}^Kv_k=1,
	\end{split}
\end{align}
or equivalently,
\begin{align}
	&\min_{\bv\in\bbR^K}
	\frac{1}{2}\bv^\top \rmI_K \bv
	-\bQ(\bg(\bx_i))^\top\bv,
\nonumber\\
	&\text{~~s.t.~}
	\left({\renewcommand{\arraycolsep}{1pt}\begin{array}{ccccccc}
	-1&&&&&&{\protect\Huge O}\\
	1&-1&&&&&\\
	&\ddots&\ddots&&&&\\
	&&1&-1&&&\\
	&&&-1&1&&\\
	&&&&\ddots&\ddots&\\
	&&&&&-1&1\\
	{\protect\Huge O}&&&&&&-1
	\end{array}}\right)\bv
	=\left({\renewcommand{\arraycolsep}{1pt}\begin{array}{c}
	-v_1\\
	v_1-v_2\\
	\vdots\\
	v_{M-1}-v_M\\
	v_{M+1}-v_M\\
	\vdots\\
	v_K-v_{K-1}\\
	-v_K
	\end{array}}\right)
	\le\bm{0}_{K+1}
\nonumber\\
	&\hphantom{	\text{~~s.t.~}}
	\text{and }
	\bm{1}_K^\top\bv=1,
\end{align}
where $\rmI_{K}$ is the $(K\times K)$-dimensional identity matrix,
$\bm{0}_{K+1}$ is the $(K+1)$-dimensional all-zero vector,
and $\bm{1}_{K}$ is the $K$-dimensional all-one vector.
Therefore, the solution of the original problem,
$\bq_i\in\argmin_{\bv\in\hat{\Delta}_{K-1}}\|\bQ(\bg(\bx_i))-\bv\|$,
can be characterized as a minimizer of sets of 
the solution of subproblem-$M$ for all $M\in[K]$.

%========================================%
\subsection{Analysis}
\label{sec:StriAnaly}
%==========%
The expectation value of the summands of the regularizer \eqref{eq:OurUPRL} 
conditioned on $\bX=\bx$ is $\Pr(\bX=\bx)\omega_\stri(\bP(\bx),\bQ(\bg(\bx)))$ 
with the expected UPRL loss 
\begin{align}
	\omega_\stri(\bp,\bq)
	\coloneq\sum_{y=1}^K p_y 
	D_\rmH(\bq,\hat{\Delta}_{K-1})
	=D_\rmH(\bq,\hat{\Delta}_{K-1}).
\end{align}

%==========%
As in Section~\ref{sec:PrevAnaly}, we also demonstrated 
behaviors of the expected UPRL loss of the proposed UPRL method; 
see Figure~\ref{fig:UDScale} \hyl{UDdu}{du} and \hyl{UDds}{ds}.
The correlation coefficient between 
the expected UPRL loss $(\omega_\stri(\bp,\bq_i))_{i\in[10^3]}$ and 
the UD $(D_\rmH(\bq_i,\hat{\Delta}_4))_{i\in[10^3]}$ is of course 1 
(although it is important that this value is high but it is not 
essential for the unimodality-promotion to be exactly 1), 
and hence the proposed UPRL method would achieve the unimodality-promotion.
More importantly, the expected UPRL loss $\omega_\stri(\bp,\bq)$ 
gets minimum at $\bq=\bp$ if $\bp$ is unimodal 
(i.e., satisfies the requirement of the UD \eqref{eq:UDrequire}), 
which indicates that the regularizer \eqref{eq:OurUPRL} 
would induce no unexpected bias at $\bX=\bx$ 
where the underlying CPD $\bP(\bx)$ is unimodal.
Finally, although the correlation coefficient between the 
expected UPRL loss $(\omega_\stri(\bp,\bq_i))_{i\in[10^3]}$ 
and the scale $(1-\max_{j\in[5]}q_{i,j})_{i\in[10^3]}$ is not 0, 
but its absolute value is much smaller than those of the previous UPRL methods.
Therefore, we expect that the proposed strict UPRL method will 
reduce a scale-related bias compared with previous UPRL methods.

%========================================%
\section{Numerical Experiments}
\label{sec:Experiments}
%========================================%
\textbf{Purposes:}
%==========%
We have theoretically shown that, 
although the previous UPRL method \eqref{eq:alb21} 
may not only achieve the unimodality-promotion, 
but also induce a smoothness-promotion bias.
On the other hand, we contend that the strict UPRL method 
\eqref{eq:OurUPRL} achieves the unimodality-promotion 
and reduces such an unexpected scale-related bias.
On the ground of the bias-variance trade-off, 
we expect that the dominance of the previous UPRL, 
strict UPRL, and non-regularized learning methods for 
small, intermediate, and large training data size, respectively.
Furthermore, we expect that the previous UPRL method 
and strict UPRL method will be more compatible with 
larger-scale data and smaller-scale data, respectively.
With the purpose to compare their prediction 
performances and to verify these expected behaviors,
we performed numerical experiments with real-world ordinal data.

%=======================================%
\textbf{Settings (Data):}
%==========%
For the experiments, we used 21 real-world ordinal datasets 
with the total data size $n_\tot\ge1000$ among those used in 
experiments of the previous OR survey study \cite{gutierrez2015ordinal}.
It has been verified in \cite{yamasaki2022unimodal, yamasaki2025approximately}
that these real-world ordinal data would be high-UR and low-UD.
We experimented with 6 settings of the training data size, 
$n=n_\tra=25,50,100,200,400,800$, 
in order to see dependence of the training data size 
on behaviors of each method.

%=======================================%
\textbf{Settings (Methods):}
%==========%
We tried the non-regularized learning method with NLL as the objective function, 
the previous UPRL method \eqref{eq:alb21} with 
$\delta=0$ and $\lambda=10^{-8},10^{-7.5},\ldots,10^8$, 
and the strict UPRL method \eqref{eq:OurUPRL} with 
$\lambda=10^{-8},10^{-7.5},\ldots,10^8$.
We tried MLR model and AUL models with 
the mixture rate $r=0.05,0.1,\ldots,0.95$, 
$\rho(u)=e^u$, and $\tau(u)=u^2$ as a likelihood model, 
and implemented all the learner models with 
a 4-layer fully-connected neural network model 
in which every hidden layer has 300 nodes activated 
with the ReLU function in addition to bias nodes.
These methods are denoted as 
Nonr-MLR, Prev-MLR, Stri-MLR, Nonr-AUL, Prev-AUL, and Stri-AUL.
We also experimented other ordinal regression methods, 
including ordinal logistic regression \cite{mccullagh1980regression} 
and UL model \cite{yamasaki2022unimodal}, 
but none were found to be particularly superior
to the six above-listed methods in terms of the overall performance;
see the supplement for experimental results for these methods.

%=======================================%
\textbf{Settings (Training and Evaluation):}
%==========%
We trained a model by full-batch Adam optimization with 
the learning rate $10^{-(3+2t/1000)}$ at $t$-th epoch during 1000 epochs.
At the end of each training epoch, we evaluated the validation errors, 
the NLL for a conditional probability estimation task and 
the MZE, MAE, and MSE with a likelihood-based classifier for OR tasks, 
with holdout data of the size $n_\val=100$.
We selected the epoch (stopping time of the learning process),
mixture rate $r$ for AUL-based methods, and
regularization parameter $\lambda$ according to the validation error, 
and calculated the test error for the selected model 
with remaining test data of the size $(n_\tot-n_\tra-n_\val)$.
We repeated the above procedure 100 trials with a random settings 
of the data split and initial parameters to obtain 100 test errors.

%==========%
\begin{figure}[!t]
\centering%
\renewcommand{\arraystretch}{0.25}%
\renewcommand{\tabcolsep}{0pt}%
\begin{tabular}{C{2.9cm}C{2.9cm}C{2.9cm}}%
\includegraphics[height=1.6cm]{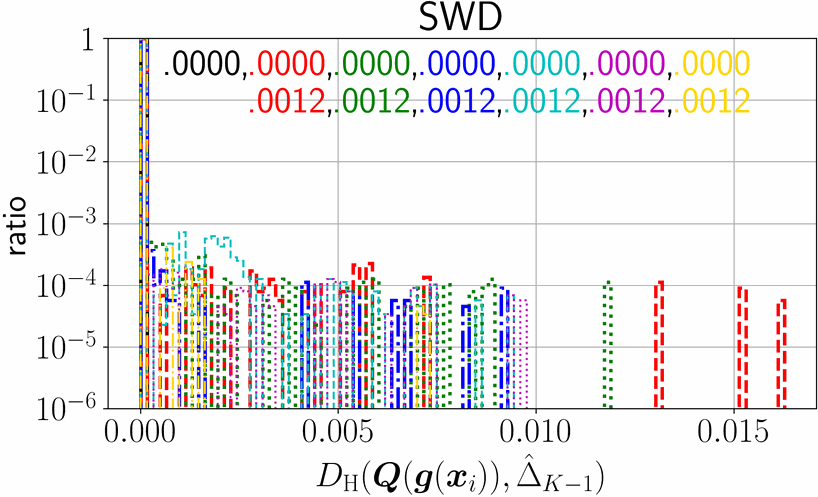}&
\includegraphics[height=1.6cm]{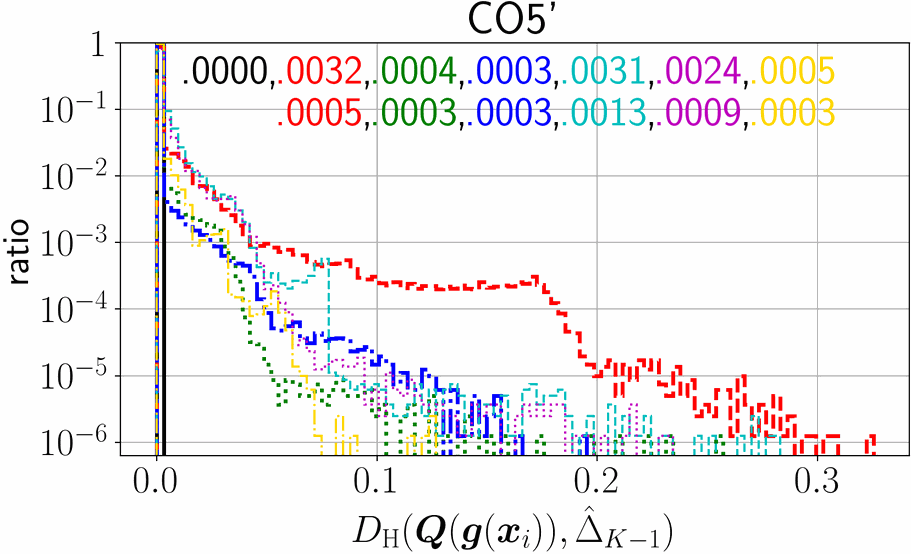}&
\includegraphics[height=1.6cm]{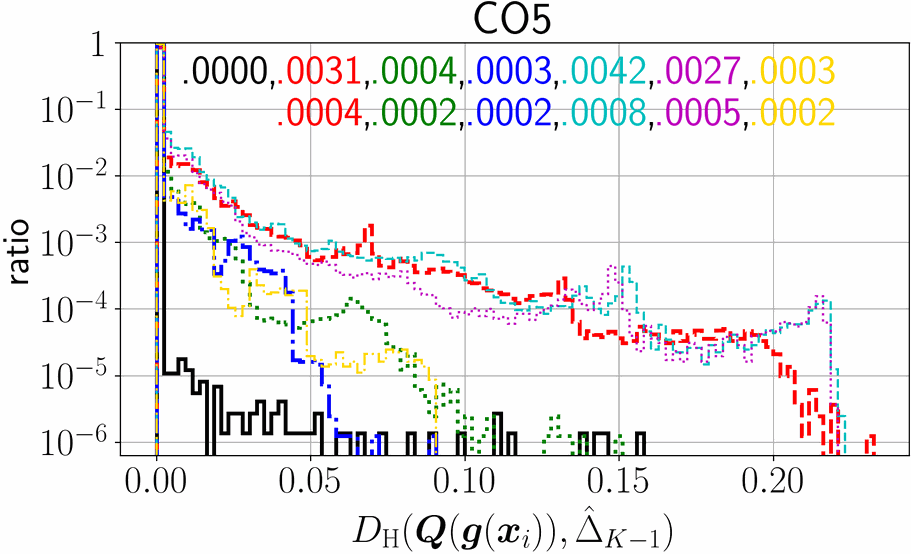}\\
\includegraphics[height=1.6cm]{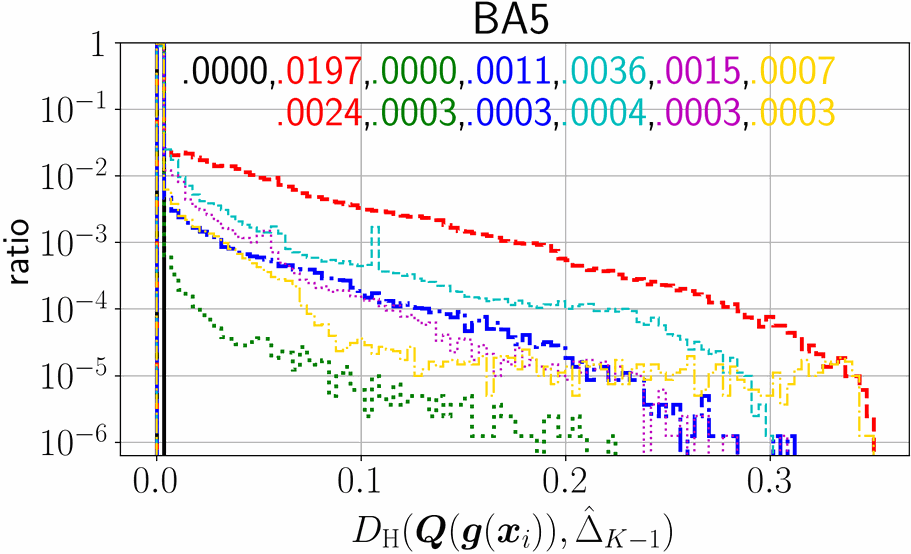}&
\includegraphics[height=1.6cm]{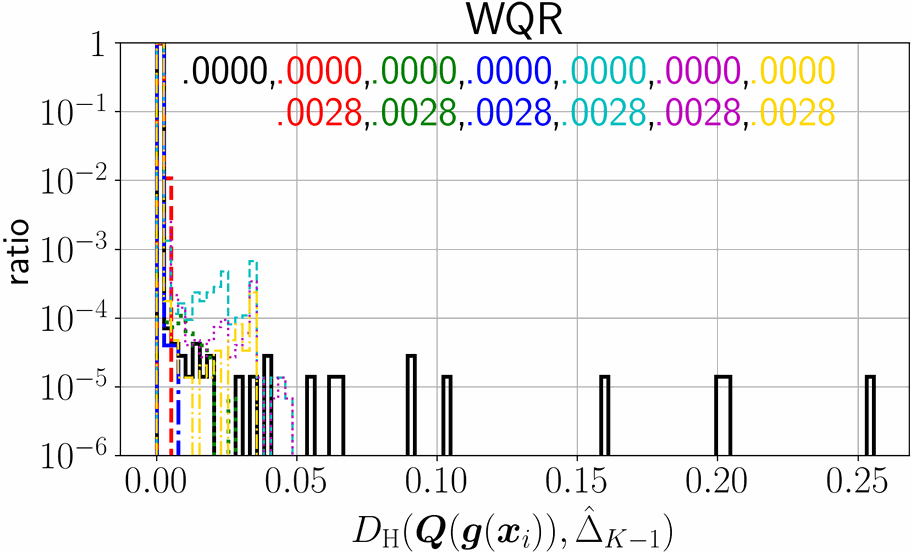}&
\includegraphics[height=1.6cm]{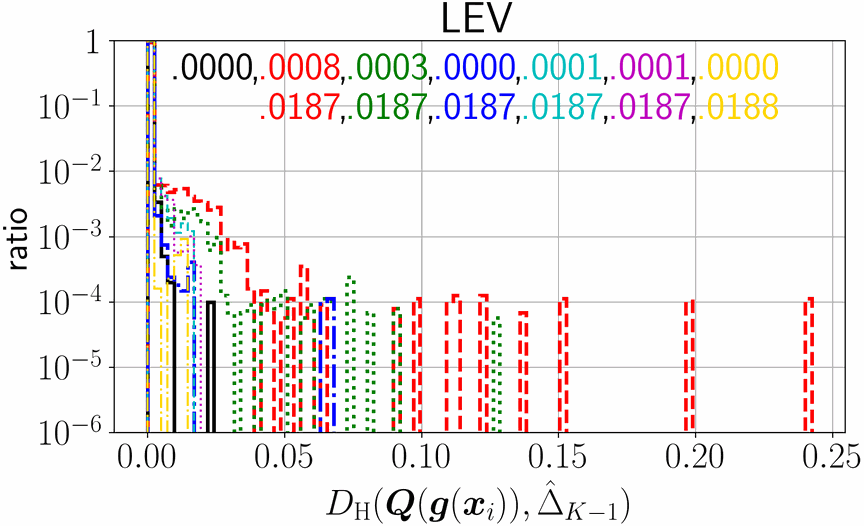}\\
\includegraphics[height=1.6cm]{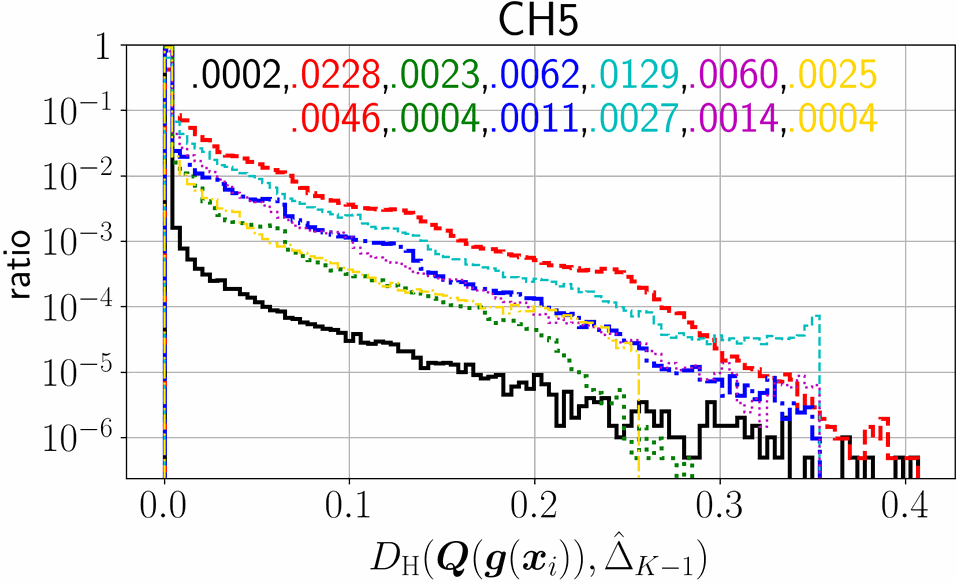}&
\includegraphics[height=1.6cm]{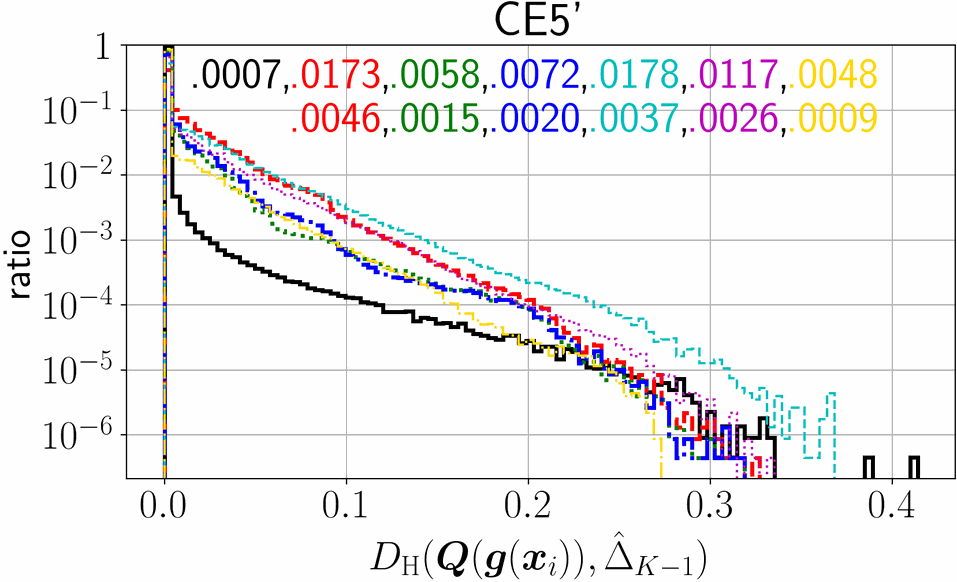}&
\includegraphics[height=1.6cm]{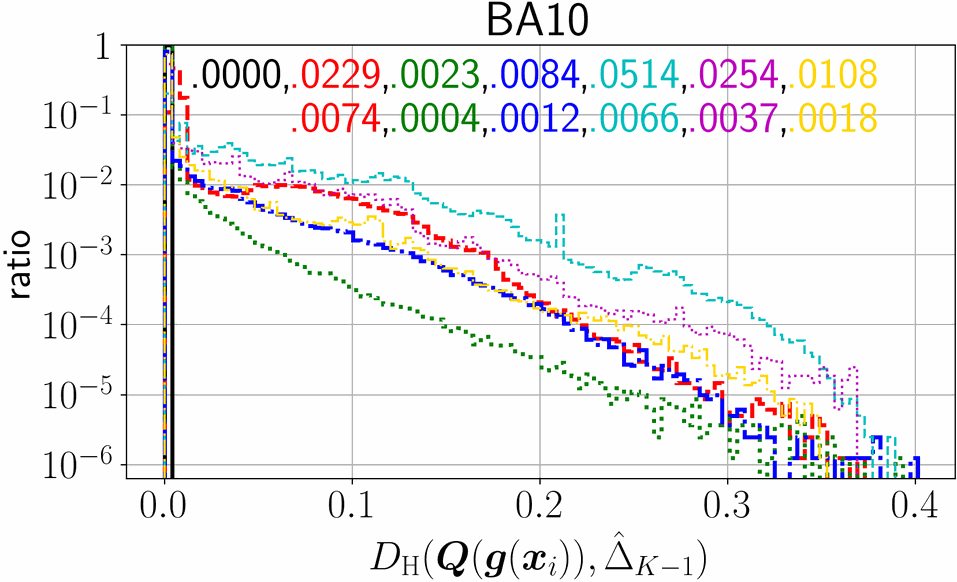}\\
\includegraphics[height=1.6cm]{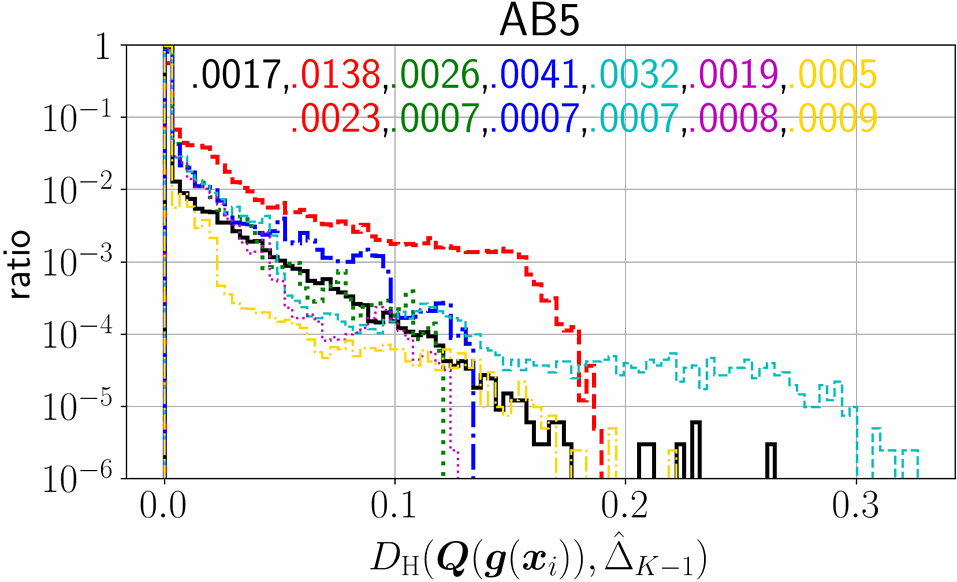}&
\includegraphics[height=1.6cm]{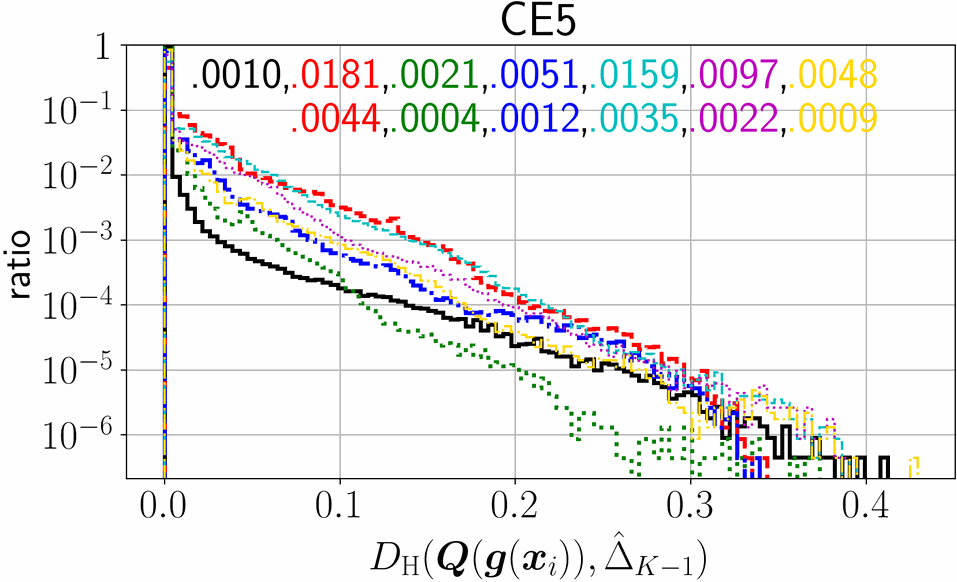}&
\includegraphics[height=1.6cm]{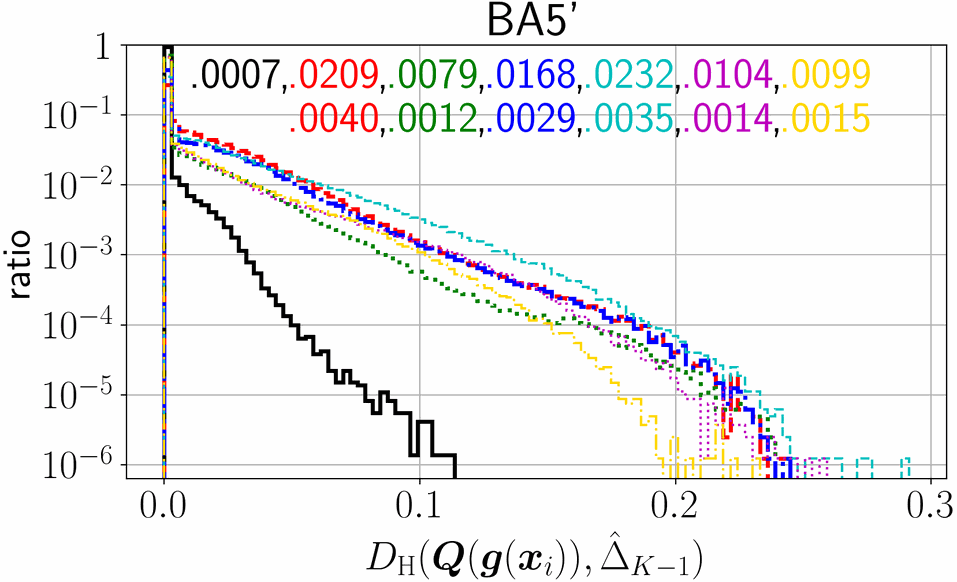}\\
\includegraphics[height=1.6cm]{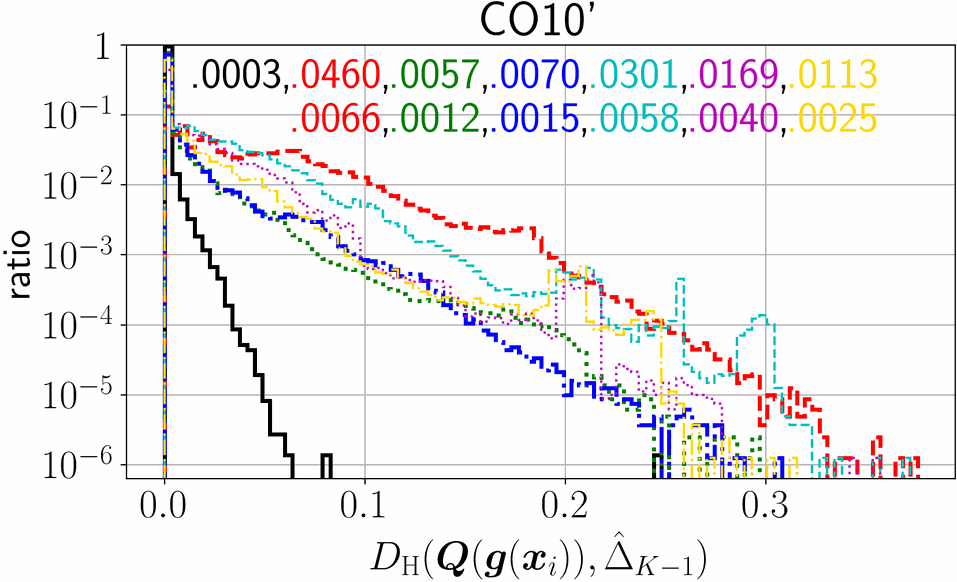}&
\includegraphics[height=1.6cm]{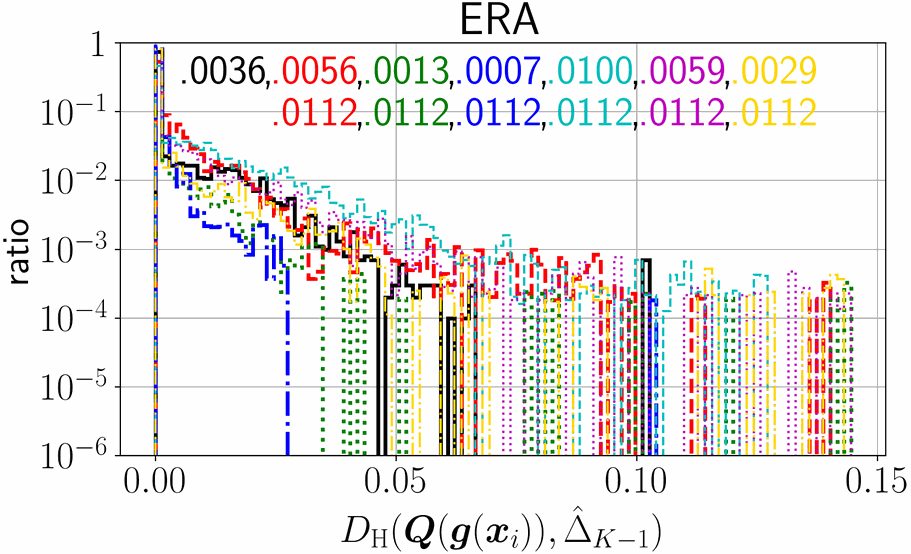}&
\includegraphics[height=1.6cm]{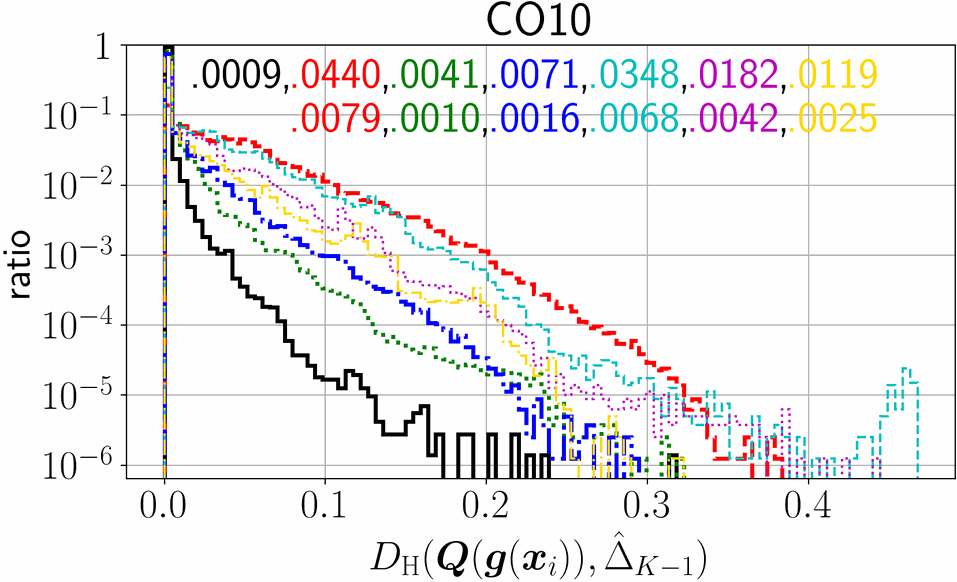}\\
\includegraphics[height=1.6cm]{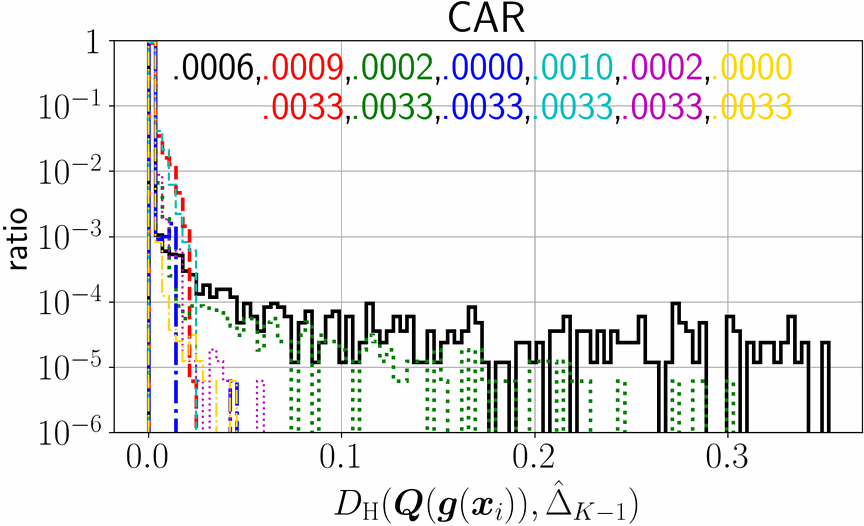}&
\includegraphics[height=1.6cm]{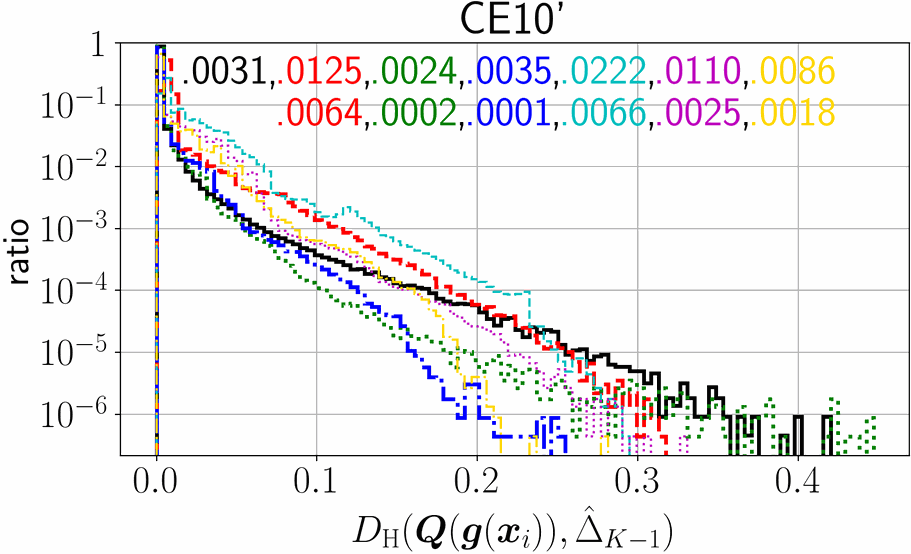}&
\includegraphics[height=1.6cm]{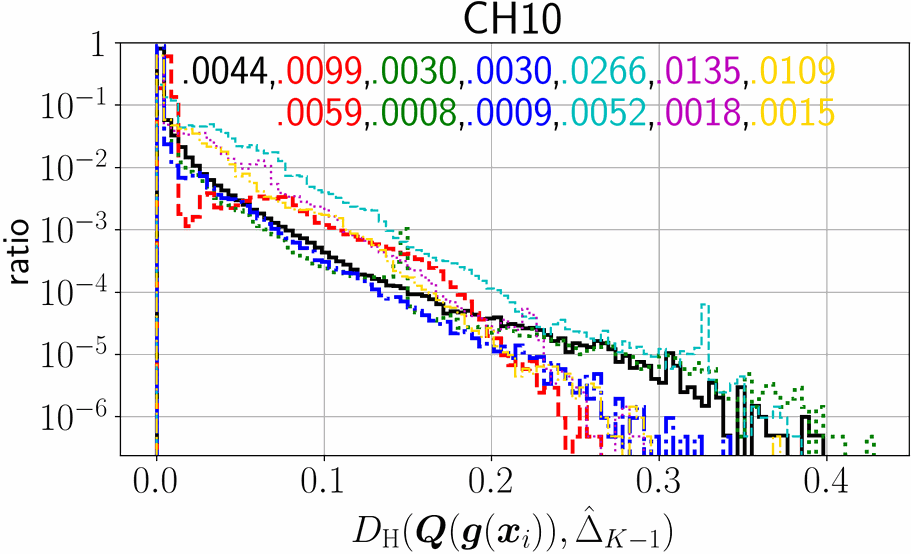}\\
\includegraphics[height=1.6cm]{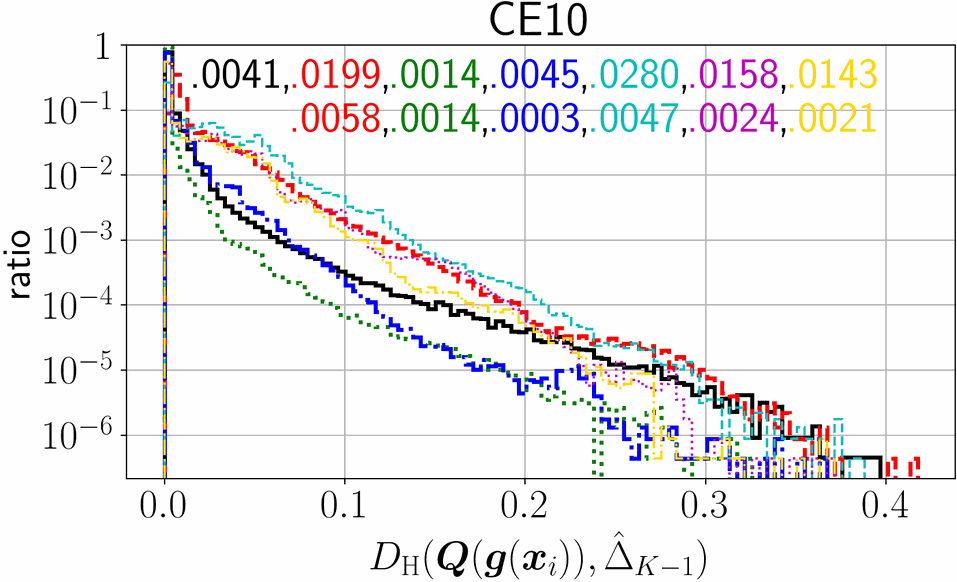}&
\includegraphics[height=1.6cm]{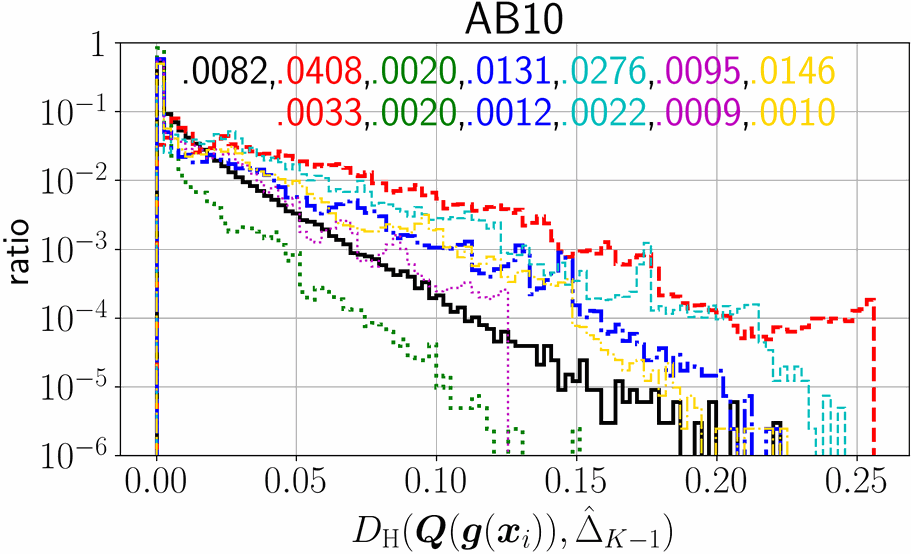}&
\includegraphics[height=1.6cm]{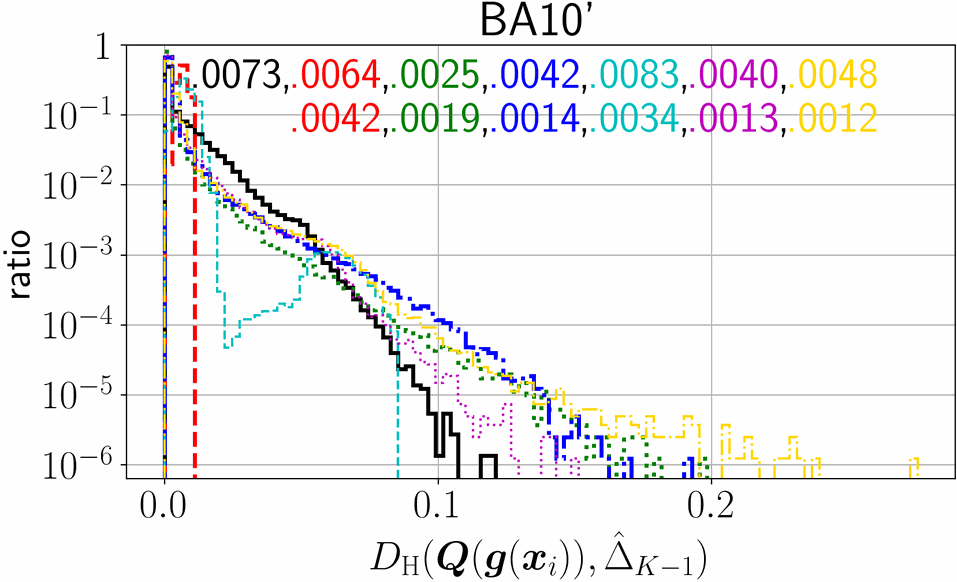}
\end{tabular}
\caption{%
Log-scaled histogram of aggregation of 100-trial 
estimates of the UD $D_\rmH(\bP(\bx),\hat{\Delta}_{K-1})$
by Nonr-MLR with $n_\tra=800$ (thick solid black), and Nonr-MLR (thick dashed red), 
Prev-MLR (thick dotted green), Stri-MLR (thick dash-dotted blue), Nonr-AUL (thin dashed cyan), 
Prev-AUL (thin dotted magenta), and Stri-AUL (thin dash-dotted yellow) with $n_\tra=25$.
Mean (upper) and $L_1$ distance from the black histogram (lower) 
are shown at the top of figure in the order and with their colors.}
\label{fig:Exp2-Unimodality}
\end{figure}

%=======================================%
\textbf{Results (Unimodality-promotion):}
%==========%
Figure~\ref{fig:Exp2-Unimodality} shows the comparison 
regarding the unimodality of a predicted CPD.

%==========%
For all the datasets,
the previous and strict UPRL methods reduced the UD 
from the non-regularized learning methods.
This result indicates that the previous and strict UPRL methods 
certainly have the unimodality-promotion.
For many real-world ordinal data (except for LEV and AB5) with a unimodal CPD,
the previous and strict UPRL methods gave a CPD prediction 
closer to the underlying CPD (estimated with $n_\tra=800$) 
than the non-regularized learning methods
in terms of the $L_1$ distance of the UD distribution.
This result suggests the contribution of the unimodality-promotion 
by UPRL to the improvement of the prediction performance.

%==========%
\begin{figure}[!t]
\centering%
\renewcommand{\arraystretch}{0.25}%
\renewcommand{\tabcolsep}{0pt}%
\begin{tabular}{C{2.9cm}C{2.9cm}C{2.9cm}}%
\includegraphics[height=1.6cm]{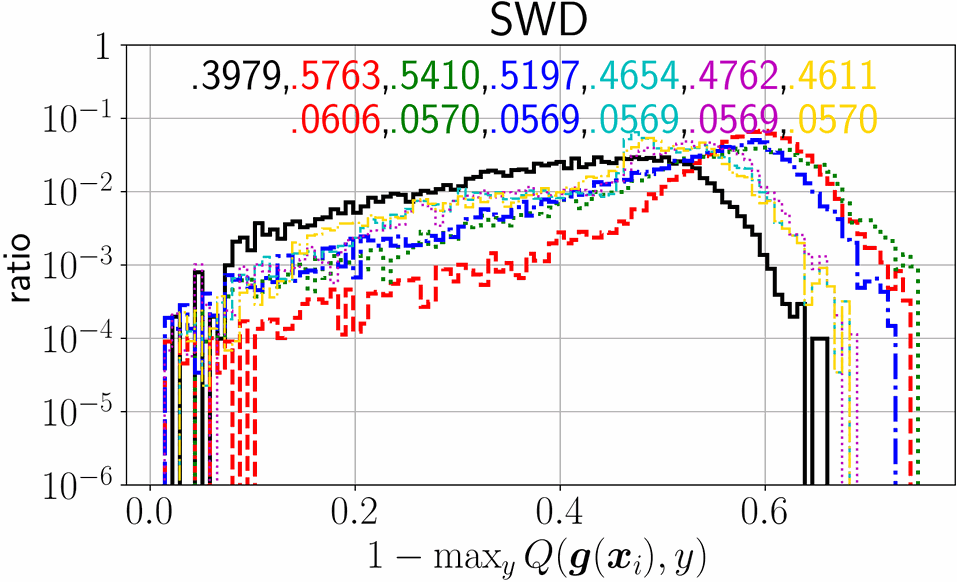}&
\includegraphics[height=1.6cm]{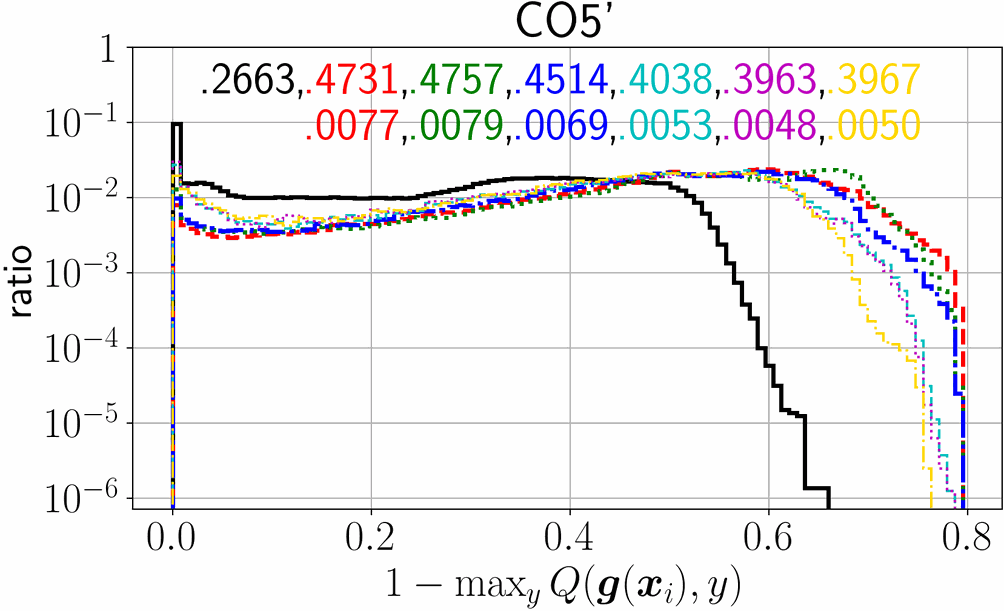}&
\includegraphics[height=1.6cm]{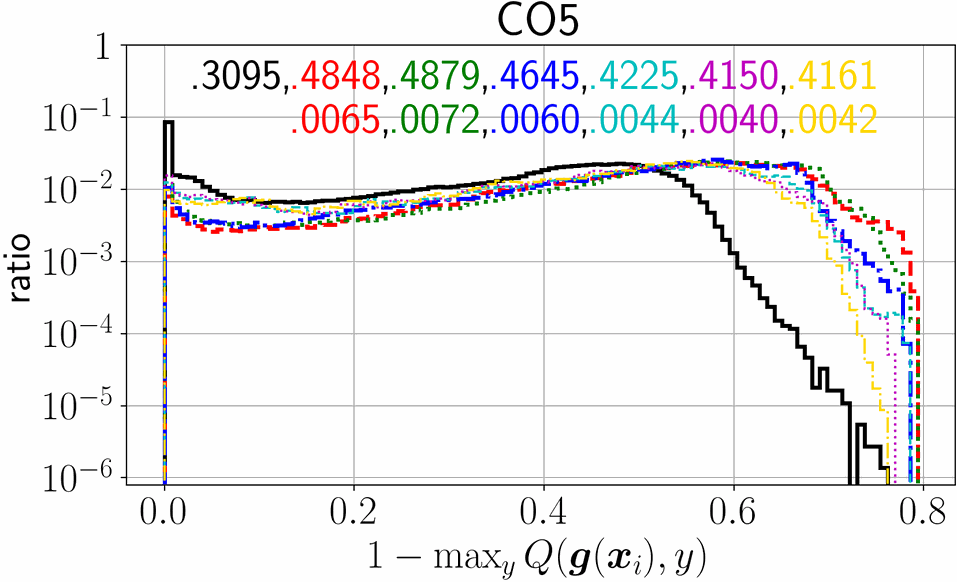}\\
\includegraphics[height=1.6cm]{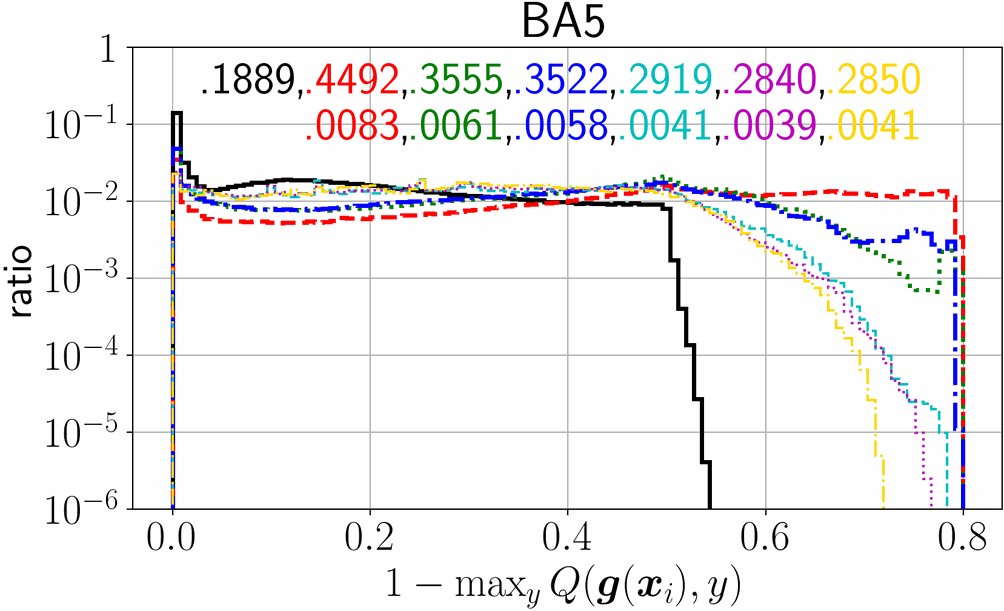}&
\includegraphics[height=1.6cm]{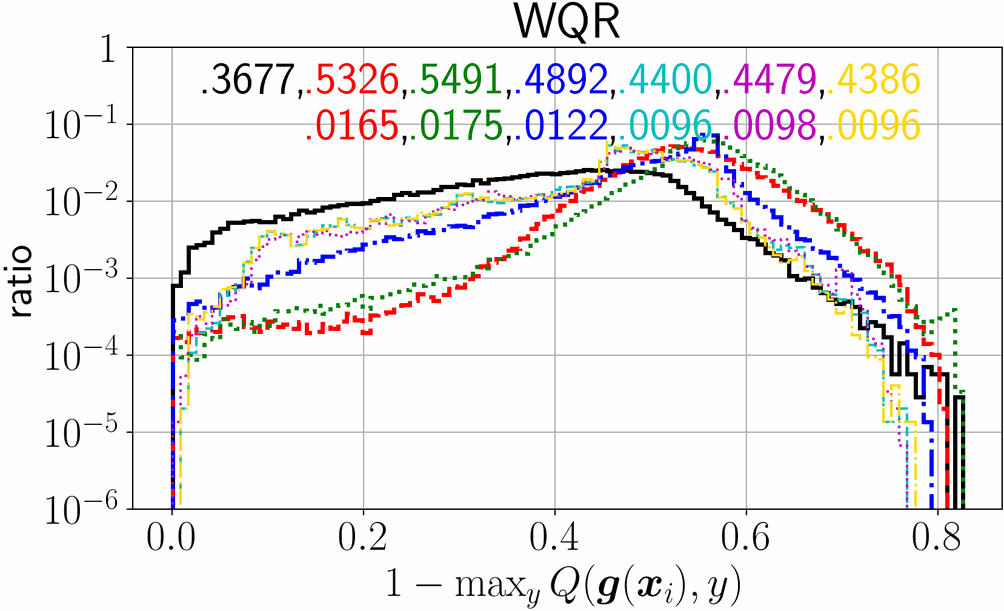}&
\includegraphics[height=1.6cm]{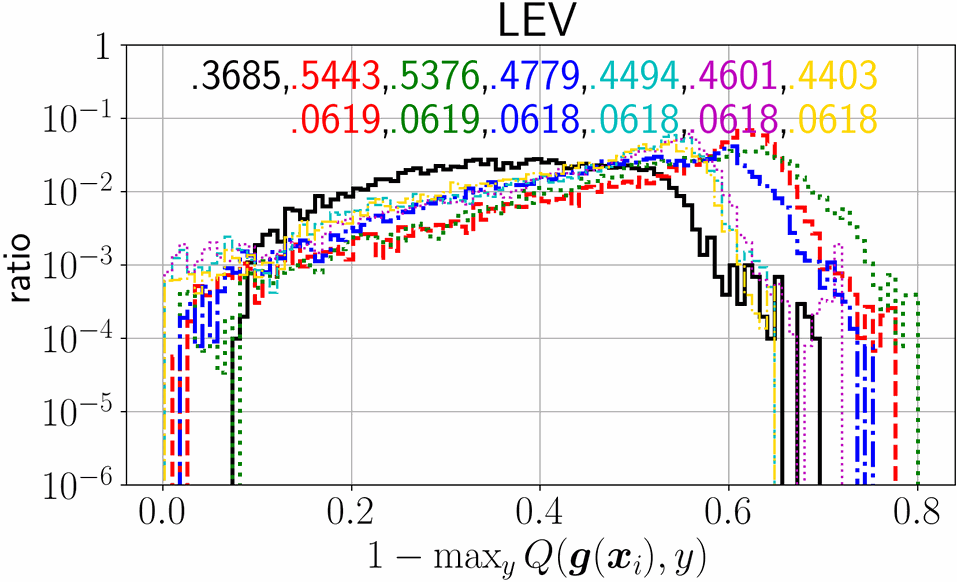}\\
\includegraphics[height=1.6cm]{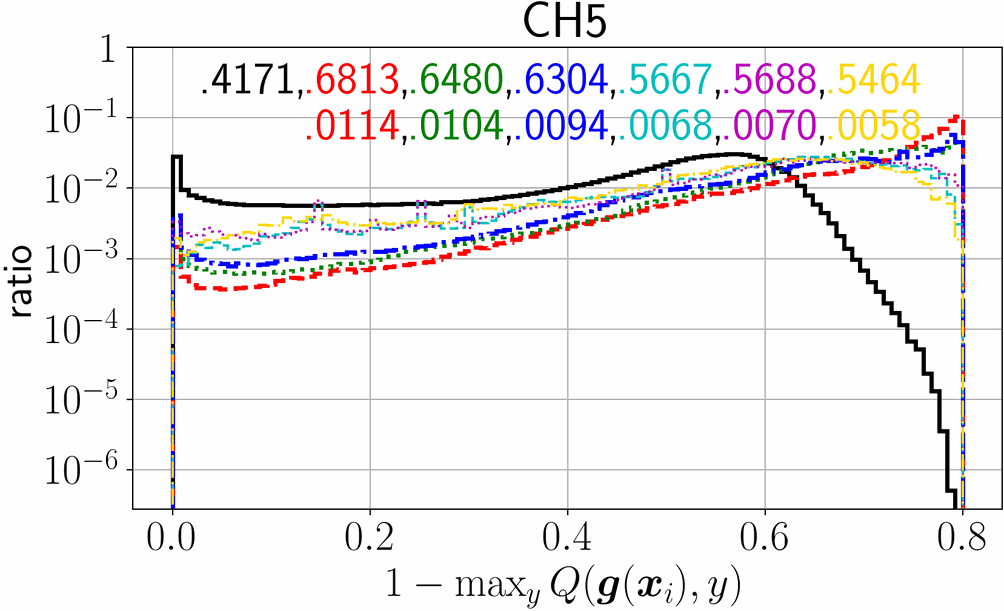}&
\includegraphics[height=1.6cm]{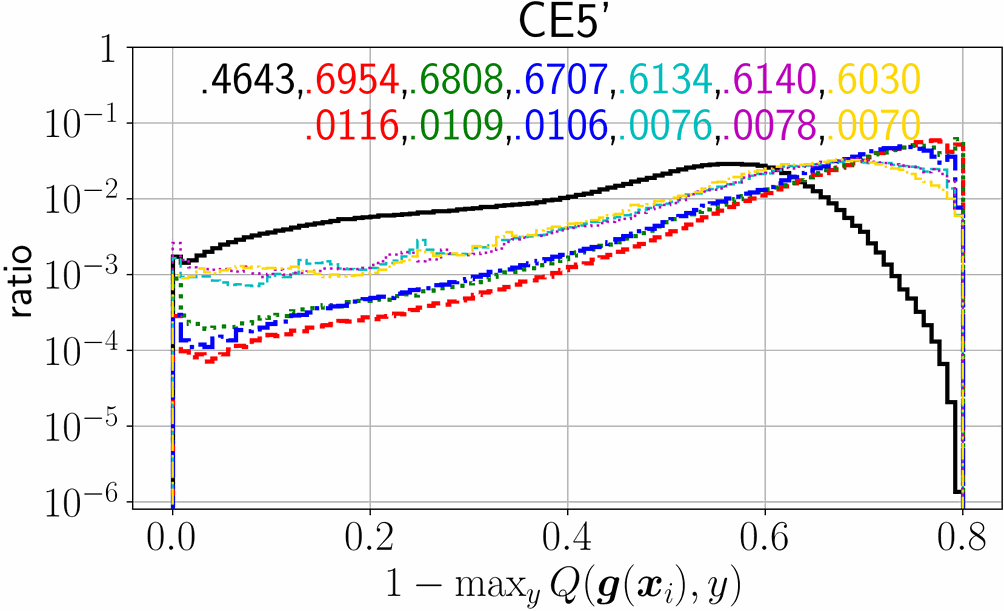}&
\includegraphics[height=1.6cm]{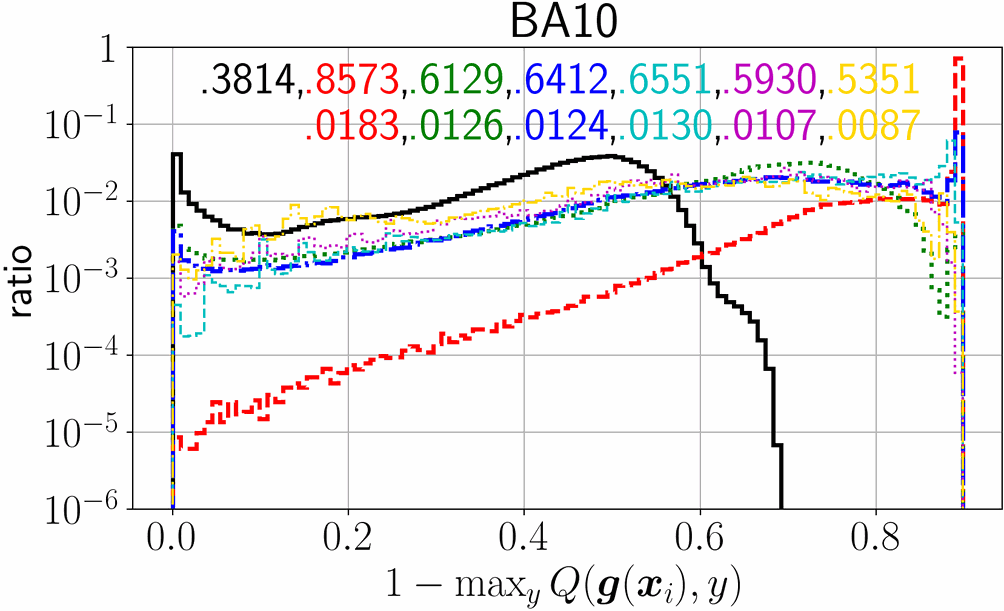}\\
\includegraphics[height=1.6cm]{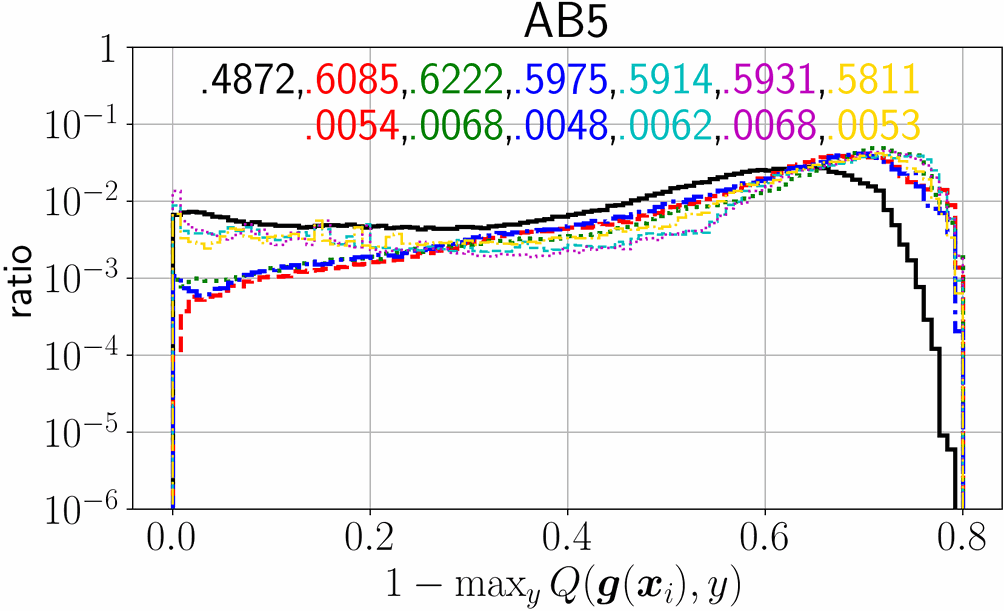}&
\includegraphics[height=1.6cm]{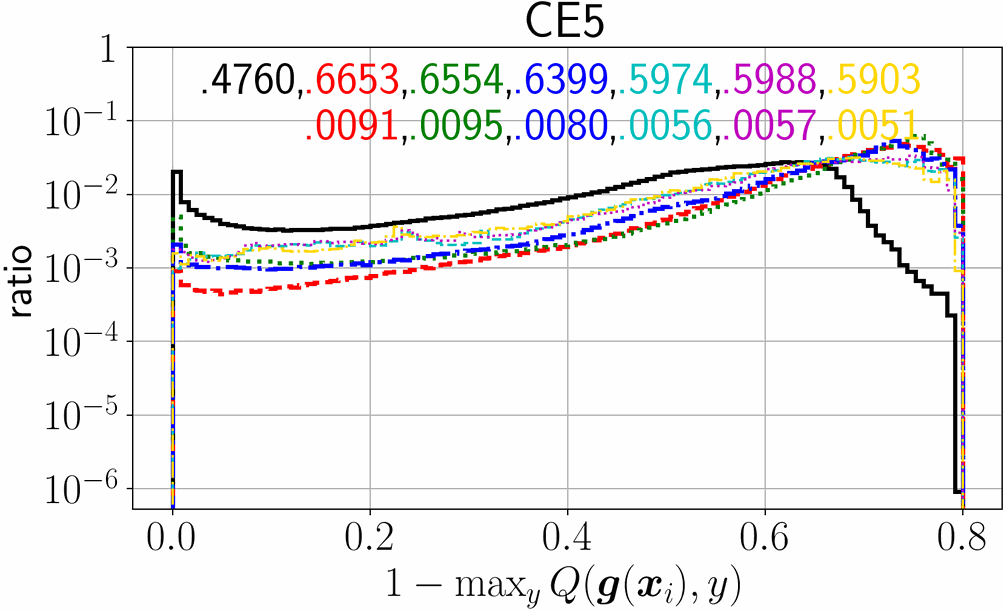}&
\includegraphics[height=1.6cm]{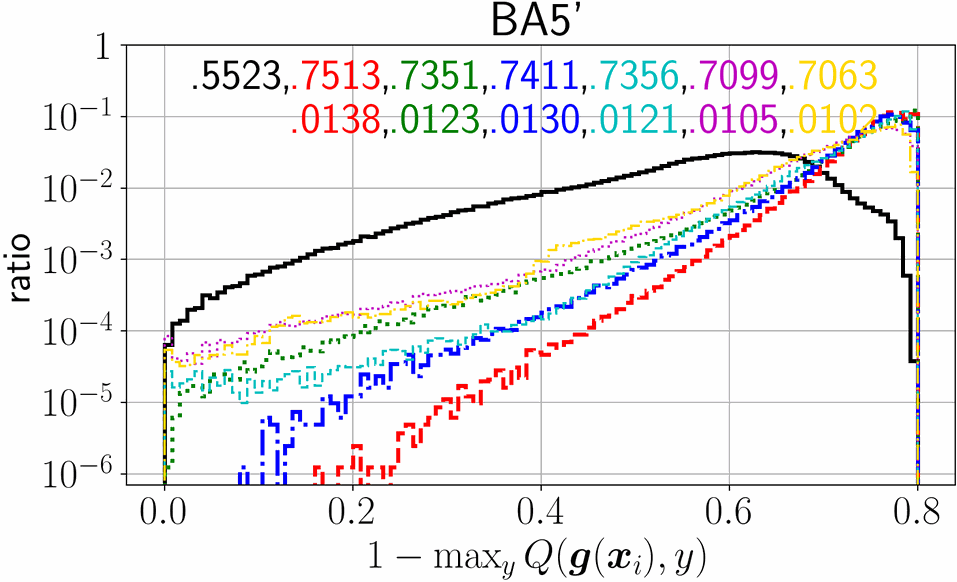}\\
\includegraphics[height=1.6cm]{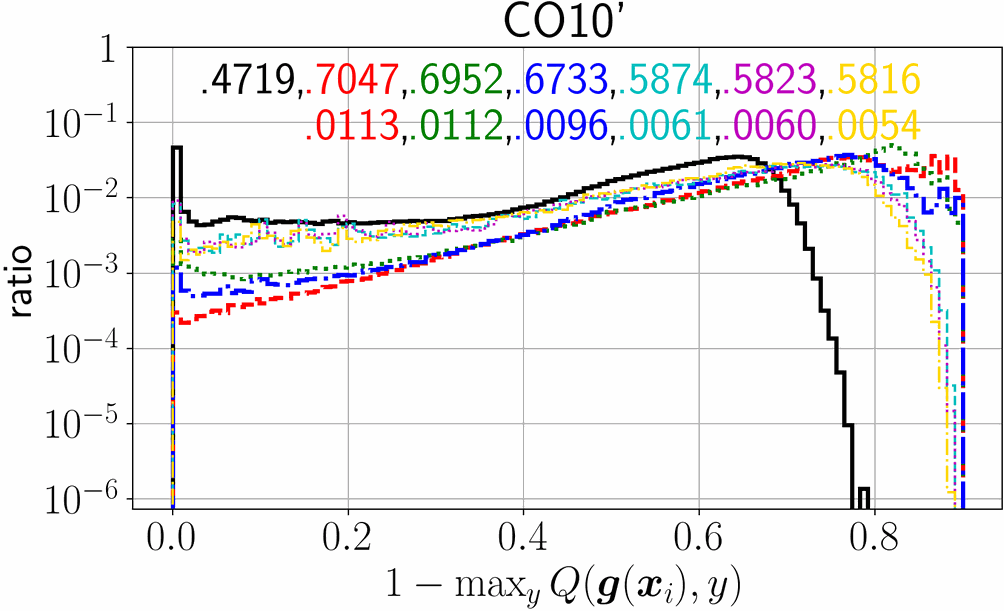}&
\includegraphics[height=1.6cm]{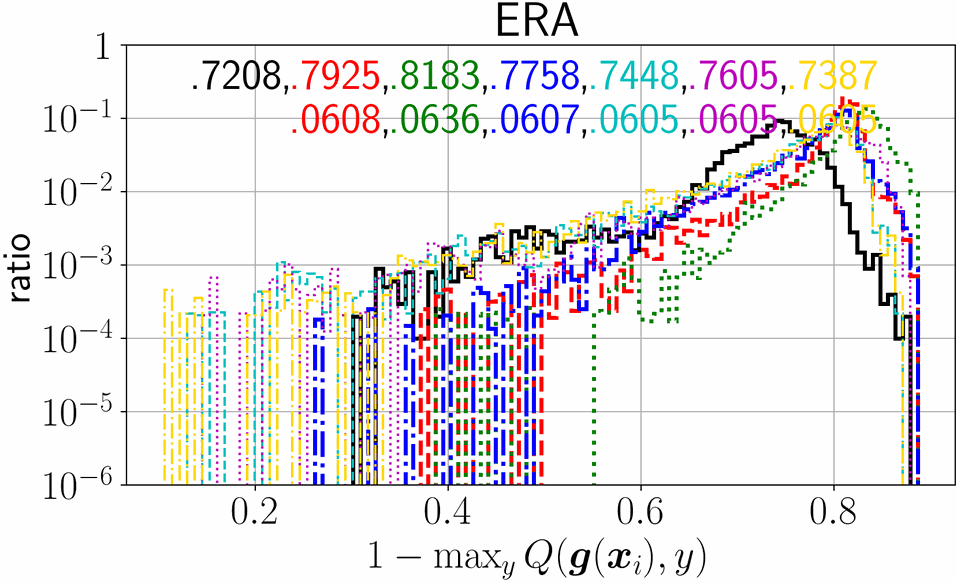}&
\includegraphics[height=1.6cm]{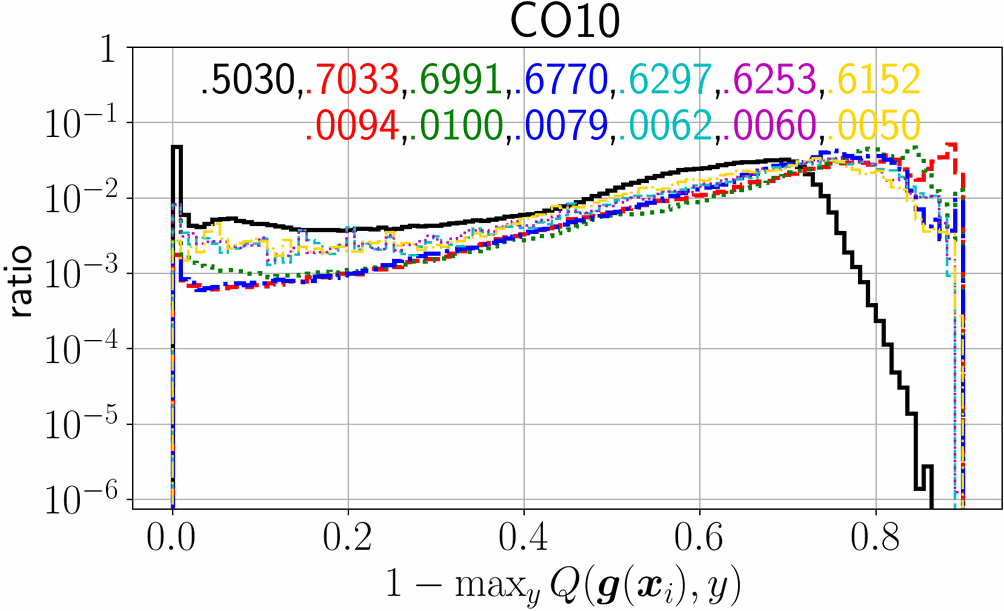}\\
\includegraphics[height=1.6cm]{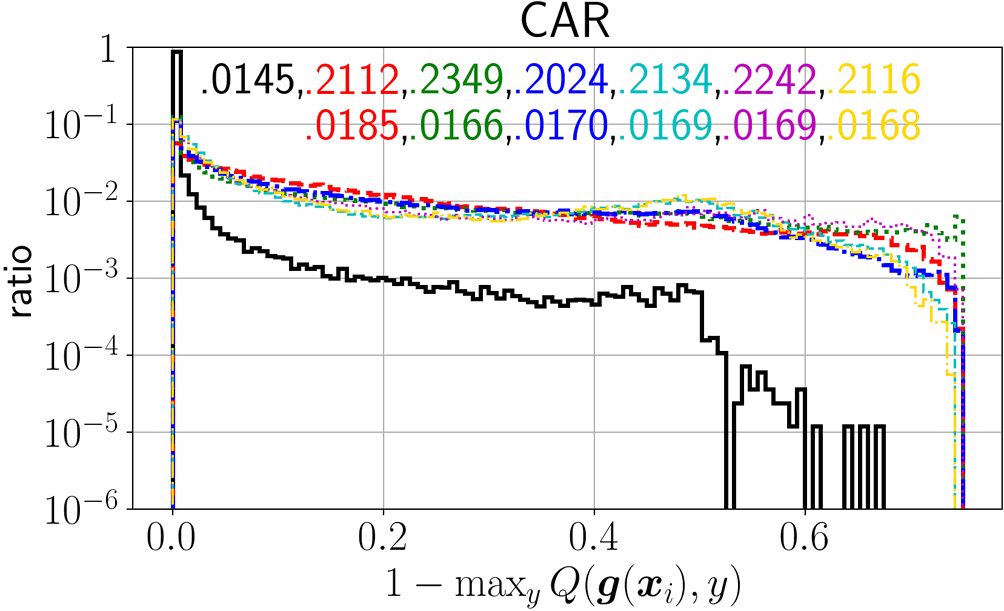}&
\includegraphics[height=1.6cm]{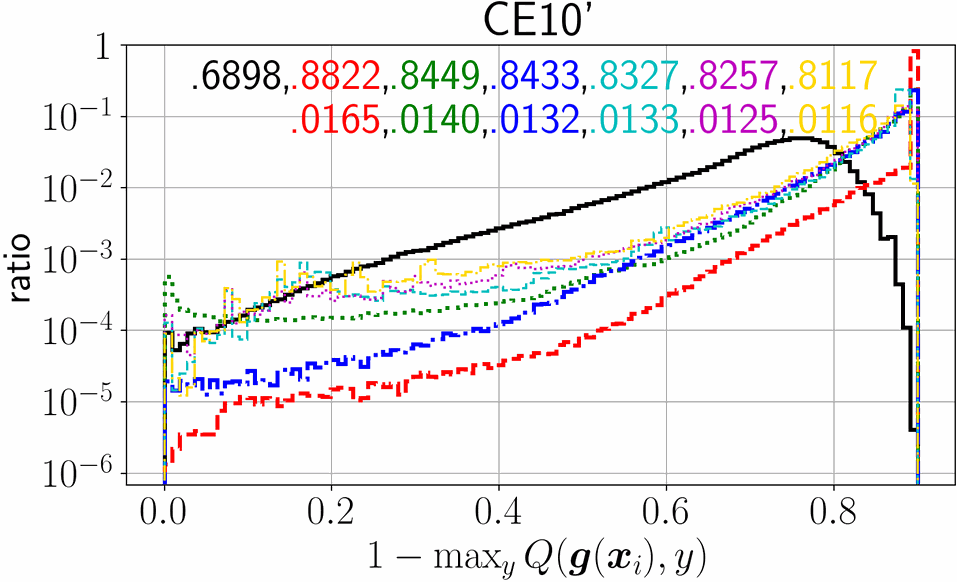}&
\includegraphics[height=1.6cm]{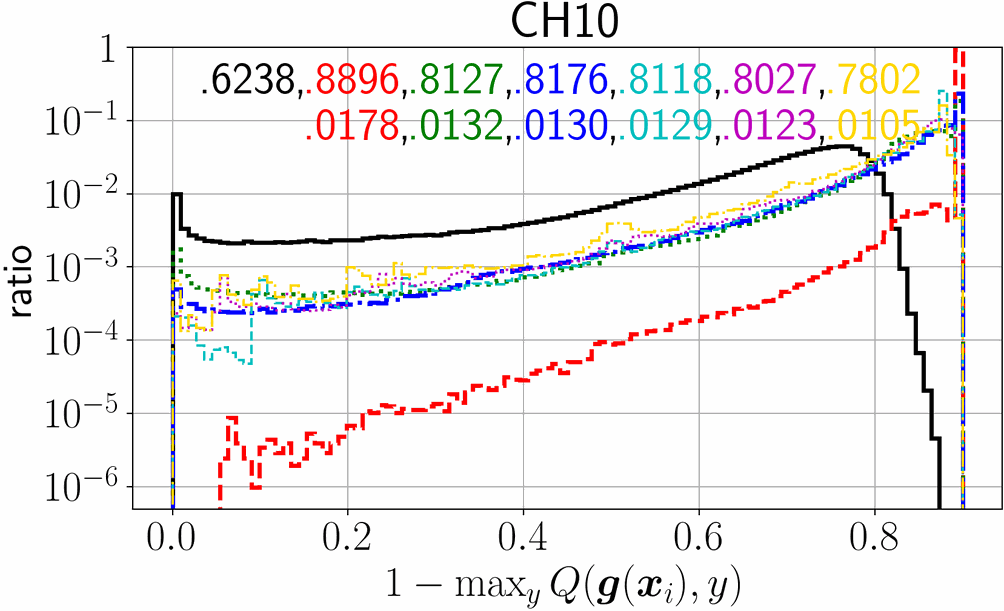}\\
\includegraphics[height=1.6cm]{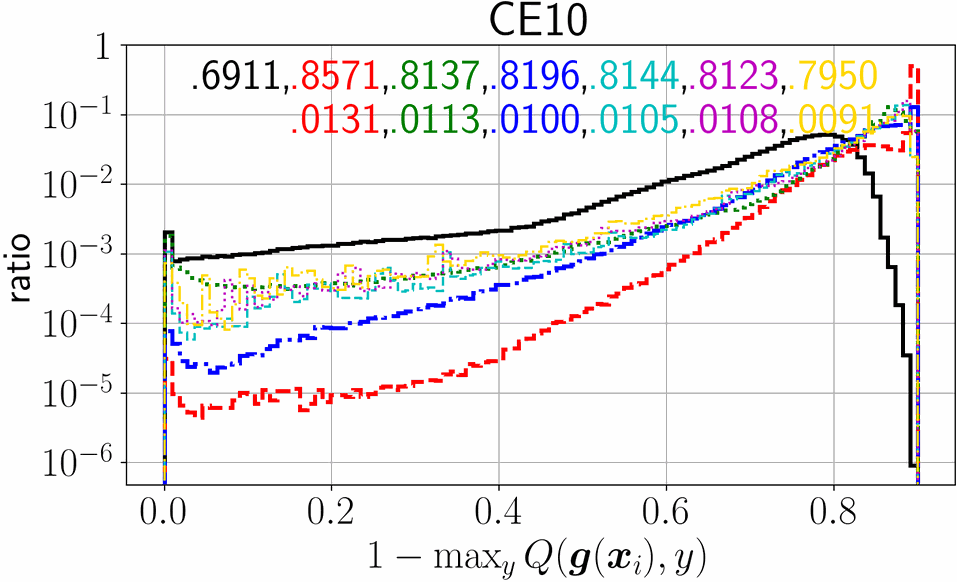}&
\includegraphics[height=1.6cm]{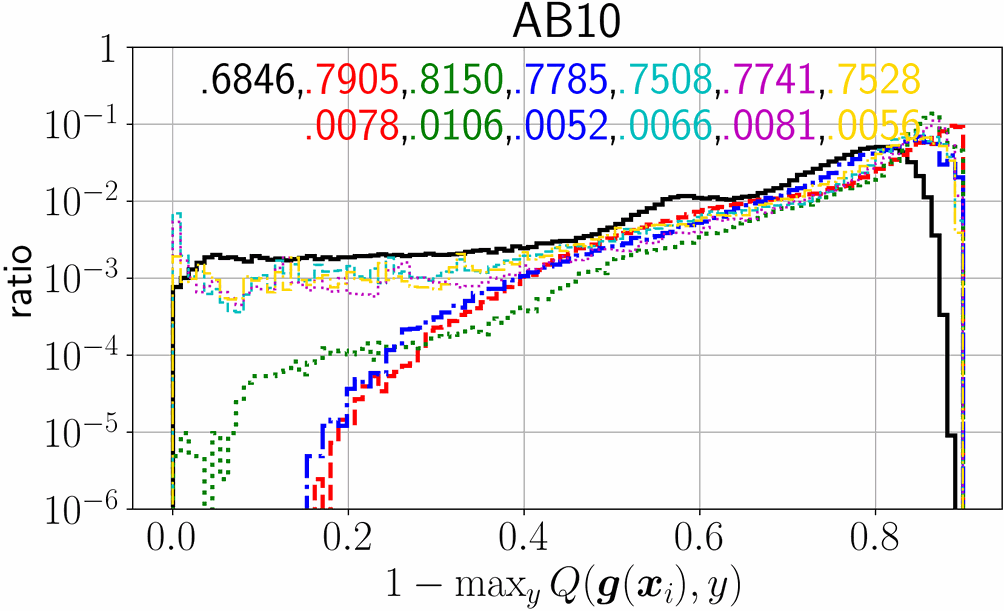}&
\includegraphics[height=1.6cm]{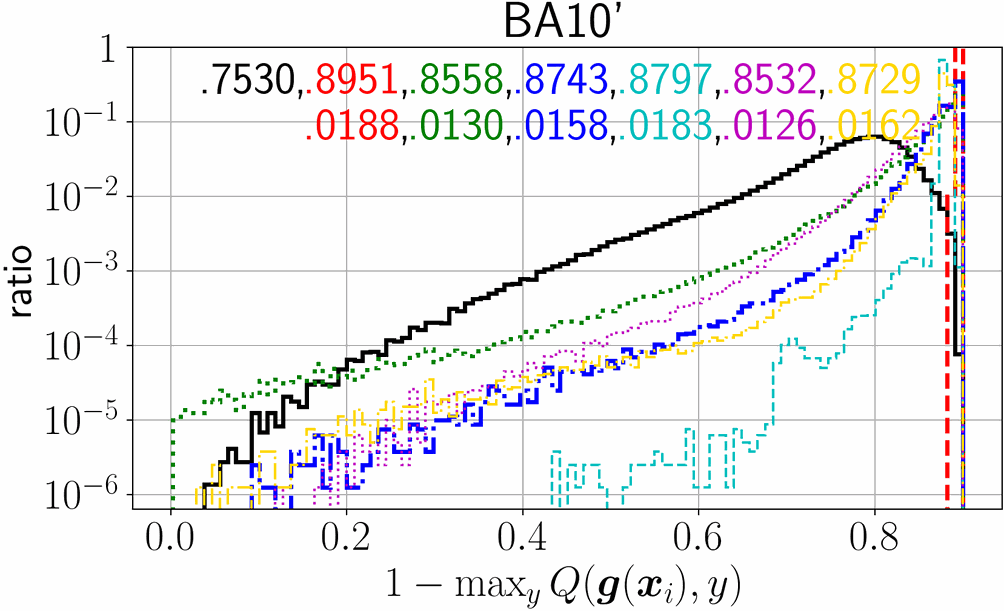}
\end{tabular}
\caption{%
Histogram of aggregation of 100-trial 
estimates of the scale $1-\max_y\Pr(Y=y|\bX=\bx)$
by Nonr-MLR with $n_\tra=800$ (thick solid black), and Nonr-MLR (thick dashed red), 
Prev-MLR (thick dotted green), Stri-MLR (thick dash-dotted blue), Nonr-AUL (thin dashed cyan), 
Prev-AUL (thin dotted magenta), and Stri-AUL (thin dash-dotted yellow) with $n_\tra=25$.
Mean (upper) and $L_1$ distance from the black histogram (lower) 
are shown at the top of figure in the order and with their colors.}
\label{fig:Exp2-Scale}
\end{figure}

%=======================================%
\textbf{Results (Smoothness-promotion):}
%==========%
Figure~\ref{fig:Exp2-Scale} shows the comparison 
regarding the scale of a predicted CPD.

%==========%
For most datasets except for 
BA10, BA5', CH10, CE10, and BA10' for MLR model and CO5', CO5, BA5, and BA10' for AUL model, 
the previous UPRL methods yielded a larger-scale CPD prediction than the strict UPRL methods.
This result indicates that the previous UPRL methods have the smoothness-promotion.
The CPD prediction by learning methods with early stopping generally tended to have a larger-scale 
than the underlying CPD (estimated with $n_\tra=800$) when the training data size is small.
Under this tendency and the smoothness-promotion of the previous UPRL methods, 
the strict UPRL methods often 
(except for BA5', CAR, and BA10' for MLR model and 
SWD, CO5', CO5, BA5, and BA10' for AUL model) 
gave a CPD prediction closer to the underlying CPD than the previous UPRL methods 
in terms of the $L_1$ distance of the scale distribution.

%==========%
\begin{figure*}[!t]
\centering%
\renewcommand{\arraystretch}{0.25}%
\renewcommand{\tabcolsep}{0pt}%
\begin{tabular}{c|c}
{\scriptsize NLL}&{\scriptsize MZE}\\&\\%
\begin{tabular}{C{2.9cm}C{2.9cm}C{2.9cm}}%
\includegraphics[height=1.6cm]{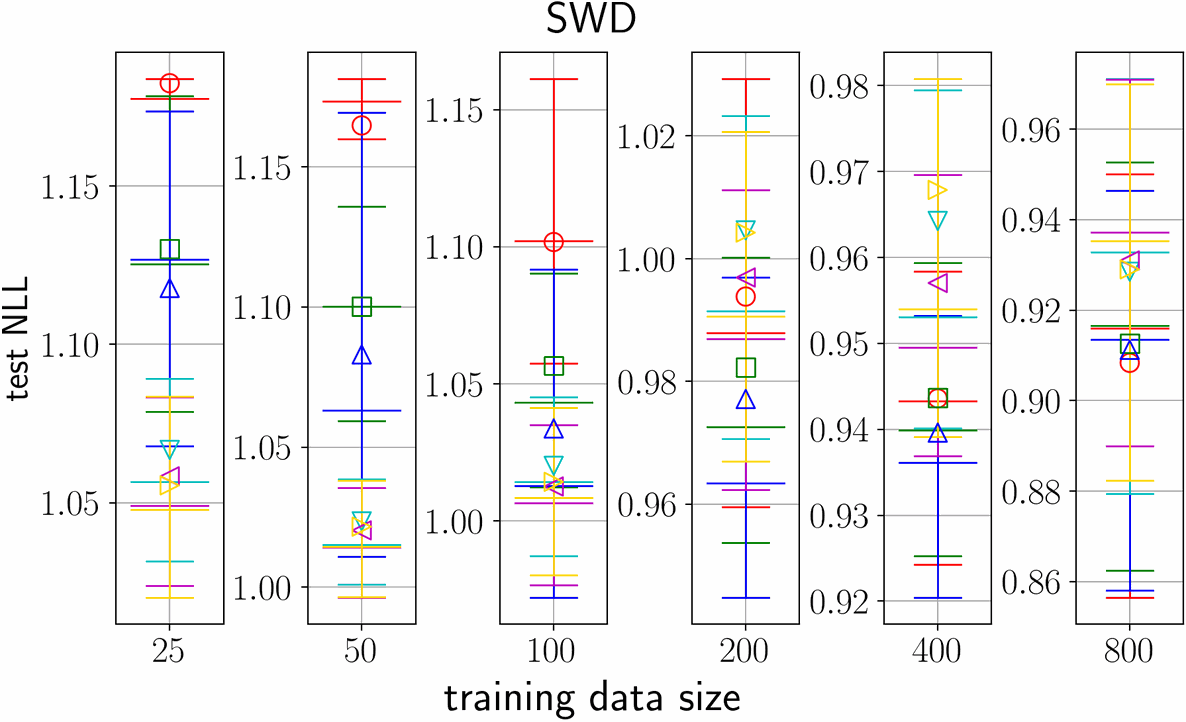}&
\includegraphics[height=1.6cm]{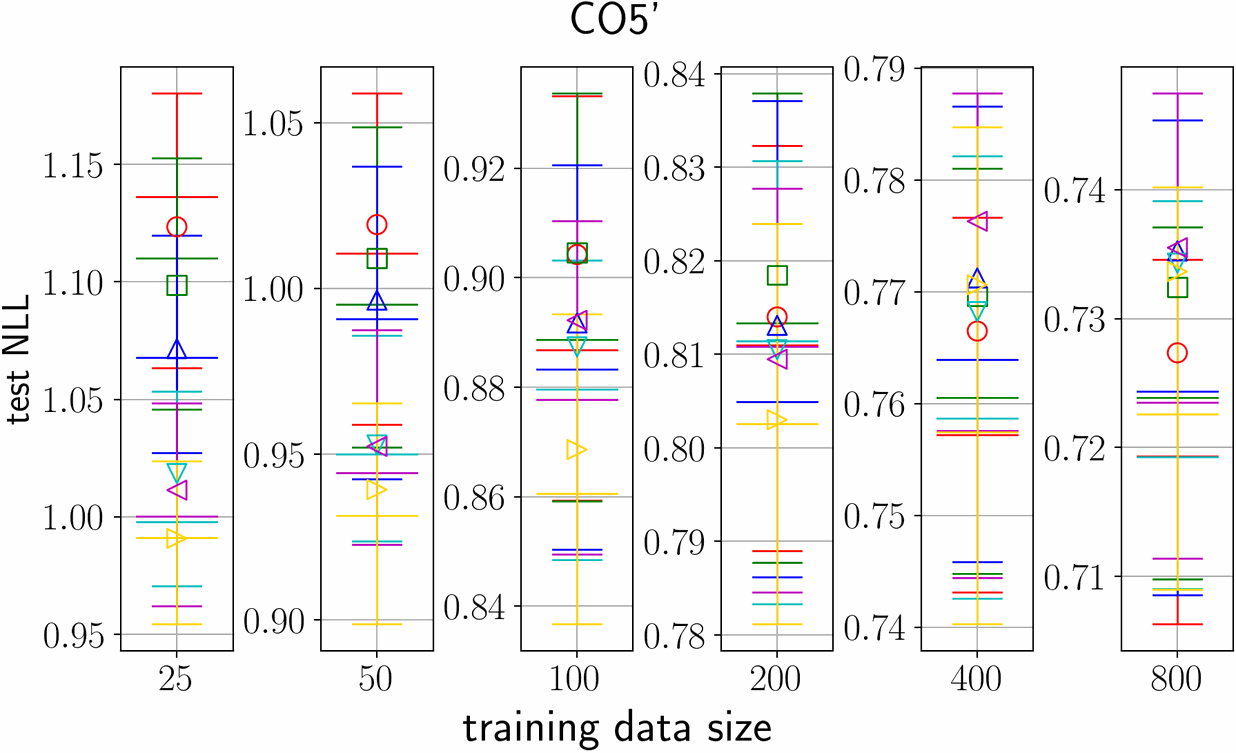}&
\includegraphics[height=1.6cm]{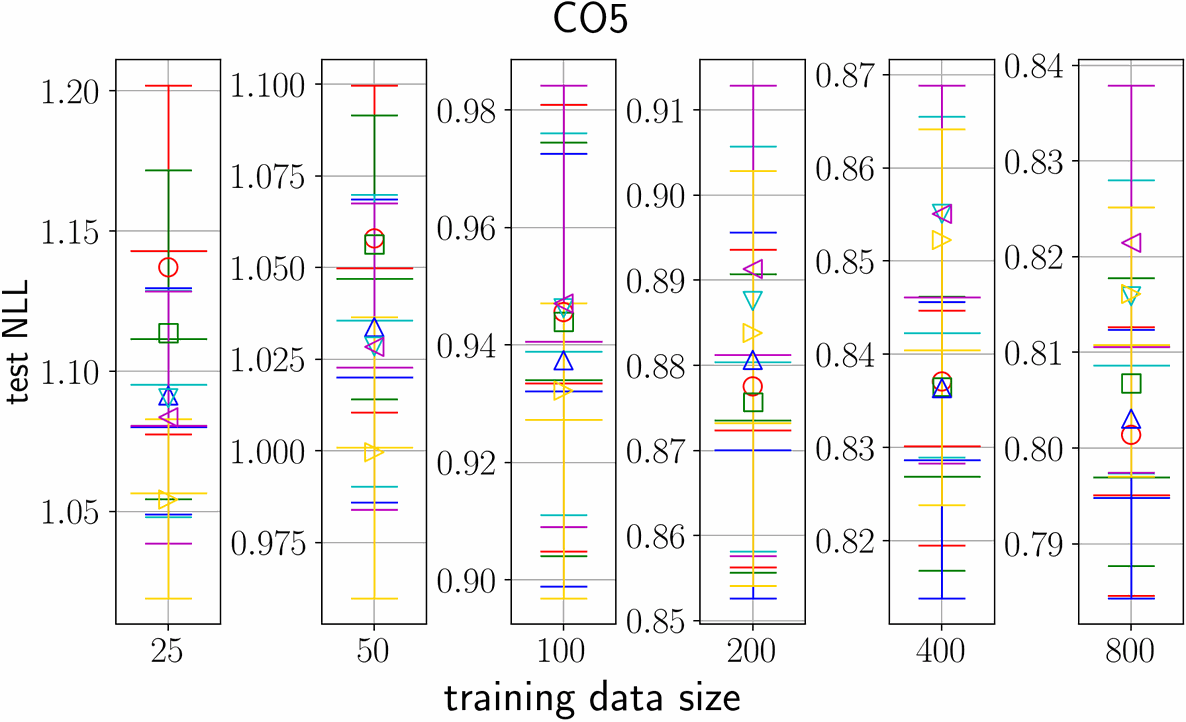}\\
\includegraphics[height=1.6cm]{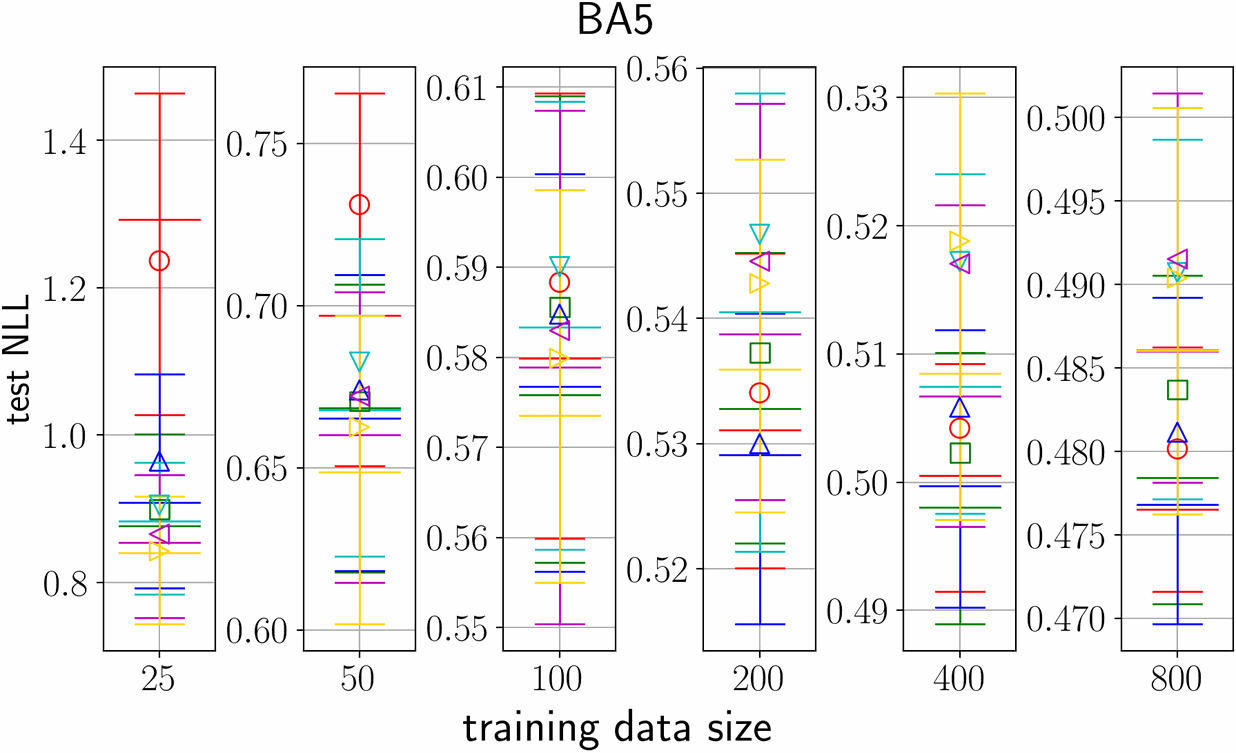}&
\includegraphics[height=1.6cm]{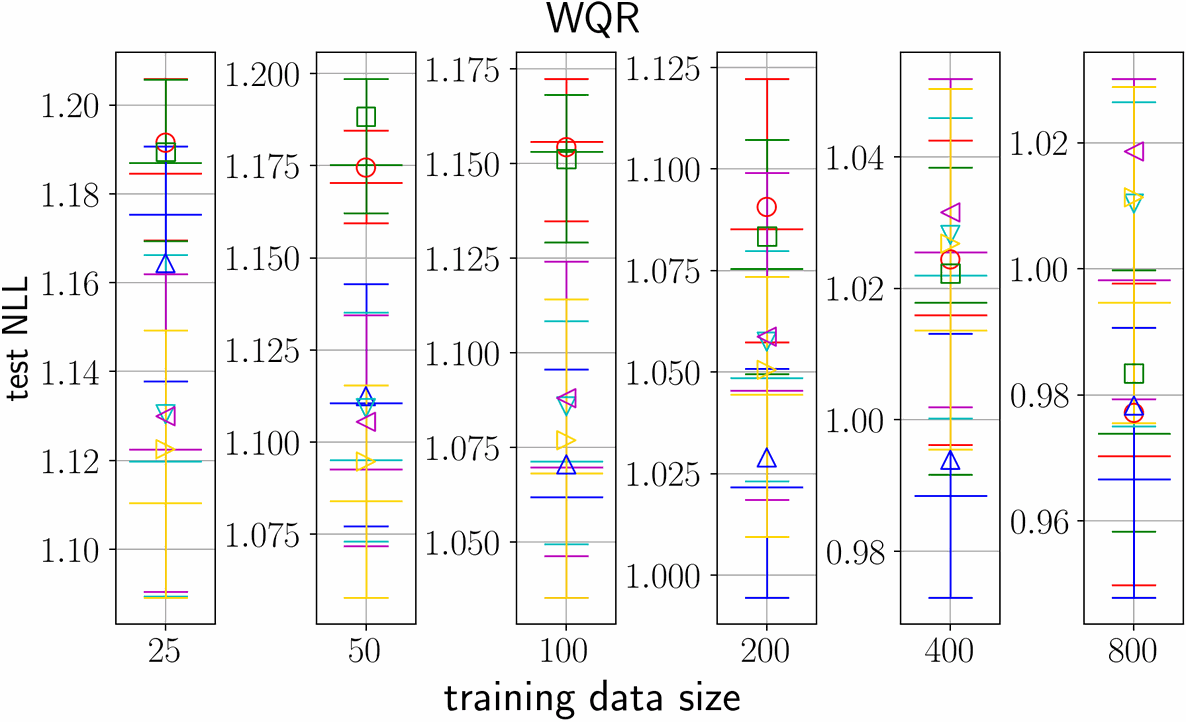}&
\includegraphics[height=1.6cm]{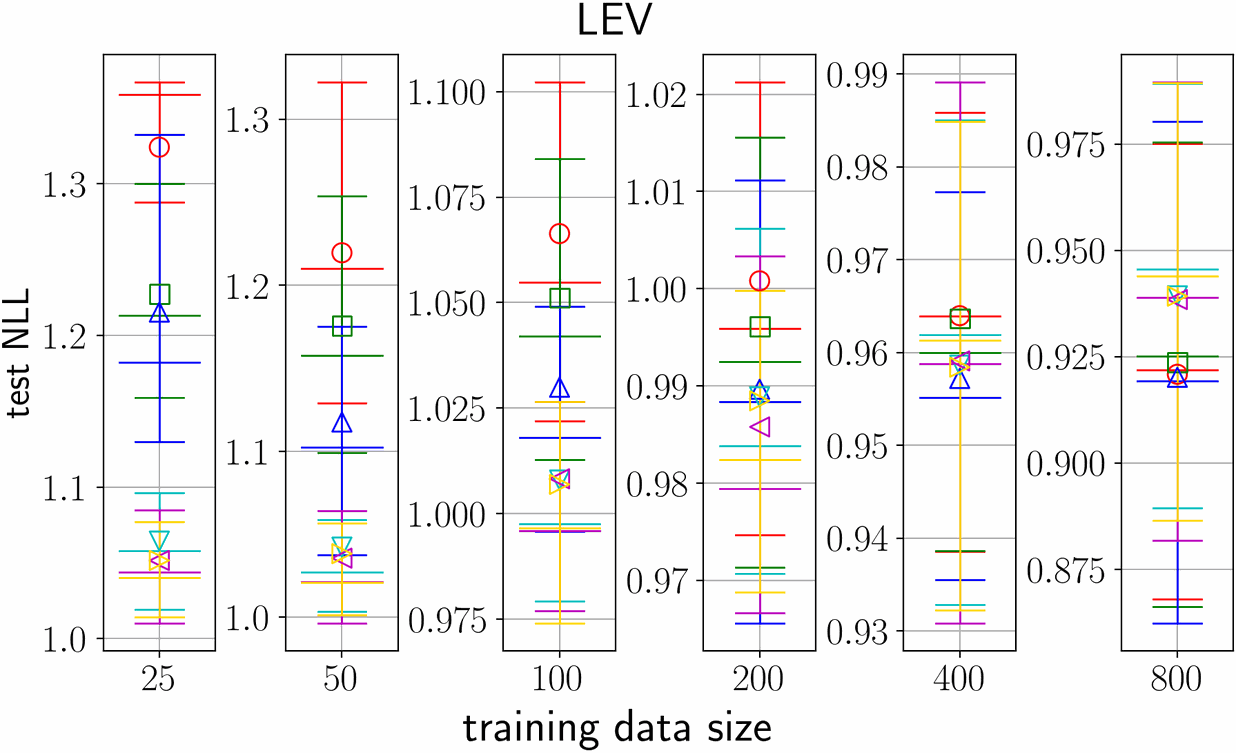}\\
\includegraphics[height=1.6cm]{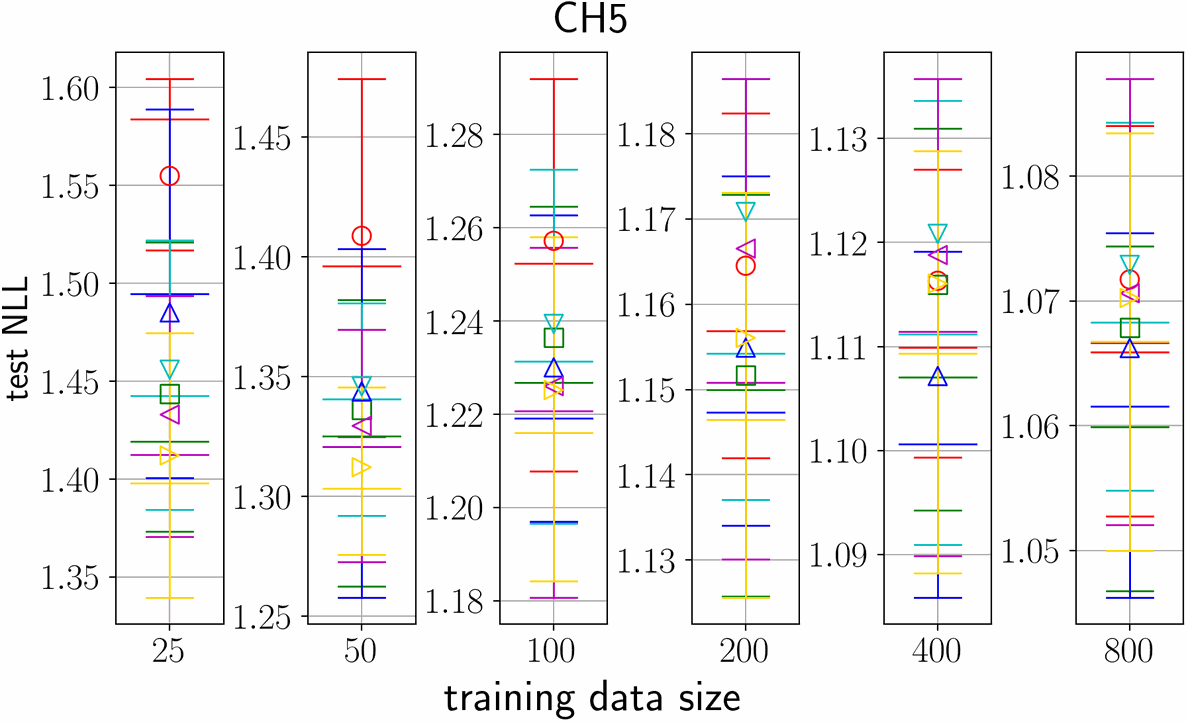}&
\includegraphics[height=1.6cm]{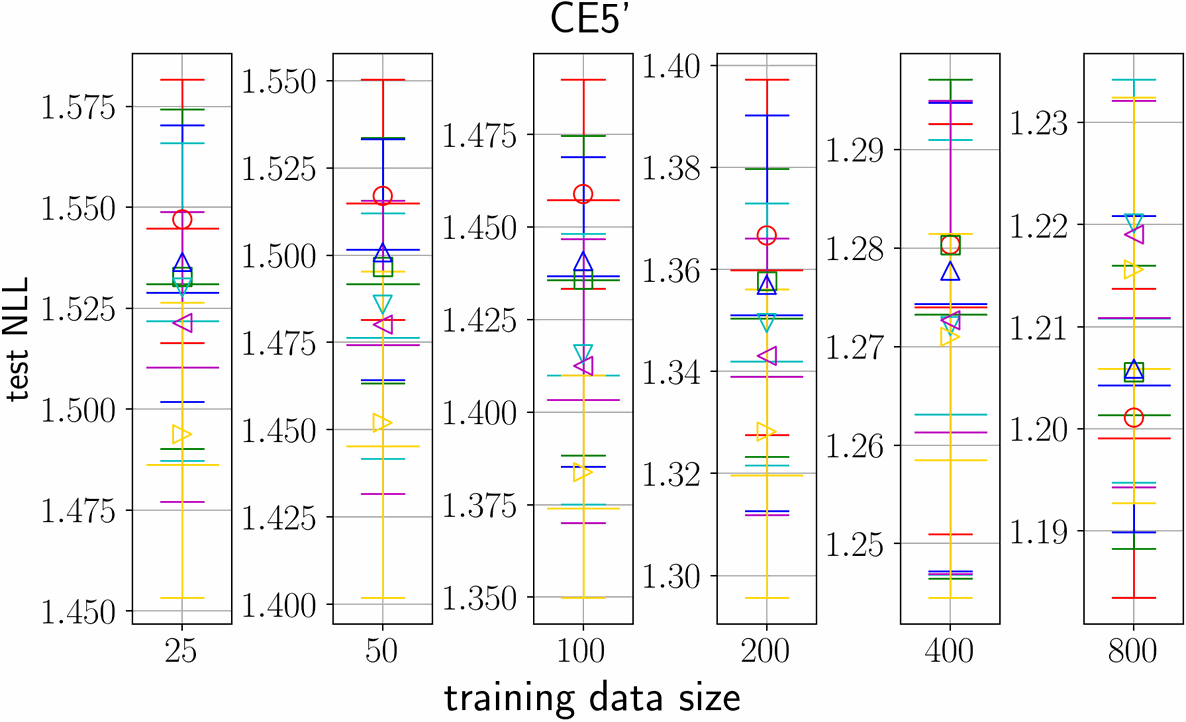}&
\includegraphics[height=1.6cm]{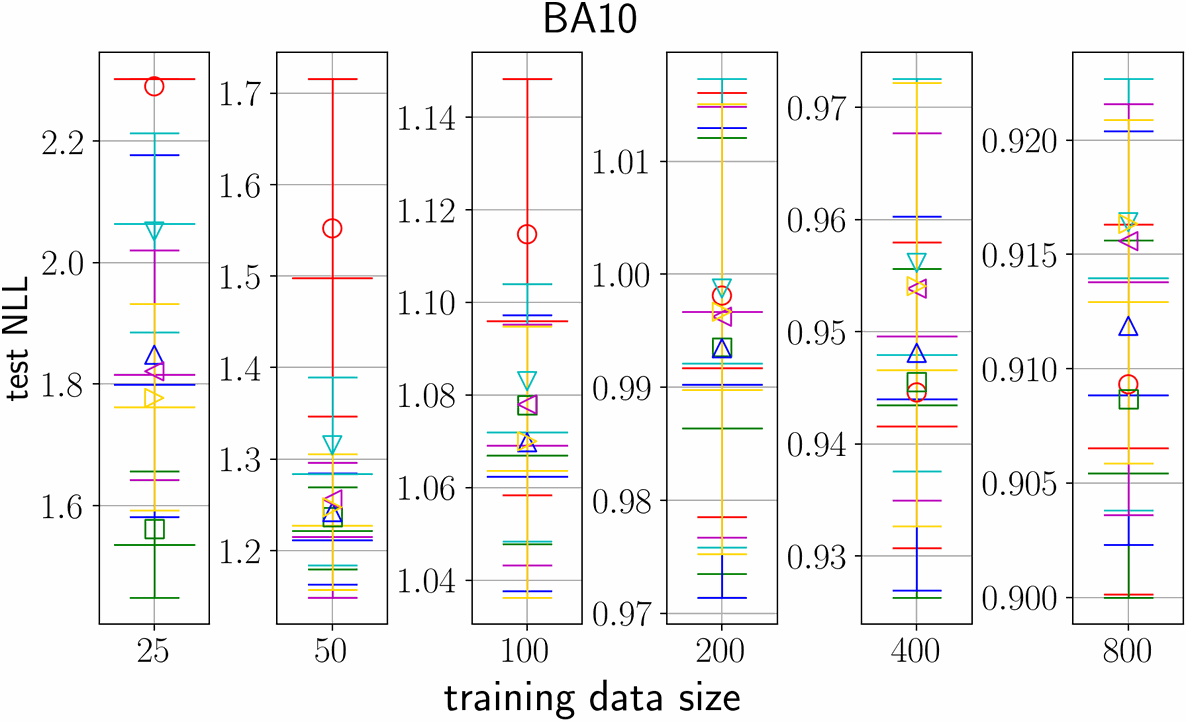}\\
\includegraphics[height=1.6cm]{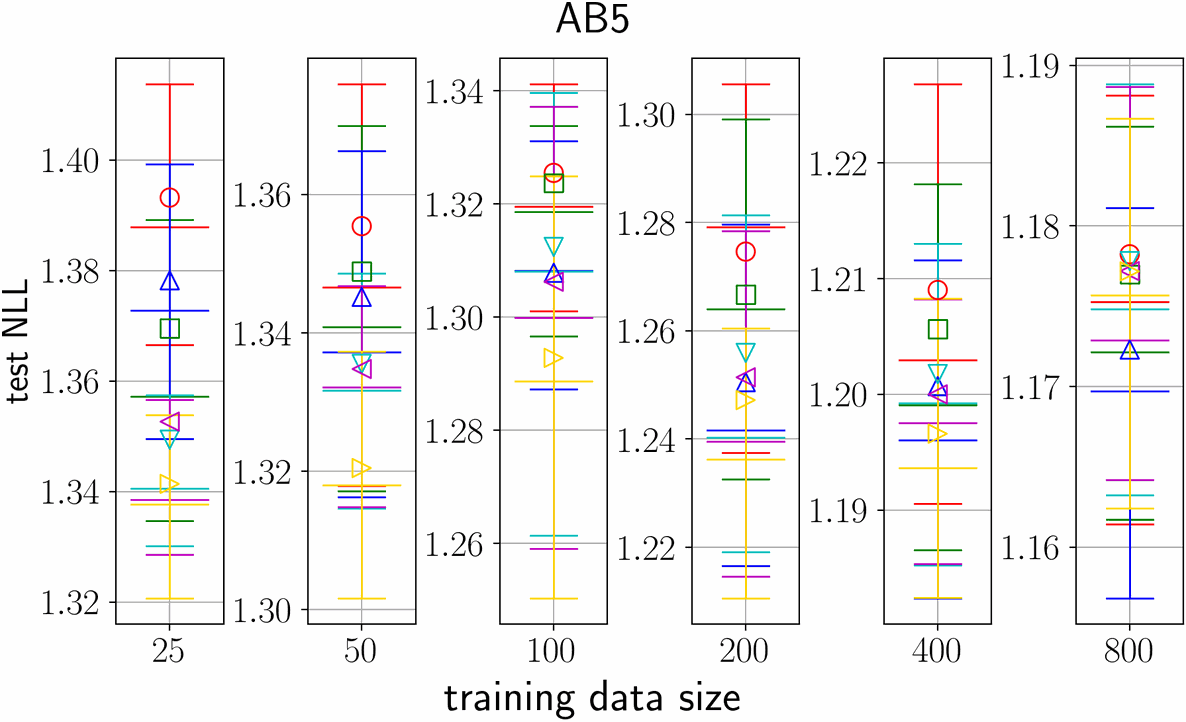}&
\includegraphics[height=1.6cm]{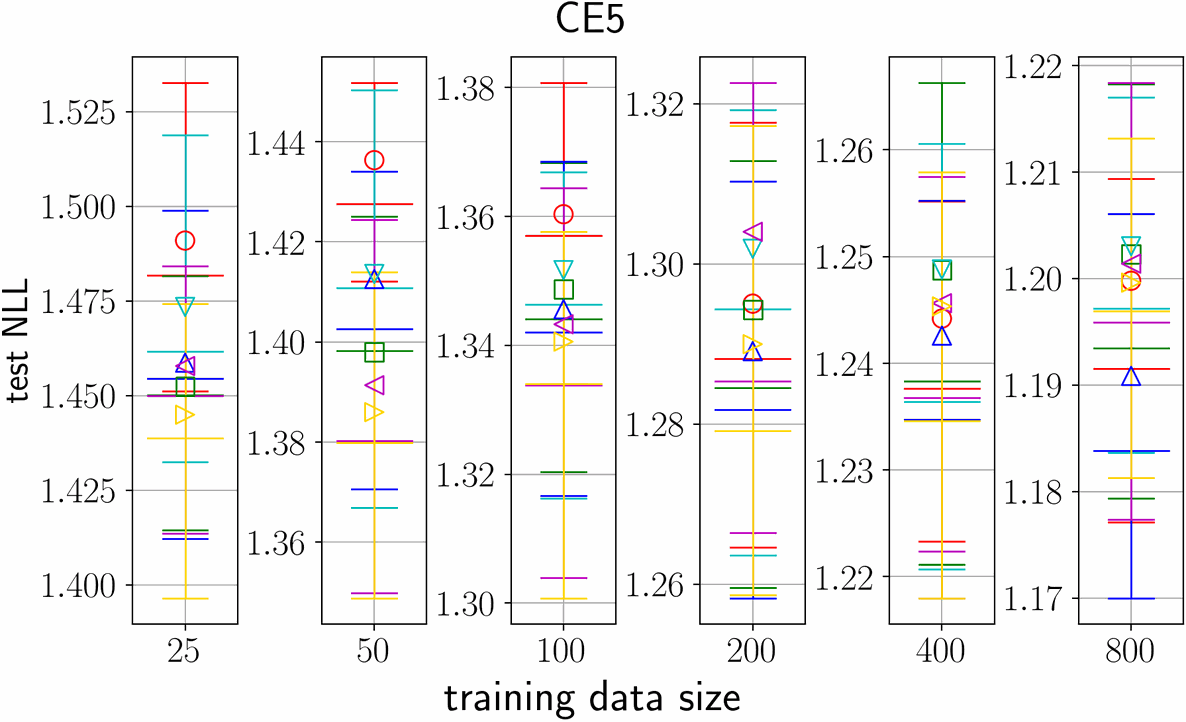}&
\includegraphics[height=1.6cm]{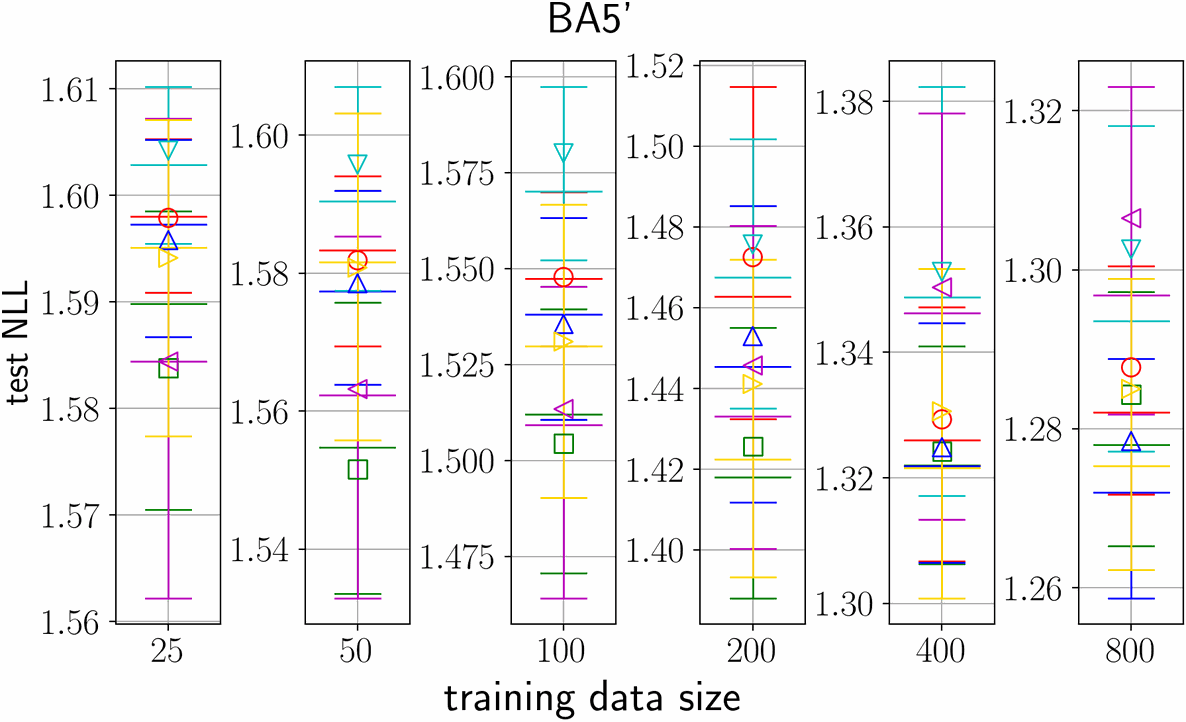}\\
\includegraphics[height=1.6cm]{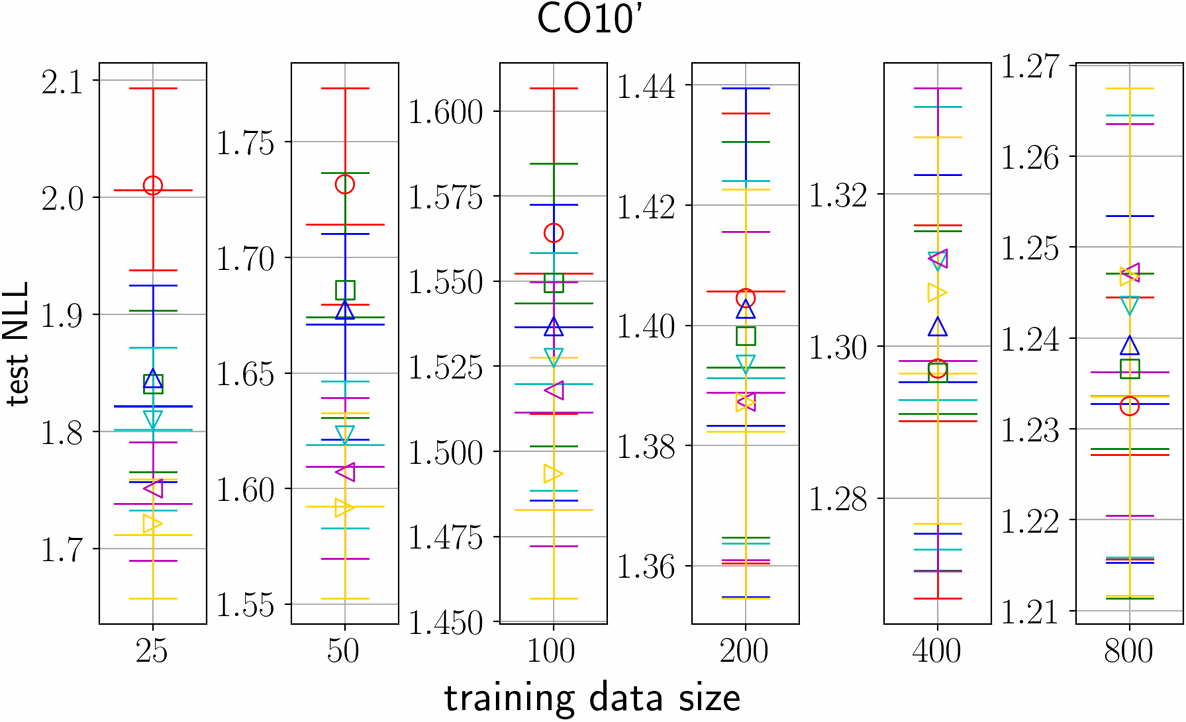}&
\includegraphics[height=1.6cm]{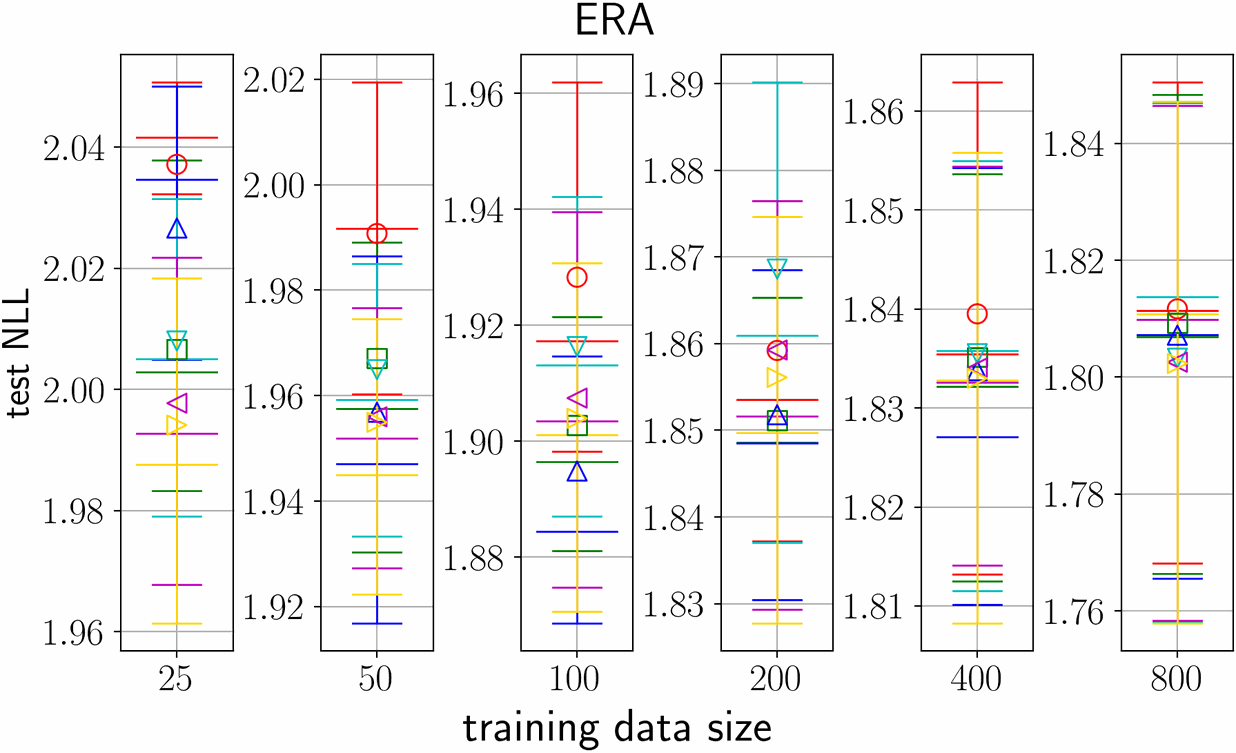}&
\includegraphics[height=1.6cm]{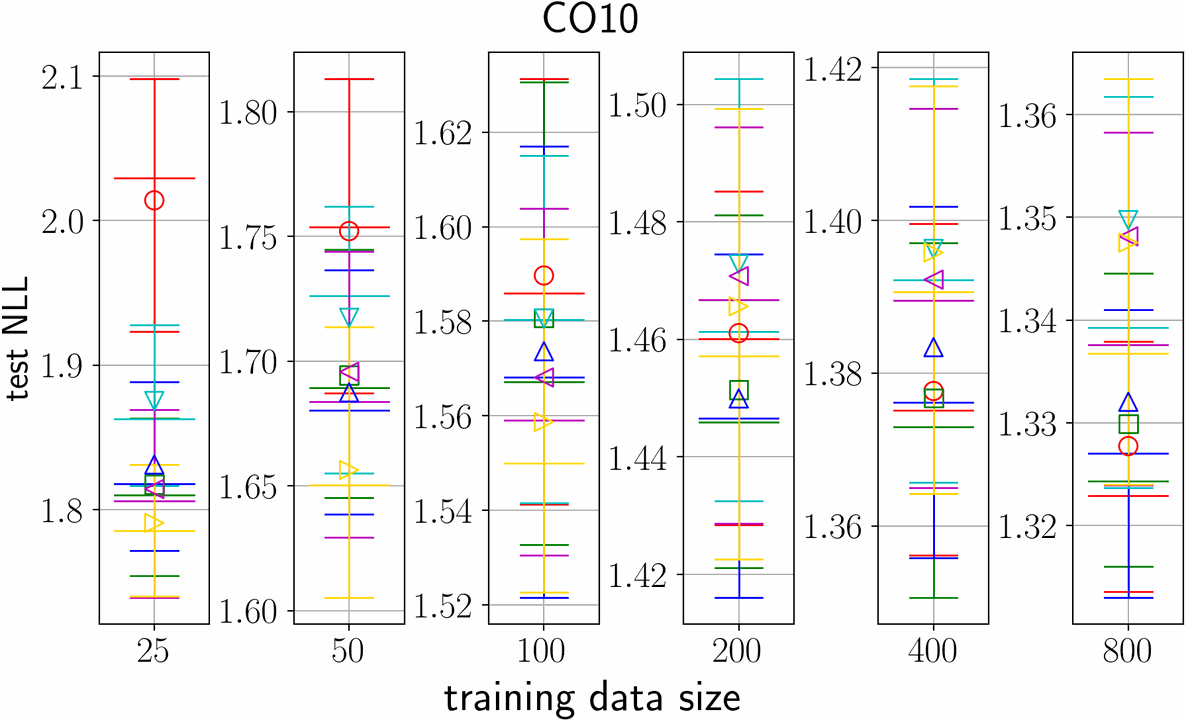}\\
\includegraphics[height=1.6cm]{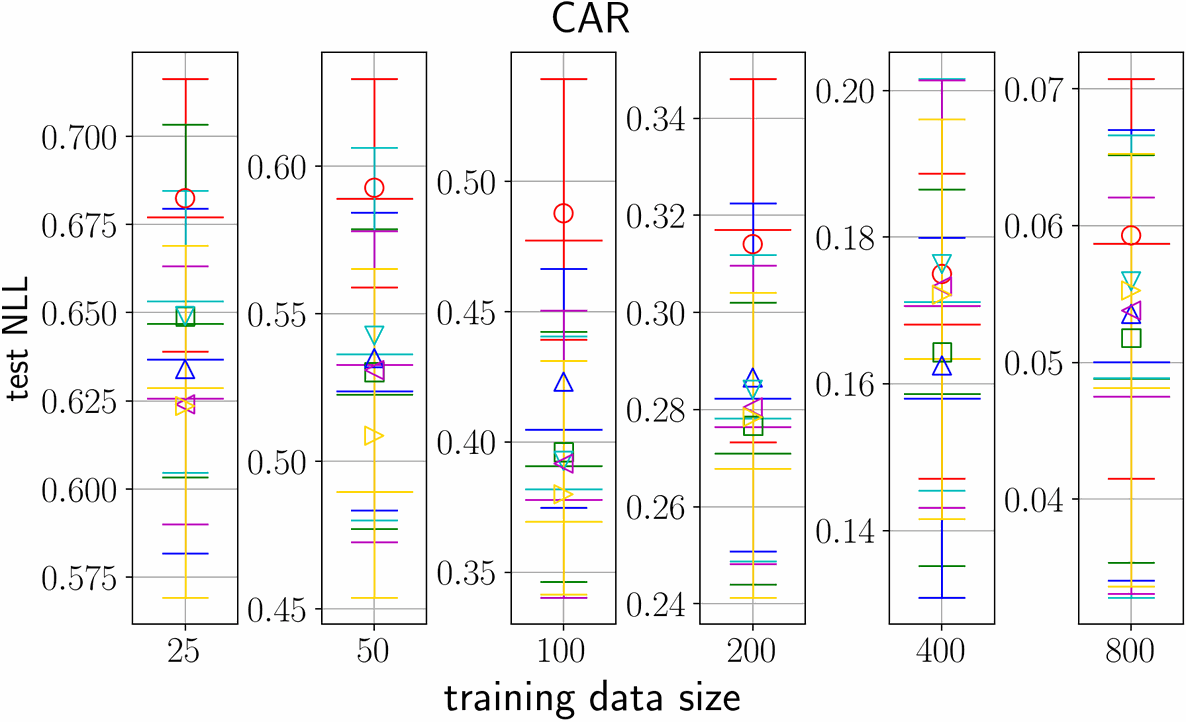}&
\includegraphics[height=1.6cm]{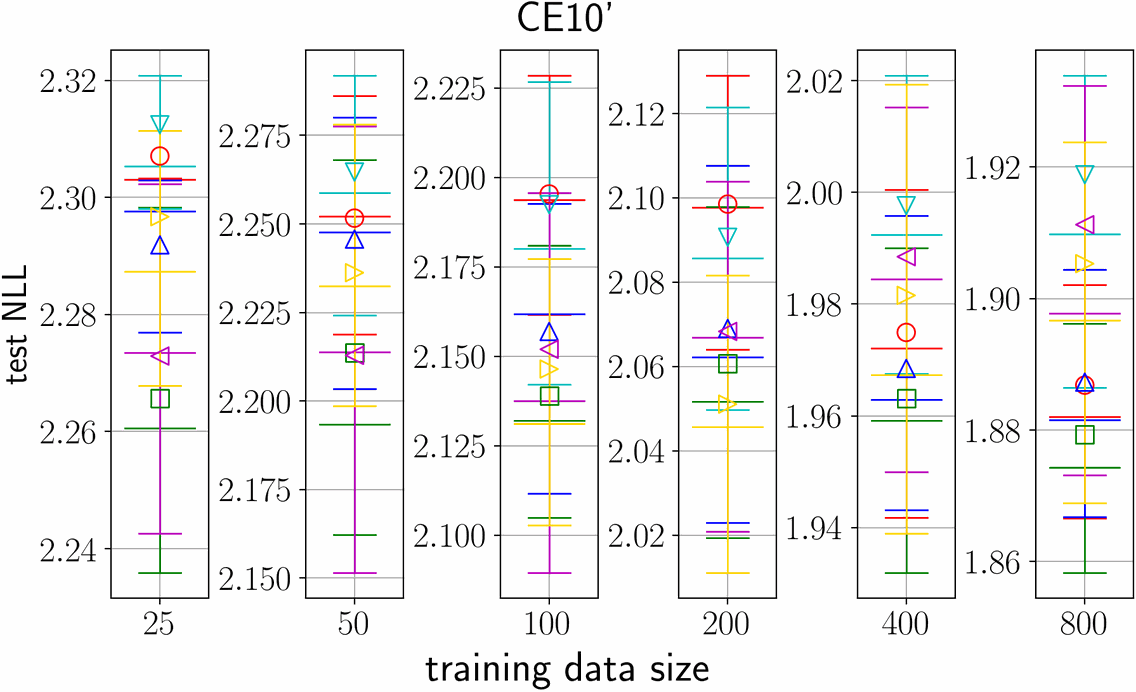}&
\includegraphics[height=1.6cm]{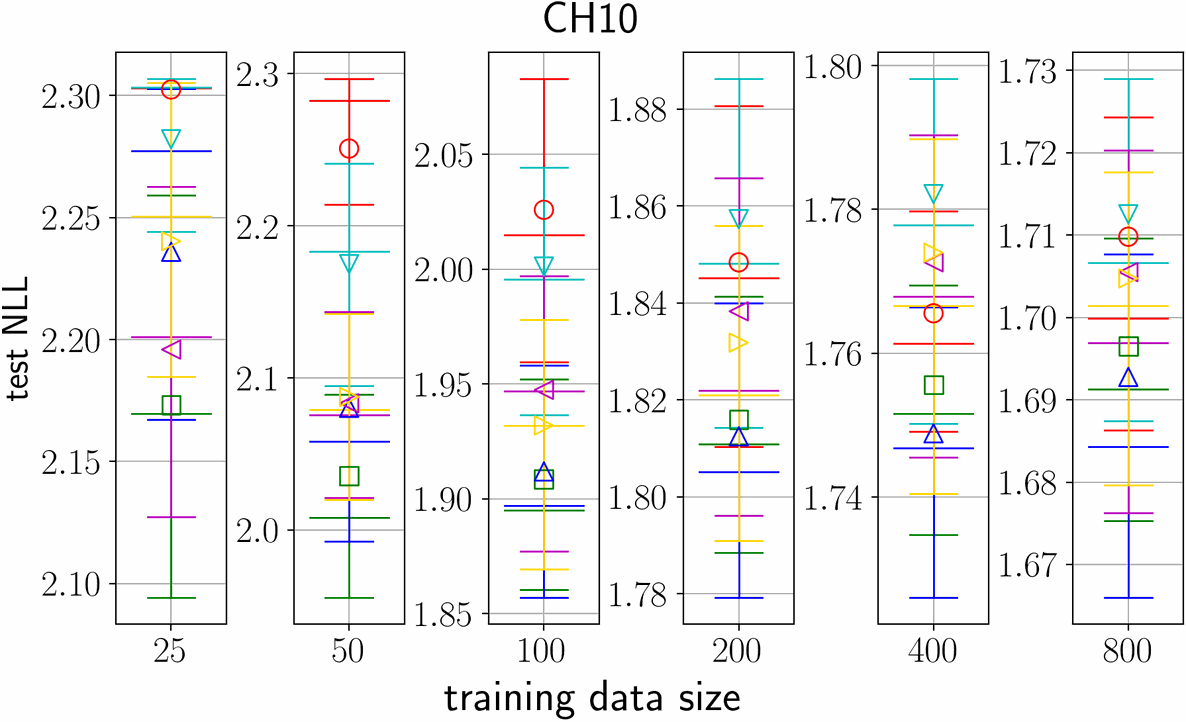}\\
\includegraphics[height=1.6cm]{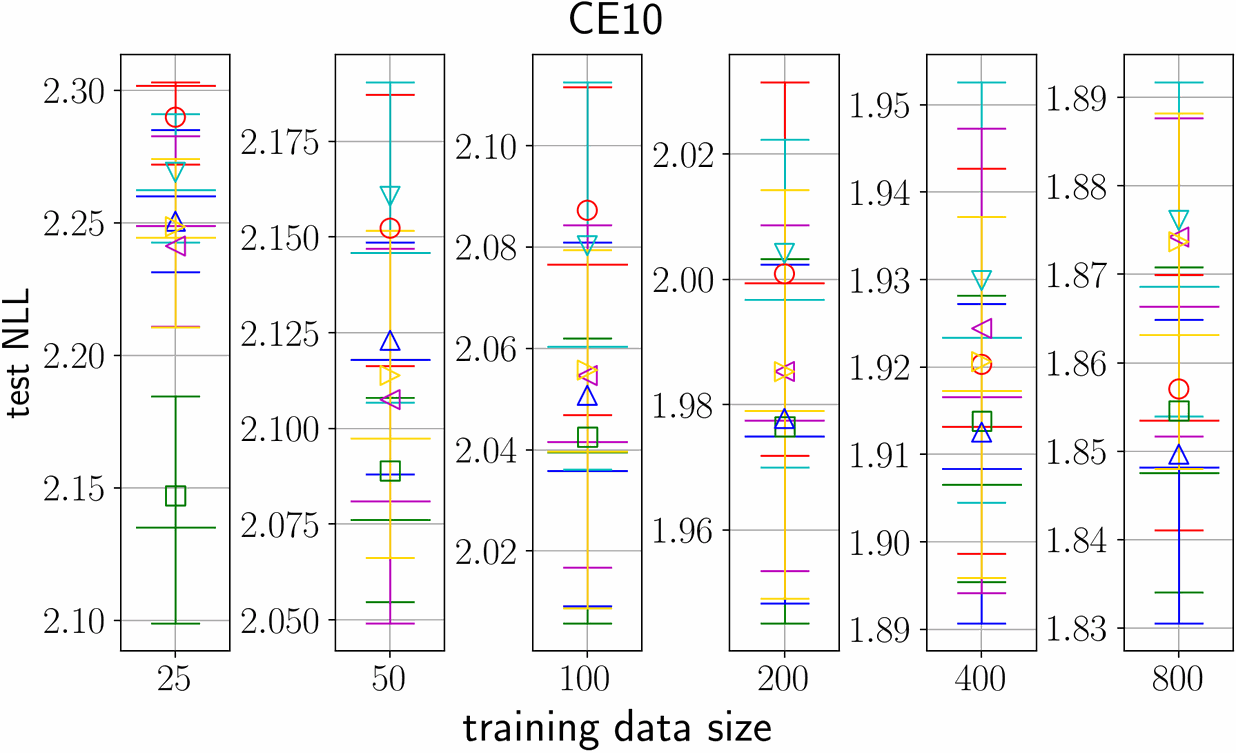}&
\includegraphics[height=1.6cm]{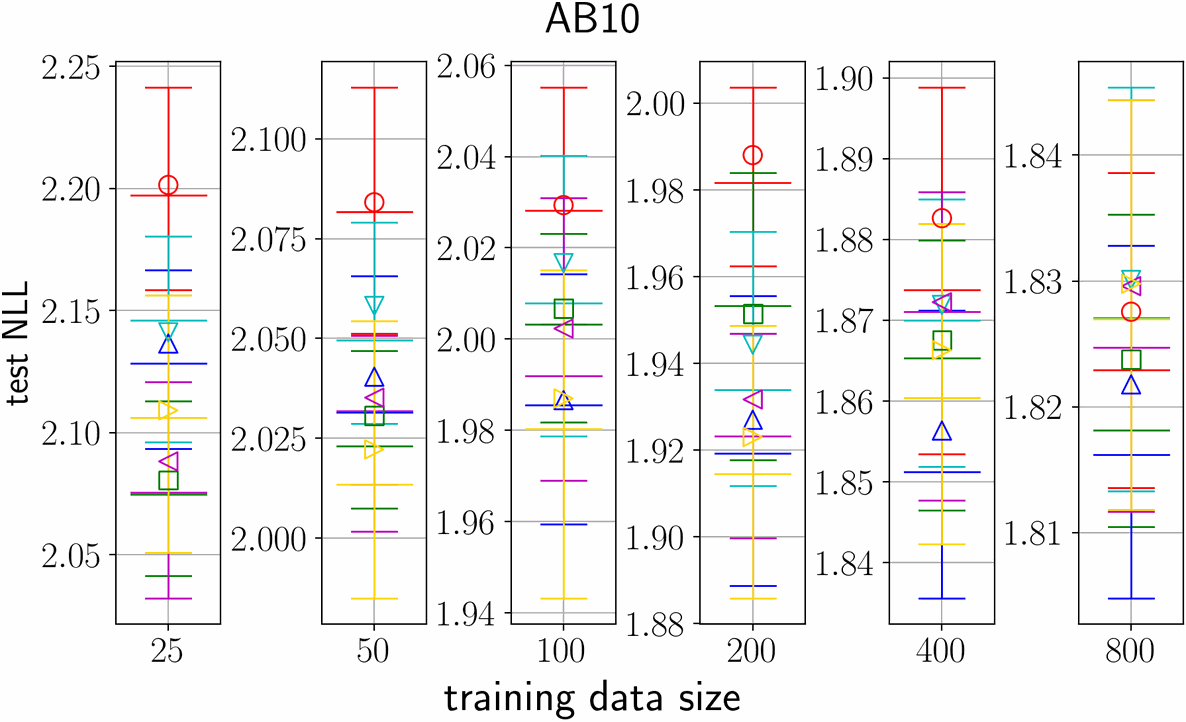}&
\includegraphics[height=1.6cm]{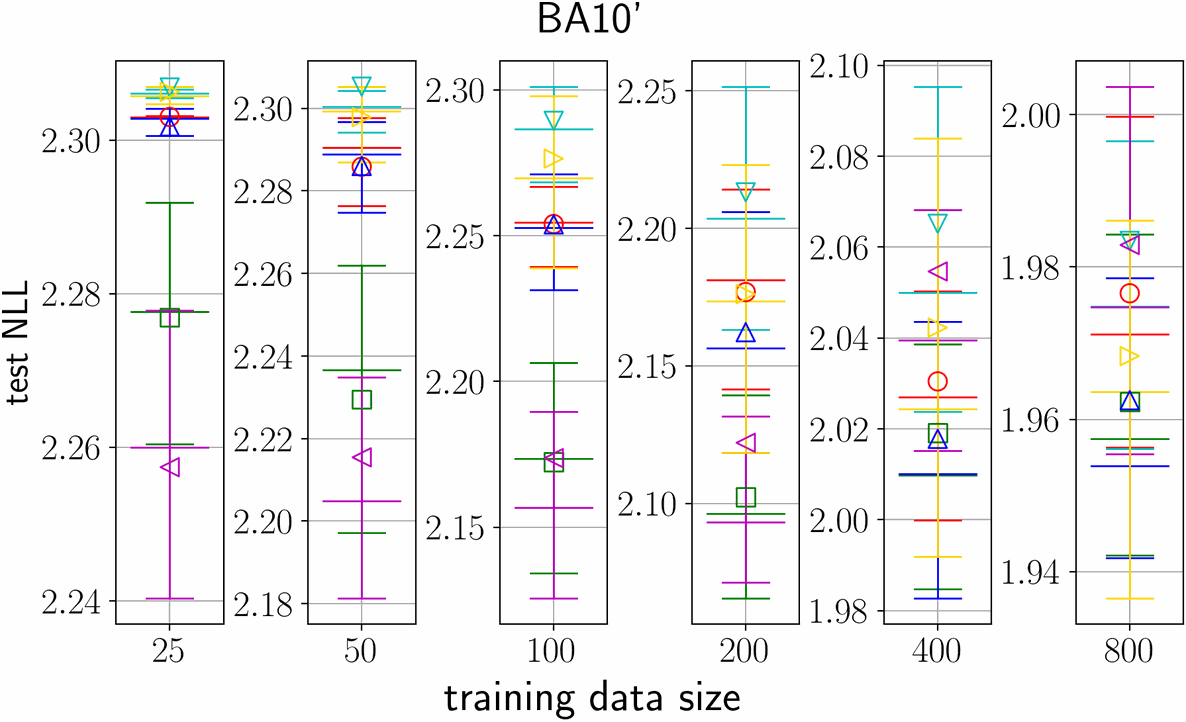}
\end{tabular}~~~&\,
\begin{tabular}{C{2.9cm}C{2.9cm}C{2.9cm}}%
\includegraphics[height=1.6cm]{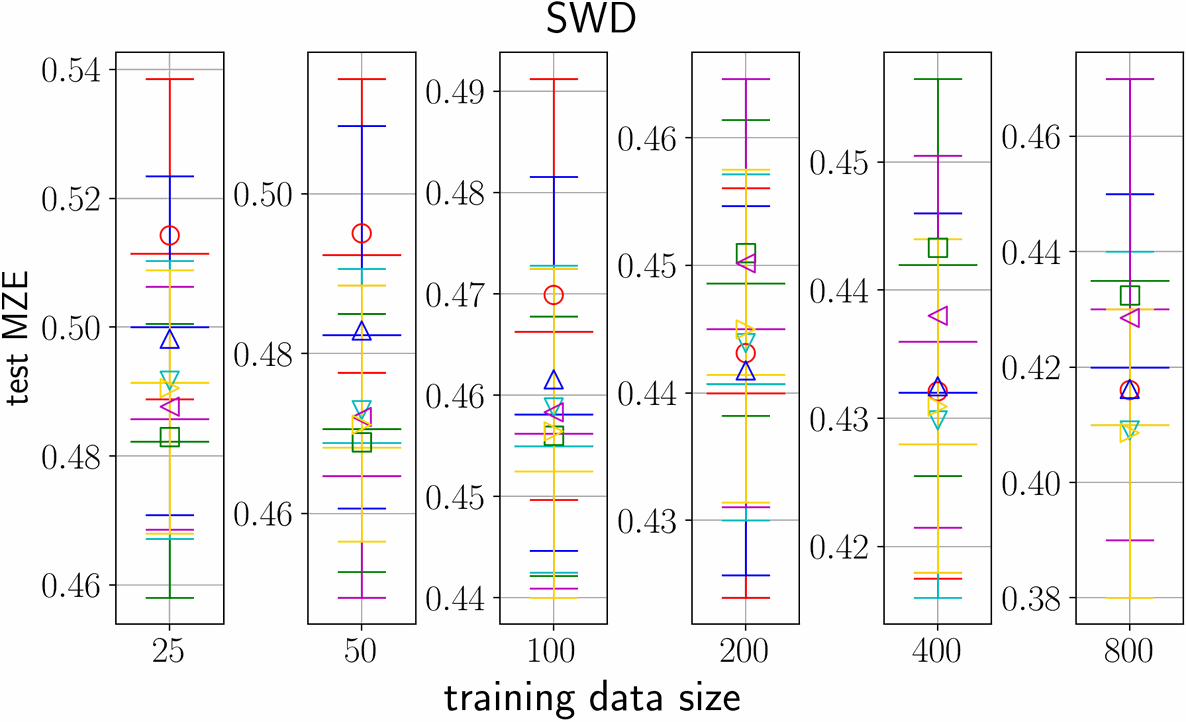}&
\includegraphics[height=1.6cm]{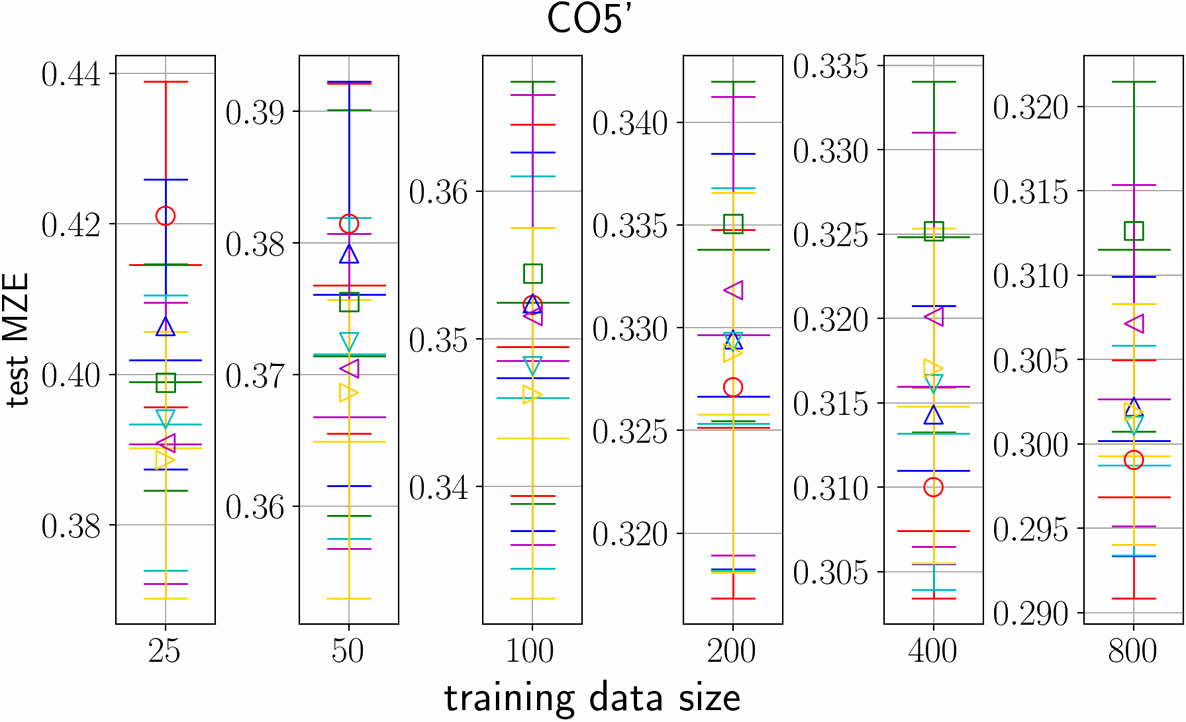}&
\includegraphics[height=1.6cm]{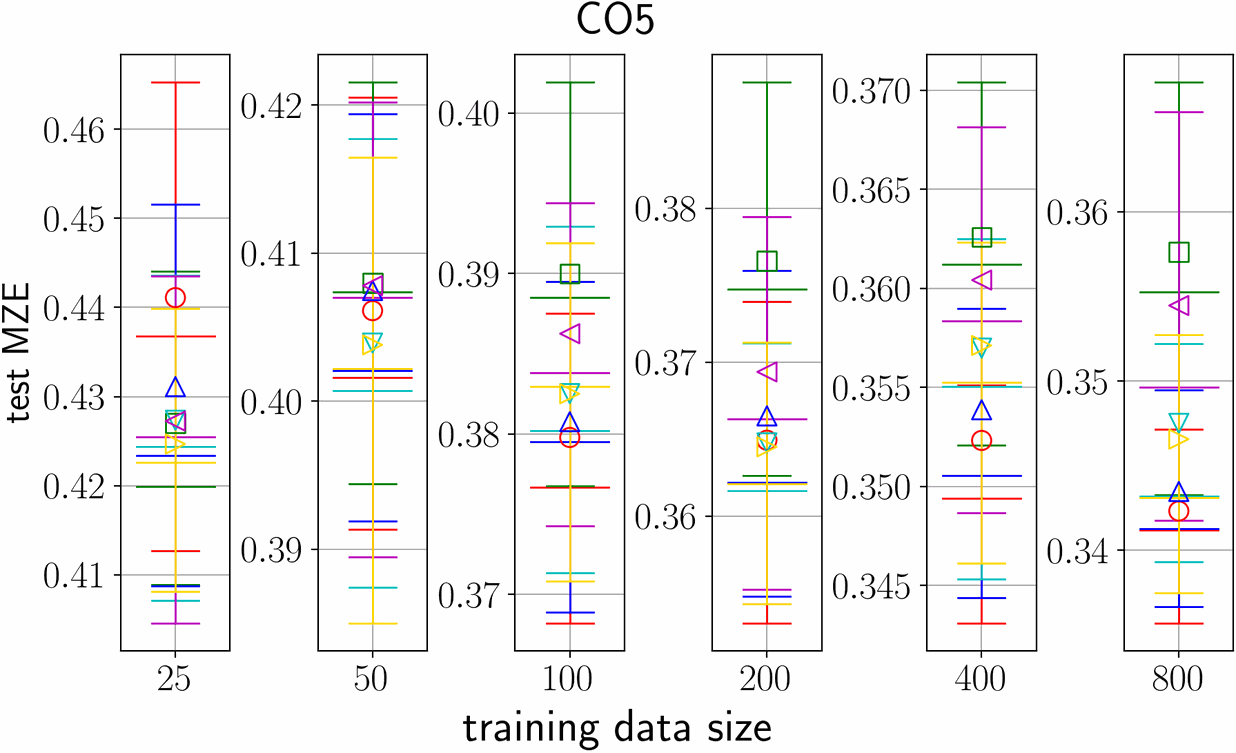}\\
\includegraphics[height=1.6cm]{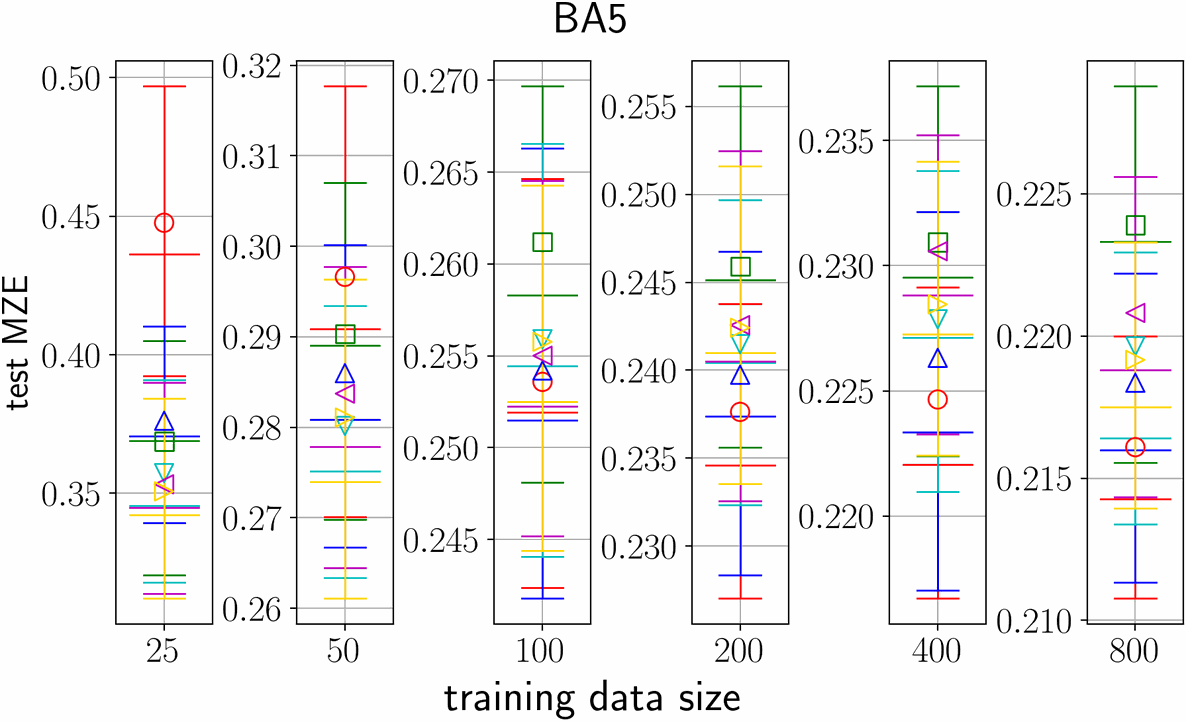}&
\includegraphics[height=1.6cm]{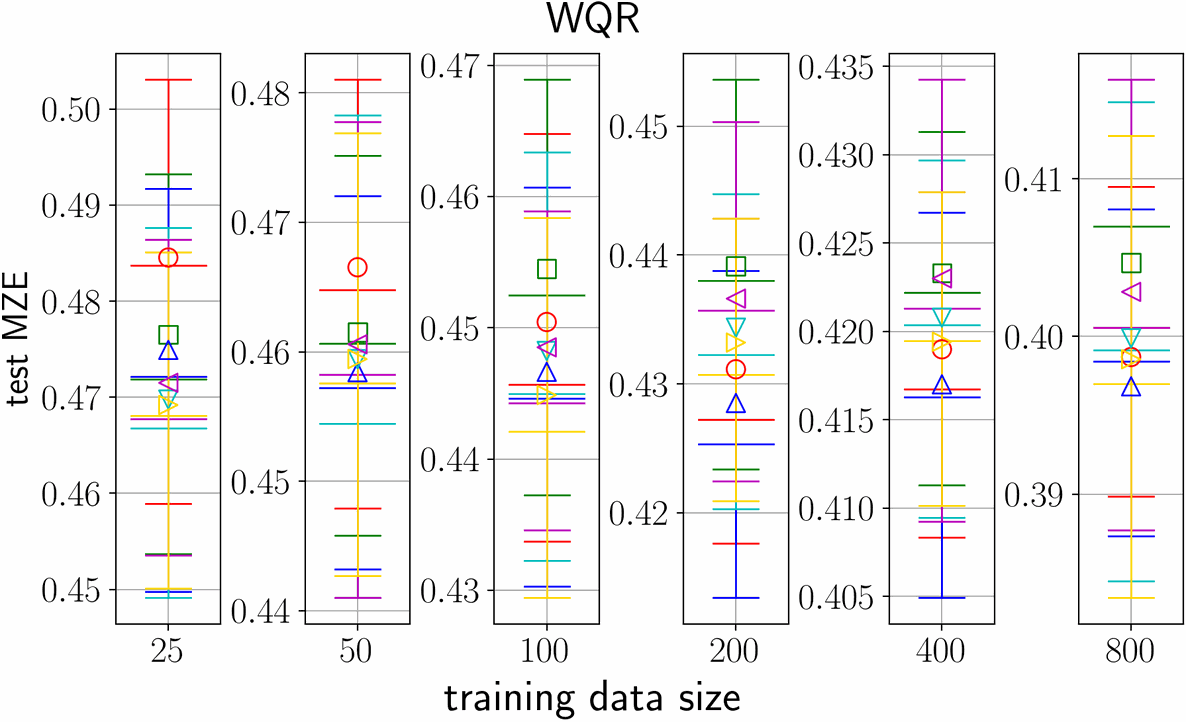}&
\includegraphics[height=1.6cm]{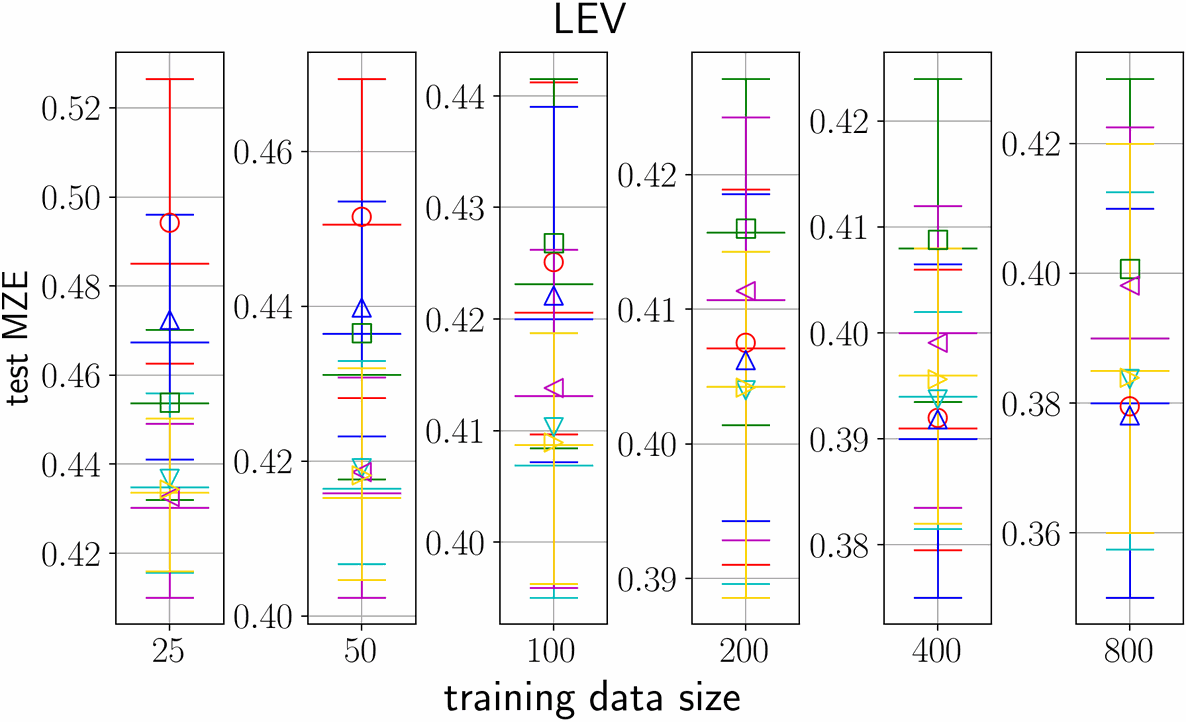}\\
\includegraphics[height=1.6cm]{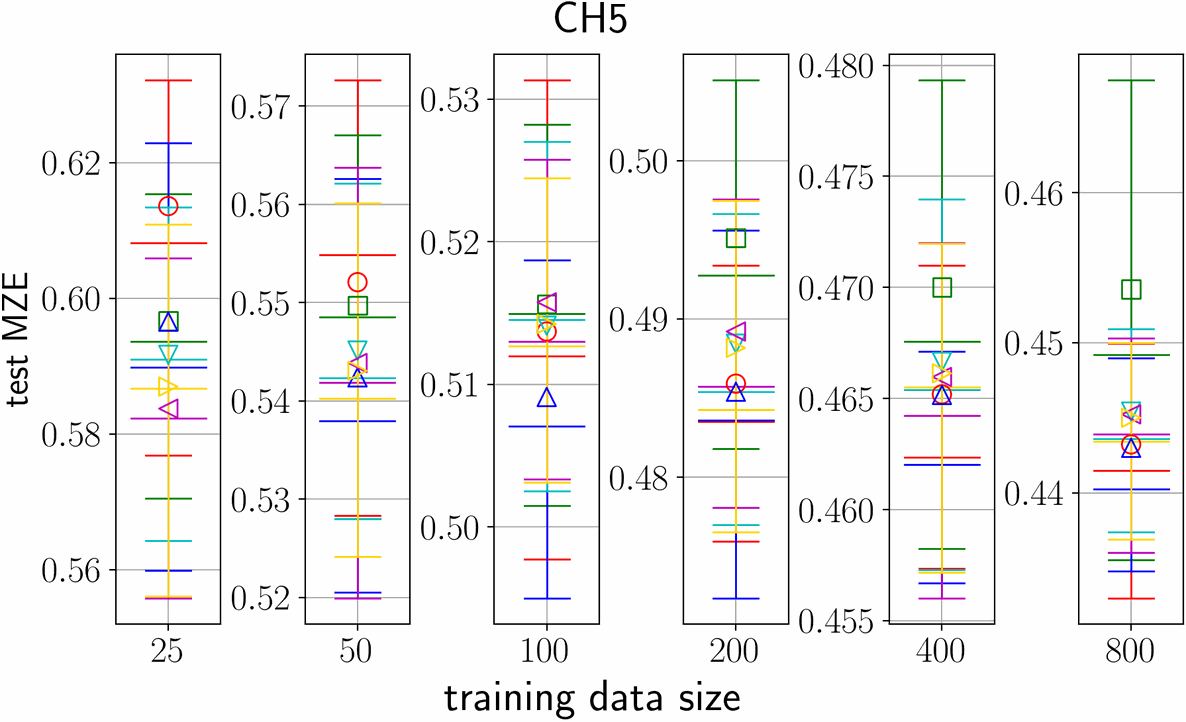}&
\includegraphics[height=1.6cm]{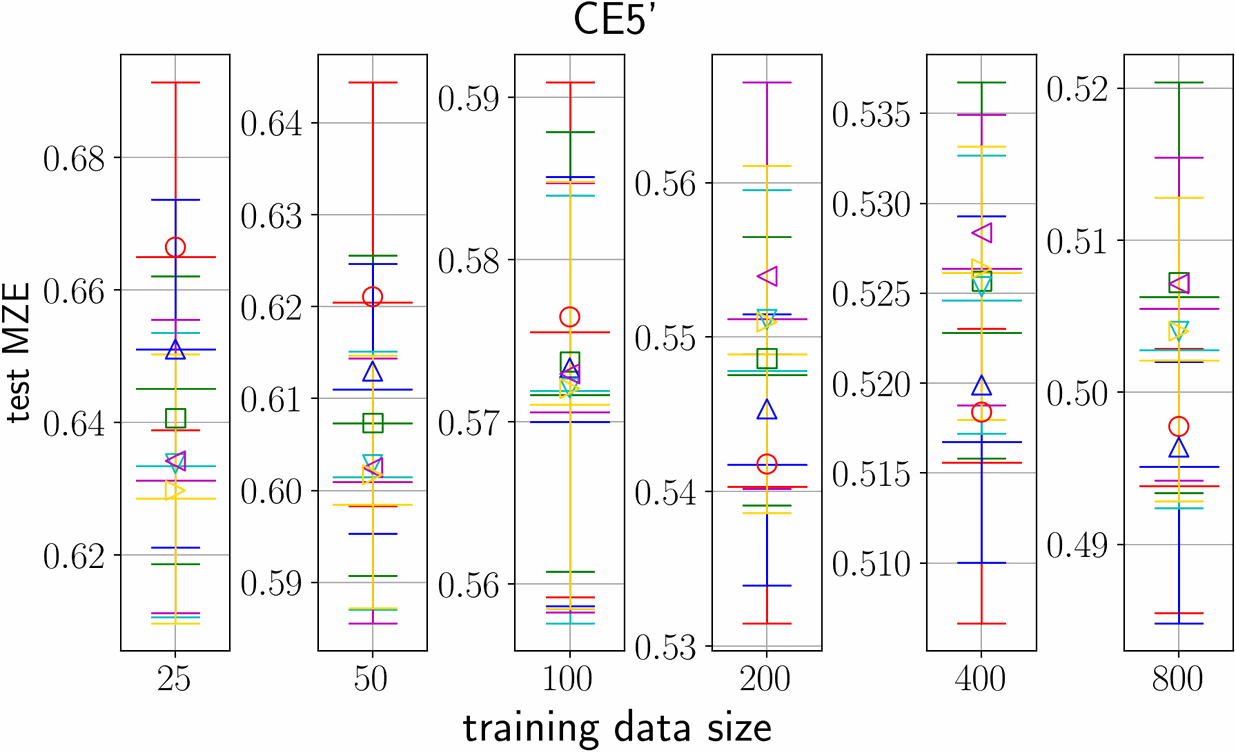}&
\includegraphics[height=1.6cm]{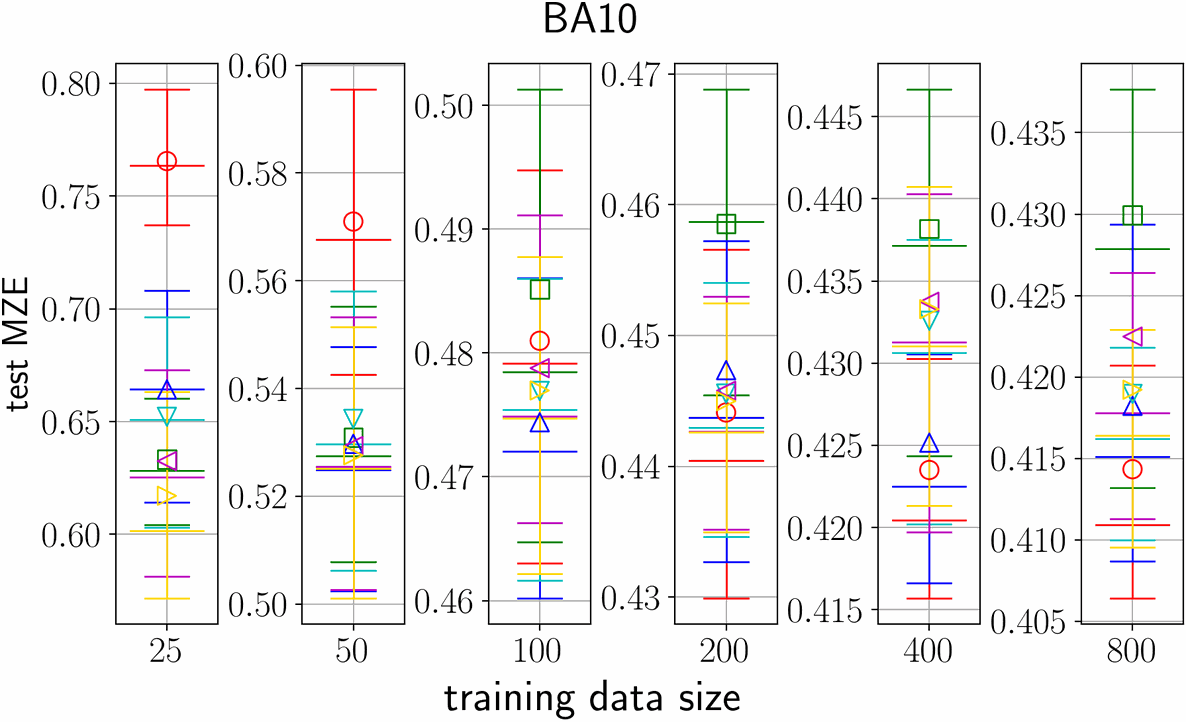}\\
\includegraphics[height=1.6cm]{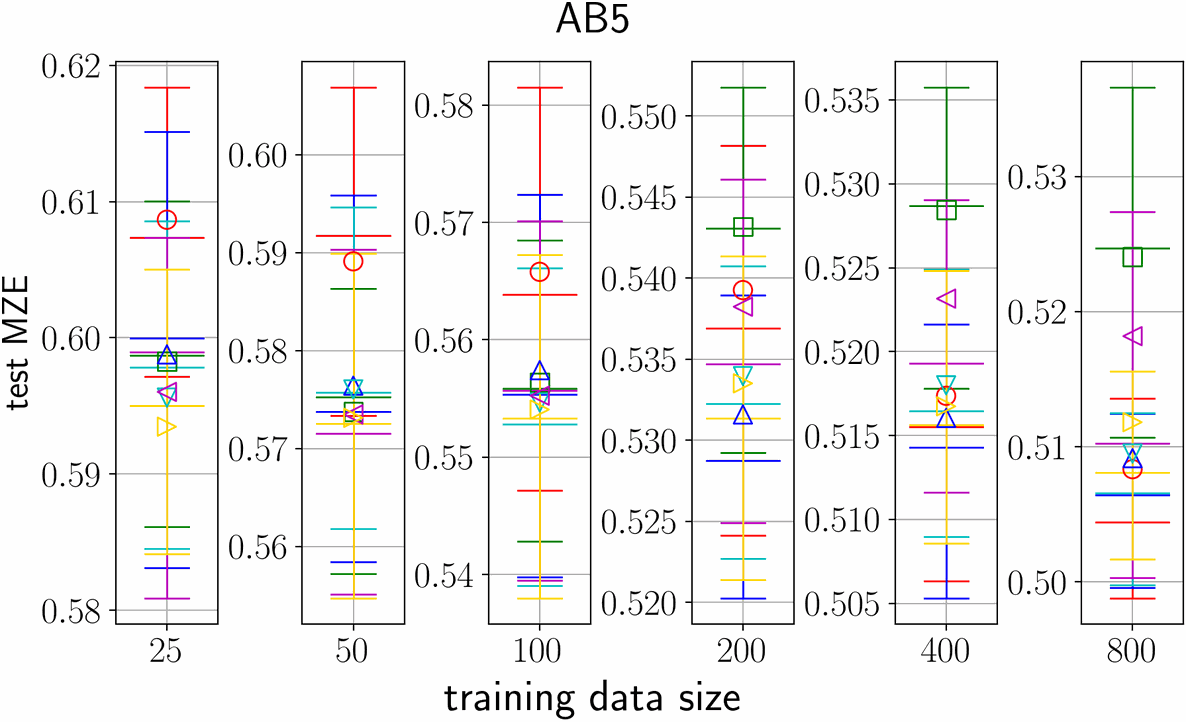}&
\includegraphics[height=1.6cm]{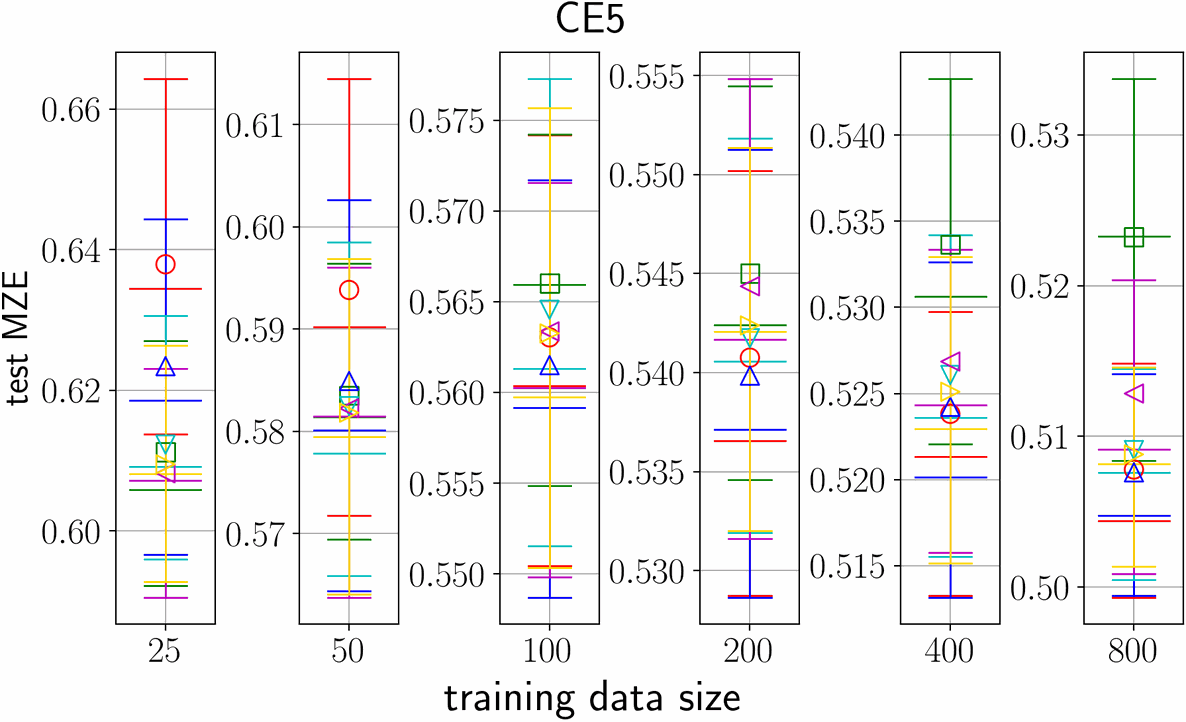}&
\includegraphics[height=1.6cm]{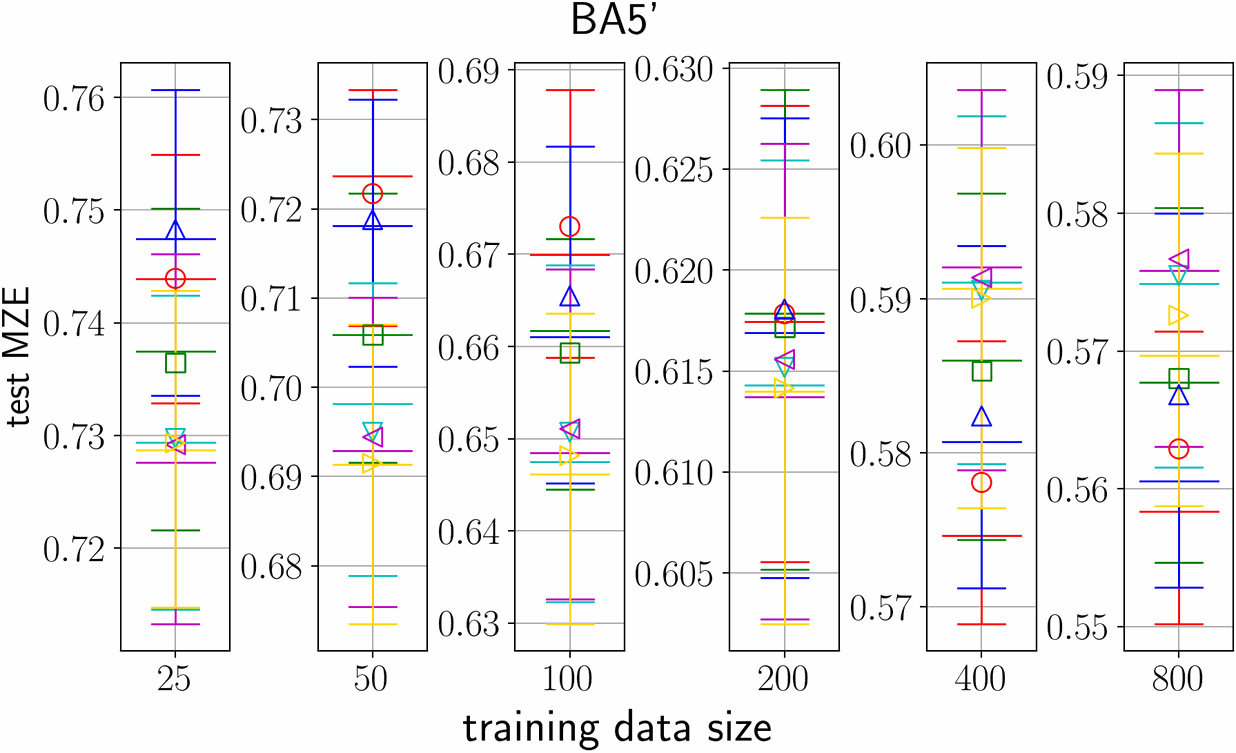}\\
\includegraphics[height=1.6cm]{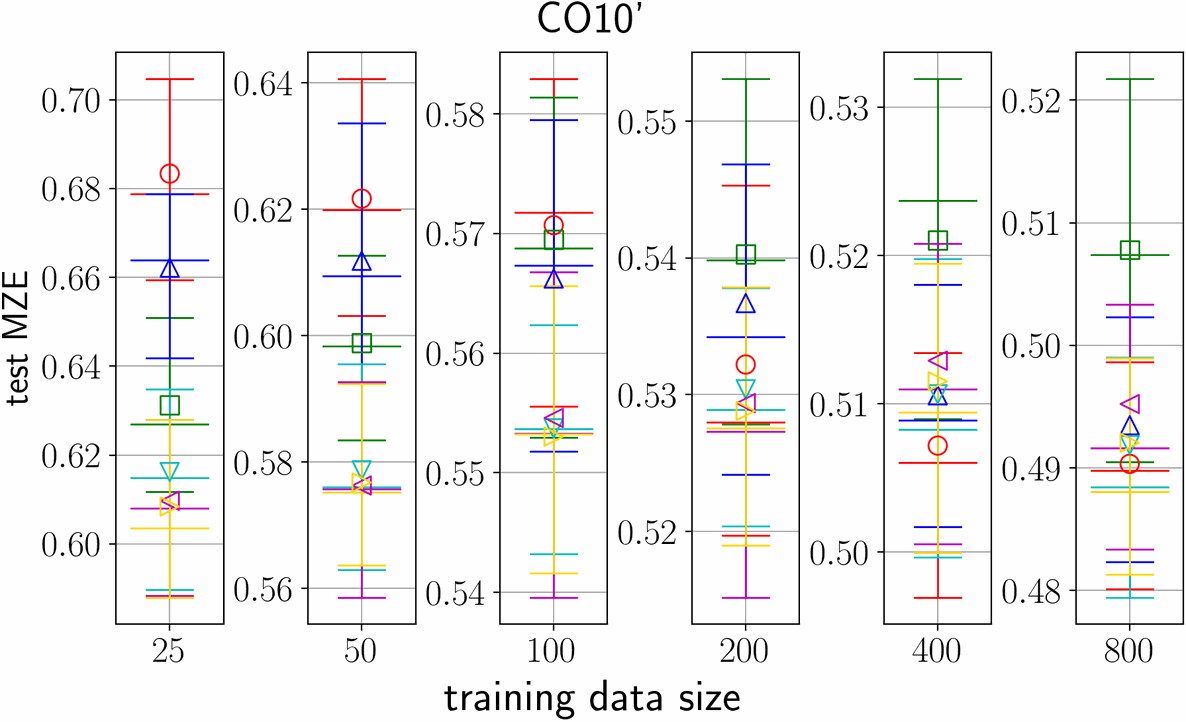}&
\includegraphics[height=1.6cm]{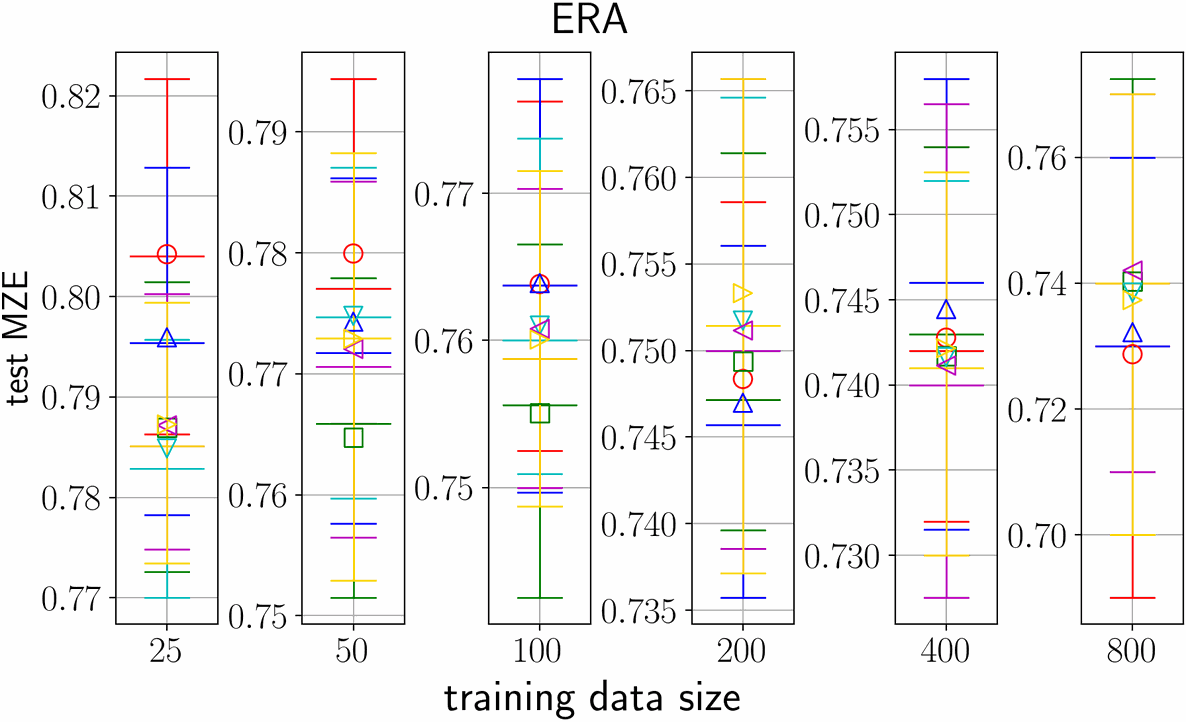}&
\includegraphics[height=1.6cm]{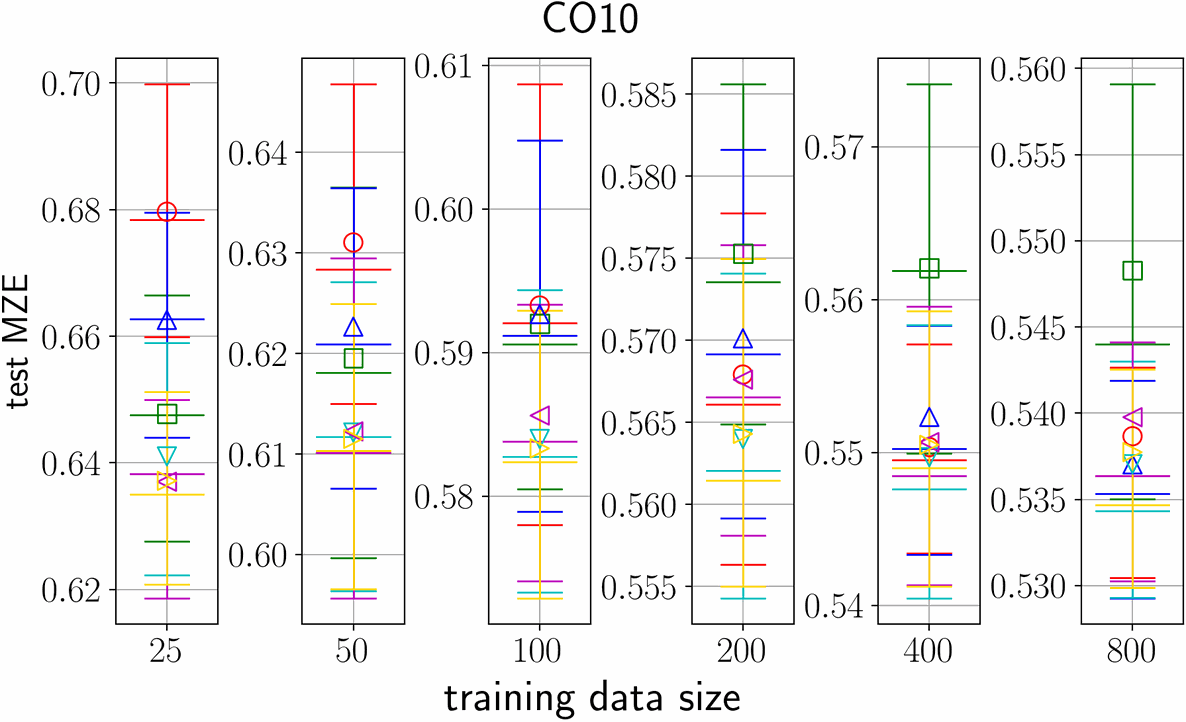}\\
\includegraphics[height=1.6cm]{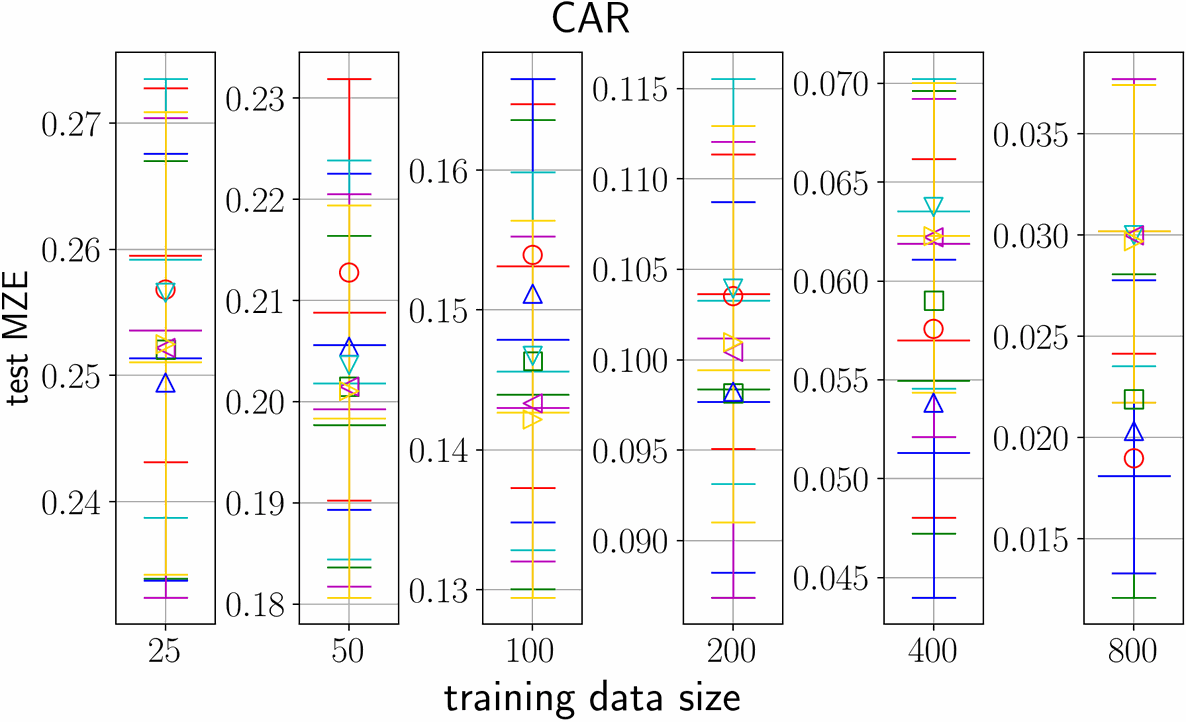}&
\includegraphics[height=1.6cm]{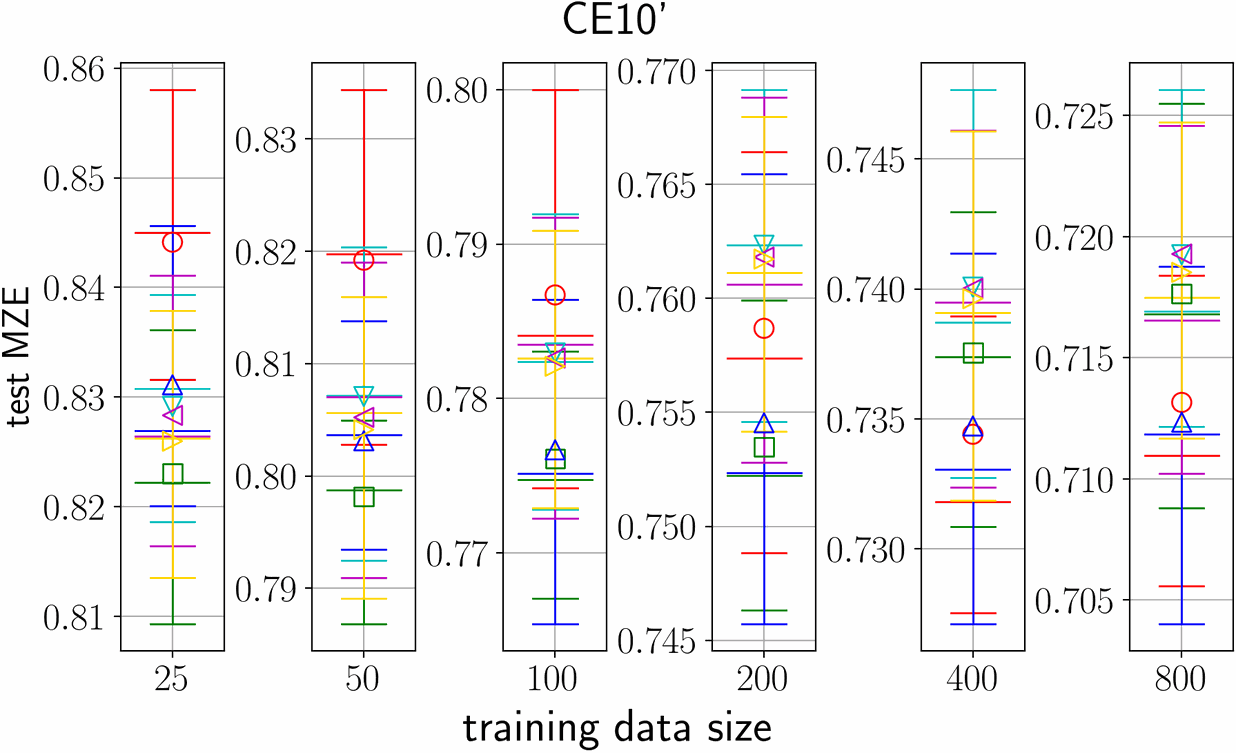}&
\includegraphics[height=1.6cm]{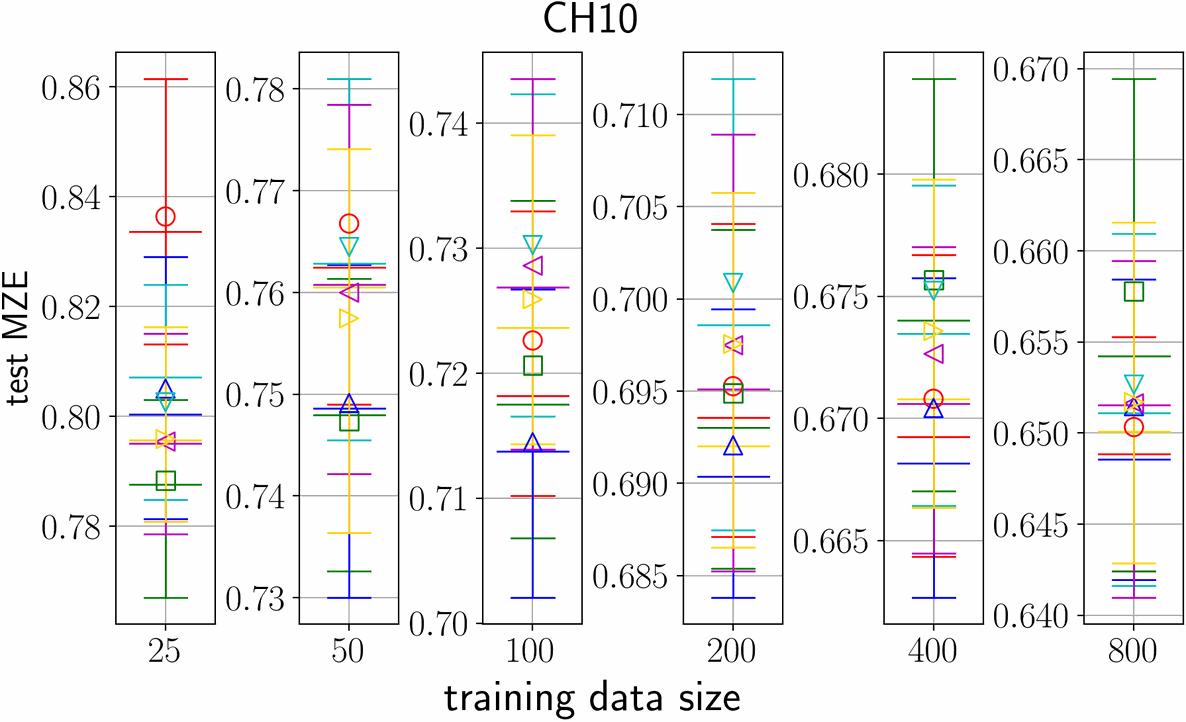}\\
\includegraphics[height=1.6cm]{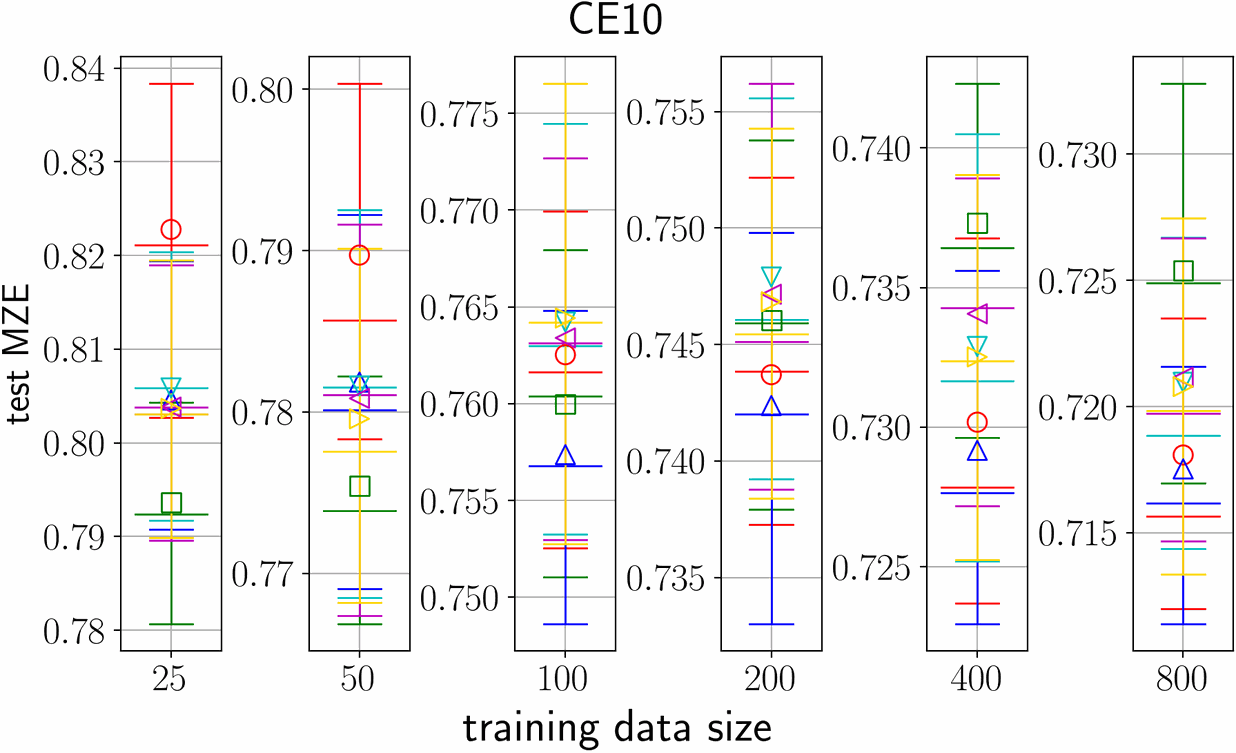}&
\includegraphics[height=1.6cm]{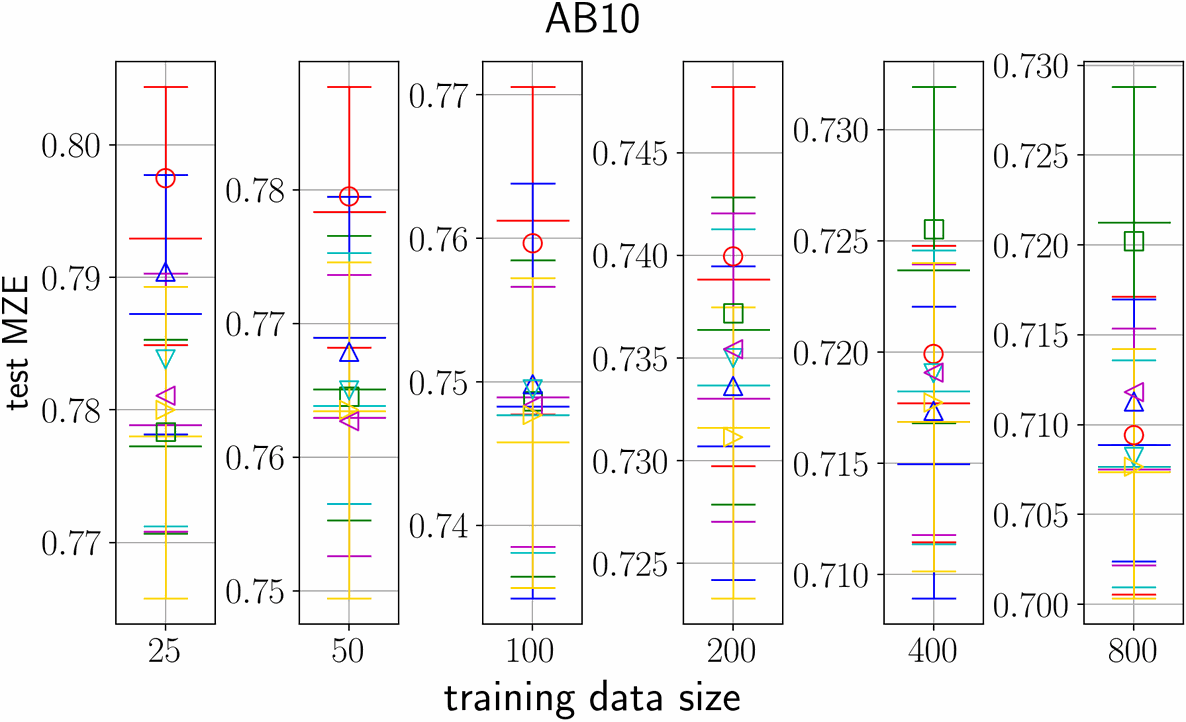}&
\includegraphics[height=1.6cm]{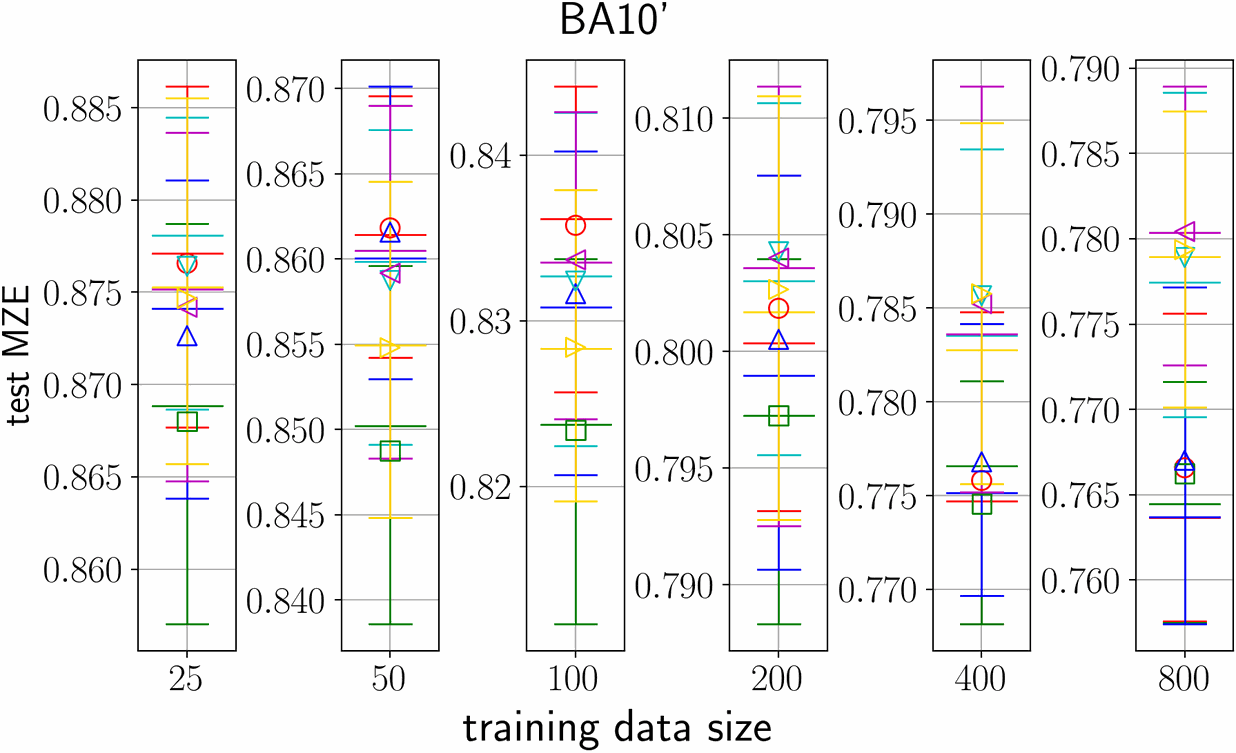}
\end{tabular}
\end{tabular}
\caption{%
Plot regarding 100-trial test NLLs (left) or MZEs (right)
by Nonr-MLR (red \protect$\circ$), Prev-MLR (green \protect$\Box$), 
Stri-MLR (blue \protect$\triangle$), Nonr-AUL (cyan \protect$\triangledown$), 
Prev-AUL (magenta \protect$\triangleleft$), and Stri-AUL (yellow \protect$\triangleright$).
Lower short, middle long, and upper short bars and marker 
represent 0.25, 0.5, and 0.75 quantiles and mean.}
\label{fig:Exp2-Performance-NLL}
\end{figure*}
%==========%
\begin{figure*}[!t]
\centering%
\renewcommand{\arraystretch}{1.5}%
\renewcommand{\tabcolsep}{4pt}%
\begin{tabular}{c|c}
{\scriptsize NLL}&{\scriptsize MZE}\\%
\begin{tabular}{cc}%
\begin{overpic}[width=4cm]{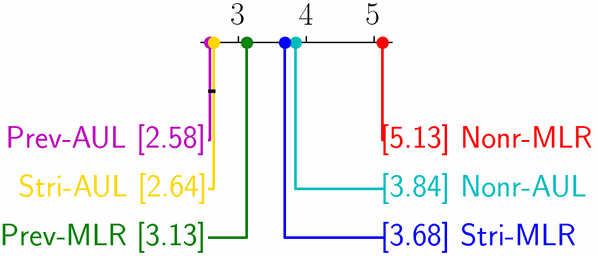}
\put(4,34){{\tiny$n_\tra\!=\!25$}}\end{overpic}&
\begin{overpic}[width=4cm]{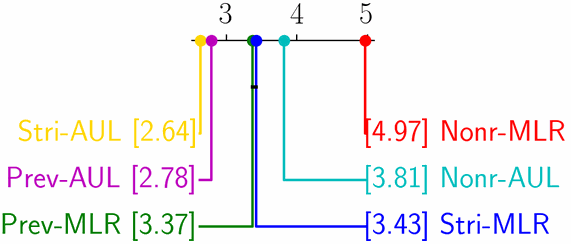}
\put(4,34){{\tiny$n_\tra\!=\!50$}}\end{overpic}\\
\begin{overpic}[width=4cm]{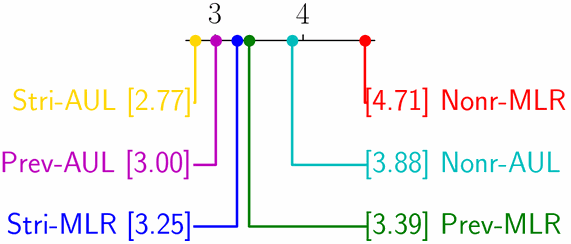}
\put(4,34){{\tiny$n_\tra\!=\!100$}}\end{overpic}&
\begin{overpic}[width=4cm]{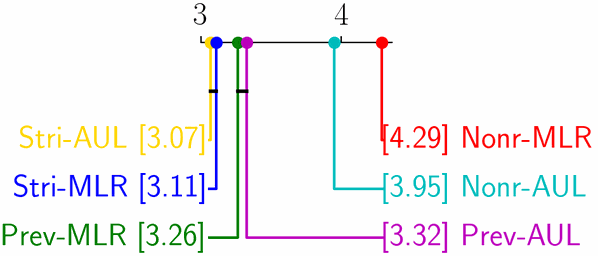}
\put(4,34){{\tiny$n_\tra\!=\!200$}}\end{overpic}\\
\begin{overpic}[width=4cm]{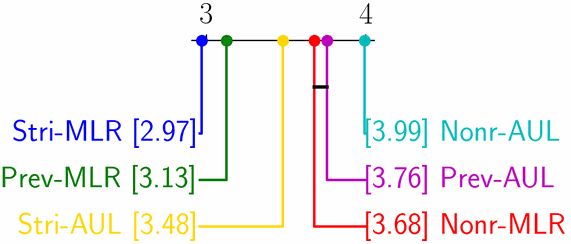}
\put(4,34){{\tiny$n_\tra\!=\!400$}}\end{overpic}&
\begin{overpic}[width=4cm]{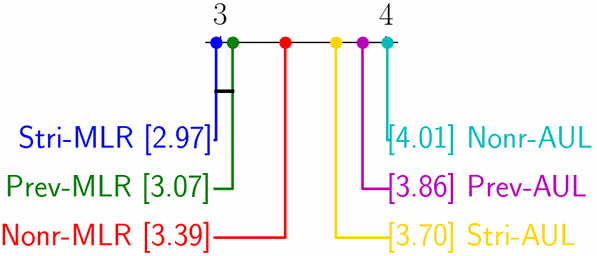}
\put(4,34){{\tiny$n_\tra\!=\!800$}}\end{overpic}\\
\end{tabular}&
\begin{tabular}{cc}%
\begin{overpic}[width=4cm]{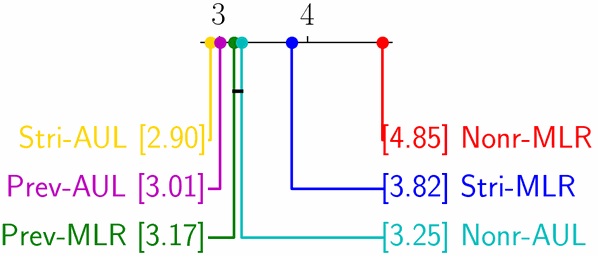}
\put(4,34){{\tiny$n_\tra\!=\!25$}}\end{overpic}&
\begin{overpic}[width=4cm]{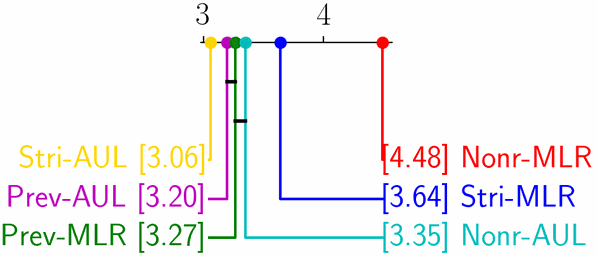}
\put(4,34){{\tiny$n_\tra\!=\!50$}}\end{overpic}\\
\begin{overpic}[width=4cm]{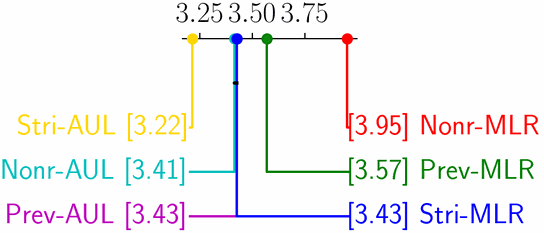}
\put(4,34){{\tiny$n_\tra\!=\!100$}}\end{overpic}&
\begin{overpic}[width=4cm]{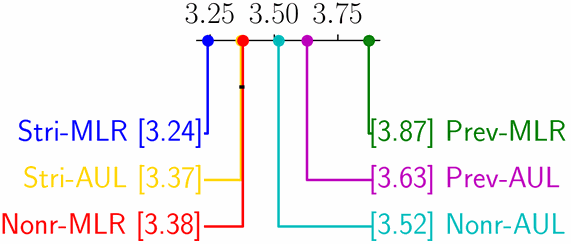}
\put(4,34){{\tiny$n_\tra\!=\!200$}}\end{overpic}\\
\begin{overpic}[width=4cm]{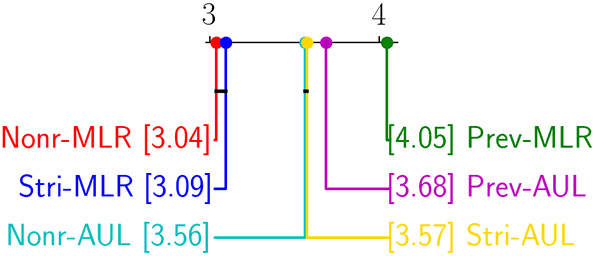}
\put(4,34){{\tiny$n_\tra\!=\!400$}}\end{overpic}&
\begin{overpic}[width=4cm]{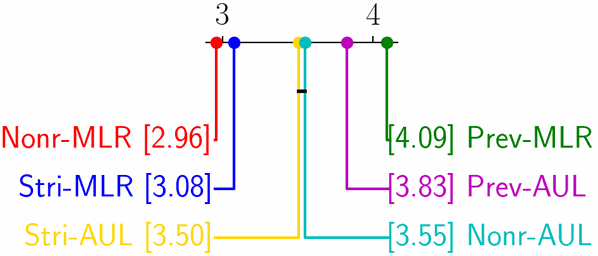}
\put(4,34){{\tiny$n_\tra\!=\!800$}}\end{overpic}\\
\end{tabular}
\end{tabular}
\caption{%
Critical difference diagram \cite{demsar06a} based on 
Conover-Friedman test \cite{conover1981rank} of 
significance level $0.05$ regarding the mean of test NLLs (left) or MZEs (right)
with $n_\tra=25,\ldots,800$.
$A[B]$ implies that the average rank of the mean test error of the method $A$ is $B$,
and the smaller the value of $B$, the better the corresponding method $A$ is.
Black horizontal line connecting methods implies that no significant difference was detected between them.}
\label{fig:CDD}
\end{figure*}
%==========%
\begin{table*}[!t]
\centering%
\renewcommand{\arraystretch}{0.75}%
\renewcommand{\tabcolsep}{2pt}%
\caption{%
A cell for each dataset and $n_\tra$ shows the best learning method 
(`${\rm n}$' in red if Nonr-MLR or cyan if Nonr-AUL, 
`${\rm p}$' in green if Prev-MLR or magenta if Prev-AUL, 
and `${\rm s}$' in blue if AStri-MLR or yellow if Stri-AUL) 
among methods with MLR or AUL models regarding the test NLLs (left) or MZEs (right)
(and methods that drew with the best method in the Mann-Whitney 
U-test with the significance level $0.05$ at the subscript if exists).}
\label{tab:Exp2-Performance-Summary2}
\begin{tabular}{ccc|cccccc|cccccc|cccccc|cccccc}
\multicolumn{3}{c}{}&\multicolumn{12}{c}{NLL}&\multicolumn{12}{c}{MZE}\\
\toprule
\multirow{2}{*}{dataset} & \multirow{2}{*}{$K$} & \multirow{2}{*}{MS} & 
\multicolumn{6}{c|}{for MLR with $n_\tra=$}& \multicolumn{6}{c|}{for AUL with $n_\tra=$}&
\multicolumn{6}{c|}{for MLR with $n_\tra=$}& \multicolumn{6}{c}{for AUL with $n_\tra=$}\\
&&&25&50&100&200&400&800&25&50&100&200&400&800&
25&50&100&200&400&800&25&50&100&200&400&800\\
\midrule
CAR & 4 & $.0146\!\pm\!{.0095}$ &
$\cs_{\cp}$&$\cp_{\cs}$&$\cp$&$\cp_{\cs}$&$\cs_{\cp}$&$\cp_{\cs}$&$\as_{\ap}$&$\as$&$\as_{\ap\an}$&$\as_{\ap\an}$&$\as_{\ap\an}$&$\ap_{\as\an}$&
$\cs_{\cp}$&$\cp_{\cs}$&$\cp_{\cs}$&$\cp_{\cs}$&$\cs$&$\cn_{\cs\cp}$&$\as_{\ap\an}$&$\as_{\ap\an}$&$\as_{\ap\an}$&$\ap_{\as\an}$&$\ap_{\as\an}$&$\as_{\an\ap}$\\
SWD & 4 & $.3980\!\pm\!{.0211}$ &
$\cs_{\cp}$&$\cs$&$\cs$&$\cs$&$\cs_{\cp\cn}$&$\cn_{\cs\cp}$&$\as_{\ap}$&$\ap_{\as\an}$&$\ap_{\as}$&$\ap_{\as\an}$&$\ap_{\an\as}$&$\an_{\as\ap}$&
$\cp$&$\cp$&$\cp_{\cs}$&$\cs_{\cn}$&$\cn_{\cs}$&$\cn_{\cs}$&$\ap_{\as\an}$&$\ap_{\as\an}$&$\as_{\ap\an}$&$\an_{\as}$&$\an_{\as}$&$\as_{\an}$\\
\midrule
BA5 & 5 & $.1890\!\pm\!{.0268}$ &
$\cp$&$\cp_{\cs}$&$\cp_{\cs\cn}$&$\cs_{\cn}$&$\cp_{\cs\cn}$&$\cs_{\cn\cp}$&$\as_{\ap}$&$\as_{\ap}$&$\as_{\ap\an}$&$\as_{\an\ap}$&$\ap_{\an\as}$&$\as_{\an\ap}$&
$\cp_{\cs}$&$\cs_{\cp}$&$\cn_{\cs}$&$\cn_{\cs}$&$\cn_{\cs}$&$\cn_{\cs}$&$\as_{\ap\an}$&$\as_{\an\ap}$&$\ap_{\as\an}$&$\an_{\ap\as}$&$\an_{\as\ap}$&$\an_{\as\ap}$\\
CO5' & 5 & $.2664\!\pm\!{.0357}$ &
$\cs$&$\cs_{\cp}$&$\cs_{\cn\cp}$&$\cs_{\cn\cp}$&$\cn_{\cp\cs}$&$\cn_{\cp\cs}$&$\as$&$\as$&$\as$&$\as_{\ap\an}$&$\as_{\an\ap}$&$\an_{\as\ap}$&
$\cp_{\cs}$&$\cp_{\cs\cn}$&$\cs_{\cn\cp}$&$\cn_{\cs}$&$\cn$&$\cn$&$\as_{\ap\an}$&$\as_{\ap\an}$&$\as_{\an}$&$\as_{\an\ap}$&$\an_{\as\ap}$&$\an_{\as}$\\
CO5 & 5 & $.3095\!\pm\!{.0329}$ &
$\cs$&$\cs$&$\cs_{\cp\cn}$&$\cs_{\cp\cn}$&$\cs_{\cp\cn}$&$\cn_{\cs\cp}$&$\as$&$\as$&$\as$&$\as_{\an\ap}$&$\as_{\an\ap}$&$\an_{\as\ap}$&
$\cp_{\cs}$&$\cn_{\cs\cp}$&$\cn_{\cs}$&$\cn_{\cs}$&$\cn_{\cs}$&$\cn_{\cs}$&$\as_{\ap\an}$&$\as_{\an\ap}$&$\as_{\an\ap}$&$\as_{\an}$&$\an_{\as}$&$\as_{\an}$\\
LEV & 5 & $.3685\!\pm\!{.0202}$ &
$\cs_{\cp}$&$\cs$&$\cs$&$\cs_{\cp}$&$\cs_{\cp\cn}$&$\cs_{\cn\cp}$&$\as_{\ap}$&$\ap_{\as\an}$&$\as_{\ap\an}$&$\ap_{\as\an}$&$\as_{\an\ap}$&$\ap_{\as\an}$&
$\cp$&$\cp_{\cs}$&$\cs_{\cn\cp}$&$\cs_{\cn}$&$\cs_{\cn}$&$\cs_{\cn}$&$\ap_{\as\an}$&$\ap_{\as\an}$&$\as_{\an\ap}$&$\as_{\an}$&$\an_{\as}$&$\an_{\as}$\\
CH5 & 5 & $.4172\!\pm\!{.0455}$ &
$\cp$&$\cp_{\cs}$&$\cs_{\cp}$&$\cp_{\cs}$&$\cs$&$\cs_{\cp}$&$\as$&$\as_{\ap}$&$\as_{\ap}$&$\as_{\ap}$&$\as_{\ap\an}$&$\as_{\ap\an}$&
$\cs_{\cp}$&$\cs$&$\cs_{\cn}$&$\cs_{\cn}$&$\cs_{\cn}$&$\cn_{\cs}$&$\ap_{\as\an}$&$\as_{\ap\an}$&$\as_{\an\ap}$&$\as_{\an\ap}$&$\ap_{\as\an}$&$\as_{\ap\an}$\\
CE5' & 5 & $.4643\!\pm\!{.0440}$ &
$\cp_{\cs}$&$\cp_{\cs}$&$\cp_{\cs}$&$\cs_{\cp\cn}$&$\cs_{\cp\cn}$&$\cn_{\cp\cs}$&$\as$&$\as$&$\as$&$\as$&$\as_{\an\ap}$&$\as_{\ap\an}$&
$\cp$&$\cp_{\cs}$&$\cs_{\cp\cn}$&$\cn_{\cs}$&$\cn_{\cs}$&$\cs_{\cn}$&$\as_{\an\ap}$&$\as_{\ap\an}$&$\an_{\as\ap}$&$\an_{\as\ap}$&$\an_{\as\ap}$&$\as_{\an\ap}$\\
CE5 & 5 & $.4760\!\pm\!{.0460}$ &
$\cp_{\cs}$&$\cp_{\cs}$&$\cs_{\cp}$&$\cs_{\cp\cn}$&$\cs_{\cn\cp}$&$\cs$&$\as$&$\as_{\ap}$&$\as_{\ap}$&$\as_{\ap\an}$&$\as_{\ap\an}$&$\as_{\ap\an}$&
$\cp$&$\cp_{\cs}$&$\cs_{\cn}$&$\cs_{\cn}$&$\cn_{\cs}$&$\cs_{\cn}$&$\ap_{\as\an}$&$\as_{\an\ap}$&$\ap_{\as\an}$&$\an_{\as\ap}$&$\as_{\ap\an}$&$\an_{\as\ap}$\\
AB5 & 5 & $.4873\!\pm\!{.0324}$ &
$\cp$&$\cs_{\cp\cn}$&$\cs$&$\cs$&$\cs_{\cp}$&$\cs_{\cp}$&$\as_{\ap}$&$\as$&$\as_{\ap}$&$\as_{\ap}$&$\as_{\ap}$&$\as_{\ap\an}$&
$\cp_{\cs}$&$\cp_{\cs}$&$\cs_{\cp}$&$\cs$&$\cs_{\cn}$&$\cn_{\cs}$&$\as_{\an\ap}$&$\ap_{\as\an}$&$\as_{\an\ap}$&$\as_{\an\ap}$&$\as_{\an}$&$\an_{\as}$\\
BA5' & 5 & $.5524\!\pm\!{.0373}$ &
$\cp$&$\cp$&$\cp$&$\cp$&$\cp_{\cs\cn}$&$\cs$&$\ap$&$\ap$&$\ap$&$\as_{\ap}$&$\as$&$\as$&
$\cp$&$\cp$&$\cp_{\cs}$&$\cs_{\cp\cn}$&$\cn$&$\cn_{\cs}$&$\ap_{\as\an}$&$\as_{\ap\an}$&$\as_{\an\ap}$&$\as_{\an\ap}$&$\as_{\an\ap}$&$\as_{\an\ap}$\\
\midrule
WQR & 6 & $.3678\!\pm\!{.0260}$ &
$\cs$&$\cs$&$\cs$&$\cs$&$\cs$&$\cs_{\cn}$&$\as_{\an\ap}$&$\as$&$\as_{\ap\an}$&$\as_{\ap\an}$&$\as_{\an\ap}$&$\an_{\as\ap}$&
$\cs_{\cp}$&$\cs_{\cp}$&$\cs_{\cn}$&$\cs_{\cn}$&$\cs_{\cn}$&$\cs_{\cn}$&$\as_{\an\ap}$&$\an_{\as\ap}$&$\as_{\an\ap}$&$\as_{\an\ap}$&$\as_{\an\ap}$&$\as_{\an\ap}$\\
\midrule
ERA & 9 & $.7208\!\pm\!{.0198}$ &
$\cp$&$\cs_{\cp}$&$\cs$&$\cp_{\cs}$&$\cs_{\cp\cn}$&$\cs_{\cp\cn}$&$\as_{\ap}$&$\as_{\ap}$&$\as_{\ap}$&$\as_{\ap}$&$\as_{\ap\an}$&$\as_{\ap\an}$&
$\cp$&$\cp$&$\cp$&$\cs_{\cn\cp}$&$\cp_{\cn\cs}$&$\cn_{\cs}$&$\an_{\ap\as}$&$\ap_{\as\an}$&$\as_{\ap\an}$&$\ap_{\an\as}$&$\ap_{\an\as}$&$\as_{\an\ap}$\\
\midrule
BA10 & 10 & $.3815\!\pm\!{.0287}$ &
$\cp$&$\cs_{\cp}$&$\cs_{\cp}$&$\cp_{\cs\cn}$&$\cp_{\cn\cs}$&$\cp_{\cn\cs}$&$\as_{\ap}$&$\ap_{\as}$&$\as_{\ap\an}$&$\as_{\ap\an}$&$\as_{\ap\an}$&$\ap_{\as\an}$&
$\cp$&$\cs_{\cp}$&$\cs$&$\cn_{\cs}$&$\cn_{\cs}$&$\cn$&$\as$&$\as_{\ap\an}$&$\as_{\an\ap}$&$\as_{\ap\an}$&$\an_{\ap\as}$&$\an_{\as\ap}$\\
CO10' & 10 & $.4720\!\pm\!{.0369}$ &
$\cp_{\cs}$&$\cs_{\cp}$&$\cs_{\cp}$&$\cs_{\cp\cn}$&$\cp_{\cn\cs}$&$\cn_{\cp\cs}$&$\as$&$\as$&$\as$&$\as_{\ap\an}$&$\as_{\an\ap}$&$\an_{\as\ap}$&
$\cp$&$\cp$&$\cs_{\cp\cn}$&$\cn$&$\cn_{\cs}$&$\cn_{\cs}$&$\as_{\ap\an}$&$\ap_{\as\an}$&$\as_{\an\ap}$&$\ap_{\as\an}$&$\an_{\as\ap}$&$\an_{\as\ap}$\\
CO10 & 10 & $.5031\!\pm\!{.0340}$ &
$\cp_{\cs}$&$\cs_{\cp}$&$\cs_{\cp}$&$\cs_{\cp}$&$\cp_{\cn\cs}$&$\cn_{\cp\cs}$&$\as_{\ap}$&$\as$&$\as_{\ap}$&$\as_{\ap\an}$&$\ap_{\as\an}$&$\as_{\ap\an}$&
$\cp$&$\cp_{\cs}$&$\cp_{\cs\cn}$&$\cn_{\cs}$&$\cn_{\cs}$&$\cs_{\cn}$&$\as_{\ap\an}$&$\as_{\an\ap}$&$\as_{\an\ap}$&$\an_{\as}$&$\an_{\as\ap}$&$\an_{\as\ap}$\\
CH10 & 10 & $.6238\!\pm\!{.0466}$ &
$\cp$&$\cp$&$\cp_{\cs}$&$\cs_{\cp}$&$\cs_{\cp}$&$\cs_{\cp}$&$\ap$&$\ap_{\as}$&$\as_{\ap}$&$\as_{\ap}$&$\as_{\ap}$&$\ap_{\as}$&
$\cp$&$\cs_{\cp}$&$\cs$&$\cs_{\cp}$&$\cs_{\cn}$&$\cn_{\cs}$&$\ap_{\as}$&$\as_{\ap}$&$\as_{\ap\an}$&$\as_{\ap\an}$&$\ap_{\as\an}$&$\as_{\ap\an}$\\
AB10 & 10 & $.6847\!\pm\!{.0239}$ &
$\cp$&$\cp$&$\cs$&$\cs$&$\cs$&$\cs_{\cp}$&$\ap$&$\as$&$\as$&$\as_{\ap}$&$\as_{\ap}$&$\ap_{\as\an}$&
$\cp$&$\cp_{\cs}$&$\cp_{\cs}$&$\cs$&$\cs$&$\cn_{\cs}$&$\as_{\ap\an}$&$\ap_{\as\an}$&$\as_{\ap\an}$&$\as_{\ap}$&$\as_{\ap\an}$&$\as_{\an\ap}$\\
CE10' & 10 & $.6898\!\pm\!{.0409}$ &
$\cp$&$\cp$&$\cp$&$\cp_{\cs}$&$\cp_{\cs}$&$\cp$&$\ap$&$\ap$&$\ap_{\as}$&$\as$&$\as_{\ap}$&$\as_{\ap}$&
$\cp$&$\cp$&$\cp_{\cs}$&$\cp_{\cs}$&$\cn_{\cs}$&$\cs_{\cn}$&$\as_{\ap\an}$&$\as_{\ap\an}$&$\as_{\an\ap}$&$\ap_{\as\an}$&$\as_{\an\ap}$&$\as_{\ap\an}$\\
CE10 & 10 & $.6911\!\pm\!{.0313}$ &
$\cp$&$\cp$&$\cp_{\cs}$&$\cp_{\cs}$&$\cs_{\cp\cn}$&$\cs_{\cp}$&$\as_{\ap}$&$\ap_{\as}$&$\as_{\ap}$&$\as_{\ap}$&$\as_{\ap}$&$\as_{\ap\an}$&
$\cp$&$\cp$&$\cs$&$\cs_{\cn}$&$\cs_{\cn}$&$\cs_{\cn}$&$\as_{\ap\an}$&$\as_{\ap\an}$&$\ap_{\an\as}$&$\as_{\ap\an}$&$\as_{\an\ap}$&$\as_{\an\ap}$\\
BA10' & 10 & $.7531\!\pm\!{.0182}$ &
$\cp$&$\cp$&$\cp$&$\cp$&$\cs_{\cp}$&$\cs_{\cp}$&$\ap$&$\ap$&$\ap$&$\ap$&$\as$&$\as$&
$\cp$&$\cp$&$\cp$&$\cp_{\cs}$&$\cp_{\cn\cs}$&$\cn_{\cp\cs}$&$\ap_{\as\an}$&$\as$&$\as$&$\as_{\ap\an}$&$\ap_{\an\as}$&$\an_{\as\ap}$\\
\midrule
\multicolumn{3}{c|}{\#win of n,p,s}&
{\tiny0,\tcg{15},6}&{\tiny0,\tcg{11},10}&{\tiny0,8,\tcb{13}}&{\tiny0,8,\tcb{13}}&{\tiny1,6,\tcb{14}}&{\tiny6,3,\tcb{12}}&{\tiny0,5,\tcy{16}}&{\tiny0,8,\tcy{13}}&{\tiny0,4,\tcy{17}}&{\tiny0,3,\tcy{18}}&{\tiny0,3,\tcy{18}}&{\tiny5,5,\tcy{11}}&
{\tiny0,\tcg{18},3}&{\tiny1,\tcg{15},5}&{\tiny2,8,\tcb{11}}&{\tiny7,3,\tcb{11}}&{\tiny\tcr{11},2,8}&{\tiny\tcr{14},0,7}&{\tiny1,7,\tcy{13}}&{\tiny1,6,\tcy{14}}&{\tiny1,3,\tcy{17}}&{\tiny5,4,\tcy{12}}&{\tiny\tcc{9},5,7}&{\tiny9,0,\tcy{12}}\\
\bottomrule\end{tabular}
\end{table*}

%=======================================%
\textbf{Results (Prediction performance):}
%==========%
Figure~\ref{fig:Exp2-Performance-NLL} shows plots regarding the NLL and MZE, 
and Figure~\ref{fig:CDD} shows the test-based post-hoc comparison of 
the overall performance across all the datasets 
to see the effectiveness of proposed methods.
Table~\ref{tab:Exp2-Performance-Summary2} shows 
a detailed comparison of the NLL and MZE for each dataset.
Note that, in Table~\ref{tab:Exp2-Performance-Summary2}, 
we report experimental evaluation of the mean scale (MS) 
$\bbE_{\bx\sim\bX}[1-\max_y\Pr(Y=y|\bX=\bx)]$, 
which was estimated following 
the same procedure as in Section~\ref{sec:Data}, as well.
The results for the MAE and MSE were similar to those for the NLL and MZE, 
and some figures and tables are provided in the supplement.

%==========%
In Figure~\ref{fig:CDD}, 
Prev-AUL was the best for NLL with $n_\tra=25$,
Stri-AUL was the best for NLL with $n_\tra=50,100,200$ and MZE with $n_\tra=25,50,100$, 
Stri-MLR was the best for NLL with $n_\tra=400,800$ and MZE with $n_\tra=200$, and
Nonr-MLR was the best for MZE with $n_\tra=400,800$.
This result indicates as overall trends for real-world ordinal data 
that the AUL model and MLR model are respectively more powerful 
with smaller-size training data and larger-size training data, 
that UPRL was effective when the training data size is not large enough,
and that previous UPRL and strict UPRL are respectively more compatible
with smaller-size training data and larger-size training data for each model.
These observations suggest that
in addition to the unimodality-promotion by UPRL, 
the smoothness-promotion of previous UPRL contributes to 
improving the prediction performance with the small training data size, 
while the bias reduction by the strict UPRL methods or non-regularized 
learning methods were effective as the training data size increases.
A more detailed look at the comparison in 
Table~\ref{tab:Exp2-Performance-Summary2} reveals 
that when the training data size is small,
the previous UPRL methods are effective for larger-scale data, 
whereas the strict UPRL methods more effective for smaller-scale data.
To summarize, the combination of either previous or strict UPRL and
the unimodality-driven AUL model is effective for small-size training data 
depending on the scale of the data, 
and the strict UPRL or non-regularized learning methods for 
the unrestricted MLR model are more promising for large-size training data 
because these methods reduce the scale-related bias associated with the previous UPRL methods.
This result can be understood from the bias-variance trade-off, 
the unimodality-promotion of UPRL, and 
the smoothness-promotion of previous UPRL.

%========================================%
\section{Conclusion}
\label{sec:Conclusion}
%==========%
Many ordinal data are high-UR and low-UD 
\cite{yamasaki2022unimodal, yamasaki2025approximately}, 
and hence UPRL methods are promising for OR tasks.
In this paper, we showed that previous UPRL methods 
\cite{belharbi2020,albuquerque2021ordinal,albuquerque2022quasi,Cardoso2025,kim2025cal} 
not only promote a predicted CPD closer to be unimodal, 
but also induce a bias promoting a smoother predicted CPD.
Therefore, we developed another UPRL method aiming 
for achieving the unimodality-promotion, but not inducing 
a bias into the predicted CPD when the underlying CPD is unimodal.
Through experimental comparison between non-regularized 
learning methods and proposed strict UPRL methods, 
we verified that the unimodality-promotion indeed contributes to 
improve the prediction performance with smaller-size training data
for many real-world ordinal data of the unimodality.
Furthermore, as suggested by the smoothness-promotion of the previous UPRL methods, 
the previous UPRL methods performed better for larger scale data
when the training data size is small, and 
the proposed UPRL methods could outperform previous UPRL methods 
for smaller-scale data or when the training data size becomes larger.
If the available data size is not large enough to favor non-regularized learning methods, 
these UPRL methods offer viable options, 
with the choice between them being 
dependent on the data scale and training data size.

%==========%
However, high computational cost is a drawback of 
the proposed UPRL method as we verify in supplement.
It is important to develop computationally efficient UPRL methods, 
but, even in such acceleration studies, statistically reasonable design of 
the regularizer remains crucial, and our analysis may serve as a guide.
Also, adjusting the balance between the strength of the unimodality- and 
smoothness-promotion may lead to further improvement in the prediction performance.
While the balance was constant in the previous UPRL method, 
the proposed strict UPRL method emphasizes the unimodality-promotion, 
so it may be possible to adjust the balance, for example, 
by the weighted combination of the previous and strict methods.
Since it would require an additional weight hyperparameter, development of 
computationally efficient strict UPRL methods is crucial for this research as well.
We consider that these are promising directions for a future research.

%========================================%
%========================================%
\clearpage
\section*{Supplementary Material}
%==========%
\renewcommand{\thesection}{S\Roman{section}}
\renewcommand{\theequation}{S\arabic{equation}}
\renewcommand{\thetable}{S\Roman{table}}
\renewcommand{\thefigure}{S\arabic{figure}}
\setcounter{section}{0}
\setcounter{equation}{0}
\setcounter{table}{0}
\setcounter{figure}{0}
%========================================%
\section{Supplementary Simulation}
\label{sec:Simulation}
%==========%
As we described in Section~\ref{sec:PrevAnaly} and \ref{sec:StriAnaly},
we performed additional simulation to study behaviors of UPRL methods with 
$K=3,4,5,6,7$, various unimodal $\bp$'s, and $\delta=0,0.05, 0.1, 0.2, 0.4$.
We show results of those simulations in Table~\ref{tab:UPRL31}--\ref{tab:UPRL71},
and got the following observations:
\begin{itemize}
\item%
The correlation coefficient between 
the expected UPRL loss $(\omega_{\prev,\delta}(\bp,\bq_i))_{i\in[10^3]}$ 
and the UD $(D_\rmH(\bq_i,\hat{\Delta}_{K-1}))_{i\in[10^3]}$ is moderately positive.
This indicates that decreasing the value of the regularizer \eqref{eq:alb21} 
leads to decreasing the UD of a predicted CPD $\bQ(\bg(\bx))$.
Thus, the previous UPRL method based on the regularizer \eqref{eq:alb21} 
would achieve the unimodality-promotion to some extent.
\item%
The correlation coefficient between 
the expected UPRL loss $(\omega_{\prev,\delta}(\bp,\bq_i))_{i\in[10^3]}$ and 
the scale $(1-\max_{j\in[K]}q_{i,j}))_{i\in[10^3]}$ is an unignorable negative value.
In other words, we can see that reducing the value of the regularizer 
\eqref{eq:alb21} is thought to have the smoothness-promotion.
\item%
By enlarging $\delta$, regularization of previous methods with a fixed 
$\lambda$ seems to weaken the unimodality- and smoothness-promotion
(the absolute value of the correlation coefficient decreases as $\delta$ increases).
\item%
The correlation coefficient between 
the expected UPRL loss $(\omega_\stri(\bp,\bq_i))_{i\in[10^3]}$ and 
the UD $(D_\rmH(\bq_i,\hat{\Delta}_{K-1}))_{i\in[10^3]}$ is of course 1, 
and hence the proposed UPRL method would achieve the unimodality-promotion.
\item%
Although the correlation coefficient between the 
expected UPRL loss $(\omega_\stri(\bp,\bq_i))_{i\in[10^3]}$ 
and the scale $(1-\max_{j\in[5]}q_{i,j})_{i\in[10^3]}$ is not 0, 
but its absolute value is much smaller than those of the previous UPRL methods.
Therefore, we expect that the proposed strict UPRL method will 
reduce a scale-related bias compared with previous UPRL methods.
\end{itemize}

%========================================%
\section{Supplementary Explanation of Ordinal Data}
\label{sec:Datasets}
%==========%
We here write supplementary explanation about the
ordinal data that we used in experiments of this study.

%==========%
The AB5 (5-class abalone) and AB10 (10-class abalone) datasets 
were generated from abalone dataset that was accessible at 
\url{https://archive.ics.uci.edu/dataset/1/abalone} \cite{nash1994population}
in UCI machine learning repository \cite{Dua2019}.
Its 10-dimensional explanatory variable has features,
`sex' in \{`male', `female', `infant'\} encoded with the one-hod encoding, 
`length of longest shell', 
`length of shell in perpendicular to longest shell', 
`height with meat in shell', 
`whole weight', 
`shucked weight (i.e., weight of meat)', 
`viscera weight (i.e., gut weight measured after bleeding)', 
and `shell weight measured after being dried' of an abalone.
Also, its target variable is `the number of rings' 
that reflects the age of an abalone.

%==========%
The BA5 (5-class bank) and BA10 (10-class bank) datasets were generated from bank8FM dataset, 
and the BA5' (5-class bank') and BA10' (10-class bank') datasets were generated from bank32NH dataset.
Both of the bank8FM and bank32NH datasets could be obtained 
from \url{https://www.dcc.fc.up.pt/~ltorgo/Regression/bank.html}
or \url{https://www.cs.toronto.edu/~delve/data/bank/desc.html}.
These datasets are synthetically generated from a simplistic simulator, 
which simulates queues in a series of banks.
There, customers come from several residential areas, 
choose their preferred bank depending on distances, 
and have tasks of varying complexity and various levels of patience,
each bank has several queues that open and close according to demand, 
the tellers have various effectivities, 
and customers may change queue if their patience expires. 
For the bank8FM and bank32NH datasets, 
8- and 32-dimensional features of their explanatory variable
are about bank, customers, tellers, and their relationship.
Also, their target variable is the rate of rejections, i.e., the fraction of customers 
that are turned away from the bank because all the open tellers have full queues.

%==========%
The CE5 (5-class census) and CE10 (10-class census) datasets were generated from House(8L) dataset, 
and the CE5' (5-class census') and CE10' (10-class census') datasets were generated from House(16H) dataset.
Both of these original datasets were accessible at 
\url{https://www.dcc.fc.up.pt/~ltorgo/Regression/census.html} or 
\url{https://www.cs.toronto.edu/~delve/data/census-house/desc.html}.
These datasets collect a part of the 1990 US census by the US Census Bureau. 
Their explanatory variable is consisted of 8- or 16-dimensional features 
concerning demographic composition and state of housing market in the region,
and their target variable is the median price of the house in the region.

%==========%
The CH5 (5-class California housing) and CH10 (10-class California housing) 
datasets were generated from a so-called California housing dataset;
refer to \url{https://lib.stat.cmu.edu/datasets/houses.zip} in StatLib repository
and to \cite{pace1997sparse}.
The California housing dataset collects information on 
all the block groups in California from the 1990 census. 
The features, 
`median income', `housing median age', 
`total rooms', `total bedrooms', `population', 
`households', `latitude', and `longitude',
are set to the 8-dimensional explanatory variable,
and `median house value' is set to the target variable.

%==========%
The CO5 (5-class computer-activity) and CO10 (10-class computer-activity) datasets were generated from CompAct dataset, 
and the CO5' (5-class computer-activity') and CO10' (10-class computer-activity') datasets were generated from CompAct(s) dataset,
both of which could be obtained from 
\url{https://www.dcc.fc.up.pt/~ltorgo/Regression/comp.html} or 
\url{https://www.cs.toronto.edu/~delve/data/comp-activ/desc.html}.
These datasets are collections of computer systems activity measures,
which were measured from a Sun SPARCstation 20/712 with 
128 Mbytes of memory running in a multi-user university department.
Users would typically be doing various tasks ranging from accessing 
the Internet, editing files, or running high CPU-bound programs.
The CompAct dataset has 21-dimensional features consisting of 
`reads (transfers per second (/s)) between system memory and user memory',
`writes (transfers /s) between system memory and user memory',
`the number of (\#) system calls of all types /s',
`\# system read calls /s',
`\# system write calls /s',
`\# system fork calls /s',
`\# system exec calls /s',
`\# characters transferred /s by system read calls',
`\# characters transferred /s by system write calls',
`\# page out requests /s',
`\# pages, paged out /s',
`\# pages /s placed on the free list',
`\# pages checked if they can be freed /s',
`\# page attaches (satisfying a page fault by reclaiming a page in memory) /s',
`\# page-in requests /s',
`\# pages paged in /s',
`\# page faults caused by protection errors (copy-on-writes)',
`\# page faults caused by address translation',
`process run queue size',
`\# memory pages available to user processes',
and `\# disk blocks available for page swapping'
as the explanatory variable.
The CompAct(s) dataset has 12-dimensional features 
excluding the paging information as the explanatory variable.
The target variable is 
`the number portion of time (\%) that CPUs run in user mode'.

%==========%
The CAR (car evaluation) dataset was accessible at
\url{https://archive.ics.uci.edu/dataset/19/car+evaluation}.
This dataset was derived from a simple hierarchical decision model 
originally developed for the demonstration of \cite{bohanec1990dex}.
Its explanatory variable is consisted of 
`buying price' in \{`low', `medium', `high', `very high'\}, 
`price of the maintenance' in \{`low', `medium', `high', `very high'\}, 
`number of doors' in \{`2', `3', `4', `5 or more'\},
`capacity in terms of persons to carry' in \{`2', `4', `5 or more'\},
`the size of luggage boot' in \{`small', `medium', `big'\},
`estimated safety of the car' in \{`low', `medium', `high'\} of a car.
\cite{gutierrez2015ordinal} encoded these features with the one-hot encoding, 
and hence the dimension of explanatory variables is 21.
Also, its 4-class target variable is `car acceptability' in 
\{`unacceptable', `acceptable', `good', `very good'\} of a car.

%==========%
The ERA, LEV, and SWD datasets were available from Weka repository \cite{frank2016weka}
or \url{https://prdownloads.sourceforge.net/weka/datasets-arie_ben_david.tar.gz}.

%==========%
The ERA (employee rejection/acceptance) dataset could also be obtained from
\url{https://www.openml.org/search?type=data&status=active&id=1030}.
This dataset was originally gathered in an academic decision-making experiment,
which aims to determine which are the most important qualities of candidates for a certain type of jobs,
during a MBA academic course. 
Its 4-dimensional explanatory variable is consisted of 
features of a candidates such as past experience, verbal skills, etc.,
and its 9-class target variable is the subjective judgment of a decision-maker to which degree 
he or she tends to accept the applicant to the job or to reject him altogether 
(the lowest score means total tendency to reject an applicant and vice versa).

%==========%
The LEV (lectures evaluation) dataset was also accessible at
\url{https://www.openml.org/search?type=data&status=active&id=1029}.
The LEV dataset contains anonymous lecturer evaluations taken at the end of MBA courses. 
Before receiving the final grades, students were asked to score their lecturers according to 
four attributes such as oral skills and contribution to their professional/general knowledge. 
Those 4 scores are adopted as features of the explanatory variable.
Also, the 5-class target variable of this dataset is 
a total evaluation of the lecturer's performance.

%==========%
The SWD (social workers decisions) dataset was also obtained from
\url{https://www.openml.org/search?type=data&status=any&id=1028}.
The SWD dataset contains real-world assessments of qualified social workers 
regarding the risk facing children if they stayed with their families at home. 
This risk assessment is often presented to judicial courts to help 
decide what is in the best interest of an alleged abused or neglected child.
Note that this dataset is often used in OR studies but 
we could not find a detailed description of the features.

%==========%
The WQR (wine quality red) dataset is a red wine version of the wine quality dataset 
obtained from \url{https://archive.ics.uci.edu/dataset/186/wine+quality};
refer to also \cite{cortez2009modeling}.
This dataset is related to red variants of the Portuguese `Vinho Verde' wine. 
Its 11-dimensional explanatory variable is consisted of `fixed acidity', 
`volatile acidity', `citric acid', `residual sugar', `chlorides', 
`free sulfur dioxide', `total sulfur dioxide', `density', `pH', 
`sulphates', `alcohol' of a wine based on physicochemical tests,
and its 6-class target variable is `quality' of a wine based on sensory data.

%==========%
We used all the datasets with standardizing the inputs.

%========================================%
\section{Supplementary Experimental Results}
\label{sec:Results}
%==========%
We also experimented with other OR methods: 
non-regularized learning methods for 
cumulative logit (CL) model \cite{agresti2010analysis},
ordinal logistic regression (proportional-odds (PO)-constrained CL; POCL) model 
\cite{mccullagh1980regression,agresti2010analysis,cao2020rank}, 
UL model \cite{yamasaki2022unimodal}, and 
PO-constrained UL (POUL) model \cite{yamasaki2022unimodal}.
CL model is consisted of
\begin{align}
\label{eq:CL}
	\begin{split}
	&Q_{\cl}(\bu,y)
\\
	&\coloneq
	\begin{cases}
	\frac{1}{1+e^{-u_1}}&\text{for }y=1,\\
	1-\frac{1}{1+e^{-u_{K-1}}}&\text{for }y=K,\\
	\frac{1}{1+e^{-u_y}}-\frac{1}{1+e^{-u_{y-1}}}&\text{for }y=2,\ldots,K-1,
	\end{cases}
	\end{split}
\end{align}
and a learner class in $\{\rho[\bg]\mid\bg\in\calG_{K-1}\}$ 
with a non-negative function $\rho$ satisfying 
\eqref{eq:RHO} to ensure $\bQ_{\cl}(\bu)\in\Delta_{K-1}$.
POCL model is representative in the OR study, and it adopts the function \eqref{eq:CL} 
and an ordered PO-constrained learner class in 
$\calG_{\text{ord-po},K-1}\coloneq\{(\acute{b}_k-a(\cdot))_{k\in[K-1]}
\mid a:\bbR^d\to\bbR,\,(b_k)_{k\in[K-1]}\in\bbR^{K-1},\,
(\acute{b}_k)_{k\in[K-1]}=\rho[(b_k)_{k\in[K-1]}]\}$.
In addition, the previous work \cite{yamasaki2022unimodal}
studied various PO-constrained models, such as POUL model
$(\bQ_\smax(-\tau(\cdot)),\calG_{\text{ord-po},K})$.
UL and POUL models are ensured to yield a unimodal likelihood,
but CL and POCL models are not.
Also, POCL and POUL models have lower representation ability 
respectively than CL and UL models.
These methods are denoted as 
Nonr-CL, Nonr-POCL, Nonr-UL, and Nonr-POUL.

%==========%
Figures~\ref{fig:Performance-BestLam-NLL}, \ref{fig:Performance-BestLam-MZE},
\ref{fig:Performance-BestLam-MAE}, and \ref{fig:Performance-BestLam-MSE} 
are augmented versions of Figure~\ref{fig:Exp2-Performance-NLL} in the main paper,
and respectively show the comparison regarding the NLL, MZE, MAE, and MSE.
Figure~\ref{fig:CDD-S} is an augmented version of Figure~\ref{fig:CDD} in the main paper,
and shows test-based post-hoc comparison of the overall performance across all the datasets.
Table~\ref{tab:best-Performance} presents NLL, MZE, MAE, and MSE version of 
Table~\ref{tab:Exp2-Performance-Summary2} in the main paper.

%==========%
Nonr-CL with the unrestricted CL model showed 
performance similar to that of the Nonr-MLR
with the unrestricted MLR model. 
Nonr-UL underperformed especially for larger-size training data,
owing to the unimodality constraint.
The PO constraint strongly restricts 
the representation ability of the likelihood model.
Presumably because of such a model constraint, 
the success or failure of PO-constrained models depended 
heavily on their compatibility with the data distribution,
and their overall performance was poor 
particularly when the training data size was large.

%========================================%
\section{Computation Cost Comparison}
\label{sec:Experiments3}
%==========%
\textbf{Purposes:}
We are also interested in computation cost of the strict UPRL method.
In this section, we investigate it experimentally.

%=======================================%
\textbf{Settings:}
%==========%
Computational costs for the methods we consider 
are proportional to the training data size $n_\tra$.
We thus calculated the computation time only for the BA10' dataset 
of the size $n_\tra=800$ with the highest $d$ and $K$ here.
We experimented under the computation environment with 
Python 3.12.3, Numpy 2.3.5, PyTorch 2.10.0, CVXOPT 1.3.2, 
and one thread of CPU AMD Ryzen 9 9950X.
Note that the computational cost of the strict UPRL methods, 
inner optimization of \eqref{eq:OurUPRL}, gets higher for larger $K$,
and that the computation time may be improved with multi-processing 
techniques but we did not use such acceleration techniques.

%=======================================%
\textbf{Results:}
%==========%
The computation time taken for 1000 epochs was 
$17.0907\pm{0.2968}$, $17.2892\pm{0.2628}$, and $633.6483\pm{697.2283}$ seconds 
(in `$\text{mean}\pm\text{STD}$' over 100 trials) for Nonr-MLR, Prev-MLR, and Stri-MLR, and 
$17.5099\pm{0.5087}$, $16.9657\pm{0.8673}$, and $1157.0944\pm{398.5733}$ seconds 
for Nonr-AUL, Prev-AUL, and Stri-AUL with the mixture rate $r=0.25$.
These results indicate the higher computation cost of the strict UPRL methods
compared with non-regularized and previous UPRL methods.
Note that, if the predicted CPD becomes unimodal, the strict UPRL 
methods skip the inner optimization of \eqref{eq:OurUPRL}.
This mechanism increases the variance of the calculation time of the strict UPRL methods.
%Furthermore, Stri-AUL was more likely to output a unimodal CPD prediction than Stri-MLR.
%We consider that this tendency led to reduce the computation time of Stri-AUL from Stri-MLR.

\clearpage
%========================================%
%==========%
\begin{table*}[p]
\centering%
\renewcommand{\arraystretch}{0.75}%
\renewcommand{\tabcolsep}{4pt}%
\caption{%
Correlation coefficient of UD or scale and expected UPRL loss for $K=3$.}
\label{tab:UPRL31}
{\footnotesize%
\begin{tabular}{c|ccccc|c|ccccc|c}\toprule
\multirow{3}{*}{$\bp$} & \multicolumn{6}{c|}{UD} & \multicolumn{6}{c}{scale}\\
& \multicolumn{5}{c}{Prev-UPRL, $\delta=$} & \multirow{2}{*}{Stri-UPRL} & \multicolumn{5}{c}{Prev-UPRL, $\delta=$} & \multirow{2}{*}{Stri-UPRL}\\
&0&0.05&0.1&0.2&\multicolumn{1}{c}{0.4}& &0&0.05&0.1&0.2&\multicolumn{1}{c}{0.4}& \\\midrule
$(0, 0, 1)$&\vc{.1192}&\vc{.1105}&\vc{.0974}&\vc{.0638}&\vc{.0192}&\vc{1.0000}&\vc{-.5092}&\vc{-.4802}&\vc{-.4519}&\vc{-.3919}&\vc{-.2546}&\vc{.0931}\\
$(0, 1, 0)$&\vc{.6921}&\vc{.6967}&\vc{.6935}&\vc{.6759}&\vc{.6328}&\vc{1.0000}&\vc{-.3712}&\vc{-.3366}&\vc{-.3044}&\vc{-.2436}&\vc{-.1313}&\vc{.0931}\\
$(0, .2, .8)$&\vc{.3142}&\vc{.3170}&\vc{.3120}&\vc{.2906}&\vc{.2558}&\vc{1.0000}&\vc{-.5805}&\vc{-.5490}&\vc{-.5170}&\vc{-.4480}&\vc{-.2851}&\vc{.0931}\\
$(.1, .1, .8)$&\vc{.2360}&\vc{.2351}&\vc{.2271}&\vc{.1993}&\vc{.1565}&\vc{1.0000}&\vc{-.6181}&\vc{-.5916}&\vc{-.5645}&\vc{-.5019}&\vc{-.3331}&\vc{.0931}\\
$(0, .8, .2)$&\vc{.6886}&\vc{.6953}&\vc{.6925}&\vc{.6741}&\vc{.6282}&\vc{1.0000}&\vc{-.4534}&\vc{-.4138}&\vc{-.3764}&\vc{-.3045}&\vc{-.1684}&\vc{.0931}\\
$(.1, .8, .1)$&\vc{.6959}&\vc{.7026}&\vc{.6999}&\vc{.6810}&\vc{.6340}&\vc{1.0000}&\vc{-.4566}&\vc{-.4165}&\vc{-.3785}&\vc{-.3054}&\vc{-.1669}&\vc{.0931}\\
$(0, .4, .6)$&\vc{.5129}&\vc{.5245}&\vc{.5252}&\vc{.5120}&\vc{.4777}&\vc{1.0000}&\vc{-.5983}&\vc{-.5612}&\vc{-.5236}&\vc{-.4445}&\vc{-.2690}&\vc{.0931}\\
$(.1, .3, .6)$&\vc{.4661}&\vc{.4805}&\vc{.4841}&\vc{.4752}&\vc{.4470}&\vc{1.0000}&\vc{-.6695}&\vc{-.6383}&\vc{-.6052}&\vc{-.5297}&\vc{-.3340}&\vc{.0931}\\
$(.2, .2, .6)$&\vc{.3912}&\vc{.4059}&\vc{.4110}&\vc{.4055}&\vc{.3868}&\vc{1.0000}&\vc{-.7320}&\vc{-.7119}&\vc{-.6900}&\vc{-.6323}&\vc{-.4316}&\vc{.0931}\\
$(0, .6, .4)$&\vc{.6414}&\vc{.6520}&\vc{.6515}&\vc{.6356}&\vc{.5925}&\vc{1.0000}&\vc{-.5421}&\vc{-.5008}&\vc{-.4604}&\vc{-.3799}&\vc{-.2180}&\vc{.0931}\\
$(.1, .6, .3)$&\vc{.6712}&\vc{.6834}&\vc{.6841}&\vc{.6686}&\vc{.6227}&\vc{1.0000}&\vc{-.5662}&\vc{-.5236}&\vc{-.4819}&\vc{-.3975}&\vc{-.2256}&\vc{.0931}\\
$(.2, .6, .2)$&\vc{.6826}&\vc{.6952}&\vc{.6965}&\vc{.6810}&\vc{.6334}&\vc{1.0000}&\vc{-.5746}&\vc{-.5314}&\vc{-.4890}&\vc{-.4026}&\vc{-.2258}&\vc{.0931}\\
$(.2, .4, .4)$&\vc{.6008}&\vc{.6228}&\vc{.6324}&\vc{.6306}&\vc{.5994}&\vc{1.0000}&\vc{-.7016}&\vc{-.6668}&\vc{-.6302}&\vc{-.5452}&\vc{-.3300}&\vc{.0931}\\
$(.3, .3, .4)$&\vc{.5384}&\vc{.5662}&\vc{.5831}&\vc{.5967}&\vc{.5916}&\vc{1.0000}&\vc{-.7780}&\vc{-.7562}&\vc{-.7325}&\vc{-.6661}&\vc{-.4355}&\vc{.0931}\\
$(.3, .4, .3)$&\vc{.6158}&\vc{.6401}&\vc{.6521}&\vc{.6534}&\vc{.6229}&\vc{1.0000}&\vc{-.7196}&\vc{-.6856}&\vc{-.6497}&\vc{-.5636}&\vc{-.3389}&\vc{.0931}\\
\bottomrule\end{tabular}}
\end{table*}
%==========%
\begin{table*}[p]
\centering%
\renewcommand{\arraystretch}{0.75}%
\renewcommand{\tabcolsep}{4pt}%
\caption{%
Correlation coefficient of UD or scale and expected UPRL loss for $K=4$.}
\label{tab:UPRL41}
{\footnotesize%
\begin{tabular}{c|ccccc|c|ccccc|c}\toprule
\multirow{3}{*}{$\bp$} & \multicolumn{6}{c|}{UD} & \multicolumn{6}{c}{scale}\\
& \multicolumn{5}{c}{Prev-UPRL, $\delta=$} & \multirow{2}{*}{Stri-UPRL} & \multicolumn{5}{c}{Prev-UPRL, $\delta=$} & \multirow{2}{*}{Stri-UPRL}\\
&0&0.05&0.1&0.2&\multicolumn{1}{c}{0.4}& &0&0.05&0.1&0.2&\multicolumn{1}{c}{0.4}& \\\midrule
$(0, 0, 0, 1)$&\vc{.1667}&\vc{.1438}&\vc{.1136}&\vc{.0447}&\vc{-.0213}&\vc{1.0000}&\vc{-.5351}&\vc{-.5115}&\vc{-.4907}&\vc{-.4375}&\vc{-.2751}&\vc{.1346}\\
$(0, 0, 1, 0)$&\vc{.5558}&\vc{.5503}&\vc{.5331}&\vc{.4884}&\vc{.4310}&\vc{1.0000}&\vc{-.4100}&\vc{-.3722}&\vc{-.3369}&\vc{-.2704}&\vc{-.1395}&\vc{.1346}\\
$(0, 0, .2, .8)$&\vc{.3008}&\vc{.2869}&\vc{.2621}&\vc{.2000}&\vc{.1379}&\vc{1.0000}&\vc{-.5927}&\vc{-.5687}&\vc{-.5462}&\vc{-.4882}&\vc{-.3028}&\vc{.1346}\\
$(0, .1, .1, .8)$&\vc{.3038}&\vc{.2899}&\vc{.2649}&\vc{.2019}&\vc{.1381}&\vc{1.0000}&\vc{-.5988}&\vc{-.5750}&\vc{-.5534}&\vc{-.4954}&\vc{-.3057}&\vc{.1346}\\
$(0, 0, .8, .2)$&\vc{.5524}&\vc{.5477}&\vc{.5296}&\vc{.4810}&\vc{.4198}&\vc{1.0000}&\vc{-.4874}&\vc{-.4483}&\vc{-.4109}&\vc{-.3365}&\vc{-.1793}&\vc{.1346}\\
$(0, .1, .8, .1)$&\vc{.5952}&\vc{.5910}&\vc{.5726}&\vc{.5228}&\vc{.4578}&\vc{1.0000}&\vc{-.4764}&\vc{-.4353}&\vc{-.3965}&\vc{-.3205}&\vc{-.1660}&\vc{.1346}\\
$(.1, .1, .8, 0)$&\vc{.6008}&\vc{.5971}&\vc{.5788}&\vc{.5290}&\vc{.4622}&\vc{1.0000}&\vc{-.4708}&\vc{-.4290}&\vc{-.3898}&\vc{-.3133}&\vc{-.1580}&\vc{.1346}\\
$(0, 0, .4, .6)$&\vc{.4305}&\vc{.4250}&\vc{.4057}&\vc{.3522}&\vc{.2933}&\vc{1.0000}&\vc{-.6067}&\vc{-.5780}&\vc{-.5494}&\vc{-.4808}&\vc{-.2841}&\vc{.1346}\\
$(0, .1, .3, .6)$&\vc{.4489}&\vc{.4454}&\vc{.4272}&\vc{.3737}&\vc{.3127}&\vc{1.0000}&\vc{-.6313}&\vc{-.6040}&\vc{-.5771}&\vc{-.5085}&\vc{-.3004}&\vc{.1346}\\
$(0, .2, .2, .6)$&\vc{.4559}&\vc{.4533}&\vc{.4358}&\vc{.3821}&\vc{.3193}&\vc{1.0000}&\vc{-.6398}&\vc{-.6129}&\vc{-.5872}&\vc{-.5182}&\vc{-.3039}&\vc{.1346}\\
$(.1, .1, .2, .6)$&\vc{.4205}&\vc{.4191}&\vc{.4022}&\vc{.3483}&\vc{.2862}&\vc{1.0000}&\vc{-.6766}&\vc{-.6593}&\vc{-.6433}&\vc{-.5902}&\vc{-.3651}&\vc{.1346}\\
$(0, 0, .6, .4)$&\vc{.5171}&\vc{.5138}&\vc{.4958}&\vc{.4452}&\vc{.3837}&\vc{1.0000}&\vc{-.5632}&\vc{-.5274}&\vc{-.4920}&\vc{-.4152}&\vc{-.2313}&\vc{.1346}\\
$(0, .1, .6, .3)$&\vc{.5782}&\vc{.5767}&\vc{.5591}&\vc{.5074}&\vc{.4404}&\vc{1.0000}&\vc{-.5619}&\vc{-.5229}&\vc{-.4847}&\vc{-.4033}&\vc{-.2175}&\vc{.1346}\\
$(0, .2, .6, .2)$&\vc{.6262}&\vc{.6252}&\vc{.6072}&\vc{.5539}&\vc{.4817}&\vc{1.0000}&\vc{-.5477}&\vc{-.5056}&\vc{-.4650}&\vc{-.3806}&\vc{-.1985}&\vc{.1346}\\
$(.1, .1, .6, .2)$&\vc{.5998}&\vc{.6006}&\vc{.5841}&\vc{.5322}&\vc{.4610}&\vc{1.0000}&\vc{-.5709}&\vc{-.5313}&\vc{-.4926}&\vc{-.4086}&\vc{-.2150}&\vc{.1346}\\
$(0, .3, .6, .1)$&\vc{.6587}&\vc{.6571}&\vc{.6382}&\vc{.5835}&\vc{.5079}&\vc{1.0000}&\vc{-.5221}&\vc{-.4777}&\vc{-.4355}&\vc{-.3507}&\vc{-.1770}&\vc{.1346}\\
$(.1, .2, .6, .1)$&\vc{.6410}&\vc{.6415}&\vc{.6239}&\vc{.5702}&\vc{.4945}&\vc{1.0000}&\vc{-.5489}&\vc{-.5060}&\vc{-.4647}&\vc{-.3783}&\vc{-.1921}&\vc{.1346}\\
$(0, .4, .6, 0)$&\vc{.6762}&\vc{.6734}&\vc{.6536}&\vc{.5985}&\vc{.5218}&\vc{1.0000}&\vc{-.4885}&\vc{-.4429}&\vc{-.4003}&\vc{-.3176}&\vc{-.1554}&\vc{.1346}\\
$(.1, .3, .6, 0)$&\vc{.6655}&\vc{.6646}&\vc{.6458}&\vc{.5907}&\vc{.5130}&\vc{1.0000}&\vc{-.5163}&\vc{-.4711}&\vc{-.4283}&\vc{-.3426}&\vc{-.1681}&\vc{.1346}\\
$(.2, .2, .6, 0)$&\vc{.6397}&\vc{.6404}&\vc{.6228}&\vc{.5692}&\vc{.4925}&\vc{1.0000}&\vc{-.5361}&\vc{-.4924}&\vc{-.4507}&\vc{-.3643}&\vc{-.1796}&\vc{.1346}\\
$(0, .2, .4, .4)$&\vc{.5758}&\vc{.5780}&\vc{.5626}&\vc{.5109}&\vc{.4414}&\vc{1.0000}&\vc{-.6229}&\vc{-.5877}&\vc{-.5526}&\vc{-.4701}&\vc{-.2585}&\vc{.1346}\\
$(.1, .1, .4, .4)$&\vc{.5473}&\vc{.5526}&\vc{.5397}&\vc{.4911}&\vc{.4244}&\vc{1.0000}&\vc{-.6574}&\vc{-.6290}&\vc{-.6001}&\vc{-.5242}&\vc{-.2962}&\vc{.1346}\\
$(0, .3, .3, .4)$&\vc{.5858}&\vc{.5894}&\vc{.5747}&\vc{.5227}&\vc{.4503}&\vc{1.0000}&\vc{-.6314}&\vc{-.5963}&\vc{-.5617}&\vc{-.4776}&\vc{-.2598}&\vc{.1346}\\
$(.1, .2, .3, .4)$&\vc{.5676}&\vc{.5763}&\vc{.5659}&\vc{.5189}&\vc{.4498}&\vc{1.0000}&\vc{-.6797}&\vc{-.6532}&\vc{-.6265}&\vc{-.5507}&\vc{-.3096}&\vc{.1346}\\
$(.2, .2, .2, .4)$&\vc{.5396}&\vc{.5542}&\vc{.5497}&\vc{.5122}&\vc{.4521}&\vc{1.0000}&\vc{-.7254}&\vc{-.7121}&\vc{-.7004}&\vc{-.6462}&\vc{-.3826}&\vc{.1346}\\
$(0, .3, .4, .3)$&\vc{.6279}&\vc{.6308}&\vc{.6149}&\vc{.5612}&\vc{.4853}&\vc{1.0000}&\vc{-.6053}&\vc{-.5656}&\vc{-.5268}&\vc{-.4388}&\vc{-.2317}&\vc{.1346}\\
$(.1, .2, .4, .3)$&\vc{.6117}&\vc{.6192}&\vc{.6068}&\vc{.5568}&\vc{.4820}&\vc{1.0000}&\vc{-.6478}&\vc{-.6140}&\vc{-.5799}&\vc{-.4947}&\vc{-.2665}&\vc{.1346}\\
$(0, .4, .4, .2)$&\vc{.6608}&\vc{.6625}&\vc{.6452}&\vc{.5893}&\vc{.5091}&\vc{1.0000}&\vc{-.5727}&\vc{-.5292}&\vc{-.4874}&\vc{-.3975}&\vc{-.2021}&\vc{.1346}\\
$(.1, .3, .4, .2)$&\vc{.6552}&\vc{.6618}&\vc{.6476}&\vc{.5939}&\vc{.5125}&\vc{1.0000}&\vc{-.6184}&\vc{-.5787}&\vc{-.5394}&\vc{-.4482}&\vc{-.2305}&\vc{.1346}\\
$(.2, .2, .4, .2)$&\vc{.6303}&\vc{.6412}&\vc{.6307}&\vc{.5818}&\vc{.5032}&\vc{1.0000}&\vc{-.6533}&\vc{-.6198}&\vc{-.5859}&\vc{-.4987}&\vc{-.2618}&\vc{.1346}\\
$(.1, .4, .4, .1)$&\vc{.6776}&\vc{.6812}&\vc{.6642}&\vc{.6076}&\vc{.5235}&\vc{1.0000}&\vc{-.5748}&\vc{-.5304}&\vc{-.4877}&\vc{-.3953}&\vc{-.1952}&\vc{.1346}\\
$(.2, .3, .4, .1)$&\vc{.6630}&\vc{.6709}&\vc{.6570}&\vc{.6031}&\vc{.5190}&\vc{1.0000}&\vc{-.6123}&\vc{-.5718}&\vc{-.5317}&\vc{-.4388}&\vc{-.2192}&\vc{.1346}\\
$(.3, .3, .4, 0)$&\vc{.6492}&\vc{.6556}&\vc{.6403}&\vc{.5860}&\vc{.5030}&\vc{1.0000}&\vc{-.5866}&\vc{-.5443}&\vc{-.5031}&\vc{-.4106}&\vc{-.1986}&\vc{.1346}\\
\bottomrule\end{tabular}}
\end{table*}
%==========%
\begin{table*}[p]
\centering%
\renewcommand{\arraystretch}{0.75}%
\renewcommand{\tabcolsep}{4pt}%
\caption{%
Correlation coefficient of UD or scale and expected UPRL loss for $K=5$.}
\label{tab:UPRL51}
{\footnotesize%
\begin{tabular}{c|ccccc|c|ccccc|c}\toprule
\multirow{3}{*}{$\bp$} & \multicolumn{6}{c|}{UD} & \multicolumn{6}{c}{scale}\\
& \multicolumn{5}{c}{Prev-UPRL, $\delta=$} & \multirow{2}{*}{Stri-UPRL} & \multicolumn{5}{c}{Prev-UPRL, $\delta=$} & \multirow{2}{*}{Stri-UPRL}\\
&0&0.05&0.1&0.2&\multicolumn{1}{c}{0.4}& &0&0.05&0.1&0.2&\multicolumn{1}{c}{0.4}& \\\midrule
$(0, 0, 0, 0, 1)$&\vc{.2433}&\vc{.2200}&\vc{.1779}&\vc{.0786}&\vc{.0051}&\vc{1.0000}&\vc{-.4570}&\vc{-.4294}&\vc{-.4109}&\vc{-.3604}&\vc{-.1552}&\vc{.1523}\\
$(0, 0, 0, 1, 0)$&\vc{.4916}&\vc{.4767}&\vc{.4437}&\vc{.3682}&\vc{.3164}&\vc{1.0000}&\vc{-.3968}&\vc{-.3546}&\vc{-.3218}&\vc{-.2572}&\vc{-.1089}&\vc{.1523}\\
$(0, 0, 1, 0, 0)$&\vc{.5432}&\vc{.5314}&\vc{.4994}&\vc{.4265}&\vc{.3634}&\vc{1.0000}&\vc{-.4413}&\vc{-.3935}&\vc{-.3560}&\vc{-.2887}&\vc{-.1362}&\vc{.1523}\\
$(0, 0, 0, .2, .8)$&\vc{.3519}&\vc{.3367}&\vc{.2983}&\vc{.1987}&\vc{.1282}&\vc{1.0000}&\vc{-.5185}&\vc{-.4917}&\vc{-.4742}&\vc{-.4211}&\vc{-.1872}&\vc{.1523}\\
$(0, 0, 0, .8, .2)$&\vc{.5070}&\vc{.4938}&\vc{.4586}&\vc{.3734}&\vc{.3150}&\vc{1.0000}&\vc{-.4600}&\vc{-.4172}&\vc{-.3833}&\vc{-.3112}&\vc{-.1312}&\vc{.1523}\\
$(0, 0, .2, .8, 0)$&\vc{.5597}&\vc{.5468}&\vc{.5117}&\vc{.4284}&\vc{.3675}&\vc{1.0000}&\vc{-.4524}&\vc{-.4064}&\vc{-.3697}&\vc{-.2972}&\vc{-.1289}&\vc{.1523}\\
$(0, 0, .8, .2, 0)$&\vc{.5977}&\vc{.5872}&\vc{.5529}&\vc{.4716}&\vc{.4020}&\vc{1.0000}&\vc{-.4850}&\vc{-.4352}&\vc{-.3954}&\vc{-.3211}&\vc{-.1484}&\vc{.1523}\\
$(0, 0, 0, .4, .6)$&\vc{.4475}&\vc{.4385}&\vc{.4036}&\vc{.3077}&\vc{.2426}&\vc{1.0000}&\vc{-.5442}&\vc{-.5134}&\vc{-.4912}&\vc{-.4274}&\vc{-.1875}&\vc{.1523}\\
$(0, 0, .2, .2, .6)$&\vc{.4812}&\vc{.4775}&\vc{.4458}&\vc{.3527}&\vc{.2853}&\vc{1.0000}&\vc{-.5822}&\vc{-.5538}&\vc{-.5346}&\vc{-.4765}&\vc{-.2190}&\vc{.1523}\\
$(0, 0, 0, .6, .4)$&\vc{.4992}&\vc{.4894}&\vc{.4542}&\vc{.3619}&\vc{.2989}&\vc{1.0000}&\vc{-.5181}&\vc{-.4797}&\vc{-.4495}&\vc{-.3759}&\vc{-.1603}&\vc{.1523}\\
$(0, 0, .2, .6, .2)$&\vc{.5799}&\vc{.5717}&\vc{.5364}&\vc{.4445}&\vc{.3765}&\vc{1.0000}&\vc{-.5260}&\vc{-.4816}&\vc{-.4456}&\vc{-.3665}&\vc{-.1592}&\vc{.1523}\\
$(0, 0, .4, .6, 0)$&\vc{.6115}&\vc{.6010}&\vc{.5651}&\vc{.4772}&\vc{.4085}&\vc{1.0000}&\vc{-.4950}&\vc{-.4463}&\vc{-.4069}&\vc{-.3290}&\vc{-.1459}&\vc{.1523}\\
$(0, .2, .2, .6, 0)$&\vc{.6129}&\vc{.6031}&\vc{.5668}&\vc{.4756}&\vc{.4057}&\vc{1.0000}&\vc{-.5152}&\vc{-.4677}&\vc{-.4289}&\vc{-.3491}&\vc{-.1570}&\vc{.1523}\\
$(0, 0, .6, .2, .2)$&\vc{.6078}&\vc{.6022}&\vc{.5684}&\vc{.4799}&\vc{.4050}&\vc{1.0000}&\vc{-.5510}&\vc{-.5044}&\vc{-.4667}&\vc{-.3877}&\vc{-.1761}&\vc{.1523}\\
$(0, 0, .6, .4, 0)$&\vc{.6259}&\vc{.6164}&\vc{.5808}&\vc{.4937}&\vc{.4215}&\vc{1.0000}&\vc{-.5073}&\vc{-.4573}&\vc{-.4168}&\vc{-.3382}&\vc{-.1532}&\vc{.1523}\\
$(0, .2, .6, .2, 0)$&\vc{.6392}&\vc{.6319}&\vc{.5964}&\vc{.5073}&\vc{.4308}&\vc{1.0000}&\vc{-.5382}&\vc{-.4885}&\vc{-.4476}&\vc{-.3670}&\vc{-.1717}&\vc{.1523}\\
$(0, 0, .2, .4, .4)$&\vc{.5615}&\vc{.5593}&\vc{.5273}&\vc{.4334}&\vc{.3637}&\vc{1.0000}&\vc{-.5814}&\vc{-.5451}&\vc{-.5167}&\vc{-.4426}&\vc{-.1966}&\vc{.1523}\\
$(0, .2, .2, .2, .4)$&\vc{.5780}&\vc{.5815}&\vc{.5532}&\vc{.4599}&\vc{.3868}&\vc{1.0000}&\vc{-.6266}&\vc{-.5960}&\vc{-.5733}&\vc{-.5048}&\vc{-.2369}&\vc{.1523}\\
$(0, 0, .4, .4, .2)$&\vc{.6210}&\vc{.6166}&\vc{.5824}&\vc{.4895}&\vc{.4148}&\vc{1.0000}&\vc{-.5631}&\vc{-.5179}&\vc{-.4810}&\vc{-.3995}&\vc{-.1779}&\vc{.1523}\\
$(0, .2, .2, .4, .2)$&\vc{.6256}&\vc{.6239}&\vc{.5906}&\vc{.4939}&\vc{.4177}&\vc{1.0000}&\vc{-.5918}&\vc{-.5502}&\vc{-.5159}&\vc{-.4336}&\vc{-.1969}&\vc{.1523}\\
$(0, .2, .4, .4, 0)$&\vc{.6518}&\vc{.6450}&\vc{.6088}&\vc{.5155}&\vc{.4387}&\vc{1.0000}&\vc{-.5484}&\vc{-.4994}&\vc{-.4587}&\vc{-.3756}&\vc{-.1723}&\vc{.1523}\\
$(.2, .2, .2, .4, 0)$&\vc{.6139}&\vc{.6119}&\vc{.5768}&\vc{.4779}&\vc{.4019}&\vc{1.0000}&\vc{-.6000}&\vc{-.5597}&\vc{-.5254}&\vc{-.4424}&\vc{-.2093}&\vc{.1523}\\
$(0, .2, .4, .2, .2)$&\vc{.6391}&\vc{.6392}&\vc{.6069}&\vc{.5128}&\vc{.4328}&\vc{1.0000}&\vc{-.6040}&\vc{-.5617}&\vc{-.5266}&\vc{-.4450}&\vc{-.2062}&\vc{.1523}\\
$(.2, .2, .2, .2, .2)$&\vc{.6047}&\vc{.6165}&\vc{.5914}&\vc{.4951}&\vc{.4184}&\vc{1.0000}&\vc{-.6787}&\vc{-.6571}&\vc{-.6416}&\vc{-.5765}&\vc{-.2858}&\vc{.1523}\\
\bottomrule\end{tabular}}
\end{table*}
%==========%
\begin{table*}[p]
\centering%
\renewcommand{\arraystretch}{0.75}%
\renewcommand{\tabcolsep}{4pt}%
\caption{%
Correlation coefficient of UD or scale and expected UPRL loss for $K=6$.}
\label{tab:UPRL61}
{\footnotesize%
\begin{tabular}{c|ccccc|c|ccccc|c}\toprule
\multirow{3}{*}{$\bp$} & \multicolumn{6}{c|}{UD} & \multicolumn{6}{c}{scale}\\
& \multicolumn{5}{c}{Prev-UPRL, $\delta=$} & \multirow{2}{*}{Stri-UPRL} & \multicolumn{5}{c}{Prev-UPRL, $\delta=$} & \multirow{2}{*}{Stri-UPRL}\\
&0&0.05&0.1&0.2&\multicolumn{1}{c}{0.4}& &0&0.05&0.1&0.2&\multicolumn{1}{c}{0.4}& \\\midrule
$(0, 0, 0, 0, 0, 1)$&\vc{.3224}&\vc{.2841}&\vc{.2206}&\vc{.0932}&\vc{.0191}&\vc{1.0000}&\vc{-.5309}&\vc{-.5076}&\vc{-.4941}&\vc{-.4323}&\vc{-.1906}&\vc{.1145}\\
$(0, 0, 0, 0, 1, 0)$&\vc{.5087}&\vc{.4806}&\vc{.4311}&\vc{.3399}&\vc{.2877}&\vc{1.0000}&\vc{-.4381}&\vc{-.3977}&\vc{-.3663}&\vc{-.2847}&\vc{-.1003}&\vc{.1145}\\
$(0, 0, 0, 1, 0, 0)$&\vc{.5555}&\vc{.5259}&\vc{.4748}&\vc{.3828}&\vc{.3300}&\vc{1.0000}&\vc{-.4547}&\vc{-.4164}&\vc{-.3848}&\vc{-.3016}&\vc{-.1177}&\vc{.1145}\\
$(0, 0, 0, 0, .2, .8)$&\vc{.4089}&\vc{.3787}&\vc{.3189}&\vc{.1939}&\vc{.1272}&\vc{1.0000}&\vc{-.5707}&\vc{-.5511}&\vc{-.5411}&\vc{-.4785}&\vc{-.2101}&\vc{.1145}\\
$(0, 0, 0, 0, .8, .2)$&\vc{.5211}&\vc{.4950}&\vc{.4428}&\vc{.3413}&\vc{.2849}&\vc{1.0000}&\vc{-.4968}&\vc{-.4598}&\vc{-.4309}&\vc{-.3440}&\vc{-.1259}&\vc{.1145}\\
$(0, 0, 0, .2, .8, 0)$&\vc{.5653}&\vc{.5395}&\vc{.4884}&\vc{.3905}&\vc{.3328}&\vc{1.0000}&\vc{-.4817}&\vc{-.4424}&\vc{-.4108}&\vc{-.3228}&\vc{-.1166}&\vc{.1145}\\
$(0, 0, 0, .8, .2, 0)$&\vc{.5970}&\vc{.5701}&\vc{.5179}&\vc{.4196}&\vc{.3617}&\vc{1.0000}&\vc{-.4935}&\vc{-.4553}&\vc{-.4235}&\vc{-.3344}&\vc{-.1285}&\vc{.1145}\\
$(0, 0, .2, .8, 0, 0)$&\vc{.6151}&\vc{.5881}&\vc{.5355}&\vc{.4370}&\vc{.3790}&\vc{1.0000}&\vc{-.4931}&\vc{-.4543}&\vc{-.4224}&\vc{-.3340}&\vc{-.1279}&\vc{.1145}\\
$(0, 0, 0, 0, .4, .6)$&\vc{.4780}&\vc{.4540}&\vc{.3987}&\vc{.2812}&\vc{.2216}&\vc{1.0000}&\vc{-.5783}&\vc{-.5565}&\vc{-.5432}&\vc{-.4703}&\vc{-.1955}&\vc{.1145}\\
$(0, 0, 0, .2, .2, .6)$&\vc{.5089}&\vc{.4899}&\vc{.4370}&\vc{.3198}&\vc{.2617}&\vc{1.0000}&\vc{-.6057}&\vc{-.5904}&\vc{-.5831}&\vc{-.5159}&\vc{-.2233}&\vc{.1145}\\
$(0, 0, 0, 0, .6, .4)$&\vc{.5149}&\vc{.4914}&\vc{.4379}&\vc{.3277}&\vc{.2692}&\vc{1.0000}&\vc{-.5492}&\vc{-.5197}&\vc{-.4976}&\vc{-.4130}&\vc{-.1598}&\vc{.1145}\\
$(0, 0, 0, .2, .6, .2)$&\vc{.5784}&\vc{.5574}&\vc{.5056}&\vc{.3988}&\vc{.3378}&\vc{1.0000}&\vc{-.5448}&\vc{-.5120}&\vc{-.4858}&\vc{-.3949}&\vc{-.1490}&\vc{.1145}\\
$(0, 0, 0, .4, .6, 0)$&\vc{.6057}&\vc{.5819}&\vc{.5302}&\vc{.4285}&\vc{.3675}&\vc{1.0000}&\vc{-.5108}&\vc{-.4728}&\vc{-.4417}&\vc{-.3498}&\vc{-.1294}&\vc{.1145}\\
$(0, 0, .2, .2, .6, 0)$&\vc{.6327}&\vc{.6117}&\vc{.5608}&\vc{.4574}&\vc{.3935}&\vc{1.0000}&\vc{-.5222}&\vc{-.4845}&\vc{-.4541}&\vc{-.3617}&\vc{-.1304}&\vc{.1145}\\
$(0, 0, 0, .6, .2, .2)$&\vc{.6033}&\vc{.5826}&\vc{.5308}&\vc{.4240}&\vc{.3626}&\vc{1.0000}&\vc{-.5551}&\vc{-.5238}&\vc{-.4979}&\vc{-.4059}&\vc{-.1593}&\vc{.1145}\\
$(0, 0, 0, .6, .4, 0)$&\vc{.6171}&\vc{.5929}&\vc{.5408}&\vc{.4390}&\vc{.3779}&\vc{1.0000}&\vc{-.5152}&\vc{-.4775}&\vc{-.4463}&\vc{-.3540}&\vc{-.1337}&\vc{.1145}\\
$(0, 0, .2, .6, .2, 0)$&\vc{.6552}&\vc{.6327}&\vc{.5805}&\vc{.4768}&\vc{.4137}&\vc{1.0000}&\vc{-.5303}&\vc{-.4928}&\vc{-.4618}&\vc{-.3685}&\vc{-.1388}&\vc{.1145}\\
$(0, .2, .2, .6, 0, 0)$&\vc{.6560}&\vc{.6316}&\vc{.5775}&\vc{.4712}&\vc{.4060}&\vc{1.0000}&\vc{-.5267}&\vc{-.4877}&\vc{-.4555}&\vc{-.3624}&\vc{-.1324}&\vc{.1145}\\
$(0, 0, 0, .2, .4, .4)$&\vc{.5636}&\vc{.5470}&\vc{.4961}&\vc{.3841}&\vc{.3246}&\vc{1.0000}&\vc{-.5930}&\vc{-.5706}&\vc{-.5547}&\vc{-.4721}&\vc{-.1901}&\vc{.1145}\\
$(0, 0, .2, .2, .2, .4)$&\vc{.6117}&\vc{.6032}&\vc{.5571}&\vc{.4480}&\vc{.3906}&\vc{1.0000}&\vc{-.6198}&\vc{-.6032}&\vc{-.5936}&\vc{-.5168}&\vc{-.2106}&\vc{.1145}\\
$(0, 0, 0, .4, .4, .2)$&\vc{.6098}&\vc{.5916}&\vc{.5406}&\vc{.4324}&\vc{.3698}&\vc{1.0000}&\vc{-.5676}&\vc{-.5375}&\vc{-.5131}&\vc{-.4208}&\vc{-.1627}&\vc{.1145}\\
$(0, 0, .2, .2, .4, .2)$&\vc{.6423}&\vc{.6294}&\vc{.5809}&\vc{.4725}&\vc{.4080}&\vc{1.0000}&\vc{-.5846}&\vc{-.5567}&\vc{-.5350}&\vc{-.4439}&\vc{-.1684}&\vc{.1145}\\
$(0, 0, .2, .4, .4, 0)$&\vc{.6632}&\vc{.6435}&\vc{.5923}&\vc{.4871}&\vc{.4216}&\vc{1.0000}&\vc{-.5420}&\vc{-.5054}&\vc{-.4754}&\vc{-.3807}&\vc{-.1406}&\vc{.1145}\\
$(0, .2, .2, .2, .4, 0)$&\vc{.6732}&\vc{.6549}&\vc{.6034}&\vc{.4941}&\vc{.4230}&\vc{1.0000}&\vc{-.5512}&\vc{-.5143}&\vc{-.4845}&\vc{-.3889}&\vc{-.1359}&\vc{.1145}\\
$(0, 0, .2, .4, .2, .2)$&\vc{.6554}&\vc{.6424}&\vc{.5936}&\vc{.4851}&\vc{.4210}&\vc{1.0000}&\vc{-.5897}&\vc{-.5623}&\vc{-.5405}&\vc{-.4487}&\vc{-.1738}&\vc{.1145}\\
$(0, .2, .2, .4, .2, 0)$&\vc{.6857}&\vc{.6664}&\vc{.6140}&\vc{.5043}&\vc{.4340}&\vc{1.0000}&\vc{-.5559}&\vc{-.5189}&\vc{-.4885}&\vc{-.3922}&\vc{-.1405}&\vc{.1145}\\
$(.2, .2, .2, .4, 0, 0)$&\vc{.6524}&\vc{.6371}&\vc{.5854}&\vc{.4713}&\vc{.3986}&\vc{1.0000}&\vc{-.5889}&\vc{-.5579}&\vc{-.5332}&\vc{-.4411}&\vc{-.1648}&\vc{.1145}\\
$(0, .2, .2, .2, .2, .2)$&\vc{.6688}&\vc{.6596}&\vc{.6123}&\vc{.4998}&\vc{.4289}&\vc{1.0000}&\vc{-.6044}&\vc{-.5791}&\vc{-.5599}&\vc{-.4686}&\vc{-.1727}&\vc{.1145}\\
$(.2, .2, .2, .2, .2, 0)$&\vc{.6645}&\vc{.6547}&\vc{.6066}&\vc{.4924}&\vc{.4159}&\vc{1.0000}&\vc{-.6069}&\vc{-.5795}&\vc{-.5590}&\vc{-.4685}&\vc{-.1720}&\vc{.1145}\\
\bottomrule\end{tabular}}
\end{table*}
%==========%
\begin{table*}[p]
\centering%
\renewcommand{\arraystretch}{0.75}%
\renewcommand{\tabcolsep}{4pt}%
\caption{%
Correlation coefficient of UD or scale and expected UPRL loss for $K=7$.}
\label{tab:UPRL71}
{\footnotesize%
\begin{tabular}{c|ccccc|c|ccccc|c}\toprule
\multirow{3}{*}{$\bp$} & \multicolumn{6}{c|}{UD} & \multicolumn{6}{c}{scale}\\
& \multicolumn{5}{c}{Prev-UPRL, $\delta=$} & \multirow{2}{*}{Stri-UPRL} & \multicolumn{5}{c}{Prev-UPRL, $\delta=$} & \multirow{2}{*}{Stri-UPRL}\\
&0&0.05&0.1&0.2&\multicolumn{1}{c}{0.4}& &0&0.05&0.1&0.2&\multicolumn{1}{c}{0.4}& \\\midrule
$(0, 0, 0, 0, 0, 0, 1)$&\vc{.3305}&\vc{.2894}&\vc{.2130}&\vc{.0609}&\vc{-.0156}&\vc{1.0000}&\vc{-.4941}&\vc{-.4753}&\vc{-.4687}&\vc{-.4104}&\vc{-.1351}&\vc{.1271}\\
$(0, 0, 0, 0, 0, 1, 0)$&\vc{.4645}&\vc{.4221}&\vc{.3569}&\vc{.2397}&\vc{.1824}&\vc{1.0000}&\vc{-.4552}&\vc{-.4050}&\vc{-.3649}&\vc{-.2763}&\vc{-.0808}&\vc{.1271}\\
$(0, 0, 0, 0, 1, 0, 0)$&\vc{.5608}&\vc{.5259}&\vc{.4713}&\vc{.3693}&\vc{.3202}&\vc{1.0000}&\vc{-.4287}&\vc{-.3813}&\vc{-.3460}&\vc{-.2628}&\vc{-.0766}&\vc{.1271}\\
$(0, 0, 0, 1, 0, 0, 0)$&\vc{.5686}&\vc{.5375}&\vc{.4833}&\vc{.3743}&\vc{.3096}&\vc{1.0000}&\vc{-.4081}&\vc{-.3635}&\vc{-.3255}&\vc{-.2394}&\vc{-.0450}&\vc{.1271}\\
$(0, 0, 0, 0, 0, .2, .8)$&\vc{.4010}&\vc{.3660}&\vc{.2914}&\vc{.1342}&\vc{.0550}&\vc{1.0000}&\vc{-.5421}&\vc{-.5266}&\vc{-.5236}&\vc{-.4674}&\vc{-.1575}&\vc{.1271}\\
$(0, 0, 0, 0, 0, .8, .2)$&\vc{.4846}&\vc{.4454}&\vc{.3760}&\vc{.2433}&\vc{.1788}&\vc{1.0000}&\vc{-.5097}&\vc{-.4664}&\vc{-.4319}&\vc{-.3387}&\vc{-.1009}&\vc{.1271}\\
$(0, 0, 0, 0, .2, .8, 0)$&\vc{.5273}&\vc{.4886}&\vc{.4232}&\vc{.2992}&\vc{.2375}&\vc{1.0000}&\vc{-.4890}&\vc{-.4403}&\vc{-.4010}&\vc{-.3067}&\vc{-.0899}&\vc{.1271}\\
$(0, 0, 0, 0, .8, .2, 0)$&\vc{.5862}&\vc{.5529}&\vc{.4952}&\vc{.3830}&\vc{.3278}&\vc{1.0000}&\vc{-.4687}&\vc{-.4214}&\vc{-.3851}&\vc{-.2950}&\vc{-.0864}&\vc{.1271}\\
$(0, 0, 0, .2, .8, 0, 0)$&\vc{.6136}&\vc{.5823}&\vc{.5266}&\vc{.4156}&\vc{.3593}&\vc{1.0000}&\vc{-.4634}&\vc{-.4166}&\vc{-.3803}&\vc{-.2898}&\vc{-.0795}&\vc{.1271}\\
$(0, 0, 0, .8, .2, 0, 0)$&\vc{.6211}&\vc{.5928}&\vc{.5372}&\vc{.4207}&\vc{.3528}&\vc{1.0000}&\vc{-.4516}&\vc{-.4068}&\vc{-.3684}&\vc{-.2752}&\vc{-.0581}&\vc{.1271}\\
$(0, 0, 0, 0, 0, .4, .6)$&\vc{.4578}&\vc{.4264}&\vc{.3547}&\vc{.2010}&\vc{.1258}&\vc{1.0000}&\vc{-.5650}&\vc{-.5458}&\vc{-.5375}&\vc{-.4694}&\vc{-.1536}&\vc{.1271}\\
$(0, 0, 0, 0, .2, .2, .6)$&\vc{.4986}&\vc{.4748}&\vc{.4102}&\vc{.2657}&\vc{.1994}&\vc{1.0000}&\vc{-.5742}&\vc{-.5595}&\vc{-.5570}&\vc{-.4974}&\vc{-.1665}&\vc{.1271}\\
$(0, 0, 0, 0, 0, .6, .4)$&\vc{.4860}&\vc{.4522}&\vc{.3812}&\vc{.2356}&\vc{.1649}&\vc{1.0000}&\vc{-.5519}&\vc{-.5209}&\vc{-.4986}&\vc{-.4120}&\vc{-.1275}&\vc{.1271}\\
$(0, 0, 0, 0, .2, .6, .2)$&\vc{.5482}&\vc{.5165}&\vc{.4500}&\vc{.3135}&\vc{.2461}&\vc{1.0000}&\vc{-.5436}&\vc{-.5051}&\vc{-.4749}&\vc{-.3796}&\vc{-.1142}&\vc{.1271}\\
$(0, 0, 0, .2, .2, .6, 0)$&\vc{.5950}&\vc{.5628}&\vc{.4990}&\vc{.3679}&\vc{.2986}&\vc{1.0000}&\vc{-.5178}&\vc{-.4719}&\vc{-.4335}&\vc{-.3329}&\vc{-.0915}&\vc{.1271}\\
$(0, 0, 0, 0, .6, .2, .2)$&\vc{.5912}&\vc{.5644}&\vc{.5053}&\vc{.3815}&\vc{.3218}&\vc{1.0000}&\vc{-.5260}&\vc{-.4877}&\vc{-.4589}&\vc{-.3668}&\vc{-.1102}&\vc{.1271}\\
$(0, 0, 0, .2, .6, .2, 0)$&\vc{.6367}&\vc{.6084}&\vc{.5504}&\vc{.4291}&\vc{.3658}&\vc{1.0000}&\vc{-.5030}&\vc{-.4576}&\vc{-.4211}&\vc{-.3239}&\vc{-.0892}&\vc{.1271}\\
$(0, 0, .2, .2, .6, 0, 0)$&\vc{.6463}&\vc{.6185}&\vc{.5607}&\vc{.4398}&\vc{.3768}&\vc{1.0000}&\vc{-.5141}&\vc{-.4700}&\vc{-.4338}&\vc{-.3367}&\vc{-.1022}&\vc{.1271}\\
$(0, 0, 0, .6, .2, .2, 0)$&\vc{.6448}&\vc{.6189}&\vc{.5609}&\vc{.4344}&\vc{.3619}&\vc{1.0000}&\vc{-.4970}&\vc{-.4530}&\vc{-.4147}&\vc{-.3145}&\vc{-.0735}&\vc{.1271}\\
$(0, 0, .2, .6, .2, 0, 0)$&\vc{.6553}&\vc{.6299}&\vc{.5720}&\vc{.4463}&\vc{.3738}&\vc{1.0000}&\vc{-.5089}&\vc{-.4662}&\vc{-.4281}&\vc{-.3282}&\vc{-.0868}&\vc{.1271}\\
$(0, 0, 0, 0, .2, .4, .4)$&\vc{.5418}&\vc{.5176}&\vc{.4532}&\vc{.3108}&\vc{.2425}&\vc{1.0000}&\vc{-.5773}&\vc{-.5533}&\vc{-.5387}&\vc{-.4579}&\vc{-.1449}&\vc{.1271}\\
$(0, 0, 0, .2, .2, .2, .4)$&\vc{.5927}&\vc{.5794}&\vc{.5241}&\vc{.3871}&\vc{.3199}&\vc{1.0000}&\vc{-.5884}&\vc{-.5708}&\vc{-.5623}&\vc{-.4842}&\vc{-.1454}&\vc{.1271}\\
$(0, 0, 0, 0, .4, .4, .2)$&\vc{.5863}&\vc{.5599}&\vc{.4979}&\vc{.3656}&\vc{.3004}&\vc{1.0000}&\vc{-.5499}&\vc{-.5137}&\vc{-.4860}&\vc{-.3918}&\vc{-.1184}&\vc{.1271}\\
$(0, 0, 0, .2, .2, .4, .2)$&\vc{.6129}&\vc{.5918}&\vc{.5315}&\vc{.3927}&\vc{.3200}&\vc{1.0000}&\vc{-.5684}&\vc{-.5362}&\vc{-.5106}&\vc{-.4130}&\vc{-.1173}&\vc{.1271}\\
$(0, 0, 0, .2, .4, .4, 0)$&\vc{.6326}&\vc{.6046}&\vc{.5440}&\vc{.4152}&\vc{.3468}&\vc{1.0000}&\vc{-.5239}&\vc{-.4794}&\vc{-.4426}&\vc{-.3417}&\vc{-.0942}&\vc{.1271}\\
$(0, 0, .2, .2, .2, .4, 0)$&\vc{.6362}&\vc{.6093}&\vc{.5464}&\vc{.4095}&\vc{.3354}&\vc{1.0000}&\vc{-.5568}&\vc{-.5154}&\vc{-.4789}&\vc{-.3739}&\vc{-.1136}&\vc{.1271}\\
$(0, 0, 0, .2, .4, .2, .2)$&\vc{.6358}&\vc{.6174}&\vc{.5611}&\vc{.4302}&\vc{.3628}&\vc{1.0000}&\vc{-.5592}&\vc{-.5267}&\vc{-.5016}&\vc{-.4058}&\vc{-.1153}&\vc{.1271}\\
$(0, 0, 0, .4, .4, .2, 0)$&\vc{.6598}&\vc{.6351}&\vc{.5774}&\vc{.4508}&\vc{.3809}&\vc{1.0000}&\vc{-.5150}&\vc{-.4713}&\vc{-.4344}&\vc{-.3334}&\vc{-.0852}&\vc{.1271}\\
$(0, 0, .2, .2, .4, .2, 0)$&\vc{.6590}&\vc{.6343}&\vc{.5746}&\vc{.4436}&\vc{.3732}&\vc{1.0000}&\vc{-.5493}&\vc{-.5078}&\vc{-.4722}&\vc{-.3693}&\vc{-.1124}&\vc{.1271}\\
$(0, 0, .2, .4, .4, 0, 0)$&\vc{.6703}&\vc{.6461}&\vc{.5886}&\vc{.4626}&\vc{.3931}&\vc{1.0000}&\vc{-.5269}&\vc{-.4846}&\vc{-.4480}&\vc{-.3472}&\vc{-.0990}&\vc{.1271}\\
$(0, .2, .2, .2, .4, 0, 0)$&\vc{.6576}&\vc{.6335}&\vc{.5738}&\vc{.4429}&\vc{.3712}&\vc{1.0000}&\vc{-.5552}&\vc{-.5149}&\vc{-.4802}&\vc{-.3785}&\vc{-.1204}&\vc{.1271}\\
$(0, 0, 0, .4, .2, .2, .2)$&\vc{.6417}&\vc{.6255}&\vc{.5698}&\vc{.4350}&\vc{.3612}&\vc{1.0000}&\vc{-.5578}&\vc{-.5267}&\vc{-.5010}&\vc{-.4023}&\vc{-.1057}&\vc{.1271}\\
$(0, 0, .2, .4, .2, .2, 0)$&\vc{.6653}&\vc{.6420}&\vc{.5822}&\vc{.4479}&\vc{.3722}&\vc{1.0000}&\vc{-.5480}&\vc{-.5073}&\vc{-.4707}&\vc{-.3657}&\vc{-.1042}&\vc{.1271}\\
$(0, 0, .2, .2, .2, .2, .2)$&\vc{.6431}&\vc{.6287}&\vc{.5718}&\vc{.4317}&\vc{.3572}&\vc{1.0000}&\vc{-.5992}&\vc{-.5738}&\vc{-.5531}&\vc{-.4559}&\vc{-.1437}&\vc{.1271}\\
$(0, .2, .2, .2, .2, .2, 0)$&\vc{.6537}&\vc{.6311}&\vc{.5693}&\vc{.4288}&\vc{.3500}&\vc{1.0000}&\vc{-.5797}&\vc{-.5421}&\vc{-.5084}&\vc{-.4022}&\vc{-.1278}&\vc{.1271}\\
\bottomrule\end{tabular}}
\end{table*}

\begin{figure*}[p]
\centering%
\renewcommand{\arraystretch}{0.25}%
\renewcommand{\tabcolsep}{10pt}%
\begin{tabular}{ccc}%
\includegraphics[height=2.8cm]{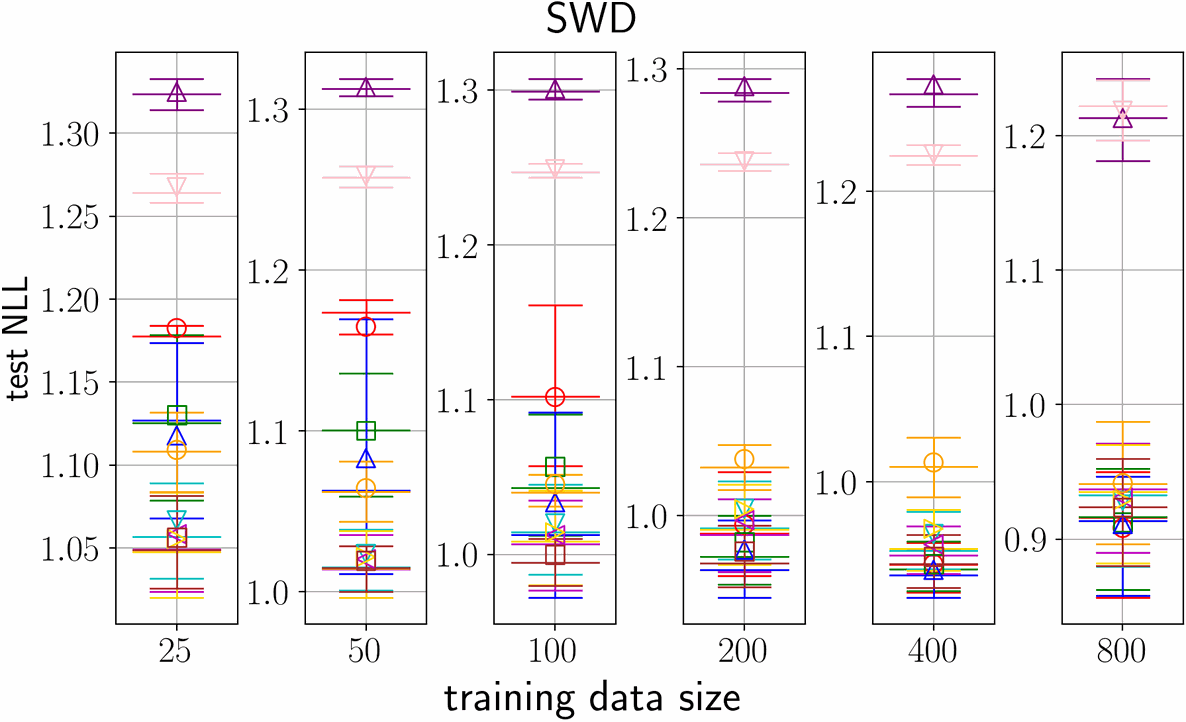}&
\includegraphics[height=2.8cm]{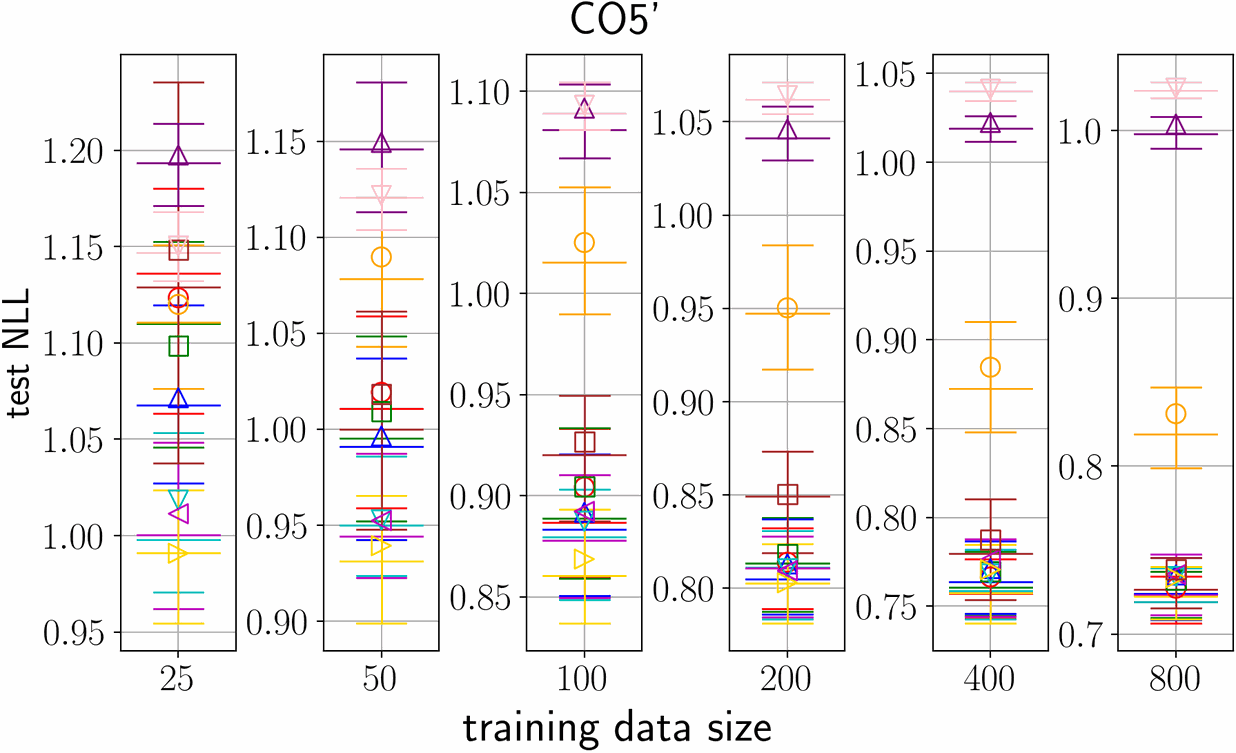}&
\includegraphics[height=2.8cm]{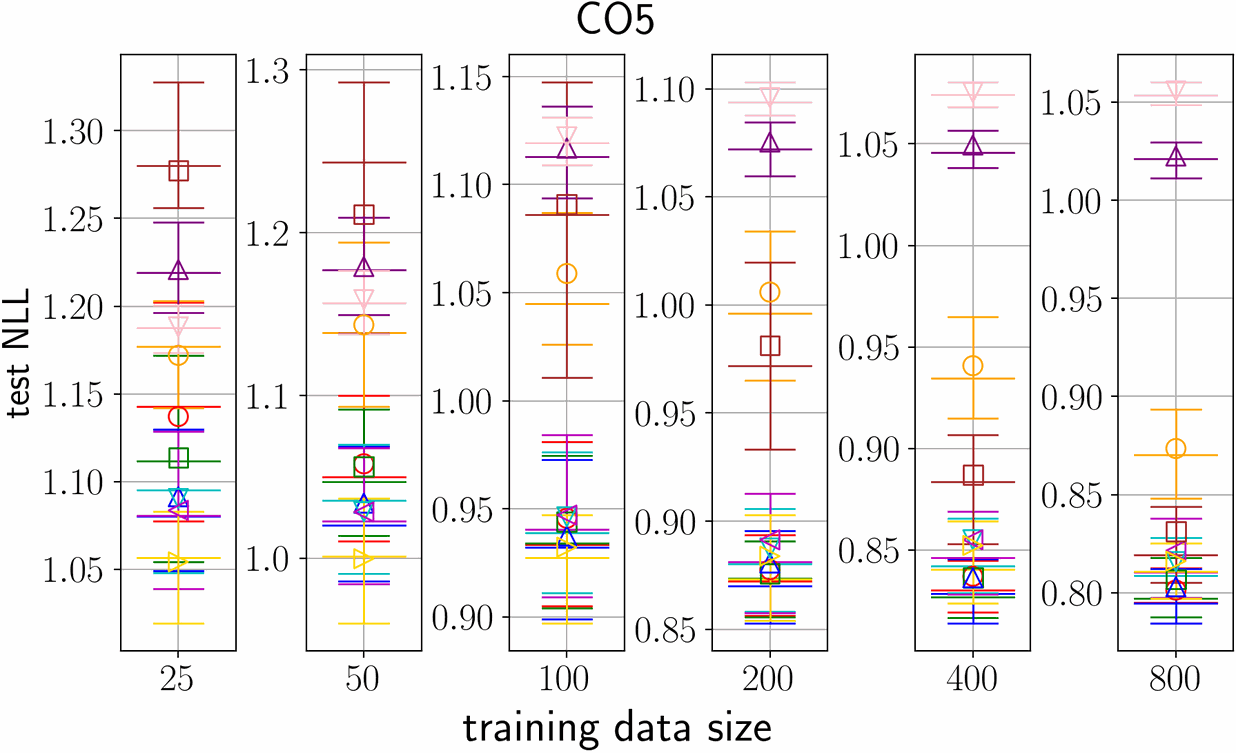}\\
\includegraphics[height=2.8cm]{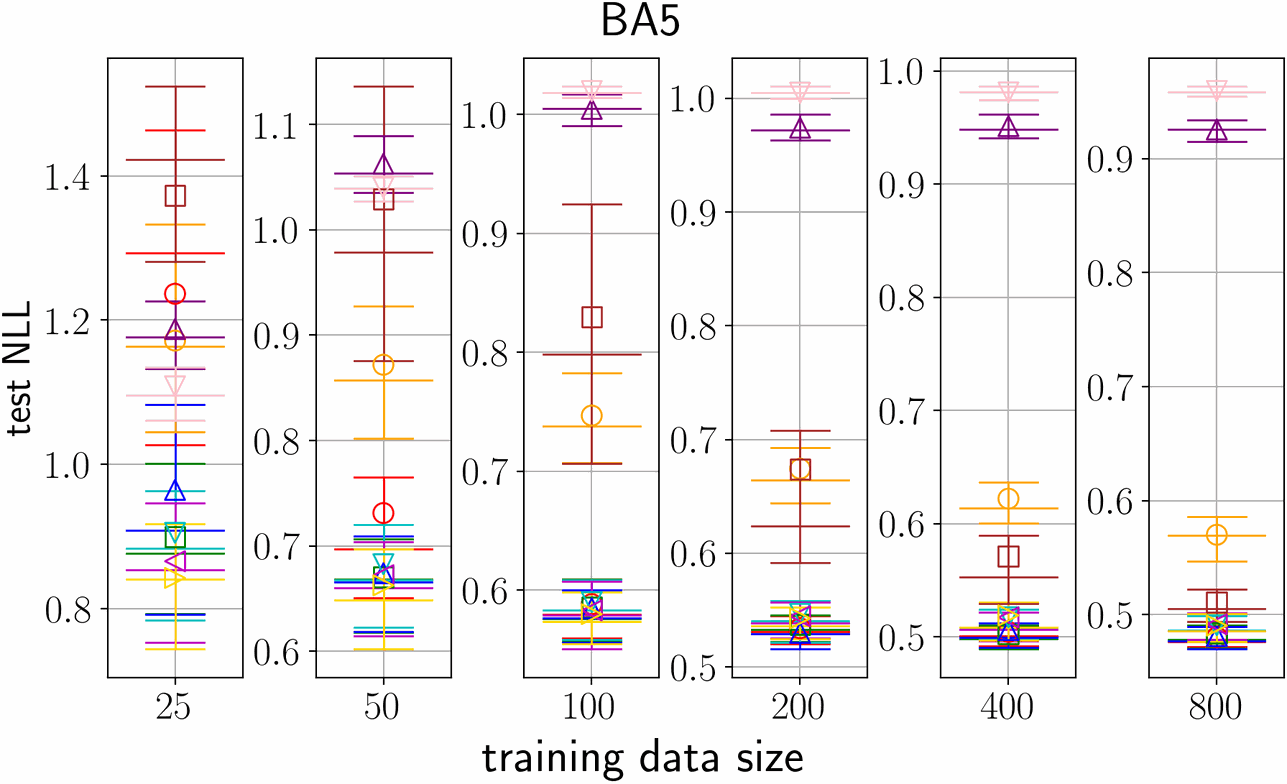}&
\includegraphics[height=2.8cm]{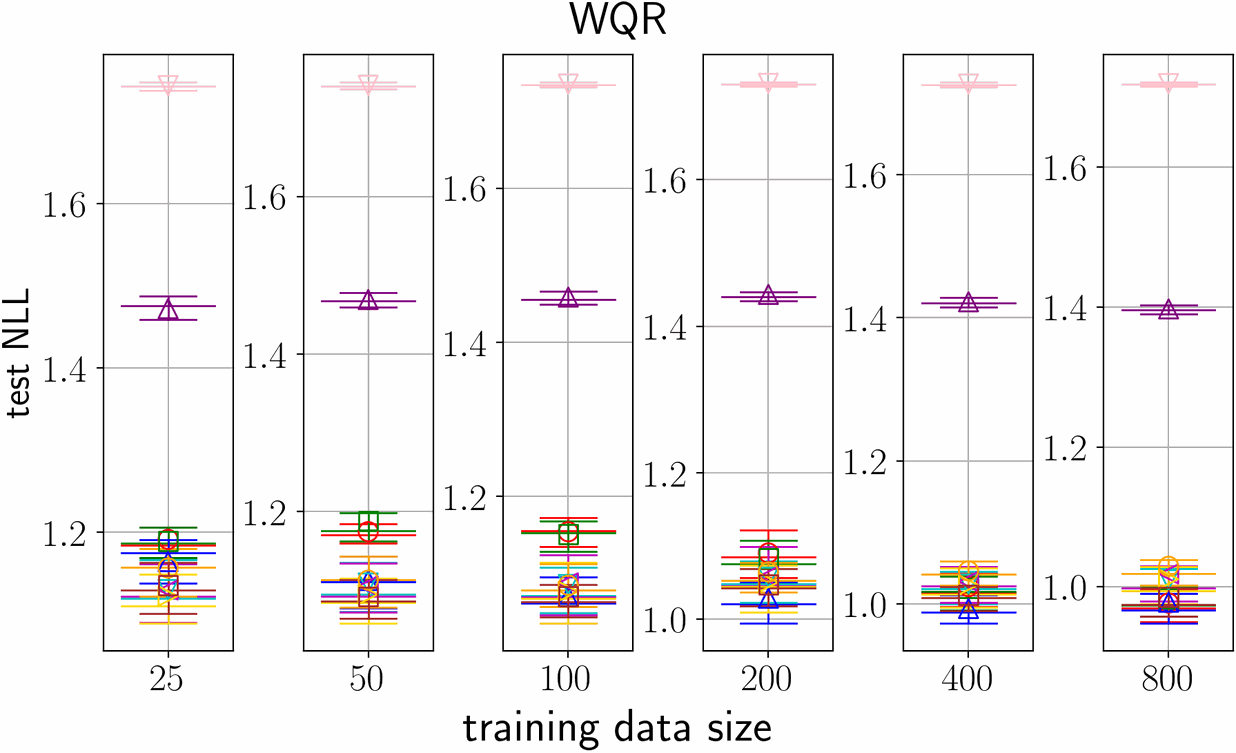}&
\includegraphics[height=2.8cm]{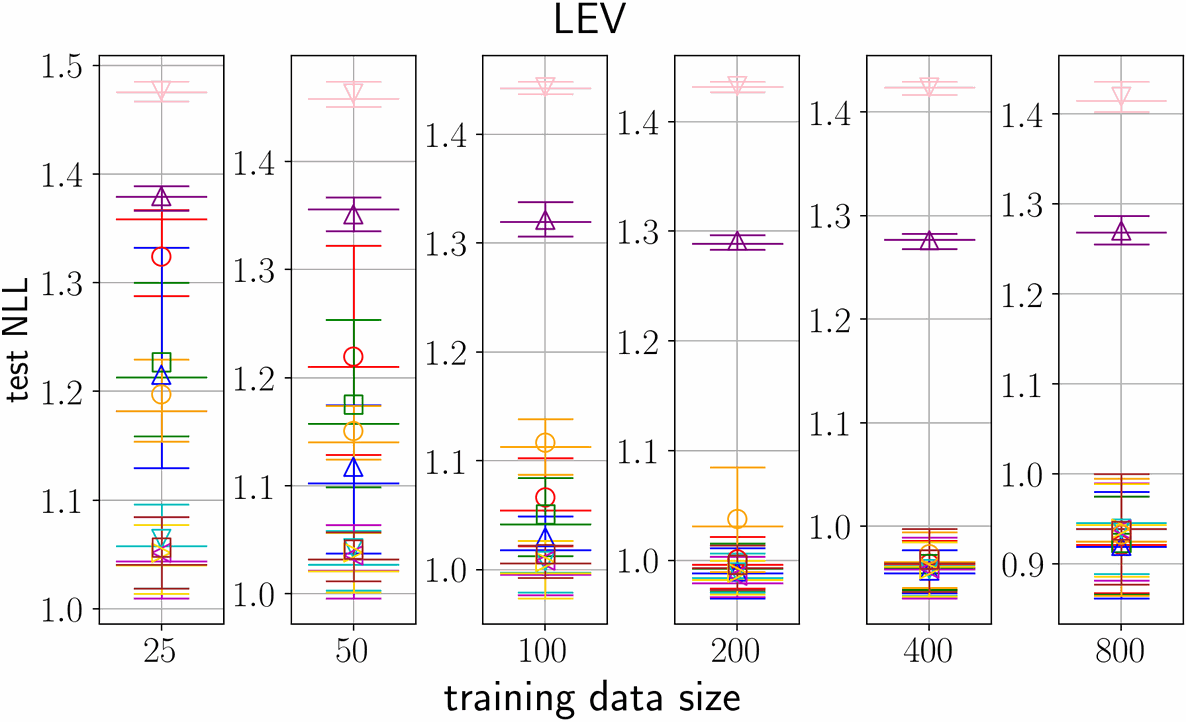}\\
\includegraphics[height=2.8cm]{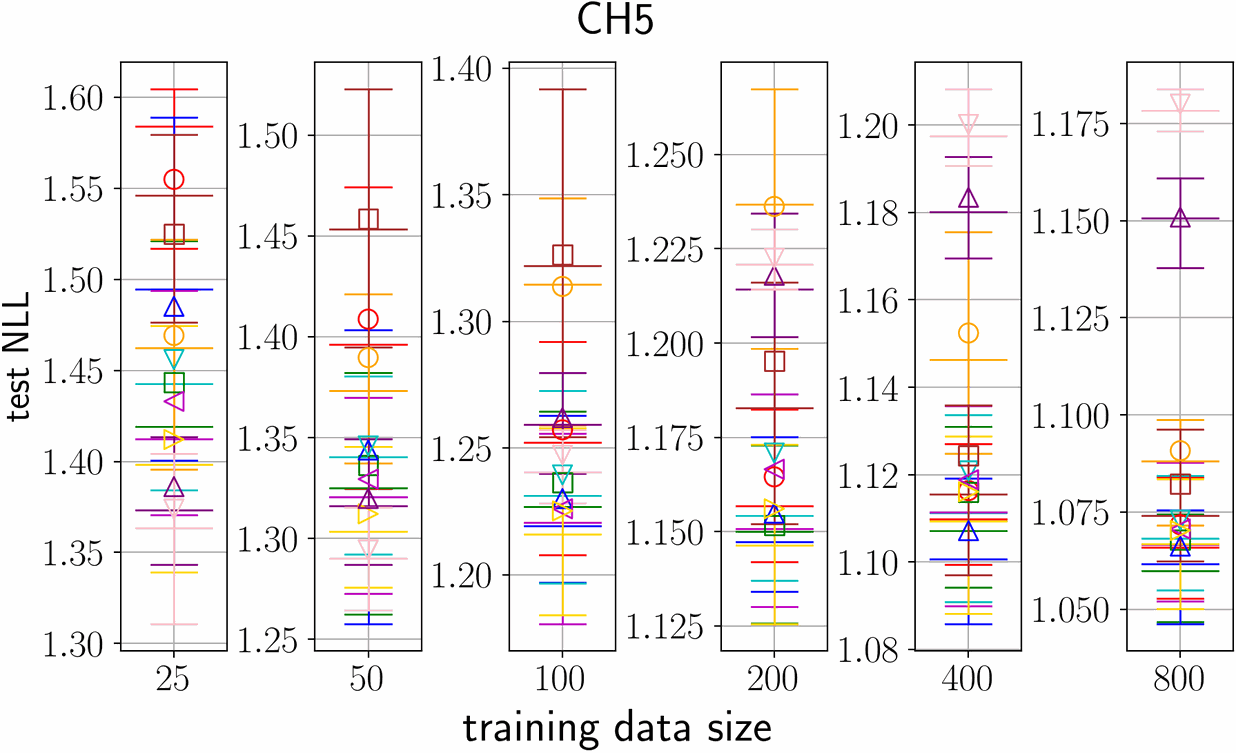}&
\includegraphics[height=2.8cm]{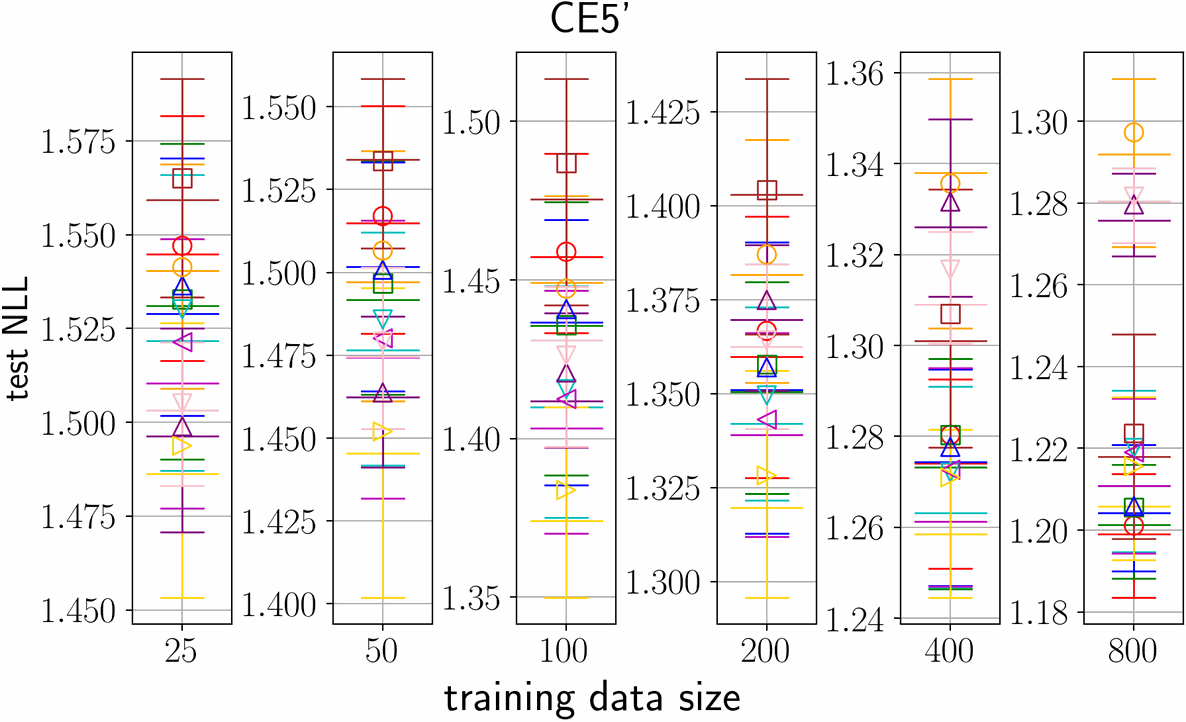}&
\includegraphics[height=2.8cm]{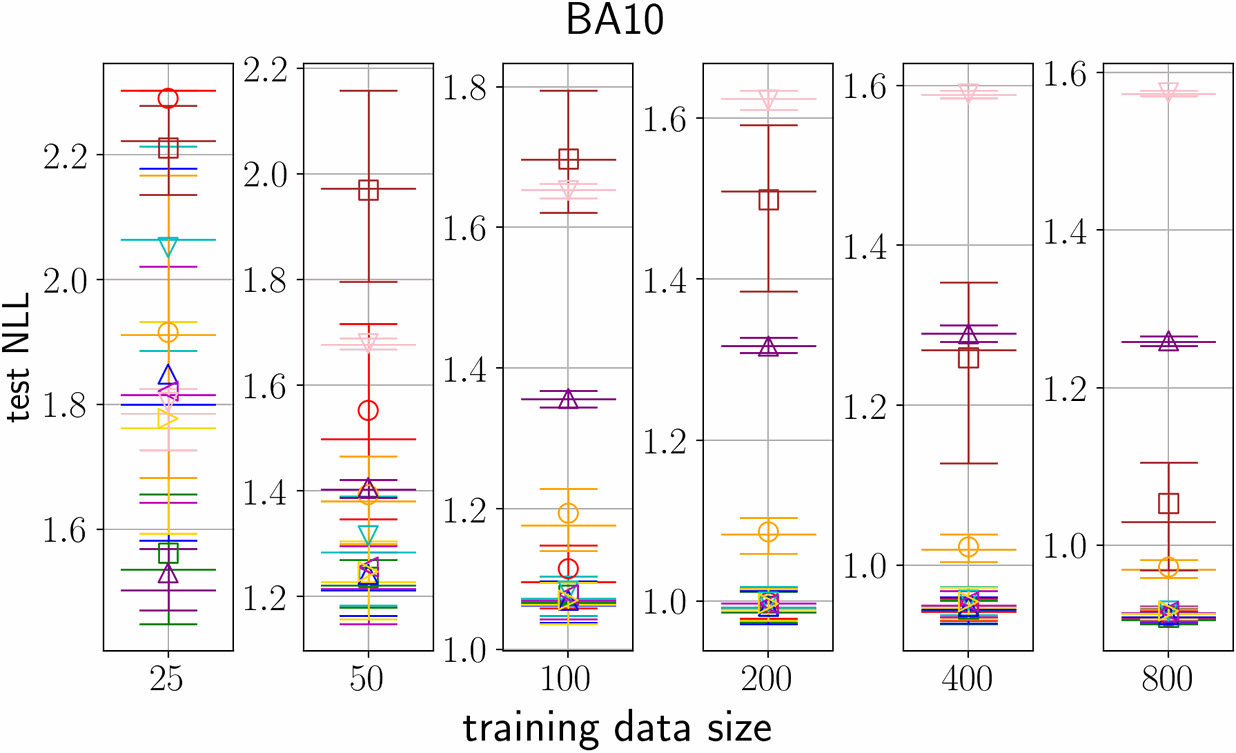}\\
\includegraphics[height=2.8cm]{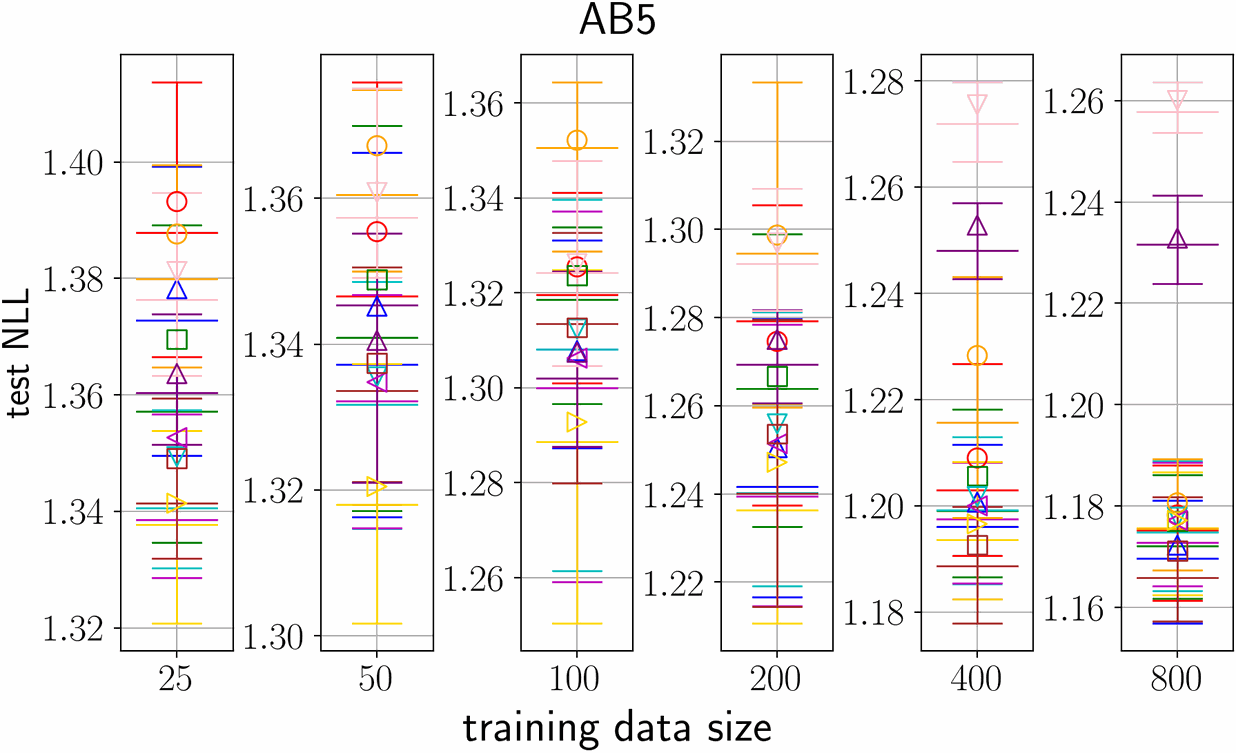}&
\includegraphics[height=2.8cm]{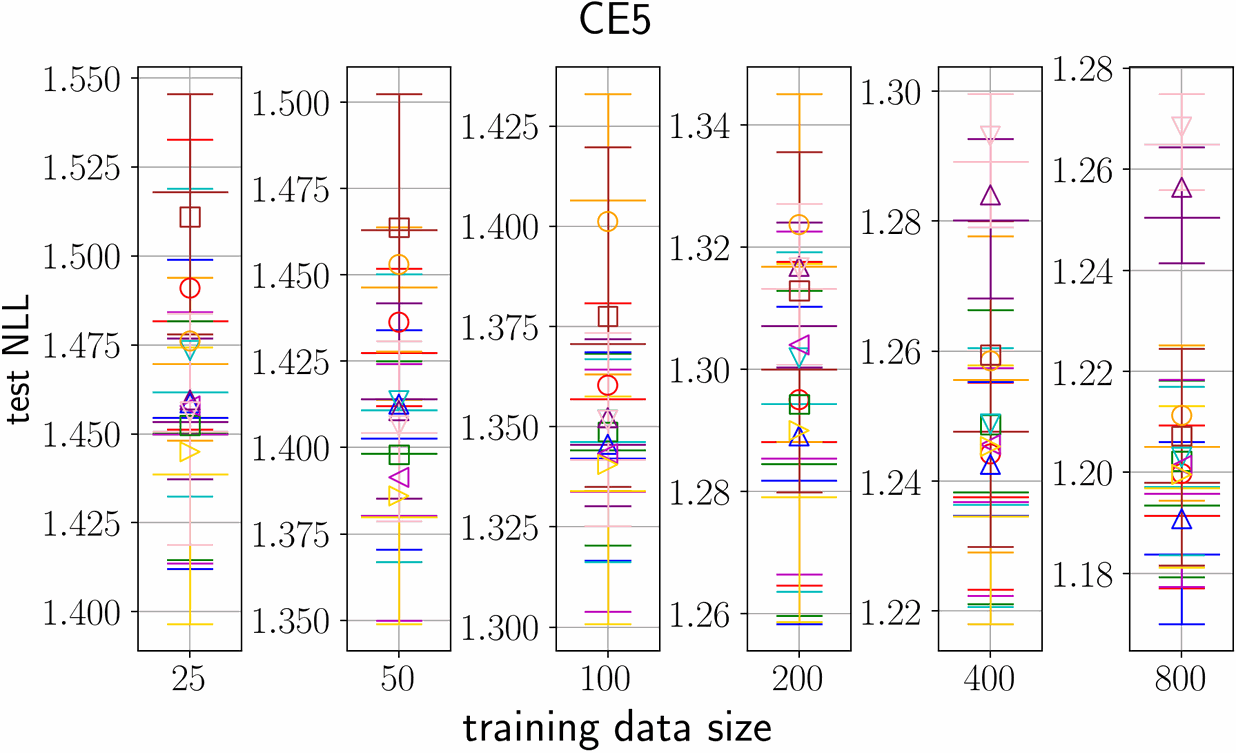}&
\includegraphics[height=2.8cm]{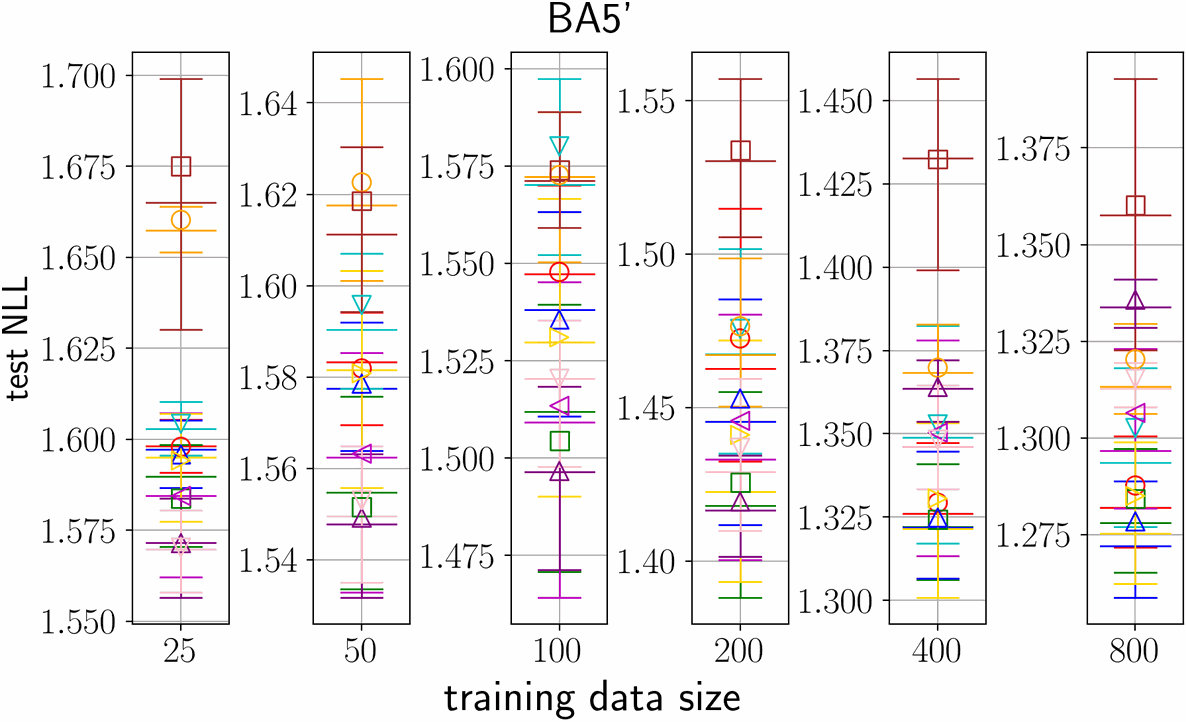}\\
\includegraphics[height=2.8cm]{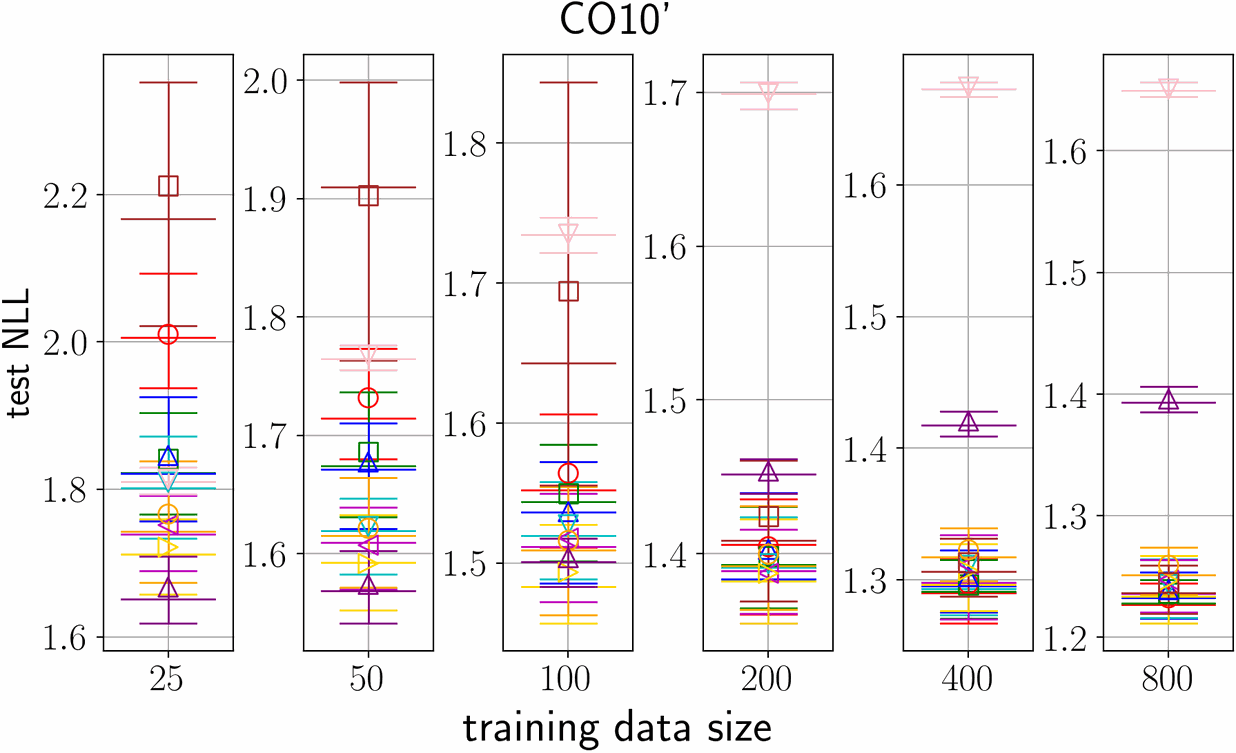}&
\includegraphics[height=2.8cm]{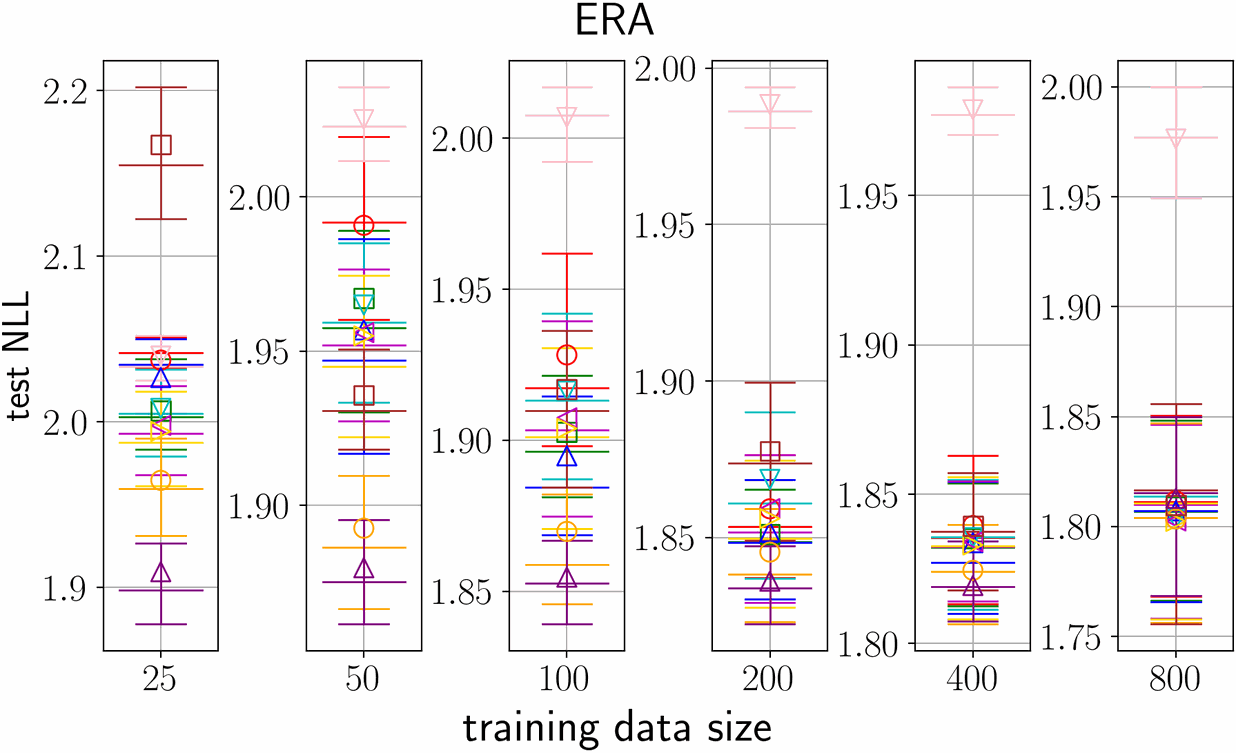}&
\includegraphics[height=2.8cm]{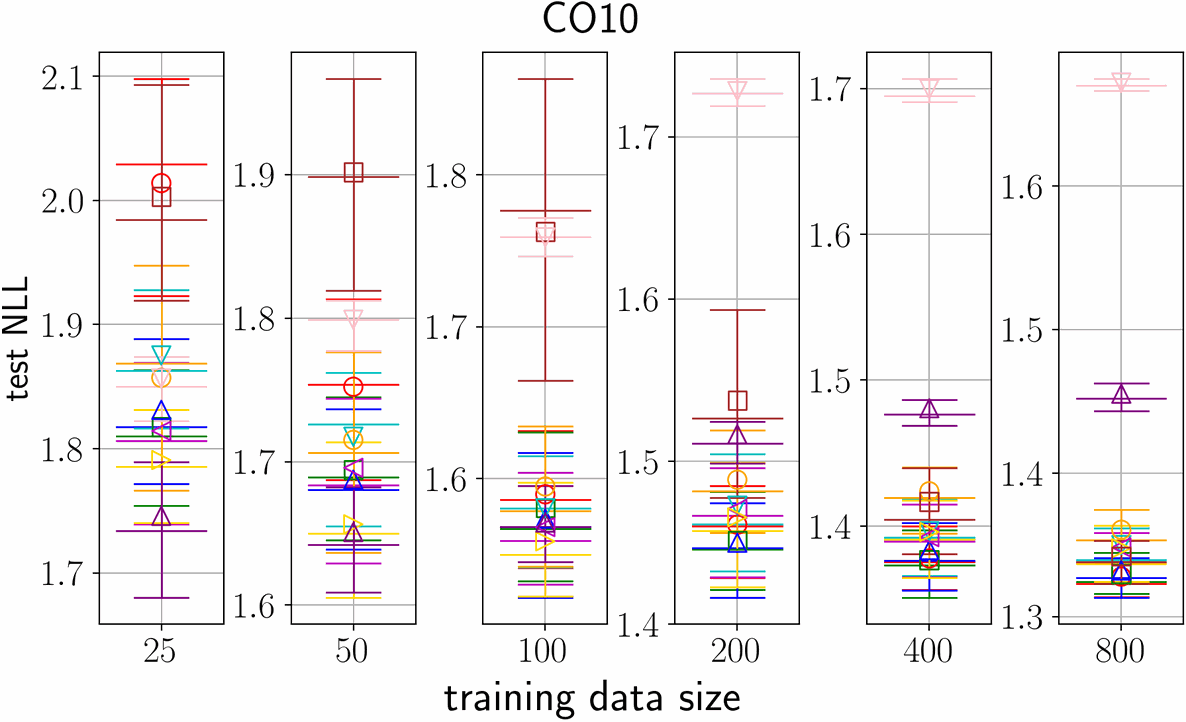}\\
\includegraphics[height=2.8cm]{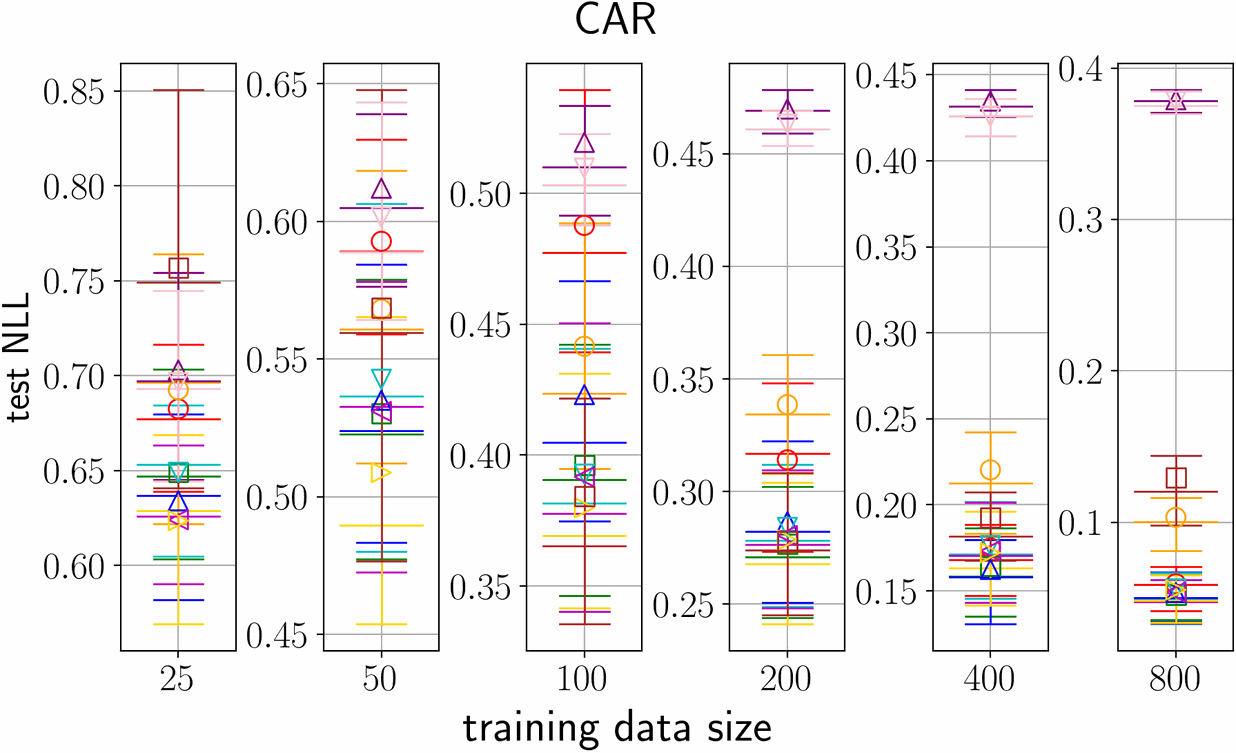}&
\includegraphics[height=2.8cm]{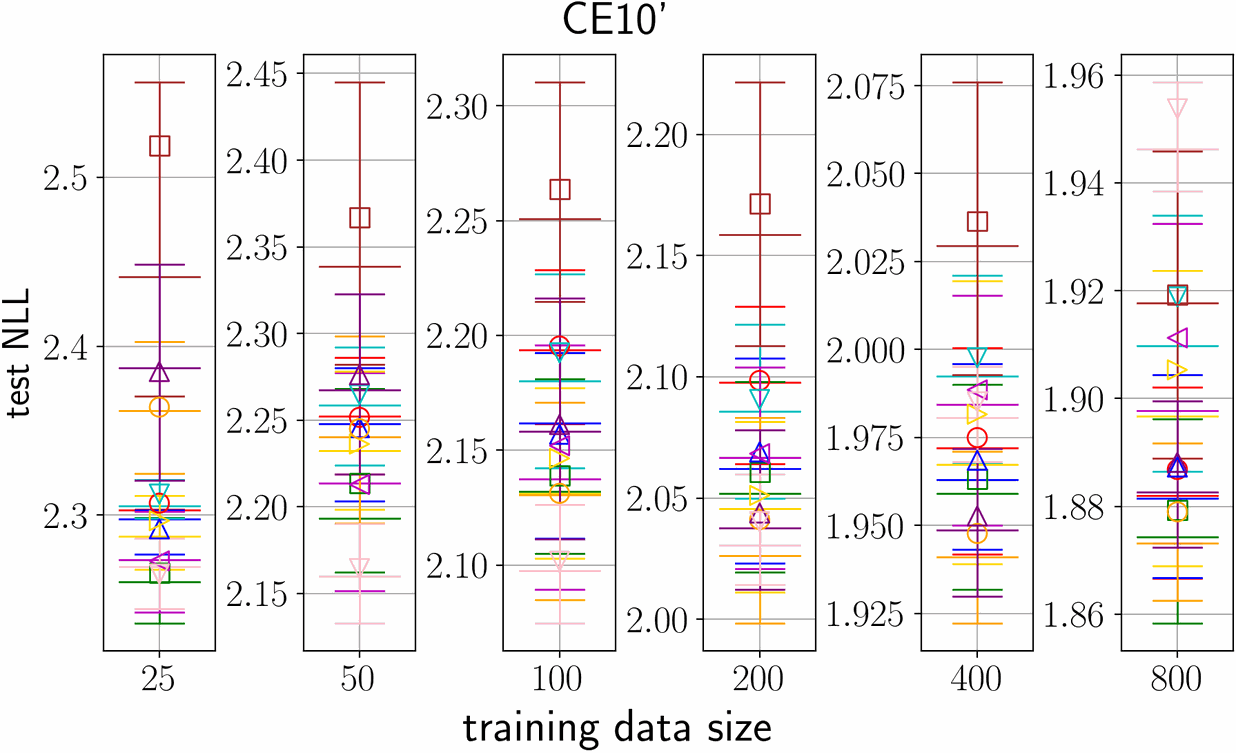}&
\includegraphics[height=2.8cm]{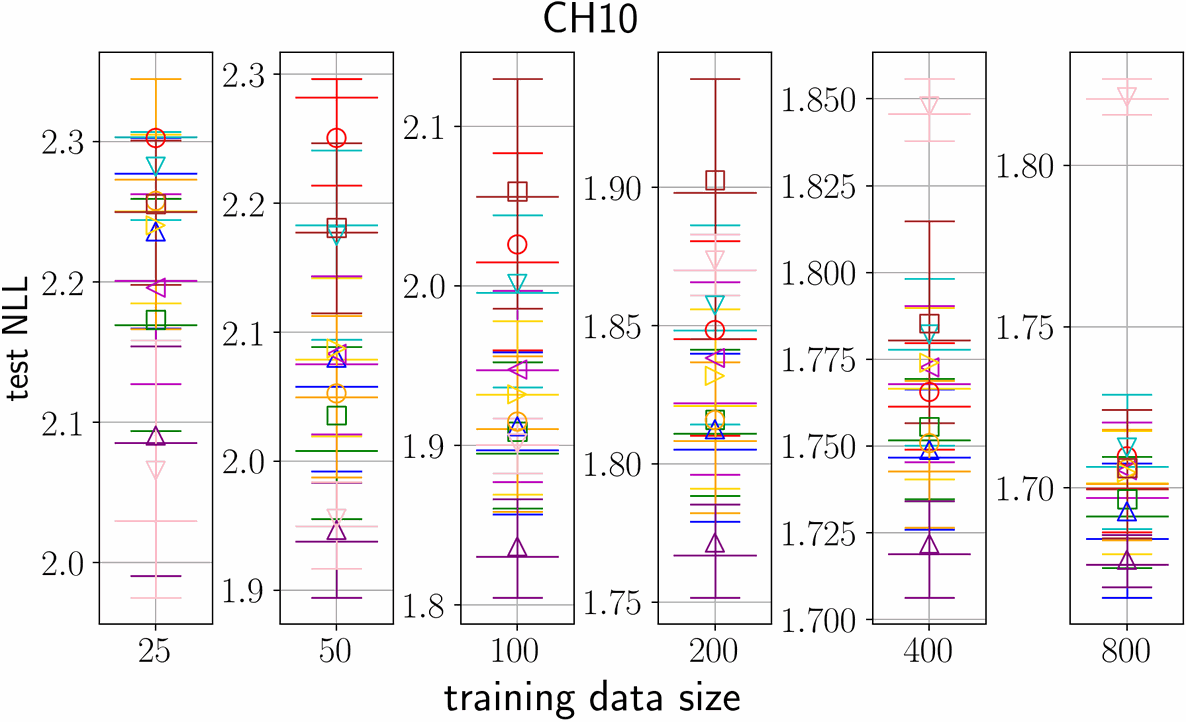}\\
\includegraphics[height=2.8cm]{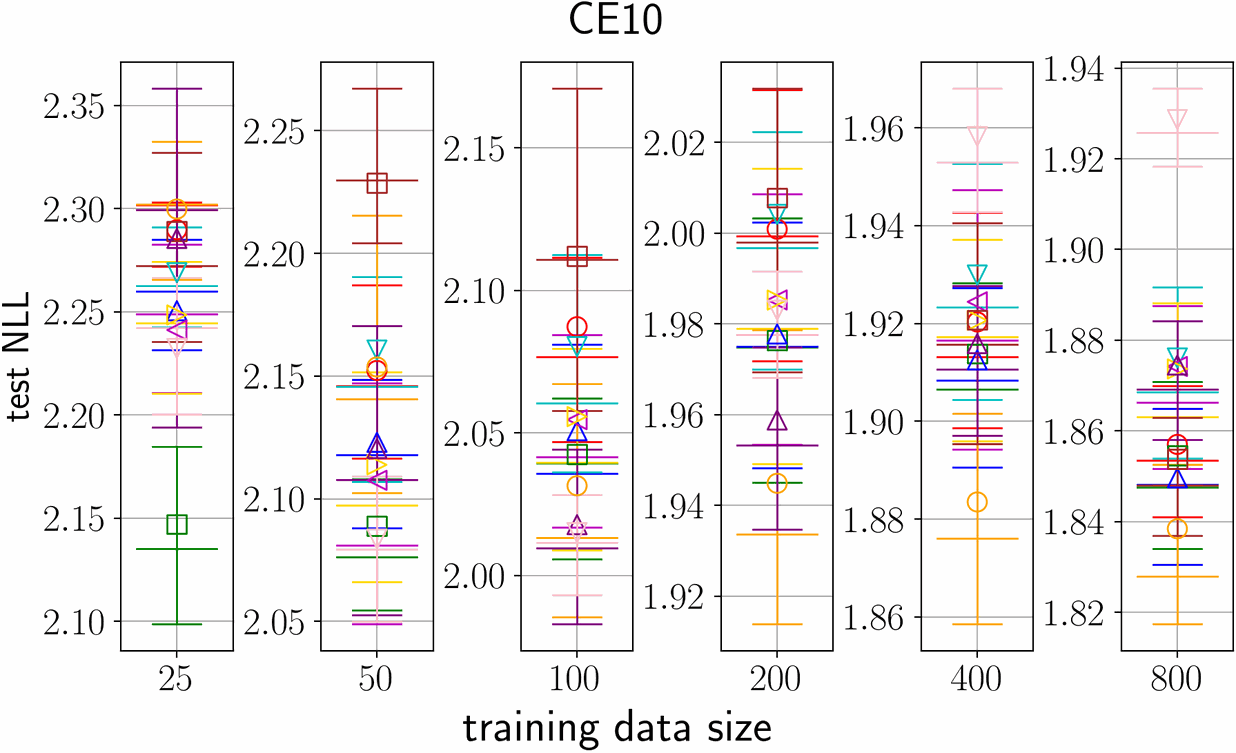}&
\includegraphics[height=2.8cm]{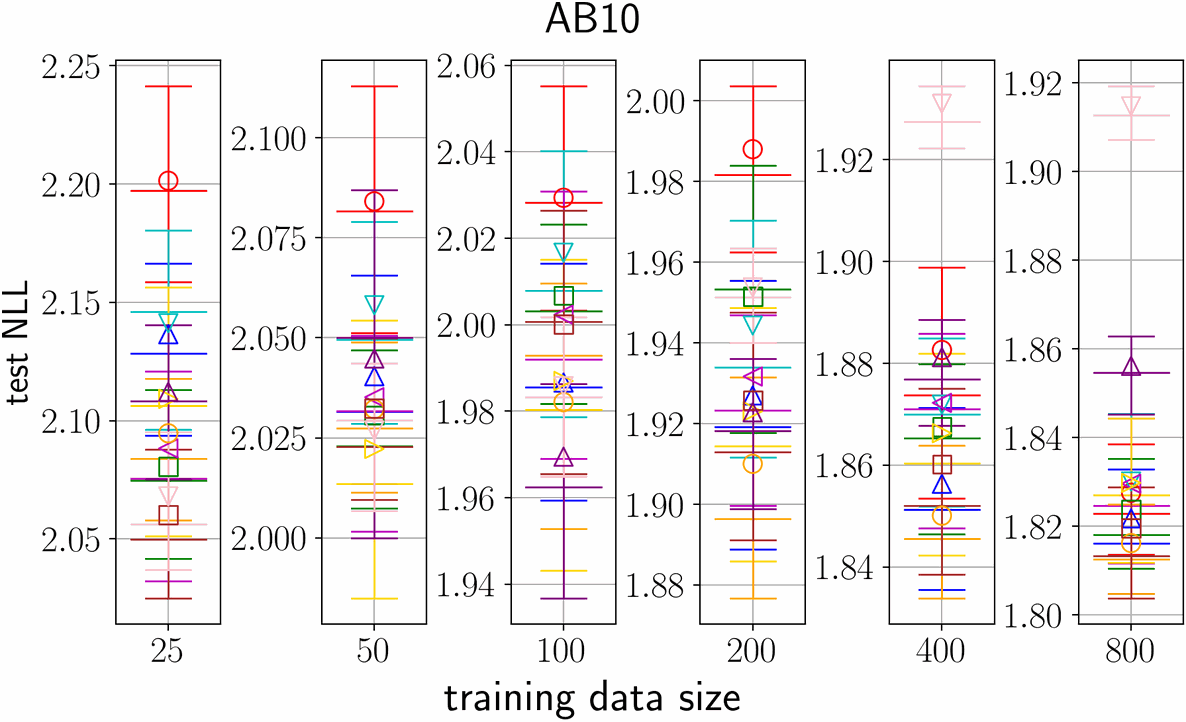}&
\includegraphics[height=2.8cm]{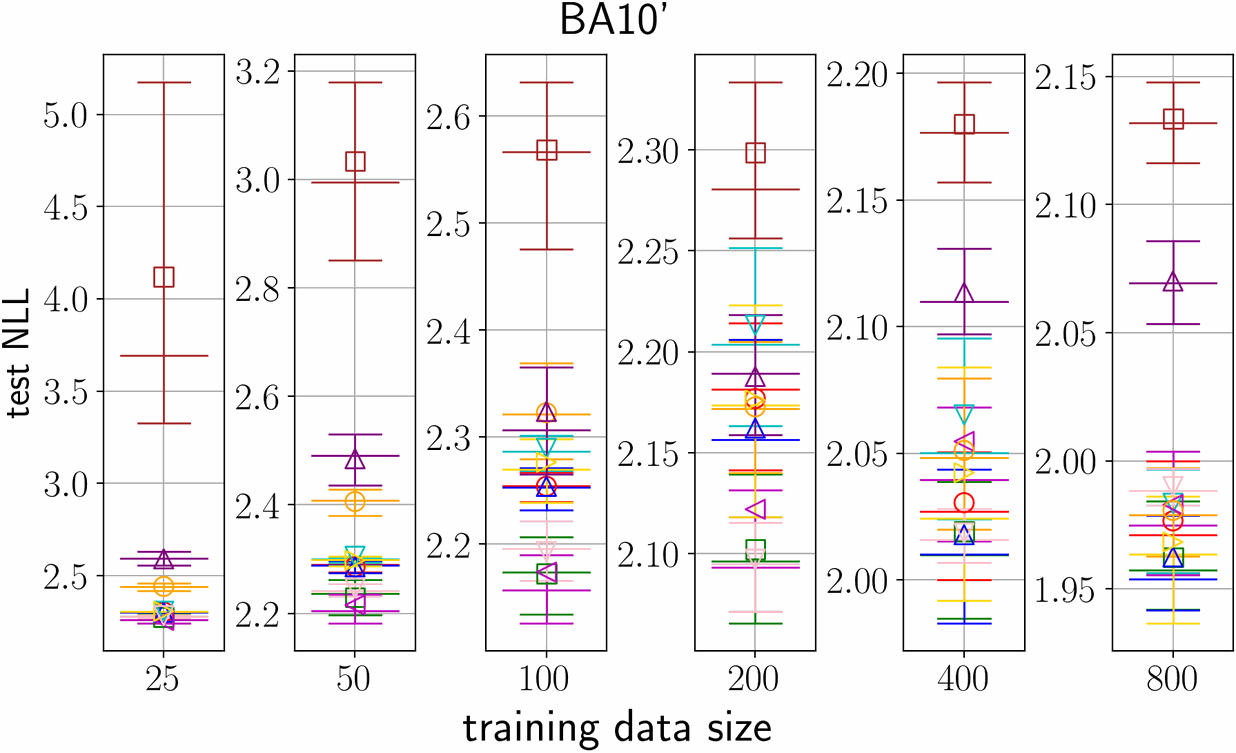}
\end{tabular}
\caption{%
Plot regarding 100-trial test NLLs
by Nonr-MLR (red solid \protect$\circ$), Prev-MLR (green solid \protect$\Box$), 
Stri-MLR (blue solid \protect$\triangle$), Nonr-AUL (cyan solid \protect$\triangledown$), 
Prev-AUL (magenta solid \protect$\triangleleft$), Stri-AUL (yellow solid \protect$\triangleright$),
Nonr-CL (orange dashed \protect$\circ$), Nonr-UL (brown dashed \protect$\Box$), 
Nonr-POCL (purple dashed \protect$\triangle$), and Nonr-POUL (pink dashed \protect$\triangledown$).
Lower short, middle long, and upper short bars and marker 
represent 0.25, 0.5, and 0.75 quantiles and mean.}
\label{fig:Performance-BestLam-NLL}
\end{figure*}
\begin{figure*}[p]
\centering%
\renewcommand{\arraystretch}{0.25}%
\renewcommand{\tabcolsep}{10pt}%
\begin{tabular}{ccc}%
\includegraphics[height=2.8cm]{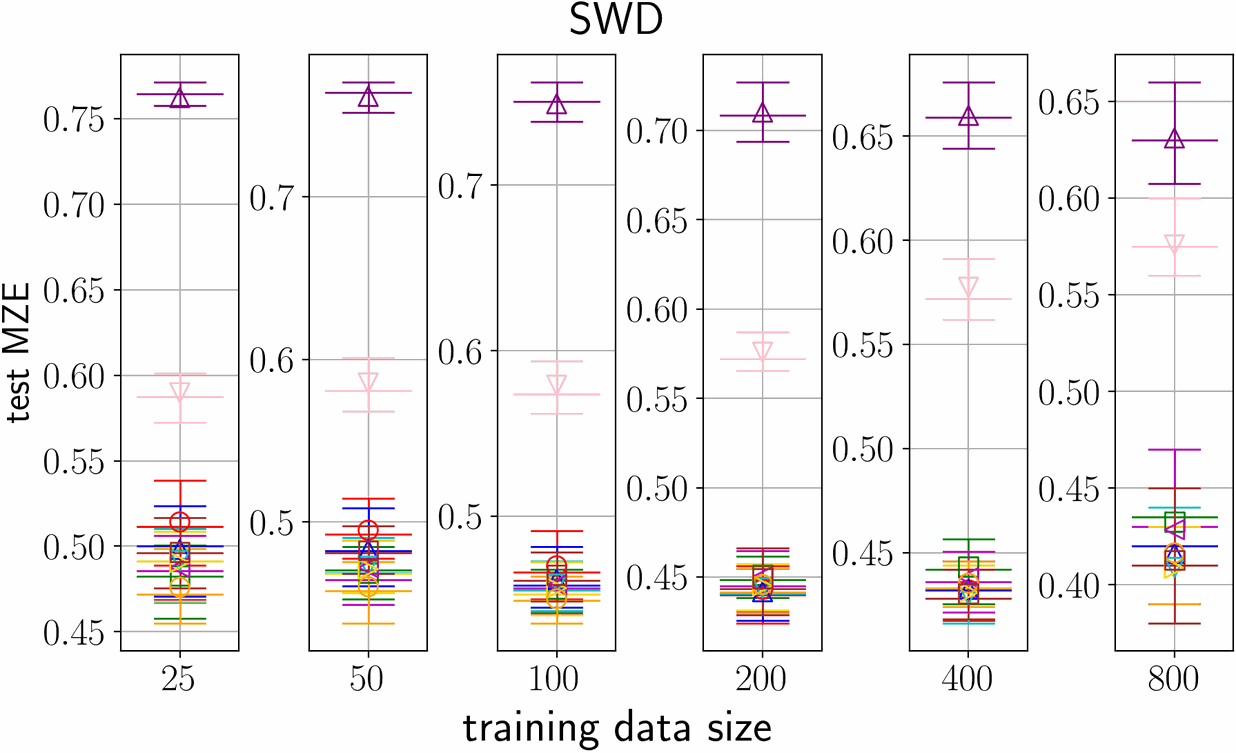}&
\includegraphics[height=2.8cm]{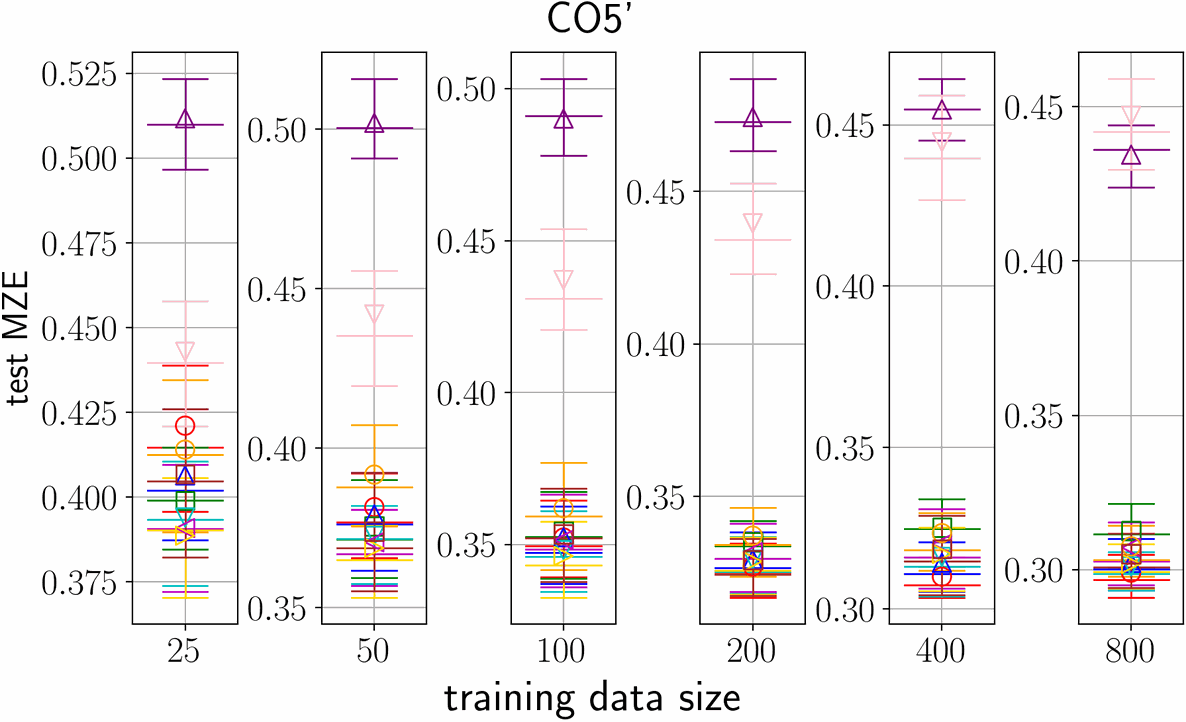}&
\includegraphics[height=2.8cm]{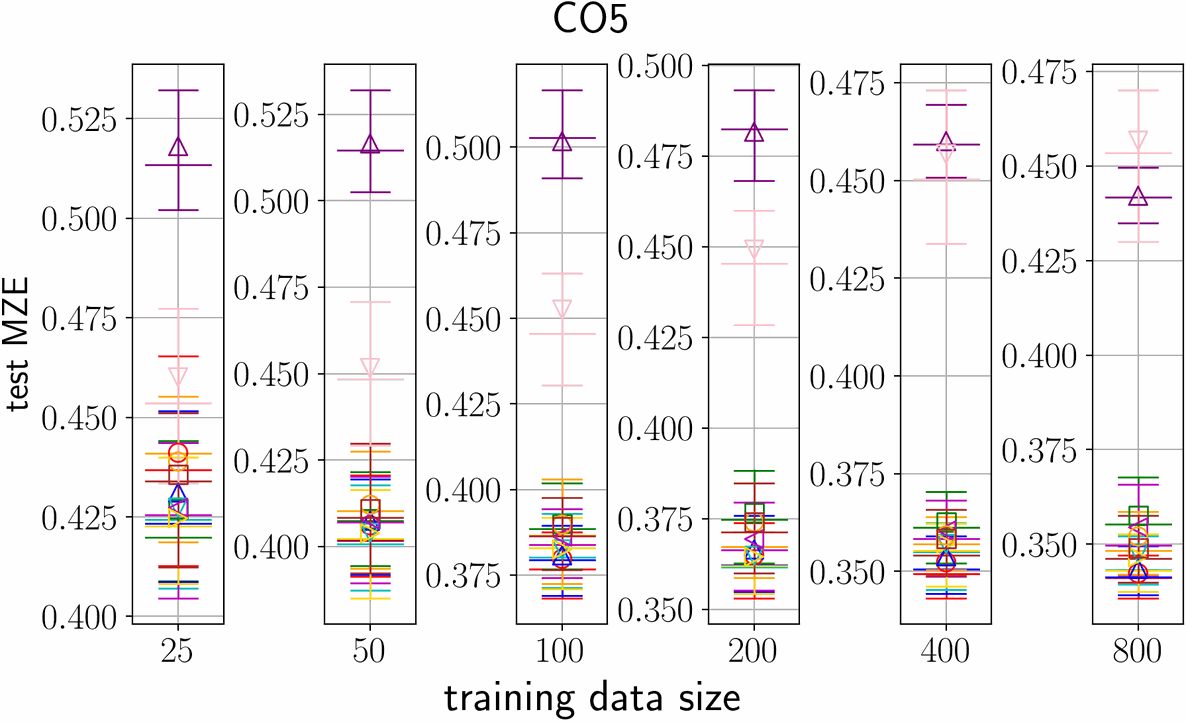}\\
\includegraphics[height=2.8cm]{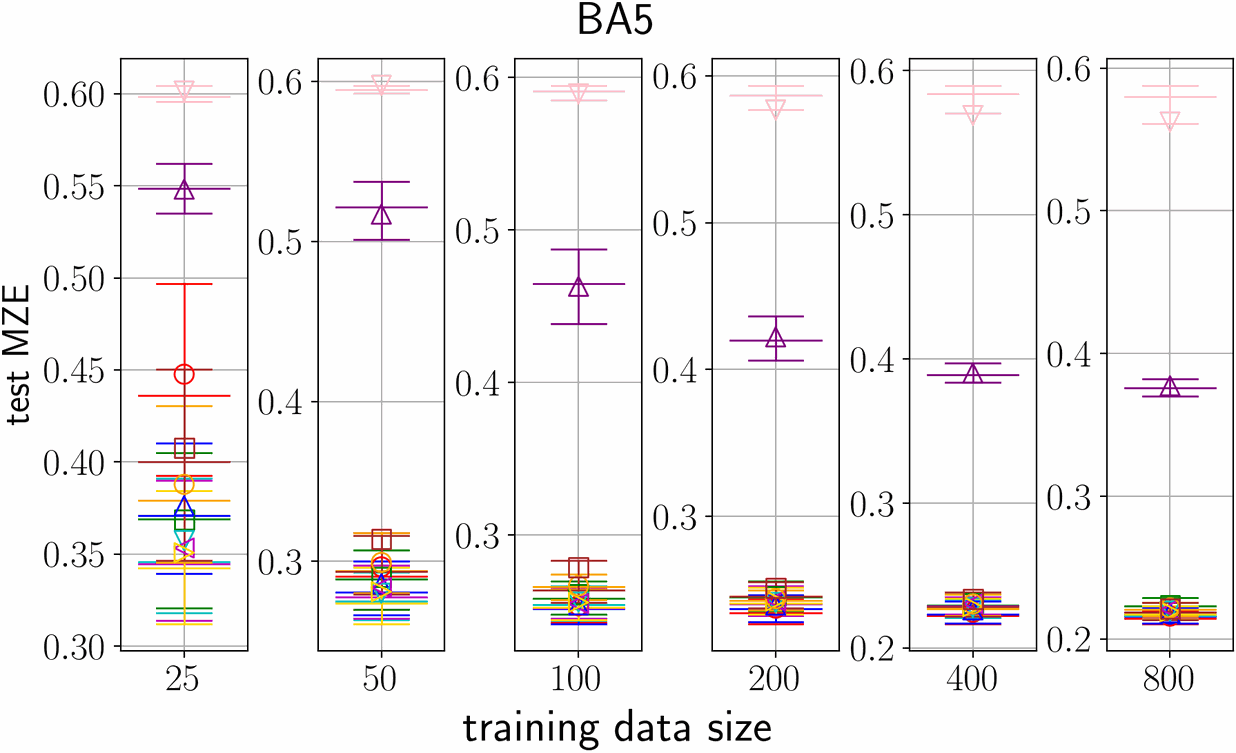}&
\includegraphics[height=2.8cm]{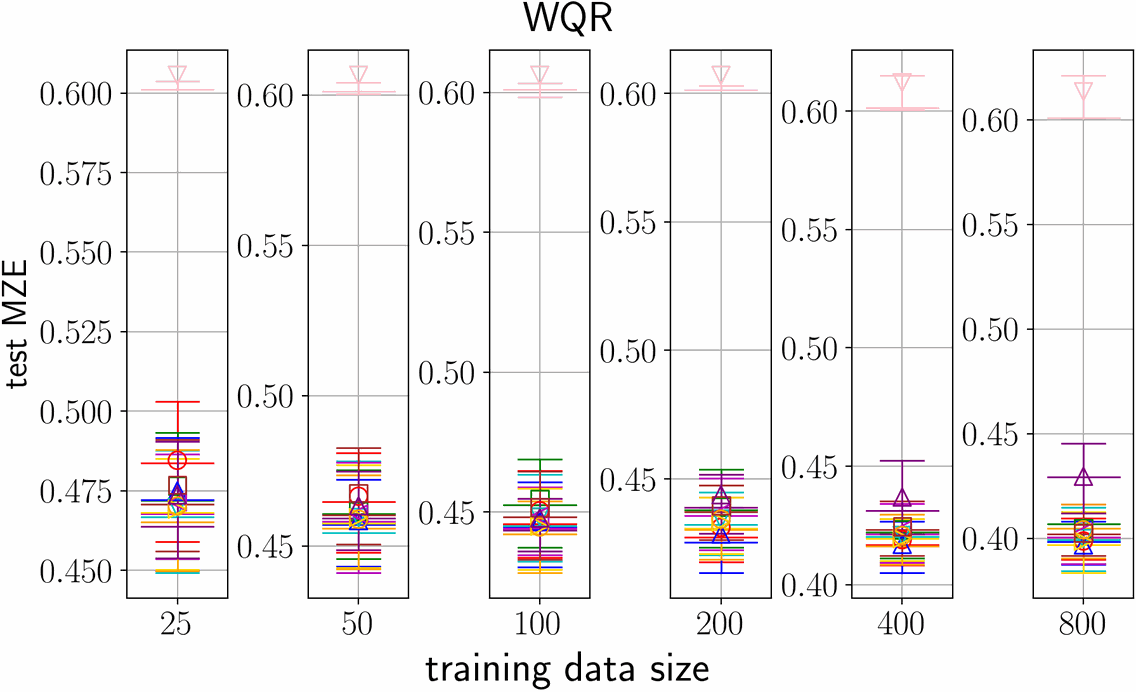}&
\includegraphics[height=2.8cm]{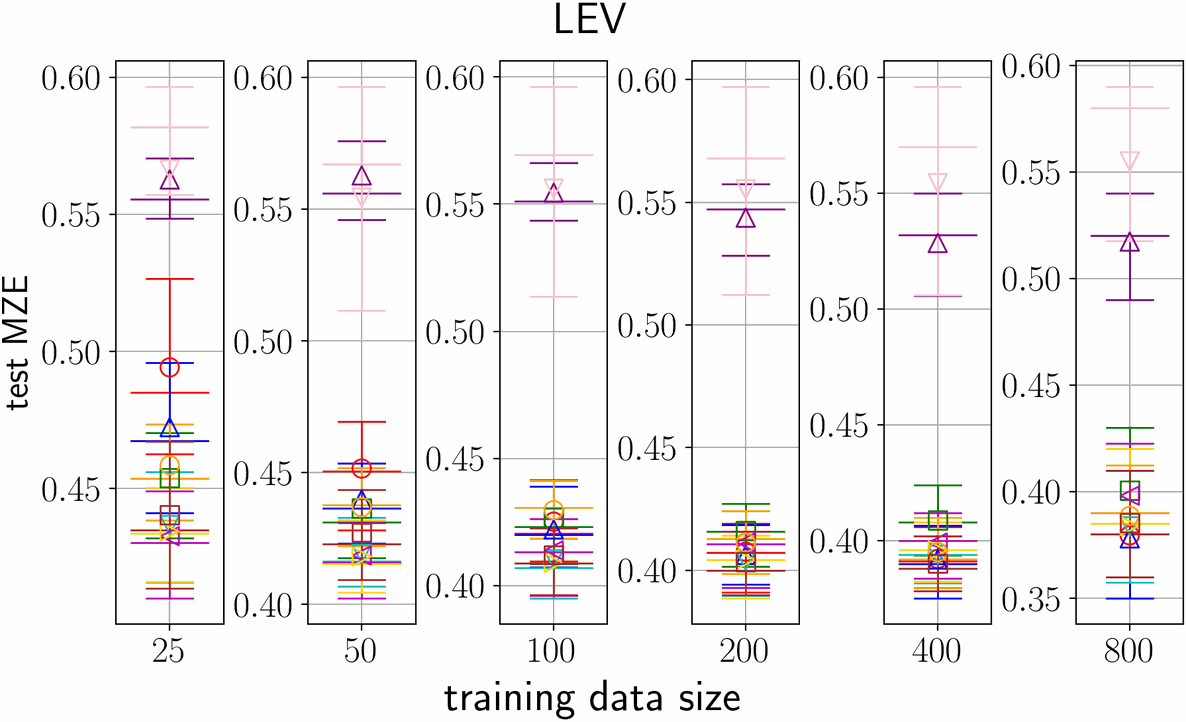}\\
\includegraphics[height=2.8cm]{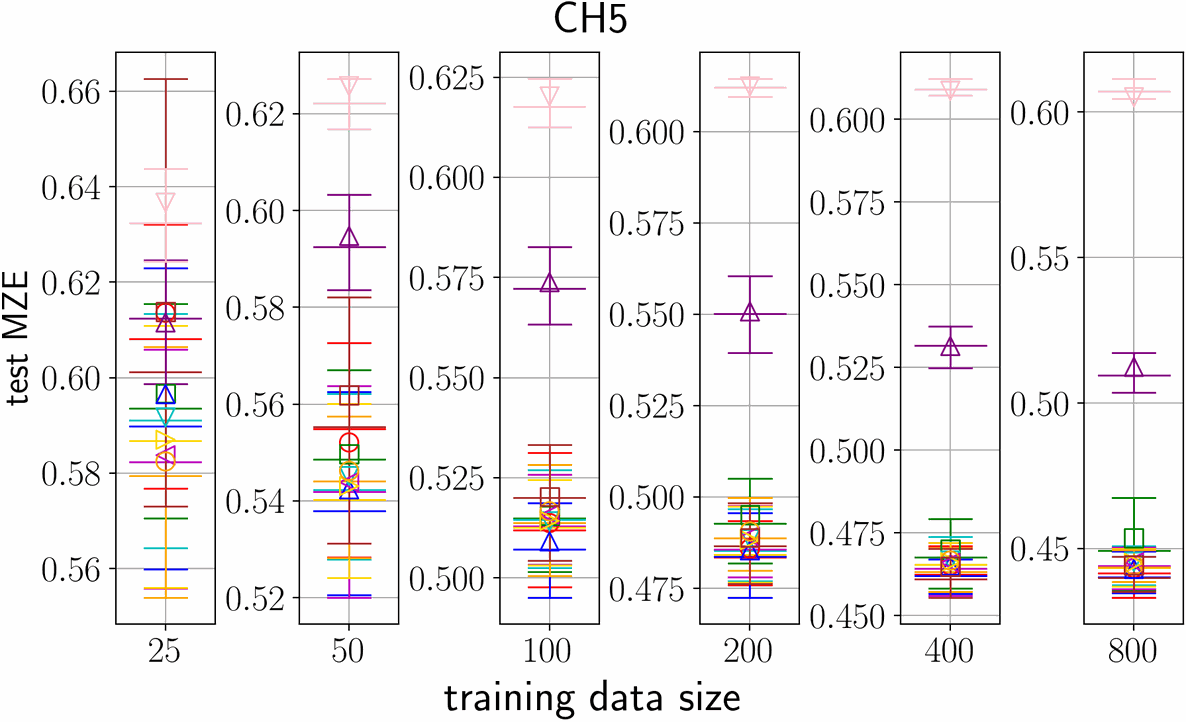}&
\includegraphics[height=2.8cm]{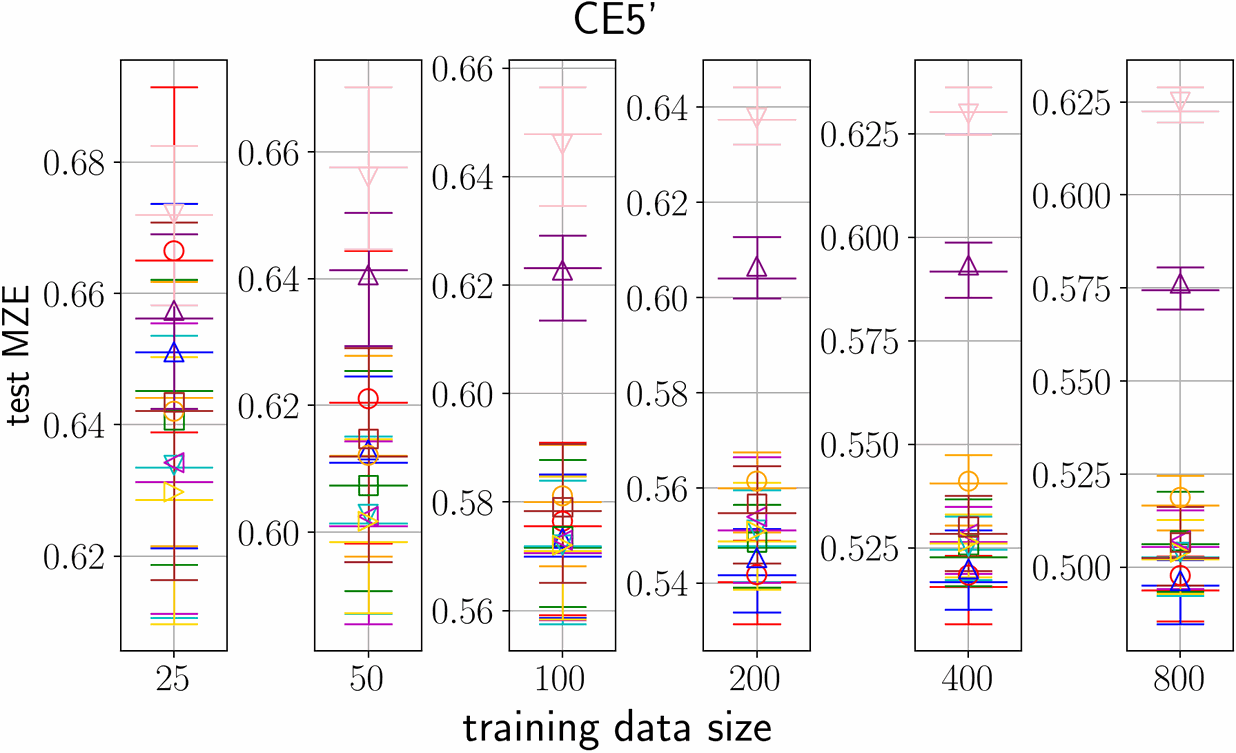}&
\includegraphics[height=2.8cm]{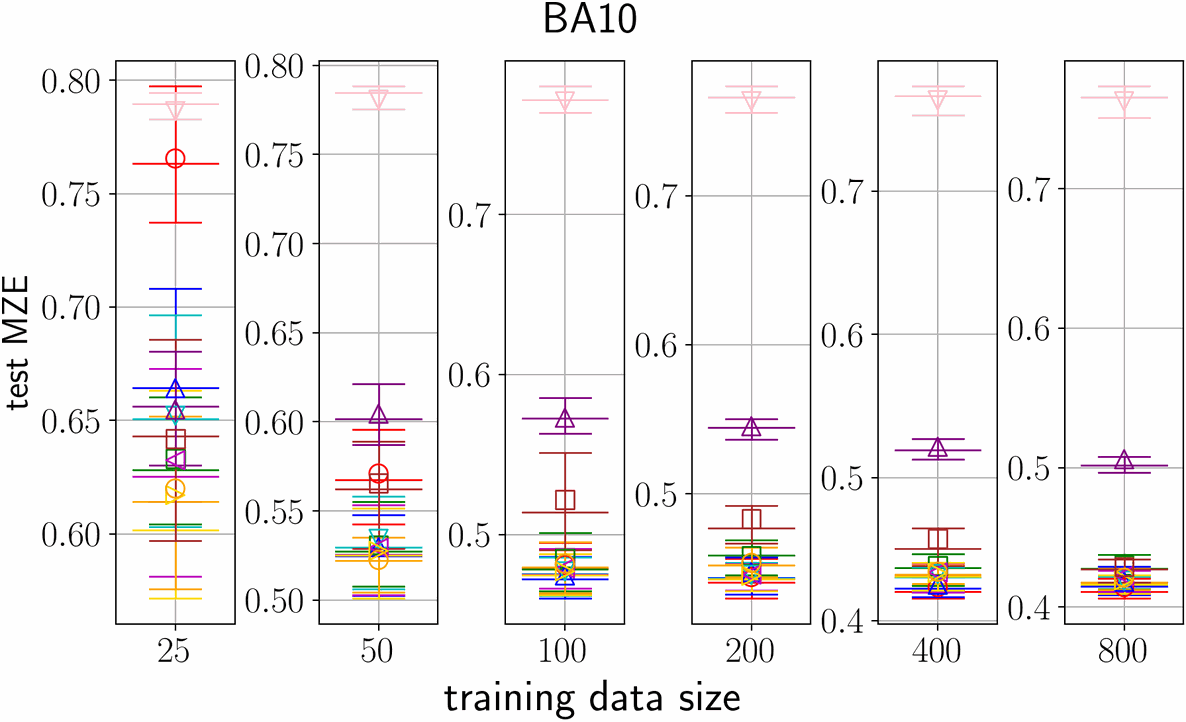}\\
\includegraphics[height=2.8cm]{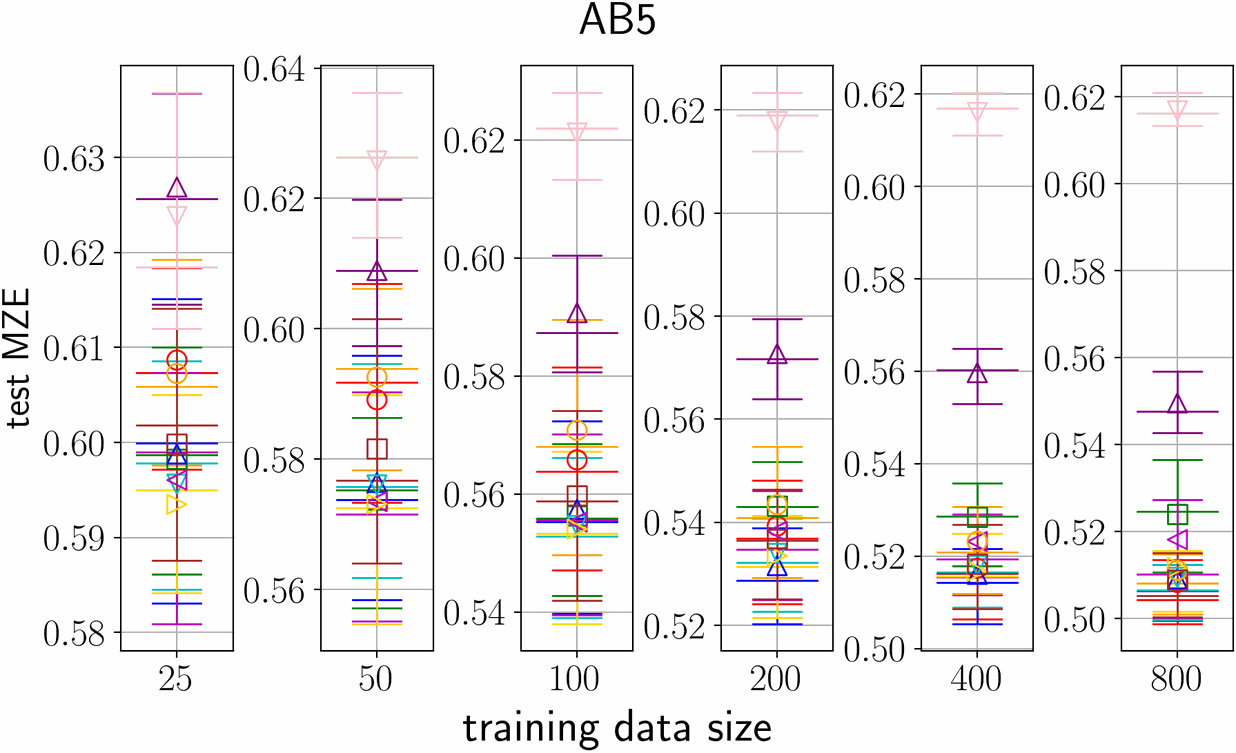}&
\includegraphics[height=2.8cm]{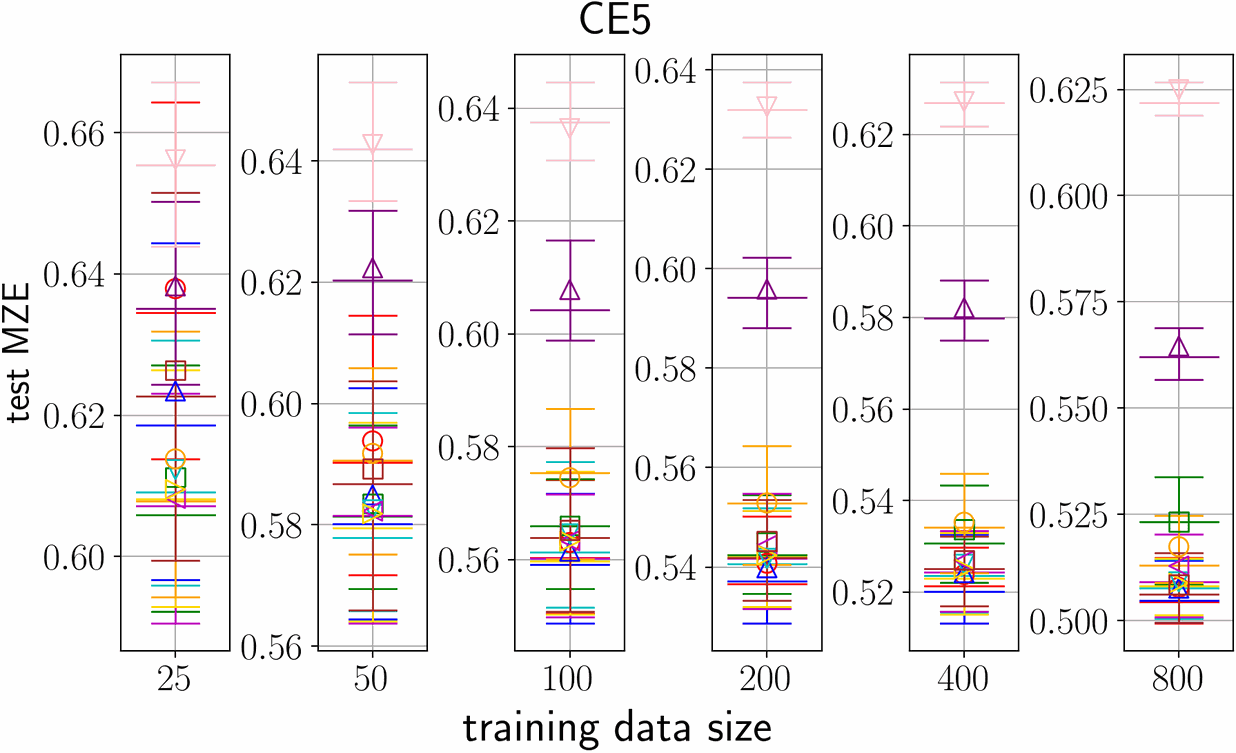}&
\includegraphics[height=2.8cm]{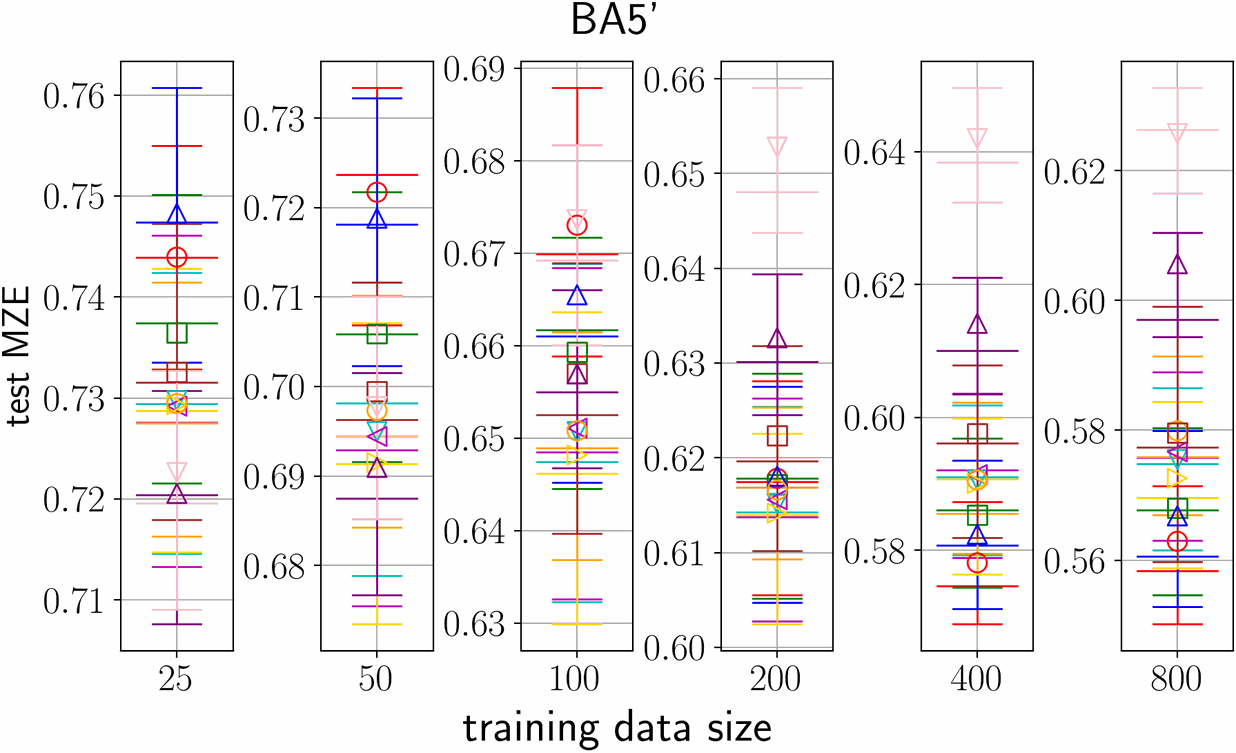}\\
\includegraphics[height=2.8cm]{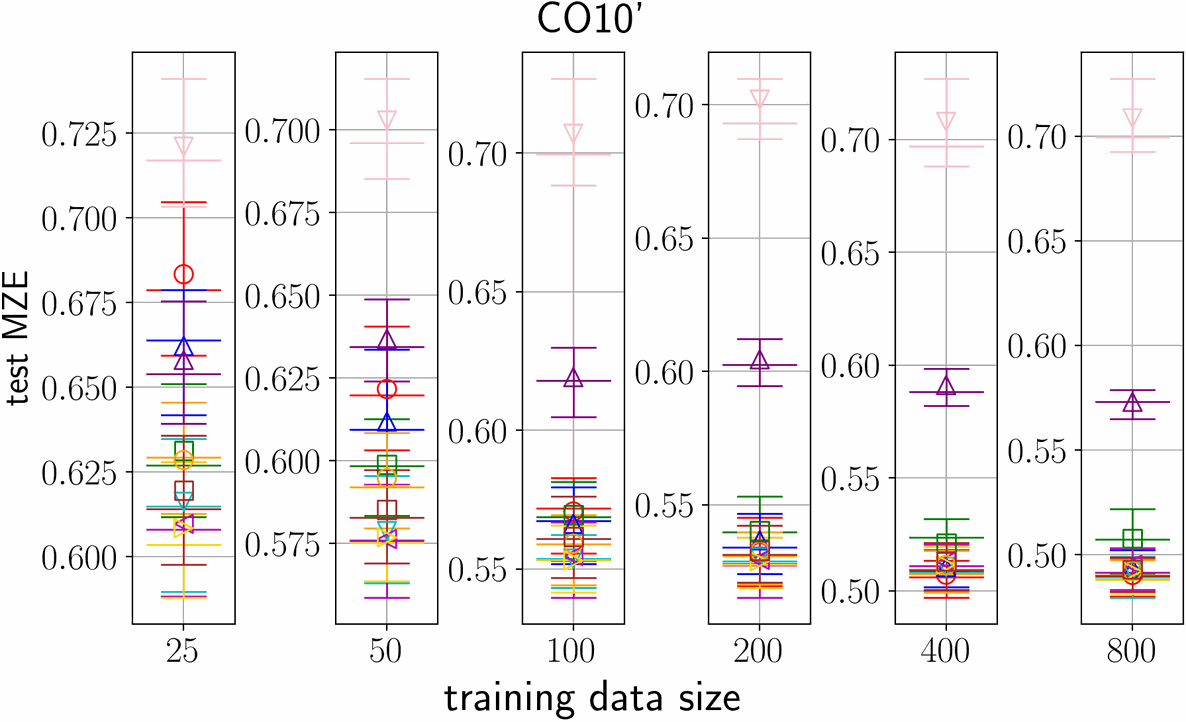}&
\includegraphics[height=2.8cm]{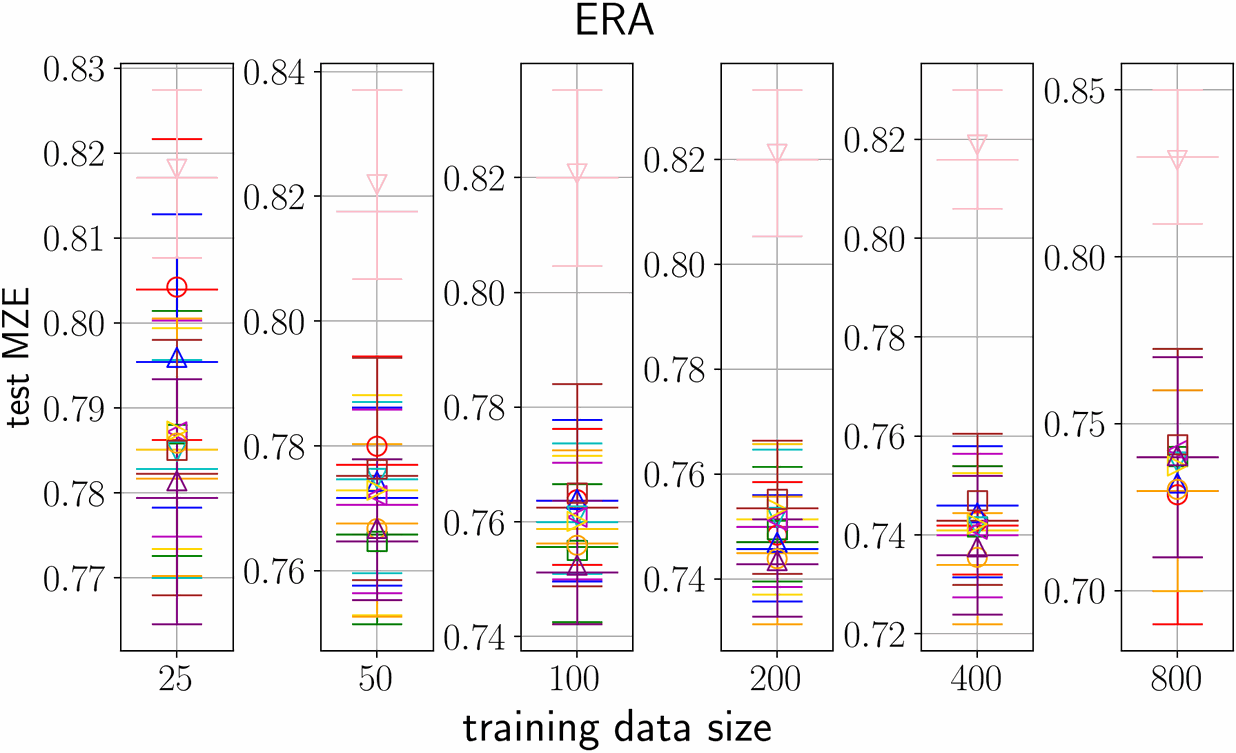}&
\includegraphics[height=2.8cm]{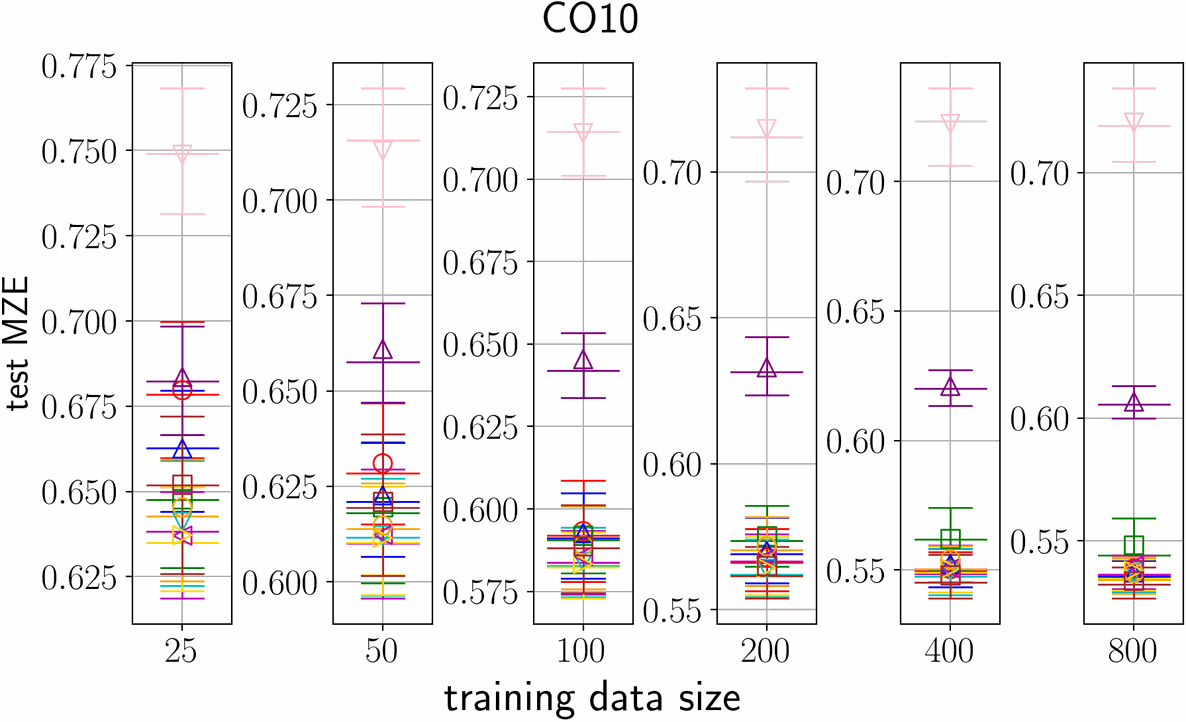}\\
\includegraphics[height=2.8cm]{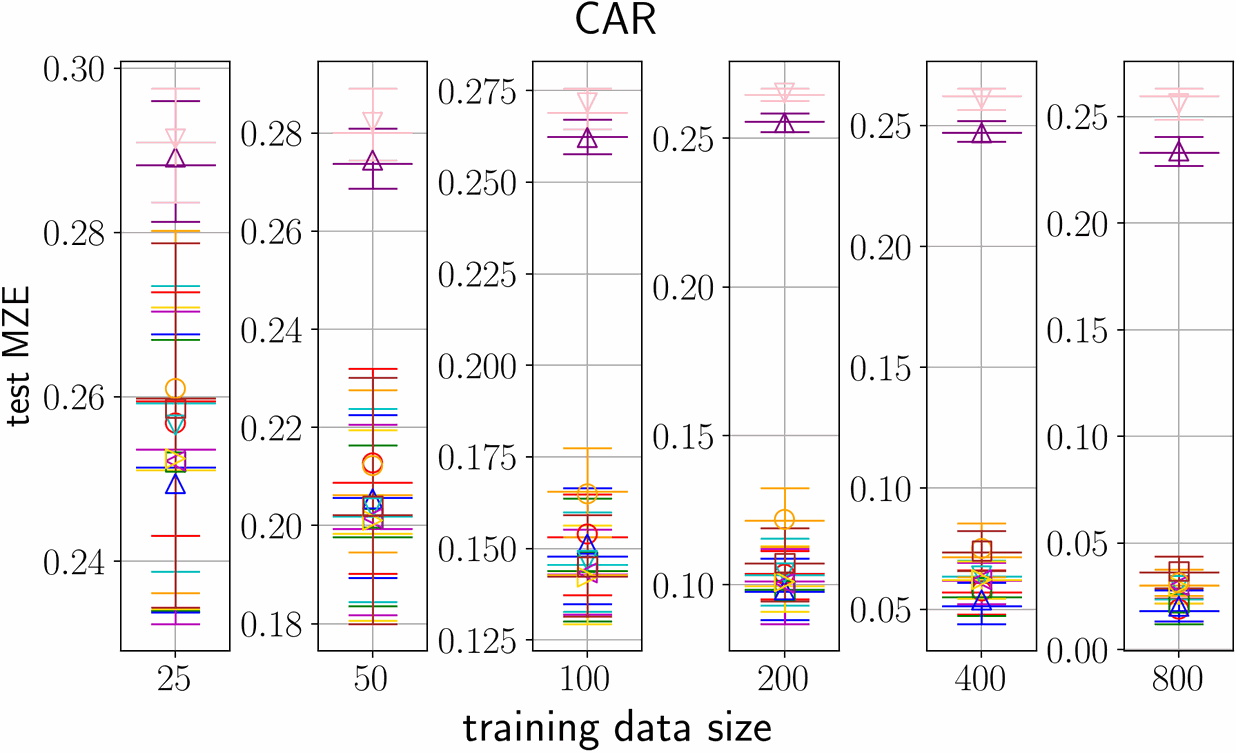}&
\includegraphics[height=2.8cm]{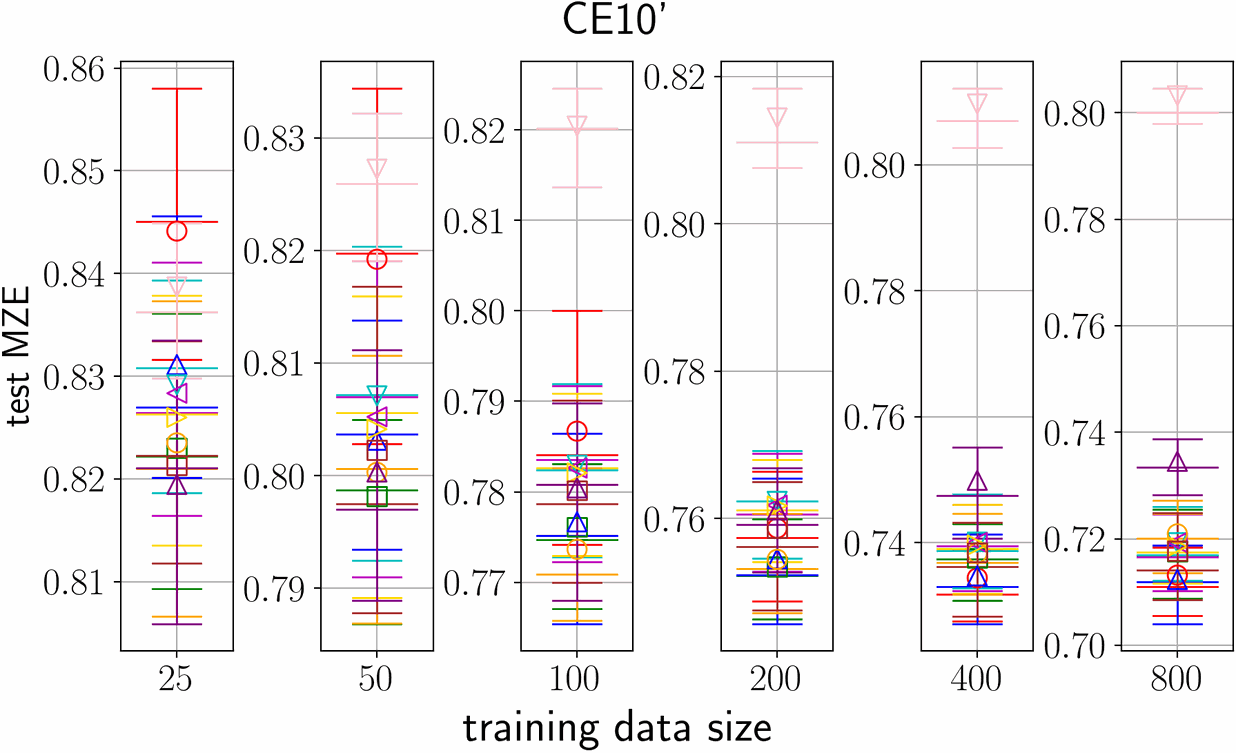}&
\includegraphics[height=2.8cm]{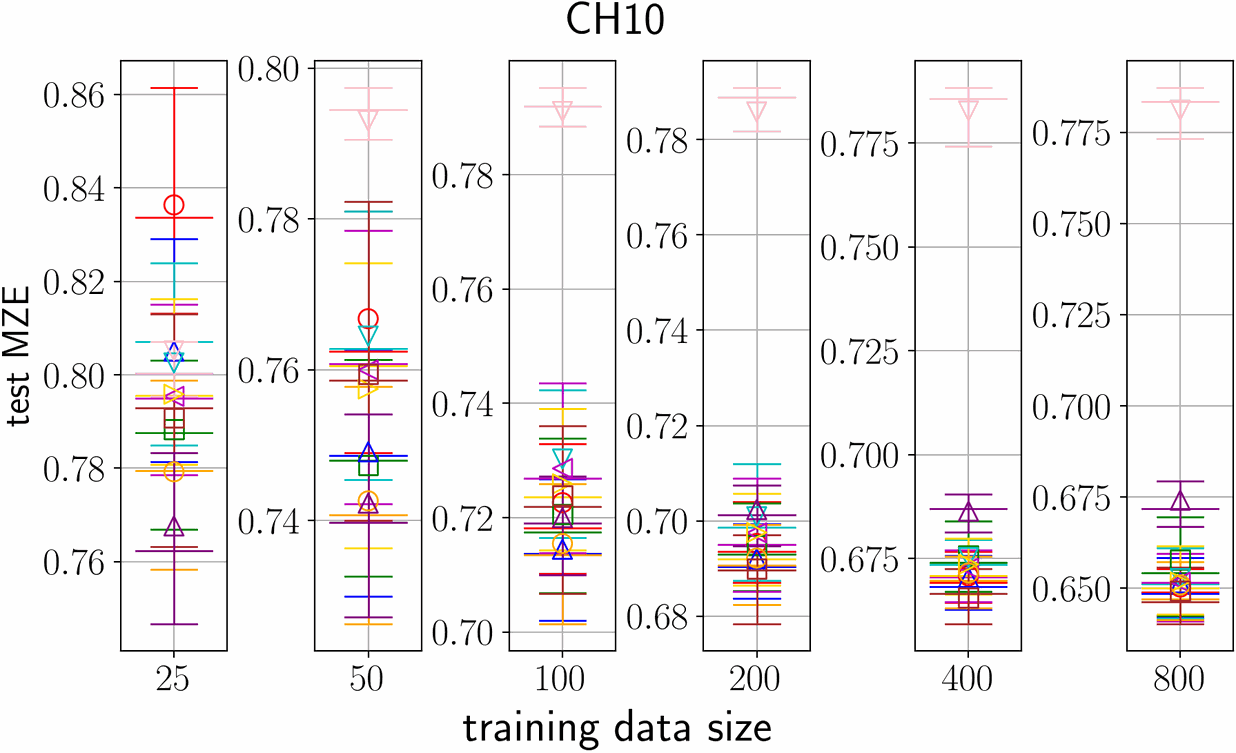}\\
\includegraphics[height=2.8cm]{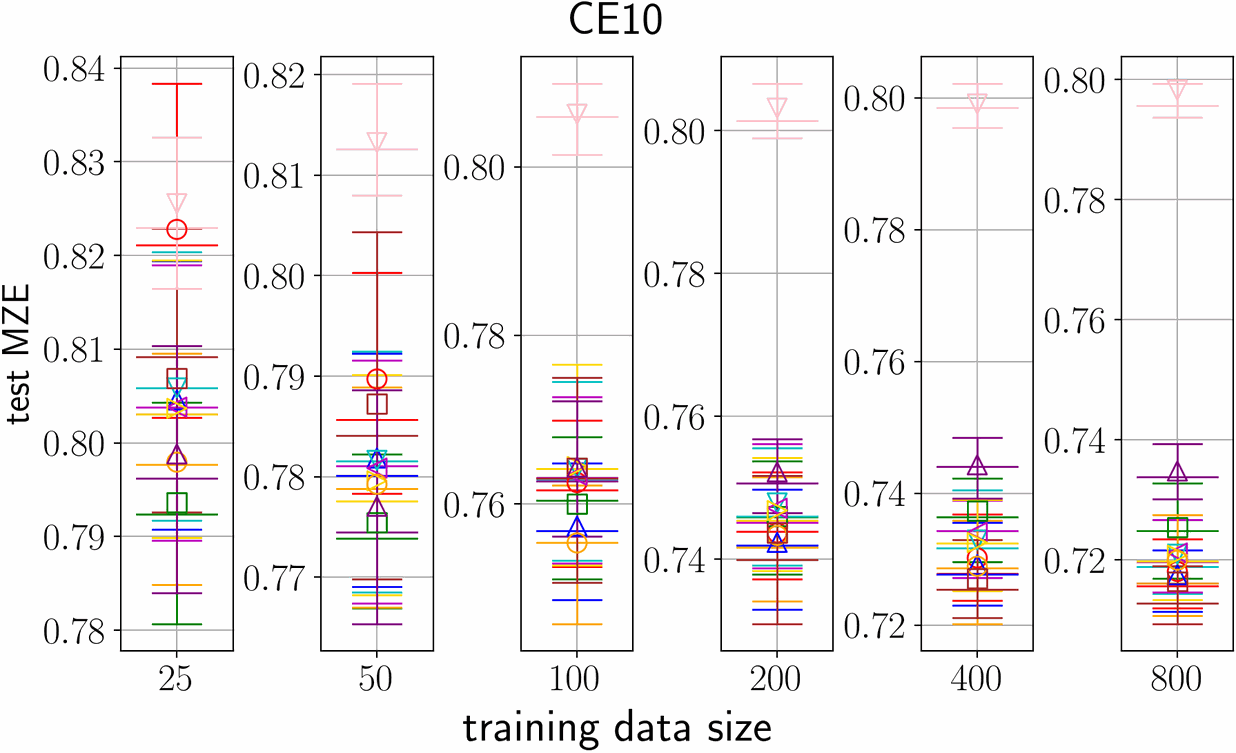}&
\includegraphics[height=2.8cm]{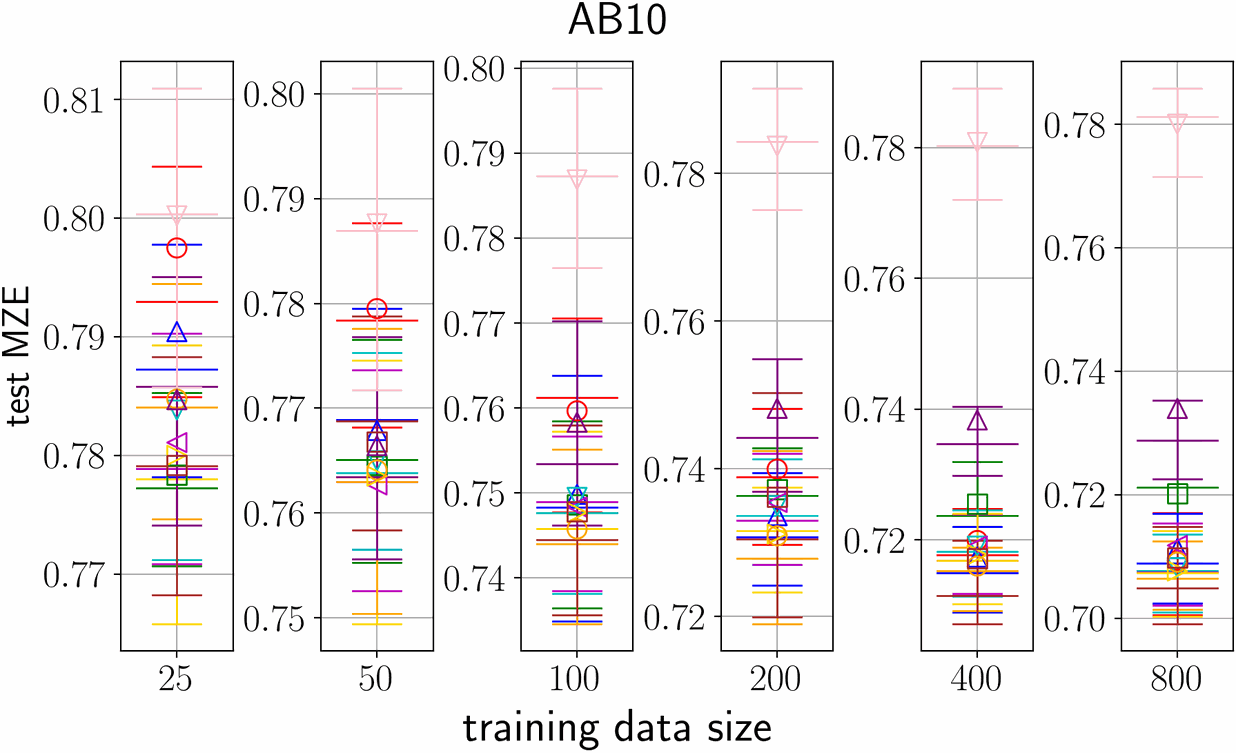}&
\includegraphics[height=2.8cm]{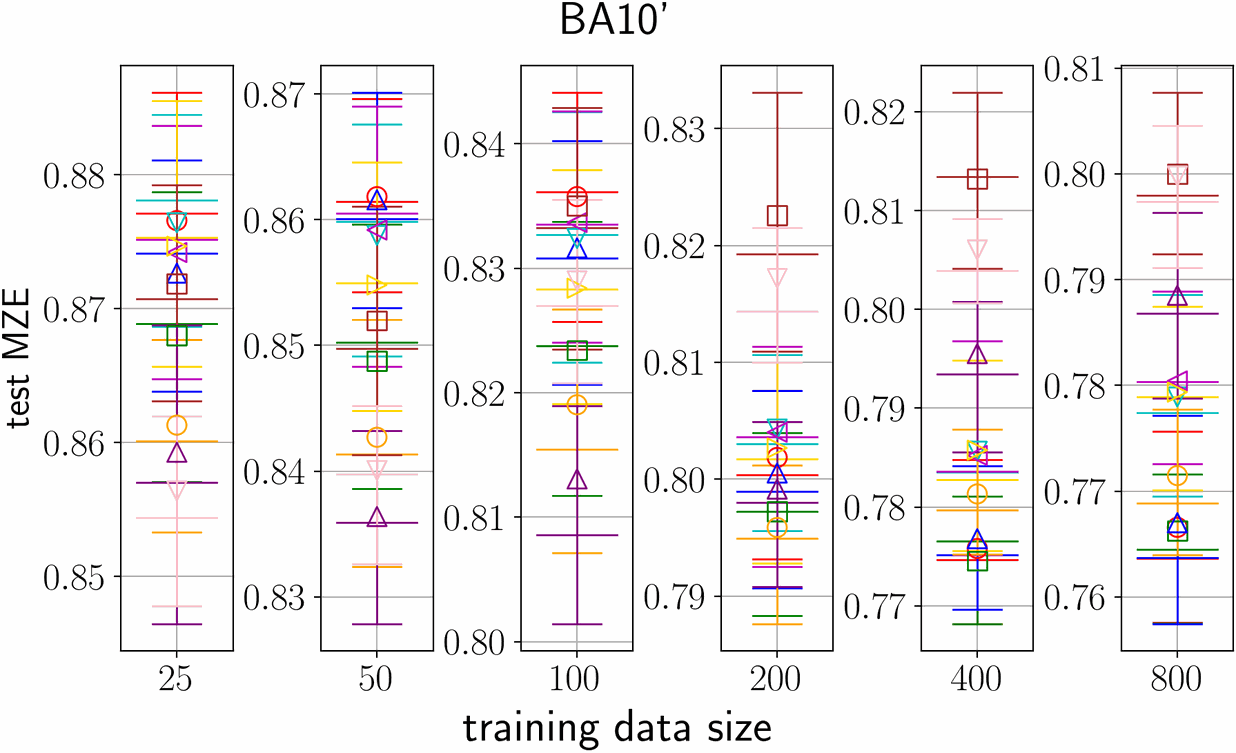}
\end{tabular}
\caption{%
Plot regarding 100-trial test MZEs
by Nonr-MLR (red solid \protect$\circ$), Prev-MLR (green solid \protect$\Box$), 
Stri-MLR (blue solid \protect$\triangle$), Nonr-AUL (cyan solid \protect$\triangledown$), 
Prev-AUL (magenta solid \protect$\triangleleft$), Stri-AUL (yellow solid \protect$\triangleright$),
Nonr-CL (orange dashed \protect$\circ$), Nonr-UL (brown dashed \protect$\Box$), 
Nonr-POCL (purple dashed \protect$\triangle$), and Nonr-POUL (pink dashed \protect$\triangledown$).
Lower short, middle long, and upper short bars and marker 
represent 0.25, 0.5, and 0.75 quantiles and mean.}
\label{fig:Performance-BestLam-MZE}
\end{figure*}
\begin{figure*}[p]
\centering%
\renewcommand{\arraystretch}{0.25}%
\renewcommand{\tabcolsep}{10pt}%
\begin{tabular}{ccc}%
\includegraphics[height=2.8cm]{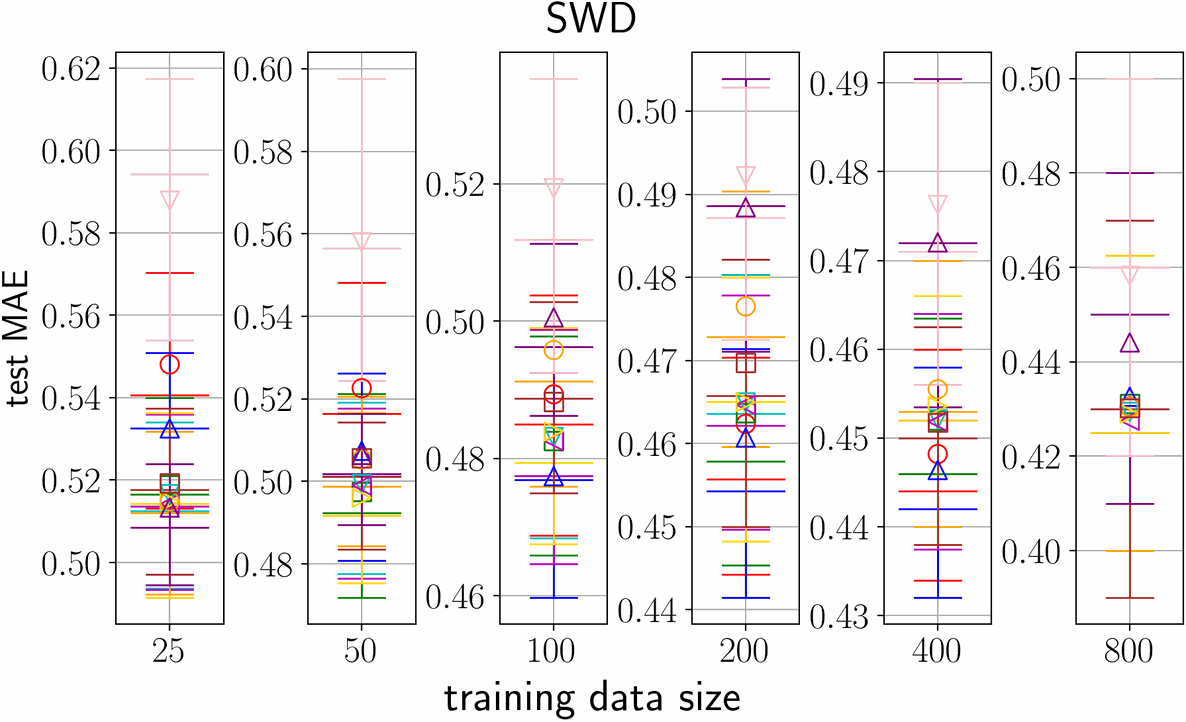}&
\includegraphics[height=2.8cm]{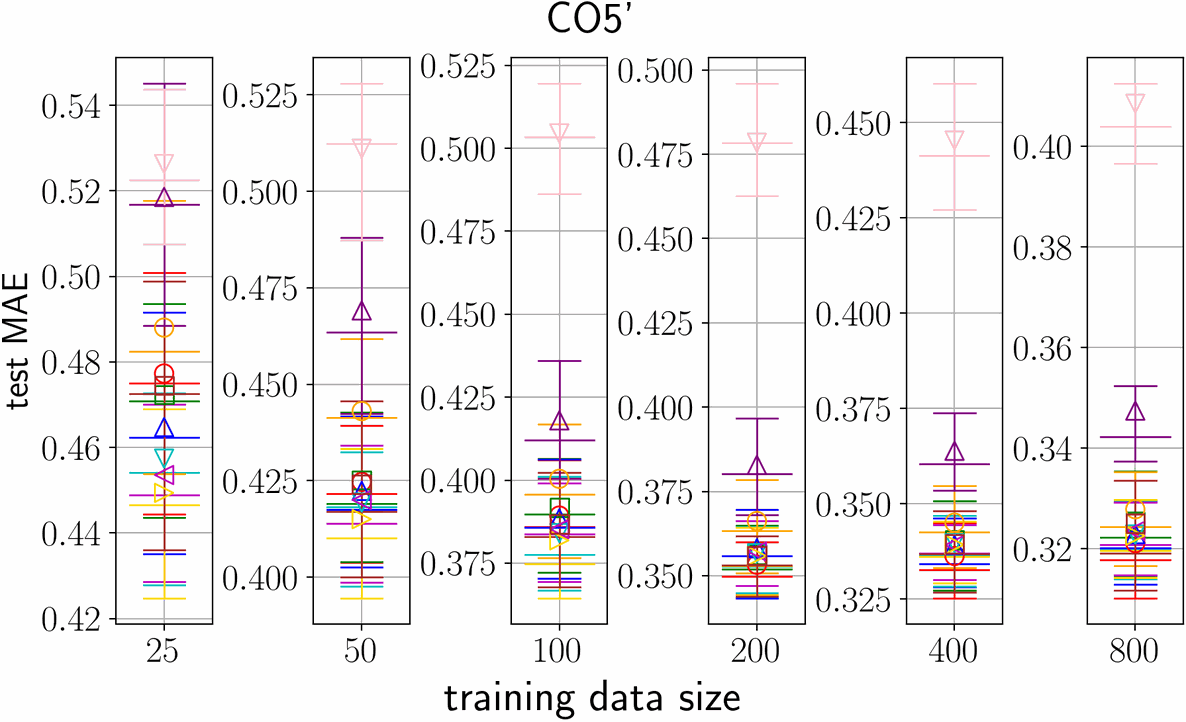}&
\includegraphics[height=2.8cm]{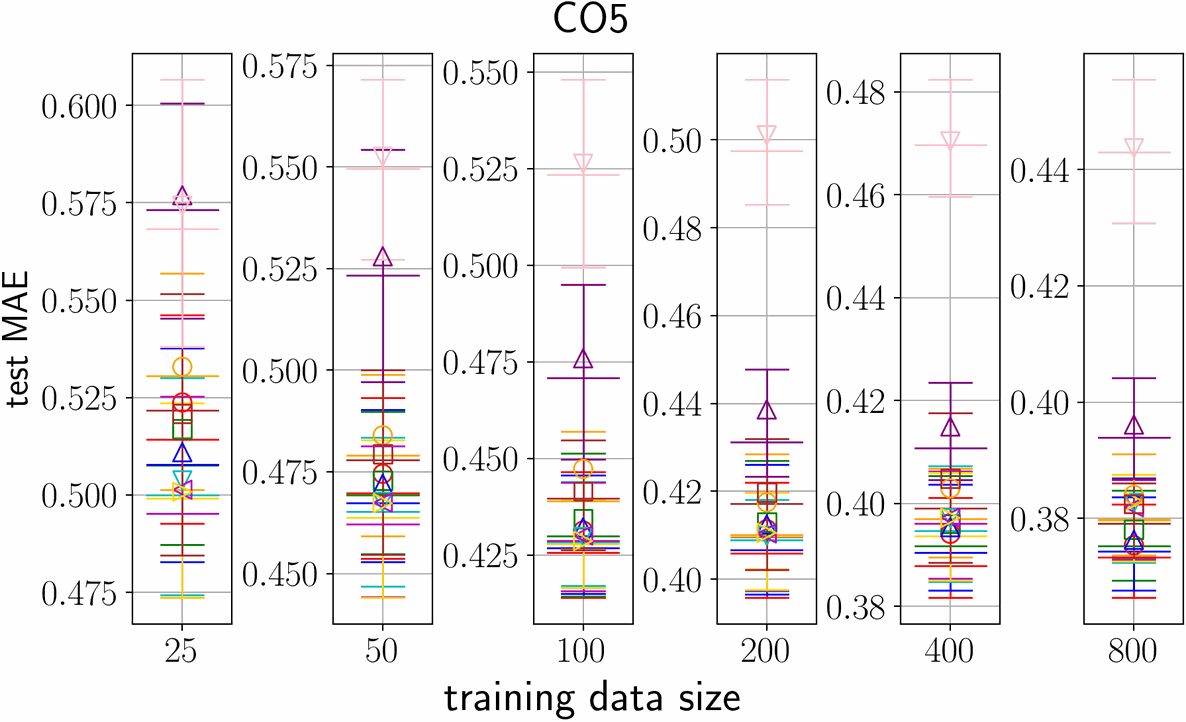}\\
\includegraphics[height=2.8cm]{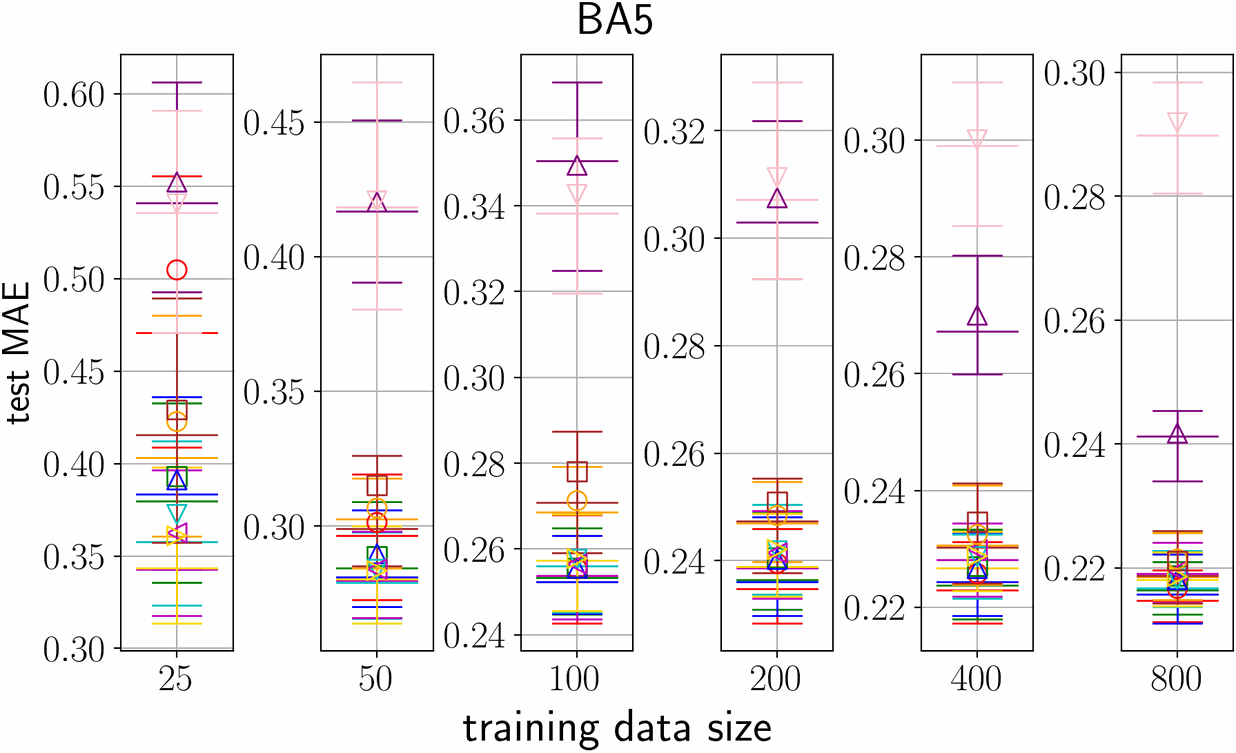}&
\includegraphics[height=2.8cm]{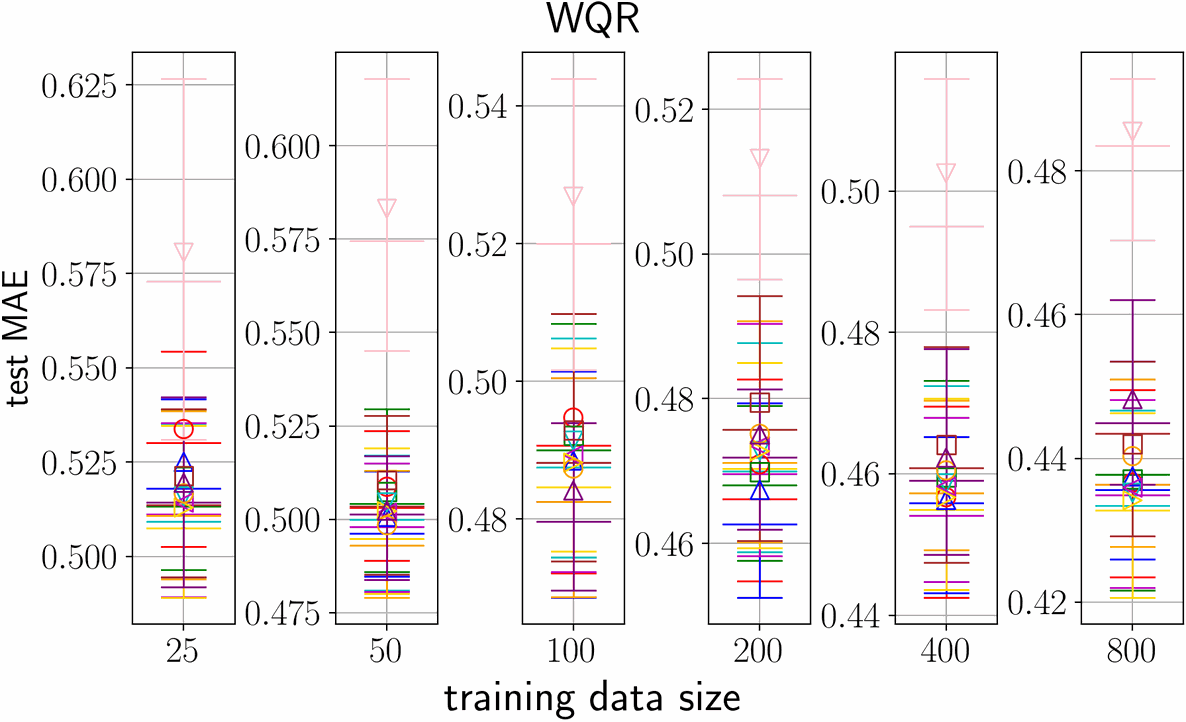}&
\includegraphics[height=2.8cm]{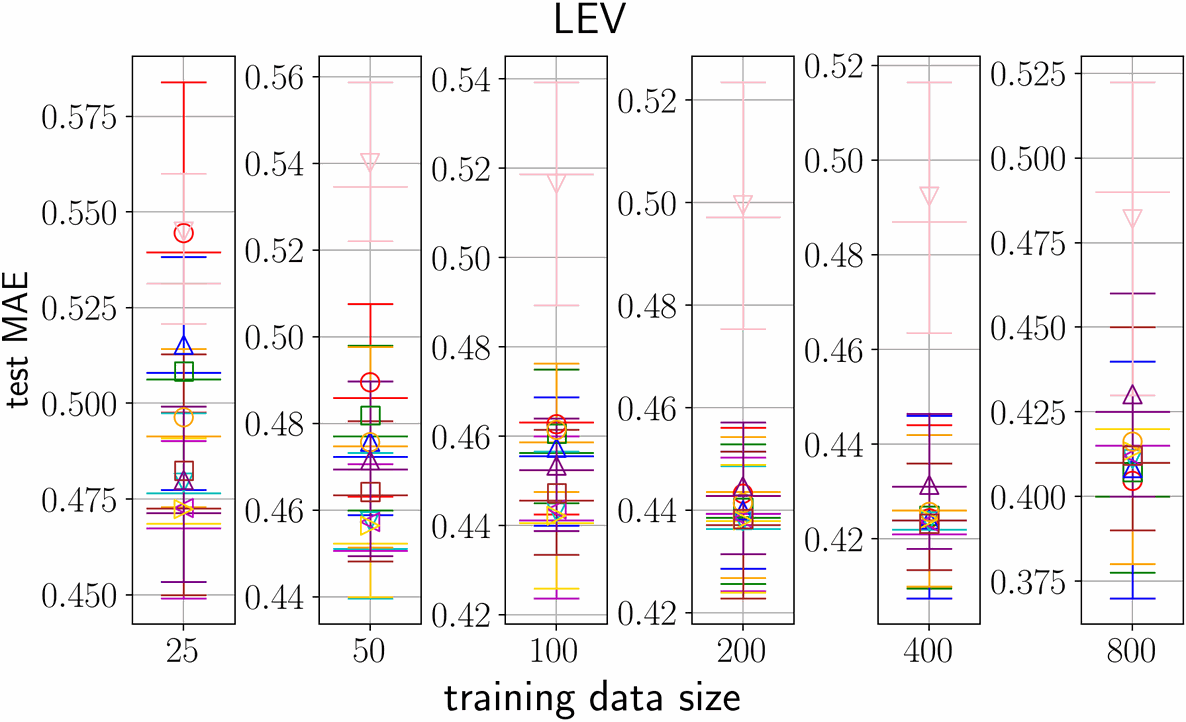}\\
\includegraphics[height=2.8cm]{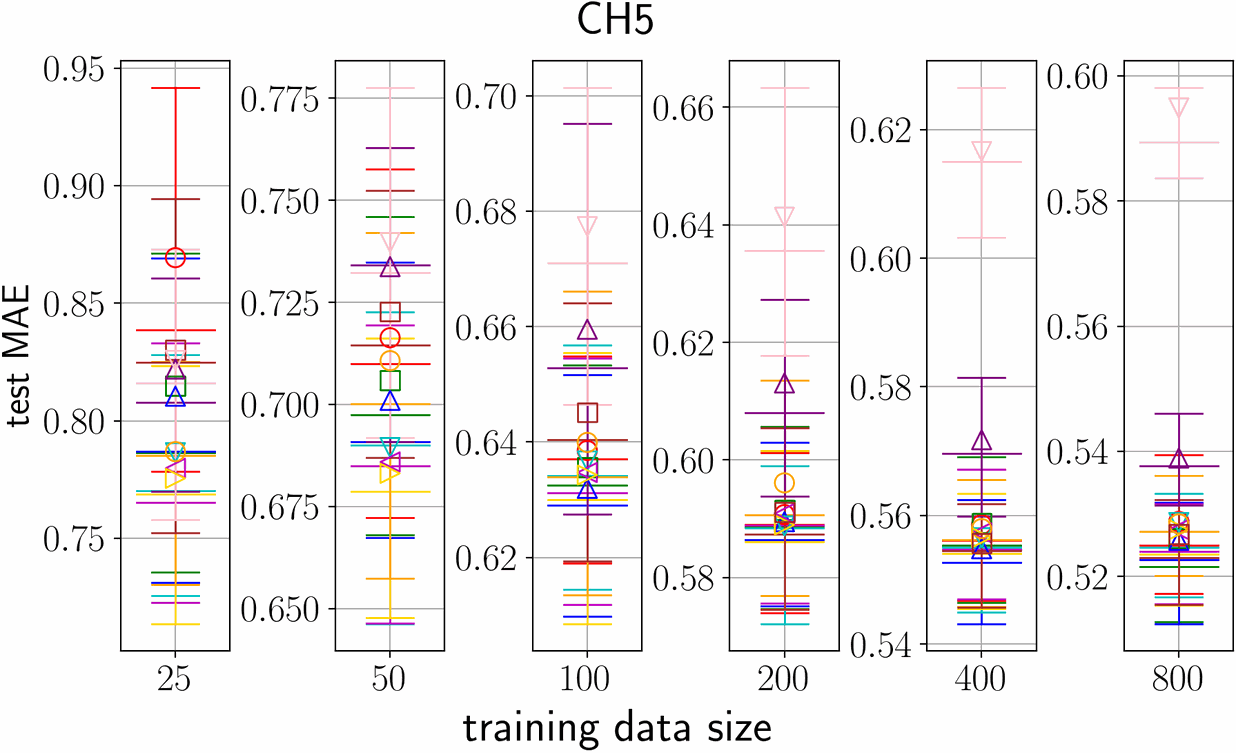}&
\includegraphics[height=2.8cm]{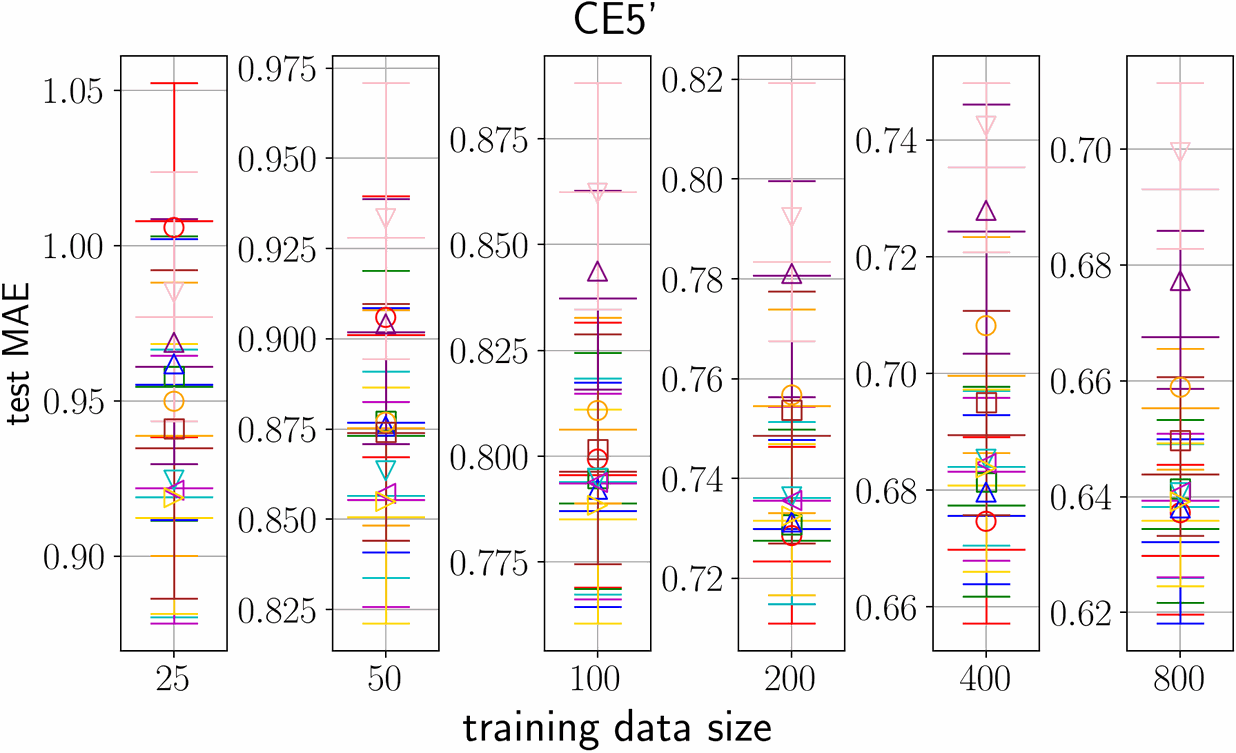}&
\includegraphics[height=2.8cm]{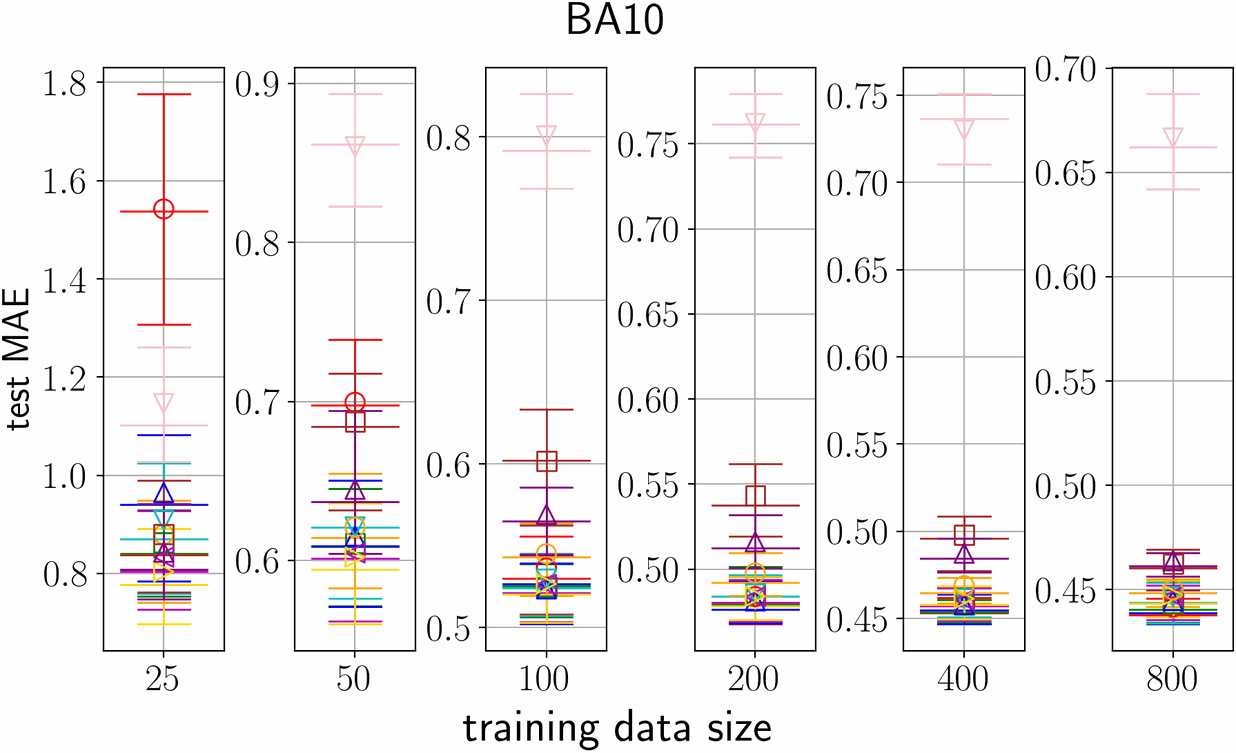}\\
\includegraphics[height=2.8cm]{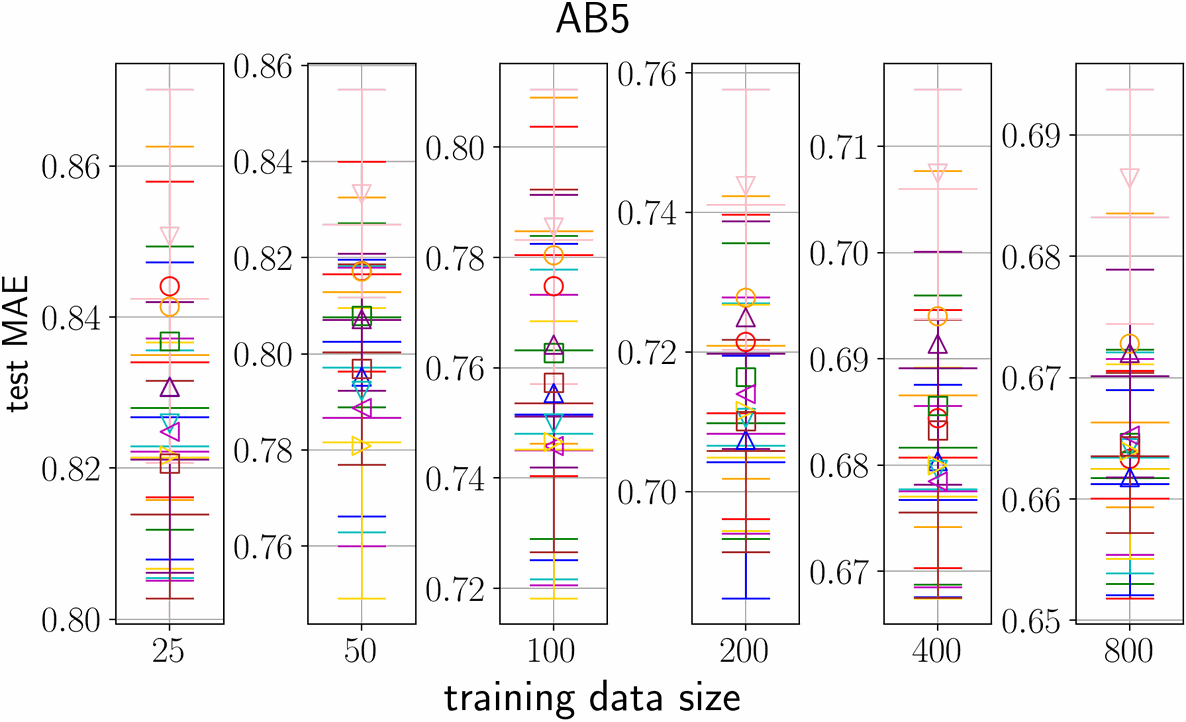}&
\includegraphics[height=2.8cm]{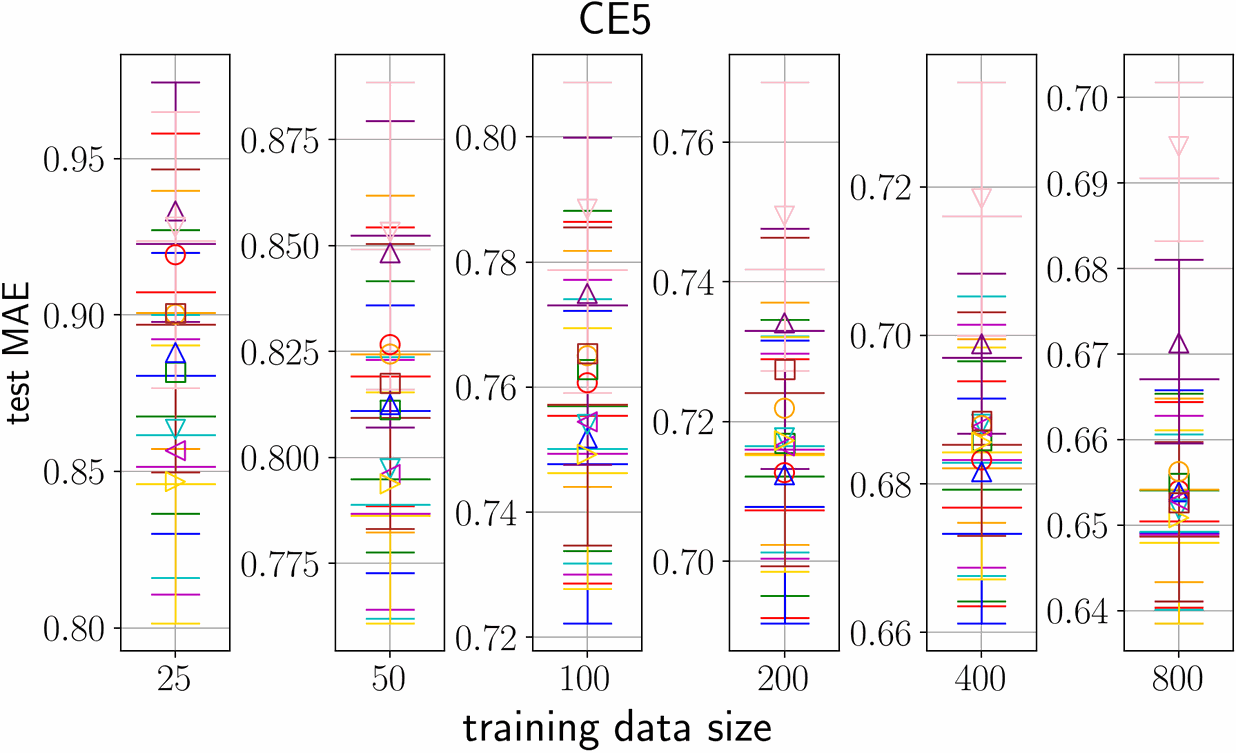}&
\includegraphics[height=2.8cm]{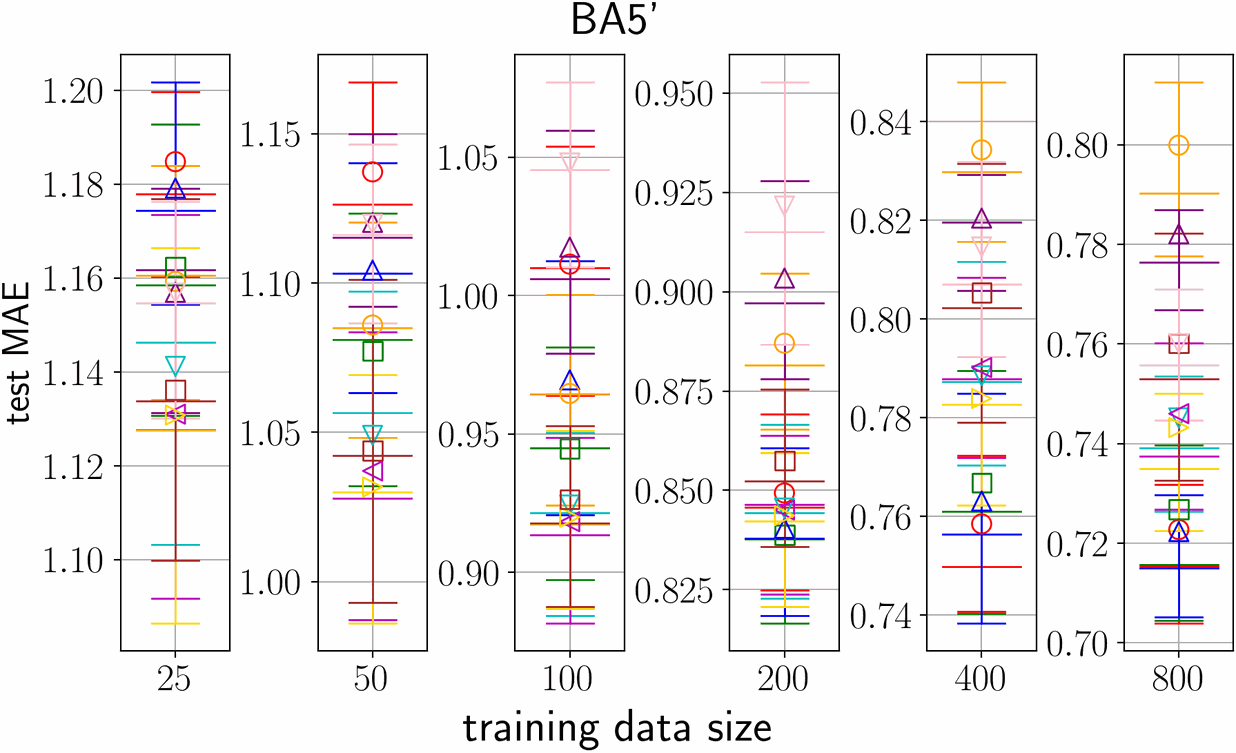}\\
\includegraphics[height=2.8cm]{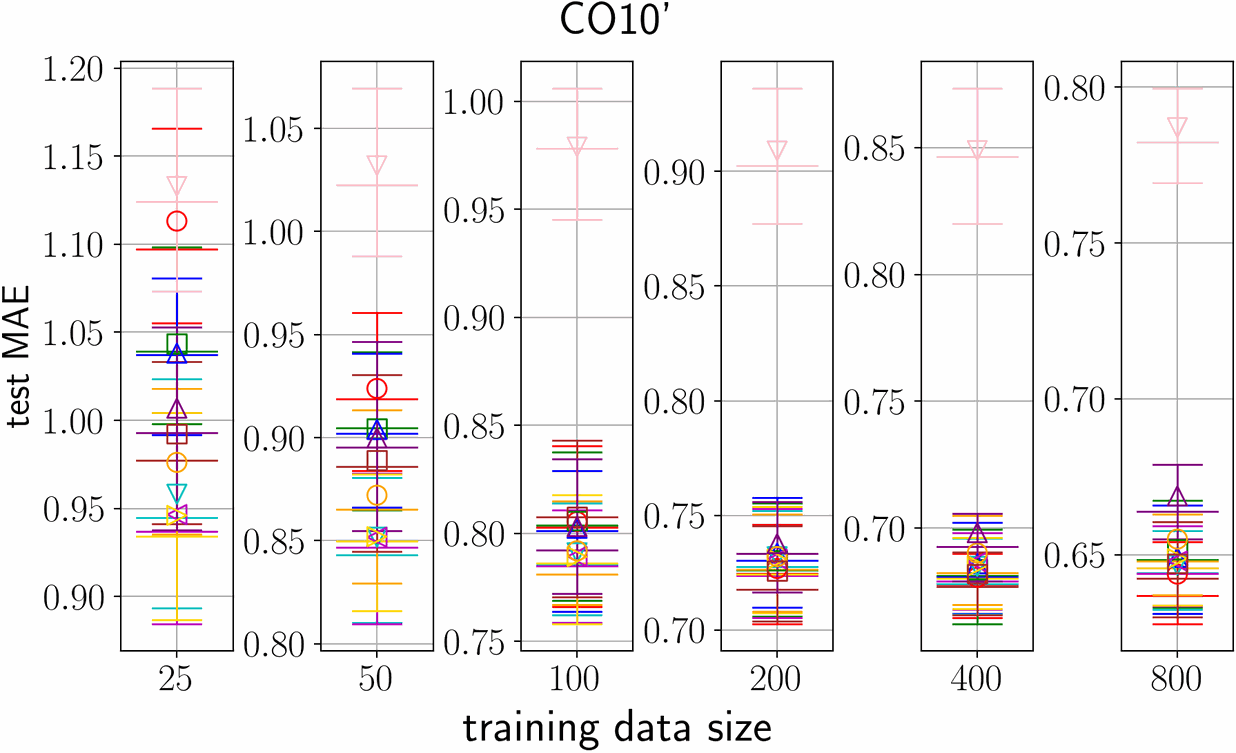}&
\includegraphics[height=2.8cm]{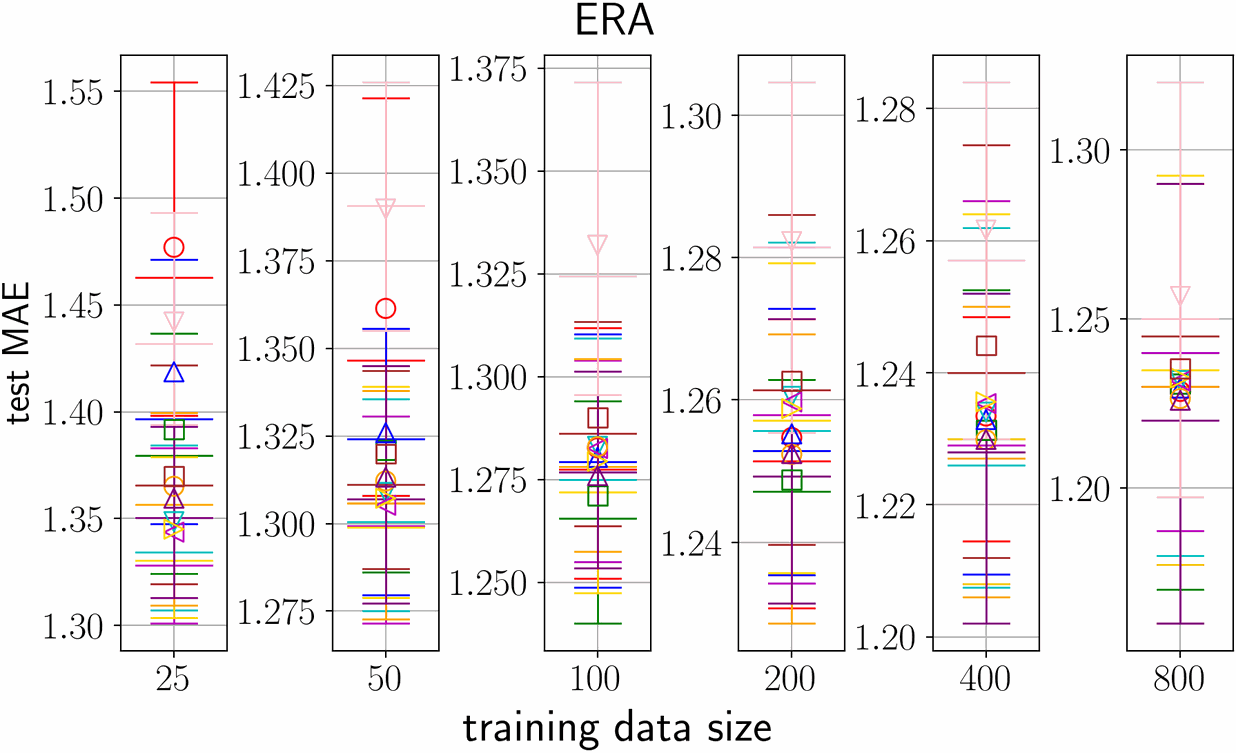}&
\includegraphics[height=2.8cm]{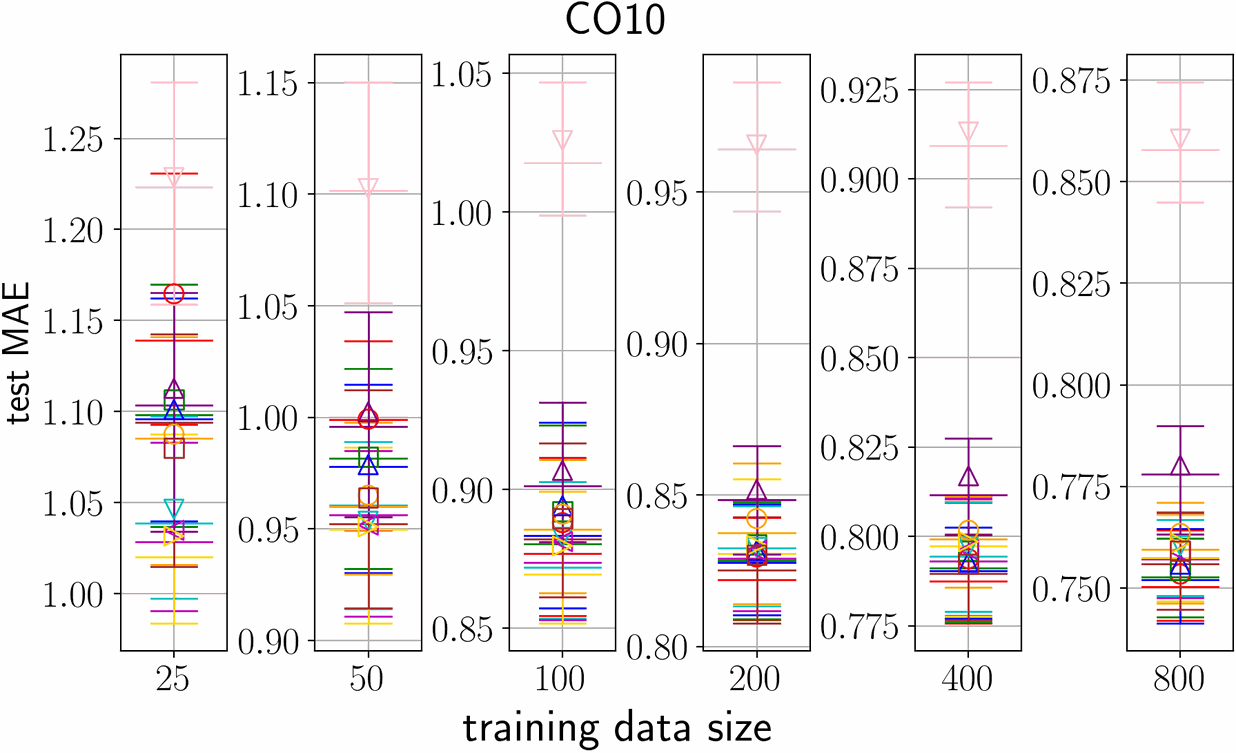}\\
\includegraphics[height=2.8cm]{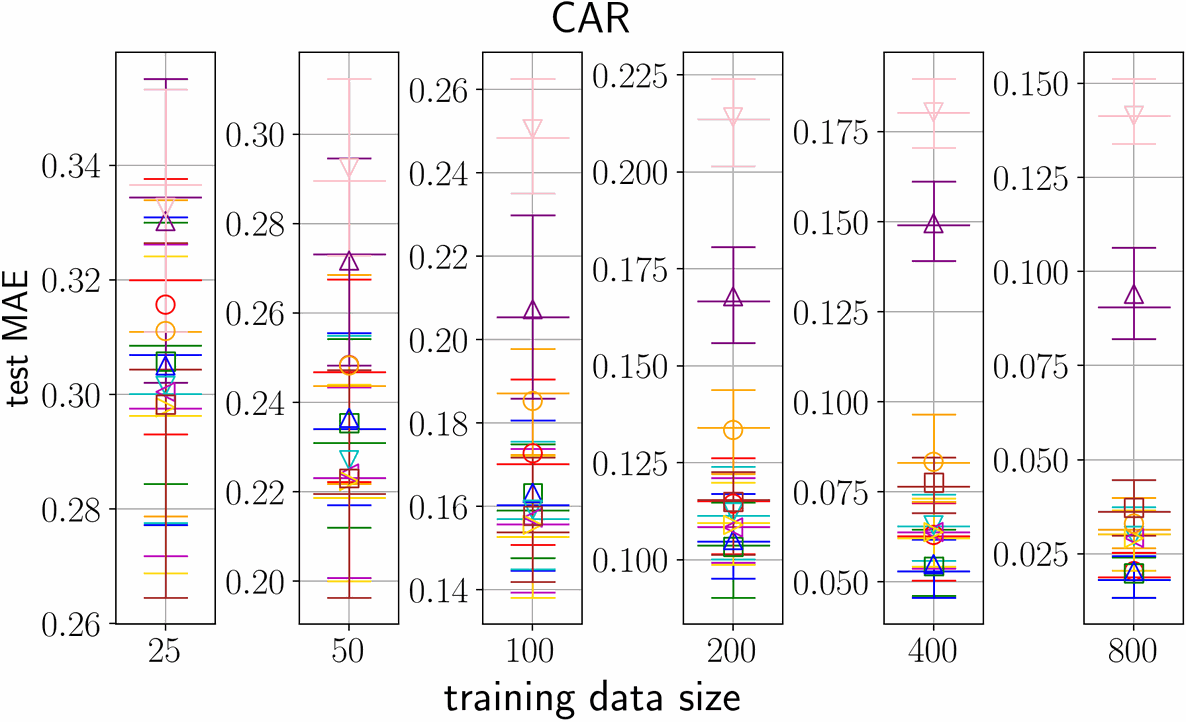}&
\includegraphics[height=2.8cm]{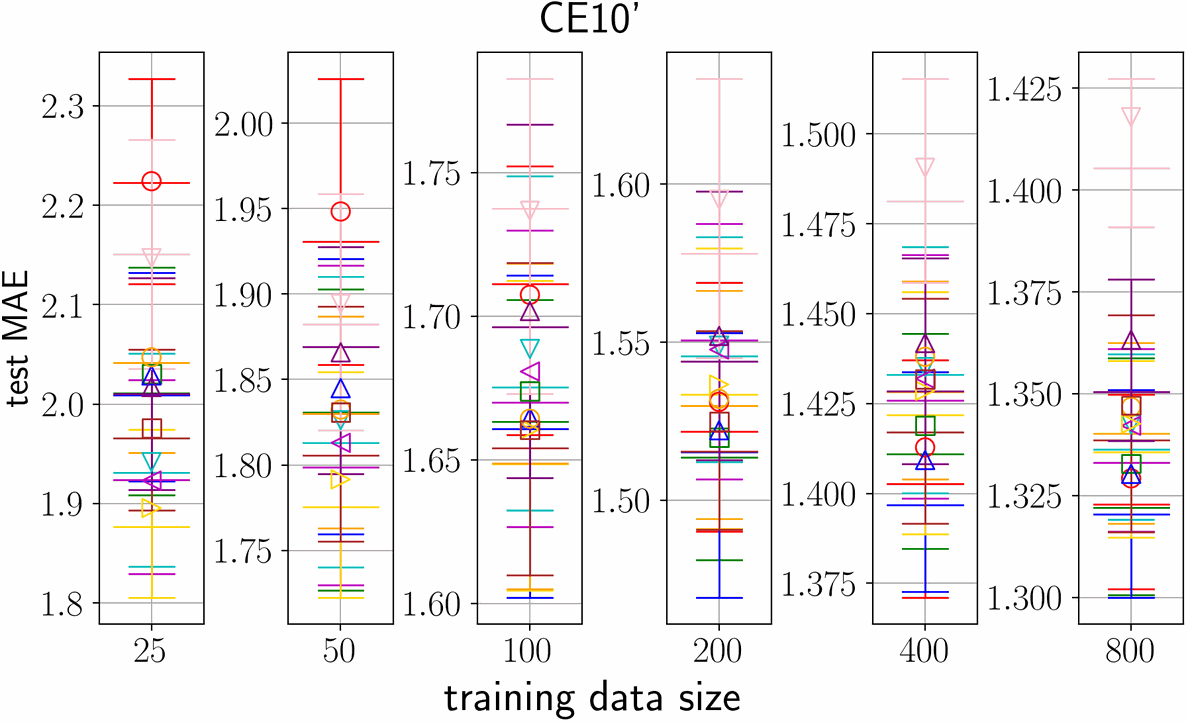}&
\includegraphics[height=2.8cm]{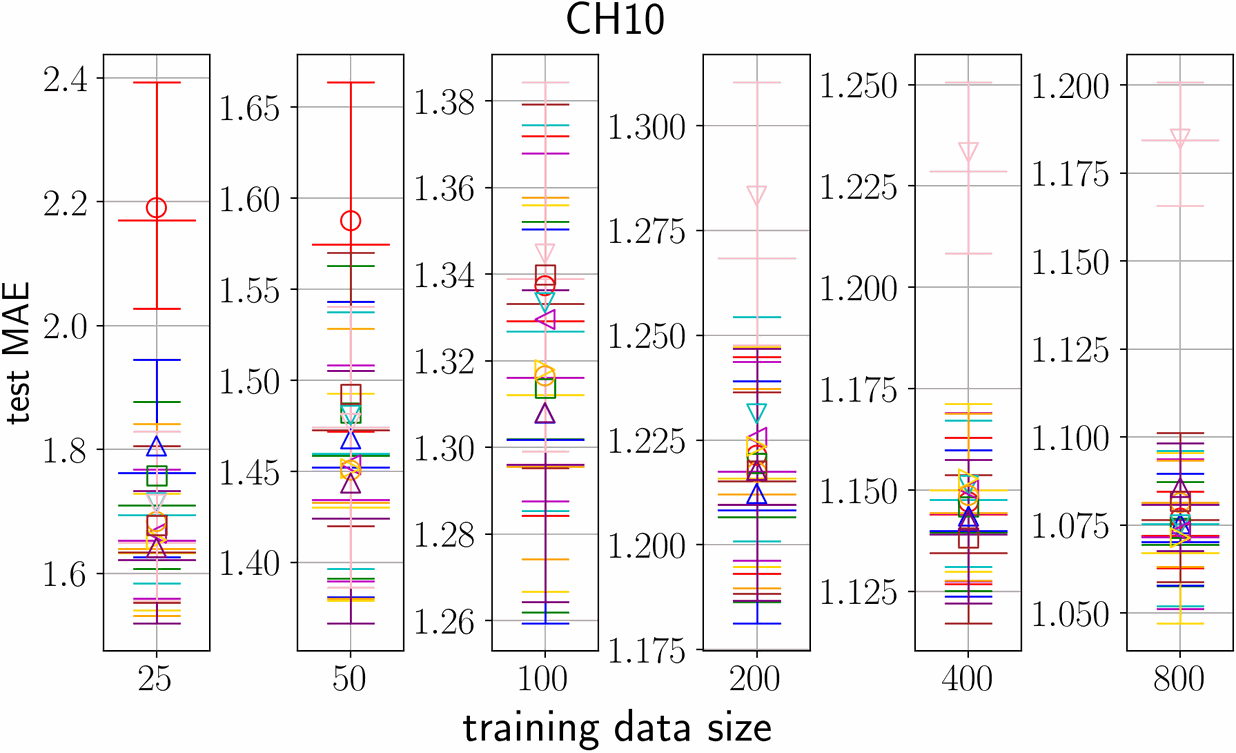}\\
\includegraphics[height=2.8cm]{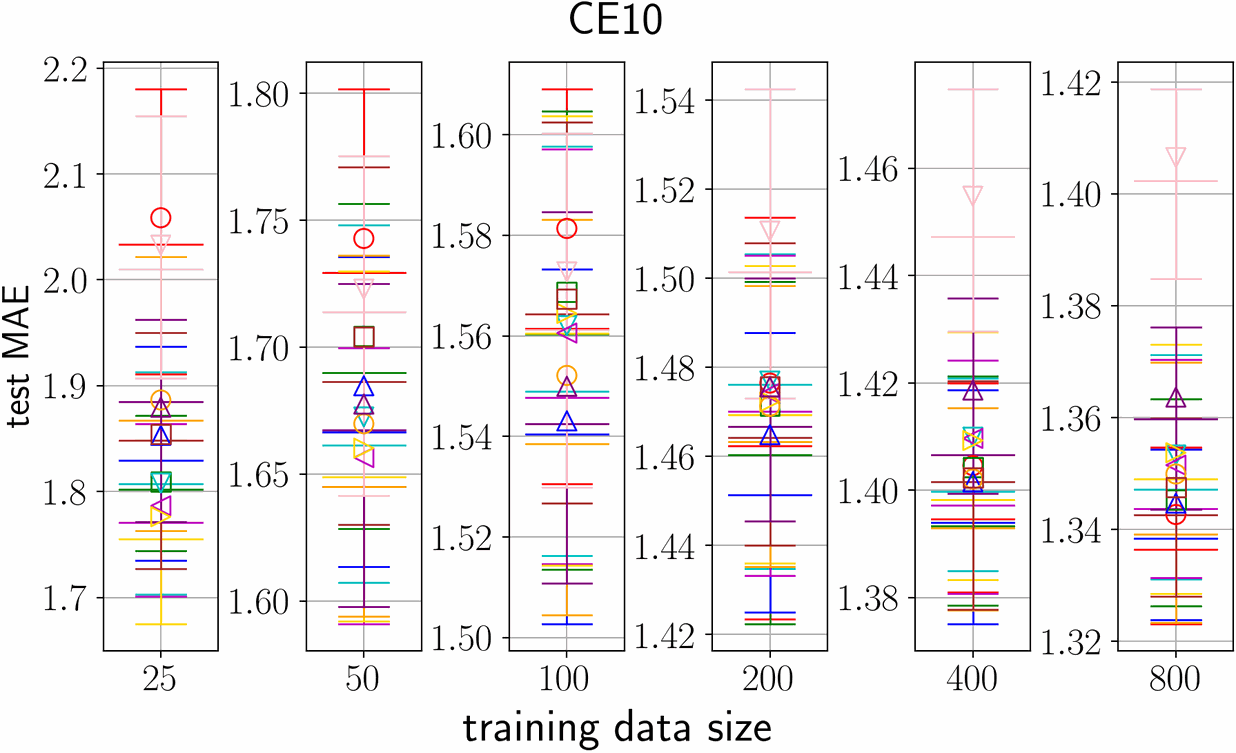}&
\includegraphics[height=2.8cm]{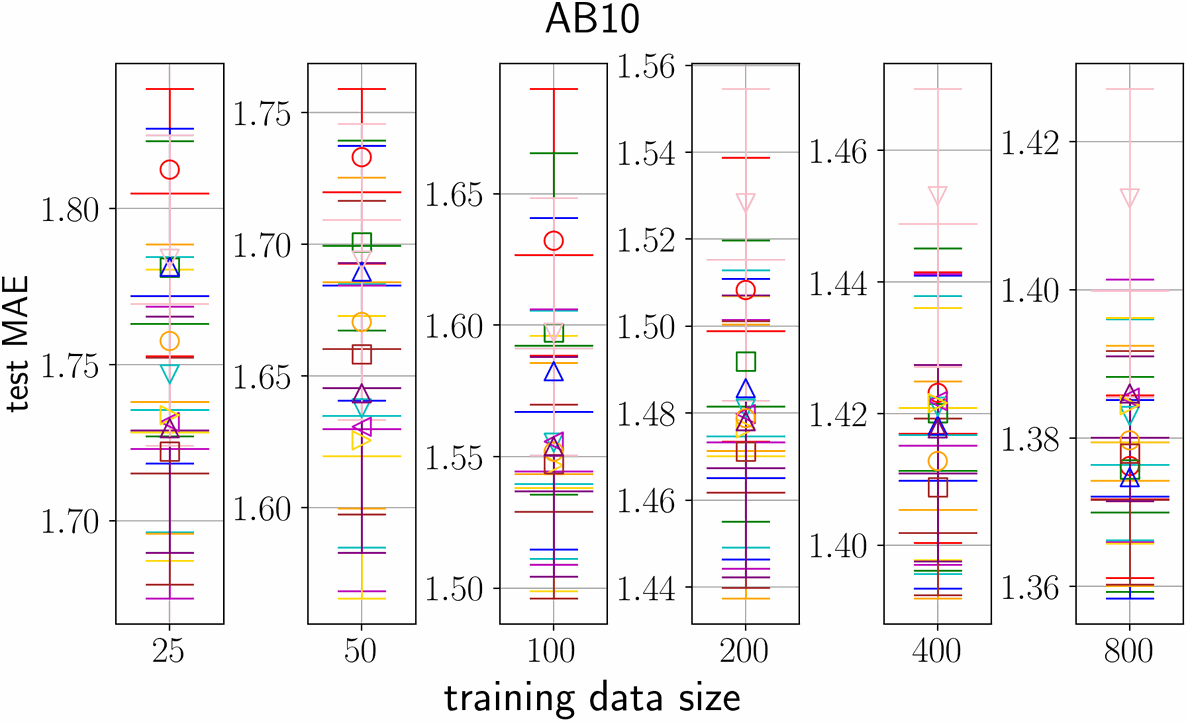}&
\includegraphics[height=2.8cm]{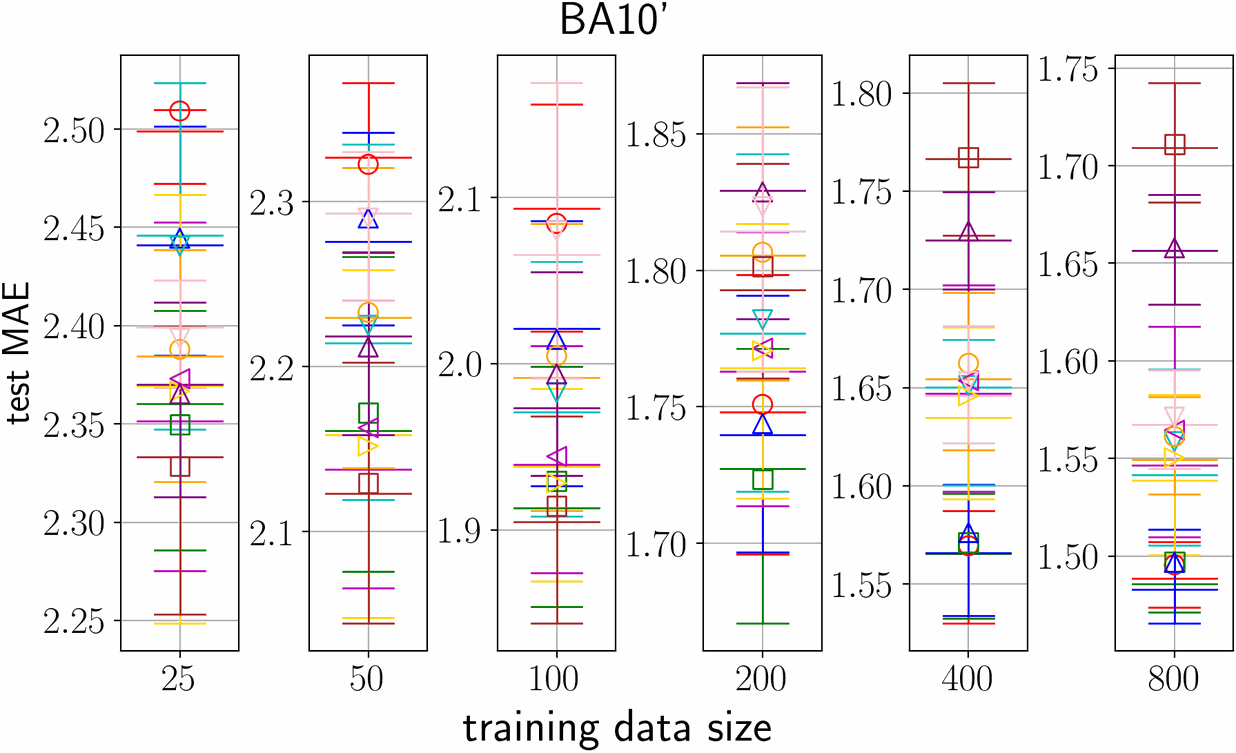}
\end{tabular}
\caption{%
Plot regarding 100-trial test MAEs
by Nonr-MLR (red solid \protect$\circ$), Prev-MLR (green solid \protect$\Box$), 
Stri-MLR (blue solid \protect$\triangle$), Nonr-AUL (cyan solid \protect$\triangledown$), 
Prev-AUL (magenta solid \protect$\triangleleft$), Stri-AUL (yellow solid \protect$\triangleright$),
Nonr-CL (orange dashed \protect$\circ$), Nonr-UL (brown dashed \protect$\Box$), 
Nonr-POCL (purple dashed \protect$\triangle$), and Nonr-POUL (pink dashed \protect$\triangledown$).
Lower short, middle long, and upper short bars and marker 
represent 0.25, 0.5, and 0.75 quantiles and mean.}
\label{fig:Performance-BestLam-MAE}
\end{figure*}
\begin{figure*}[p]
\centering%
\renewcommand{\arraystretch}{0.25}%
\renewcommand{\tabcolsep}{10pt}%
\begin{tabular}{ccc}%
\includegraphics[height=2.8cm]{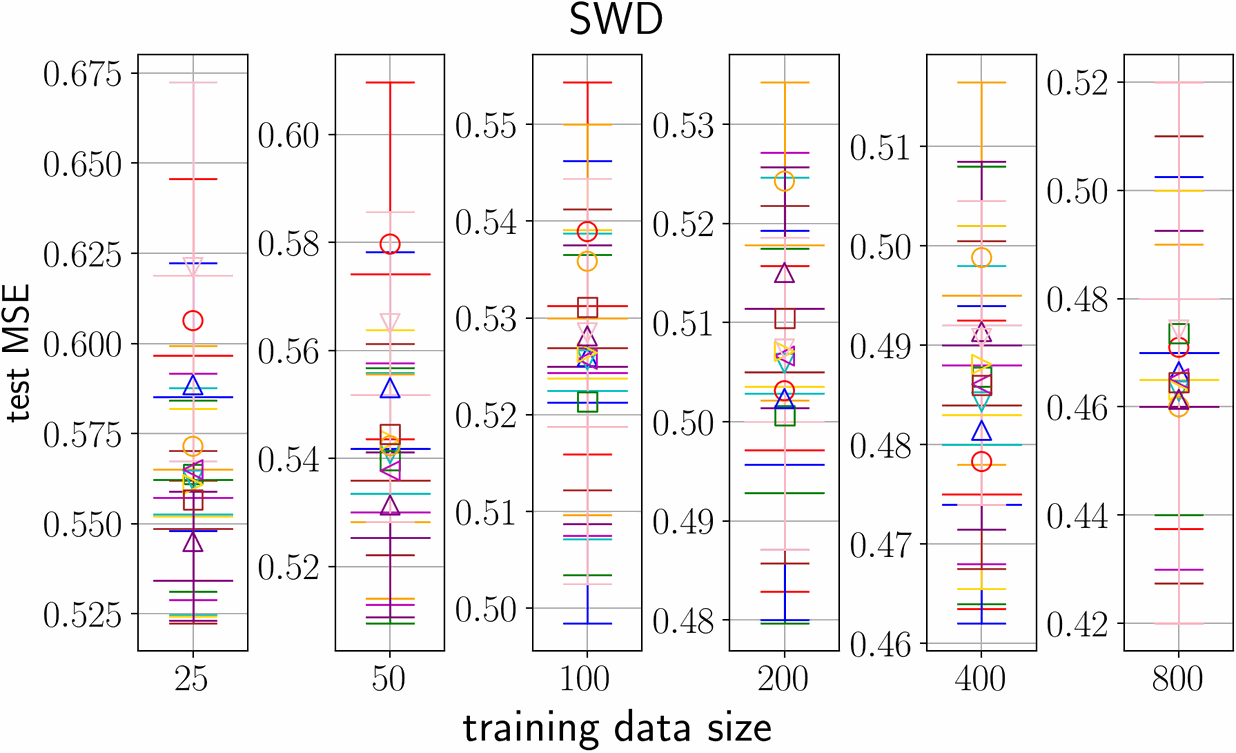}&
\includegraphics[height=2.8cm]{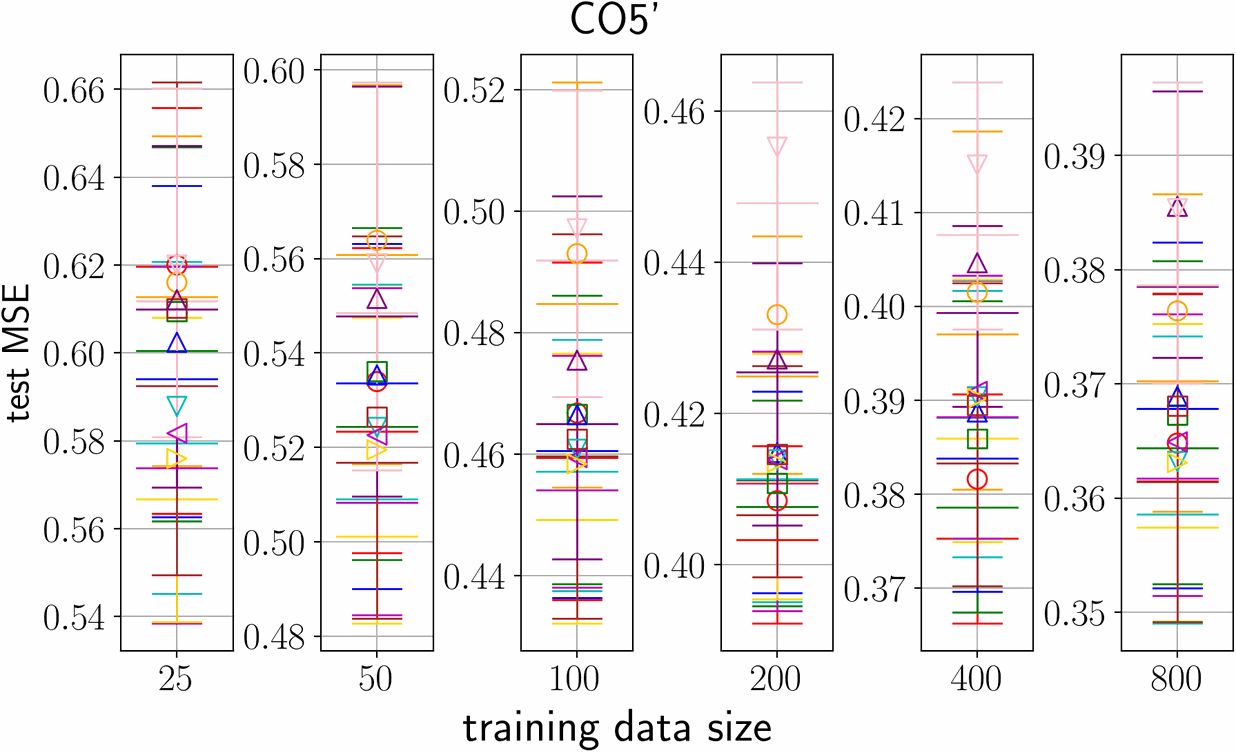}&
\includegraphics[height=2.8cm]{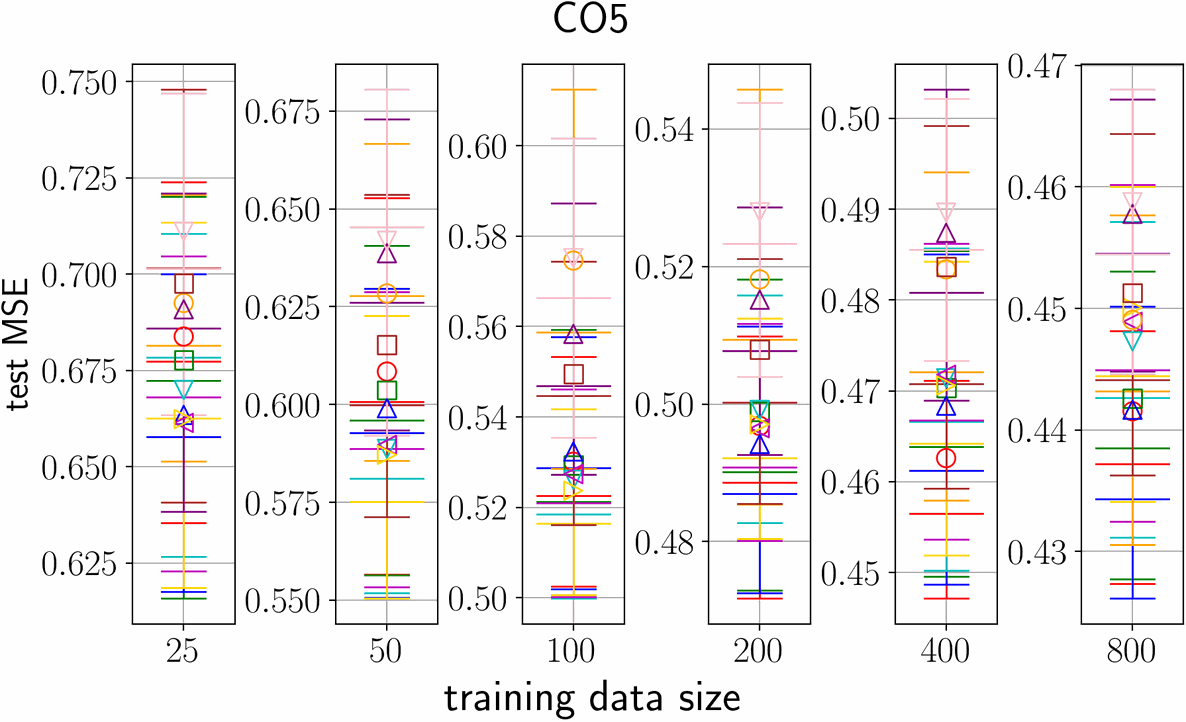}\\
\includegraphics[height=2.8cm]{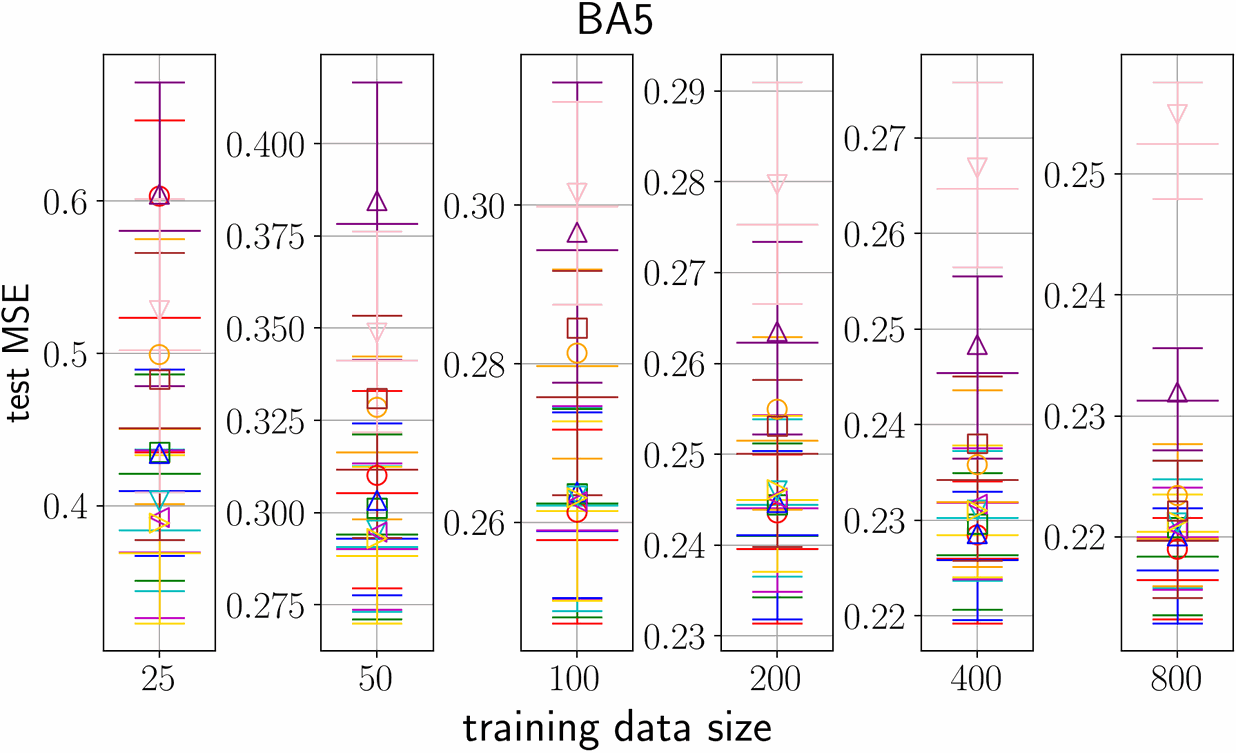}&
\includegraphics[height=2.8cm]{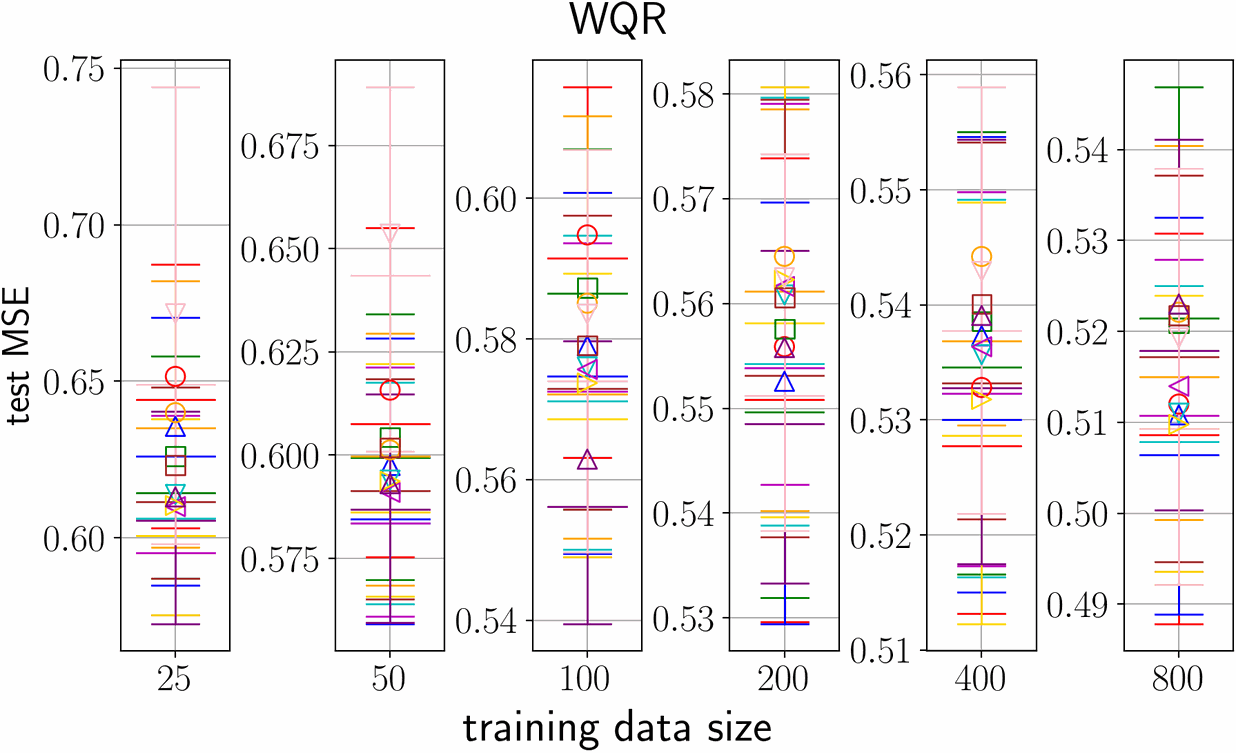}&
\includegraphics[height=2.8cm]{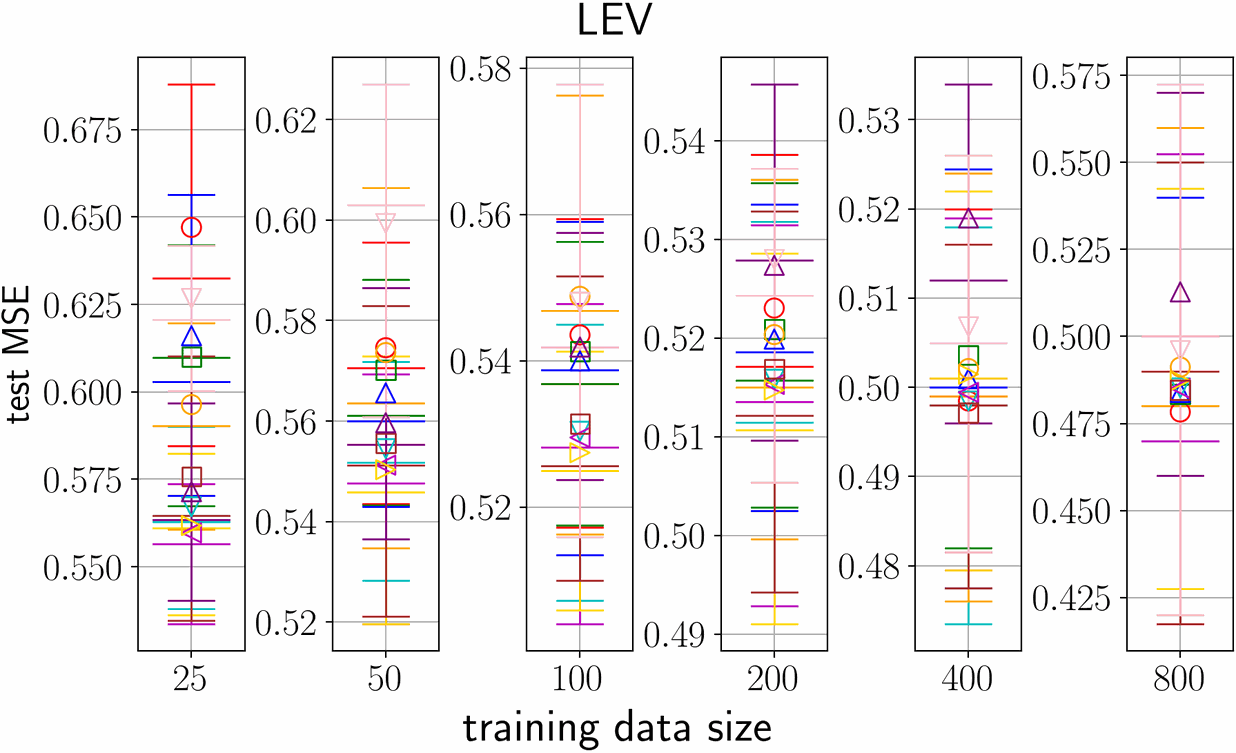}\\
\includegraphics[height=2.8cm]{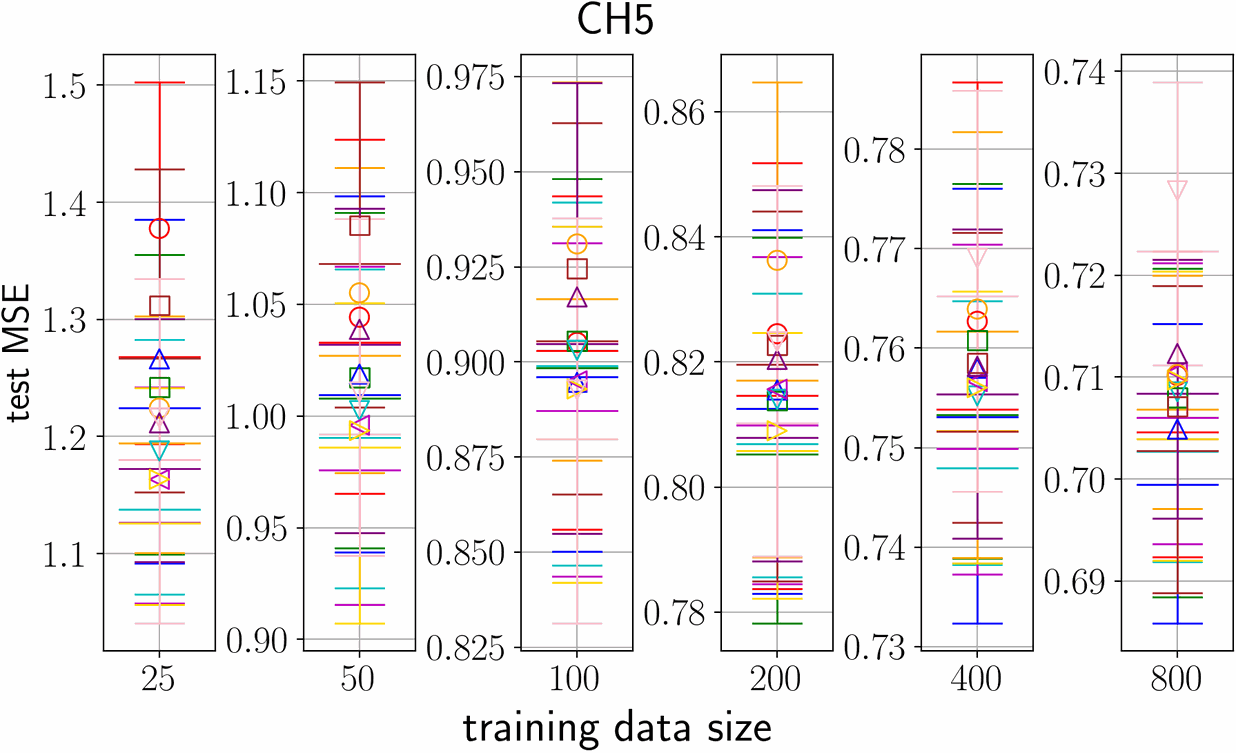}&
\includegraphics[height=2.8cm]{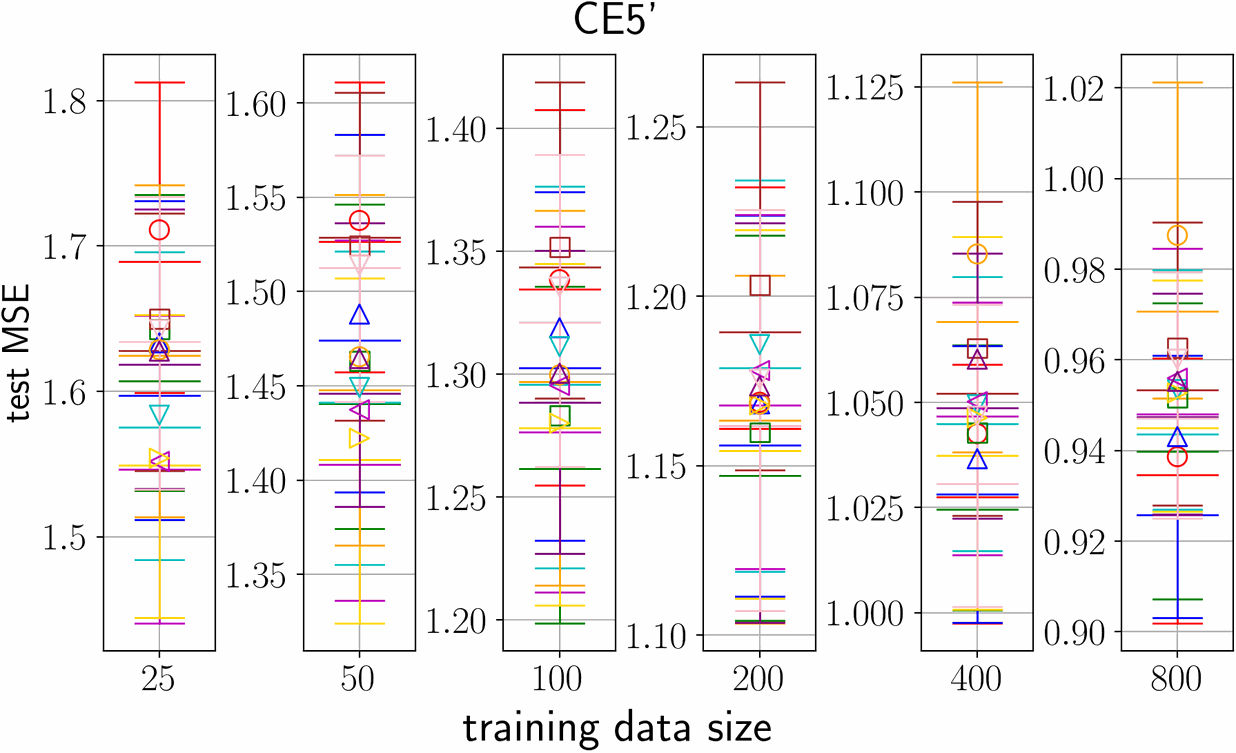}&
\includegraphics[height=2.8cm]{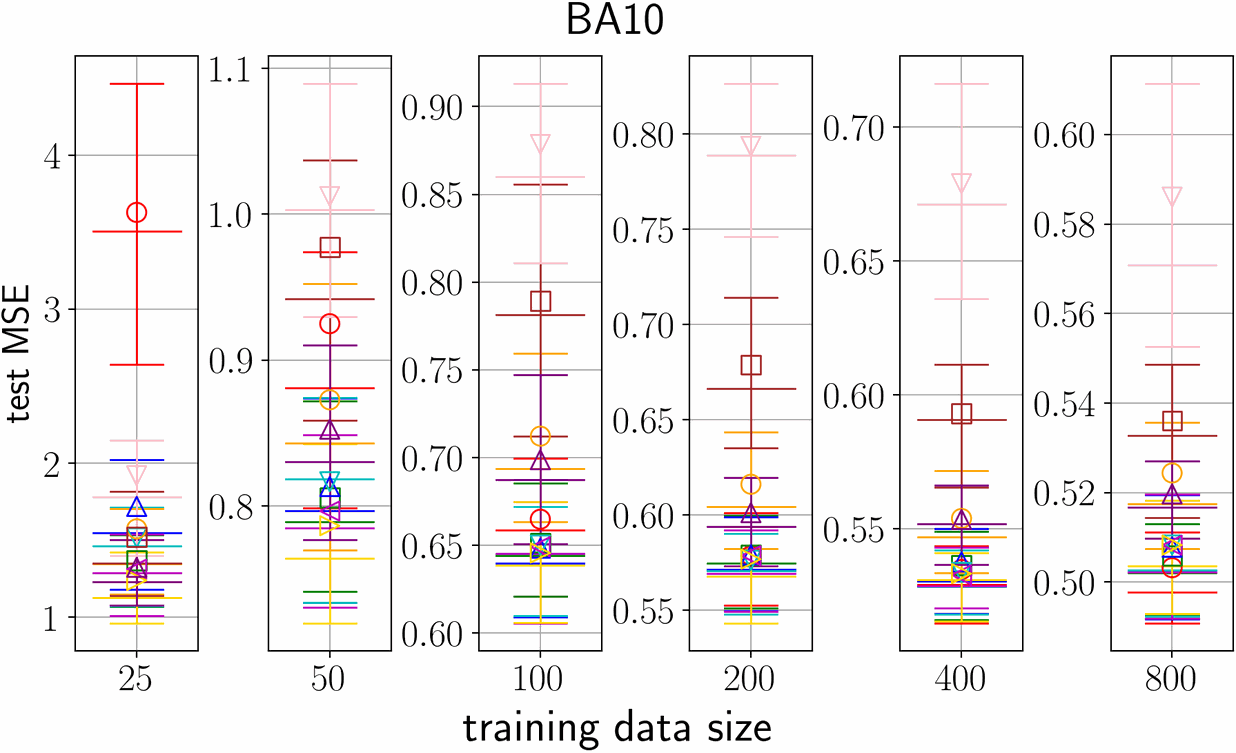}\\
\includegraphics[height=2.8cm]{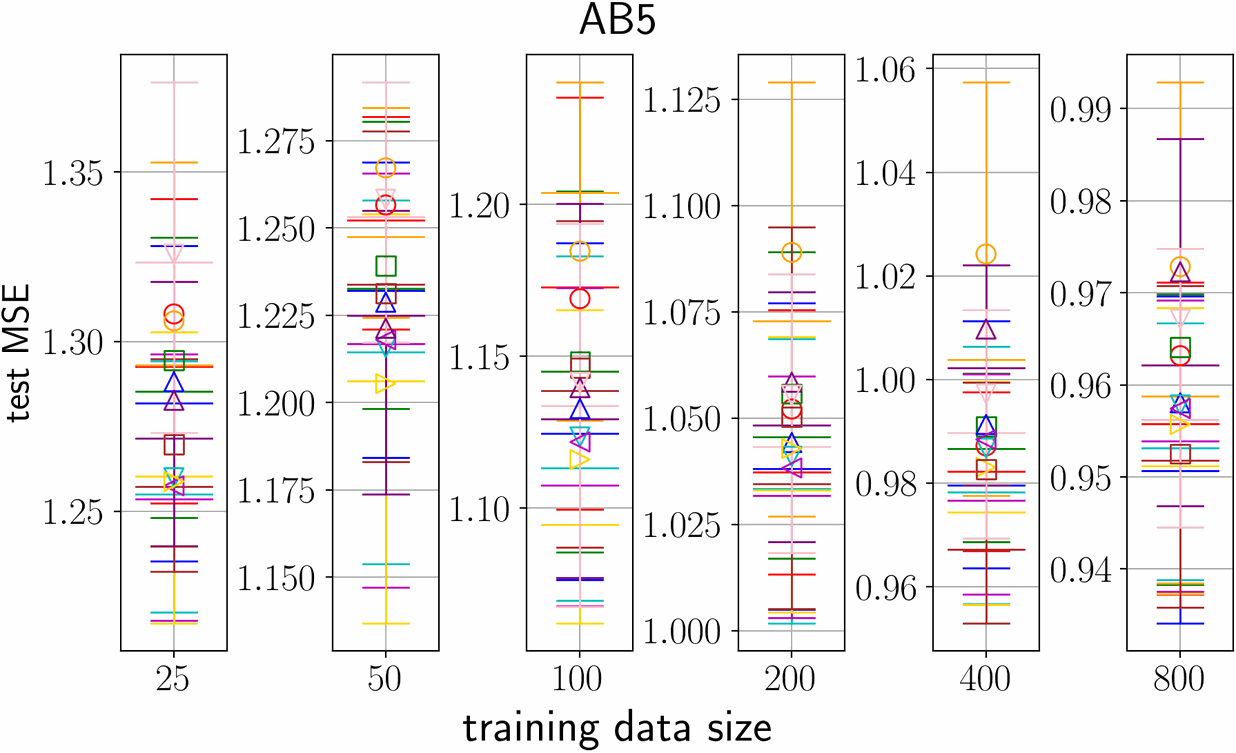}&
\includegraphics[height=2.8cm]{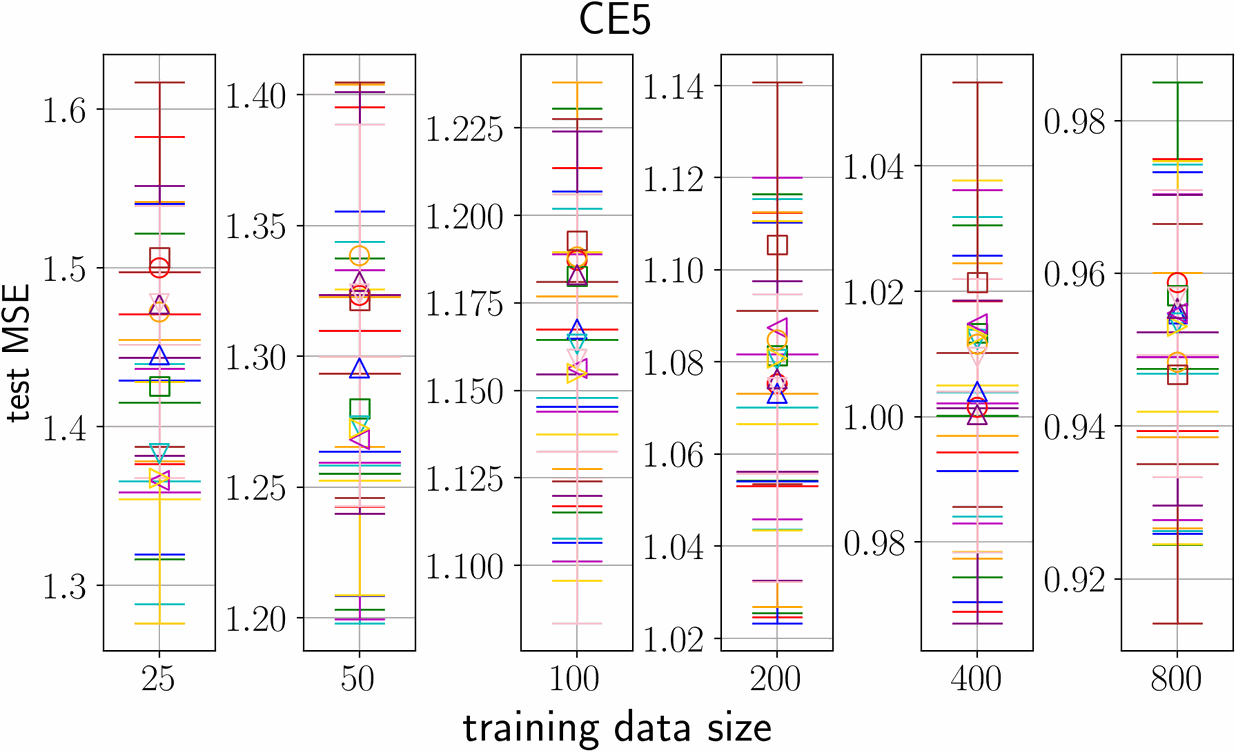}&
\includegraphics[height=2.8cm]{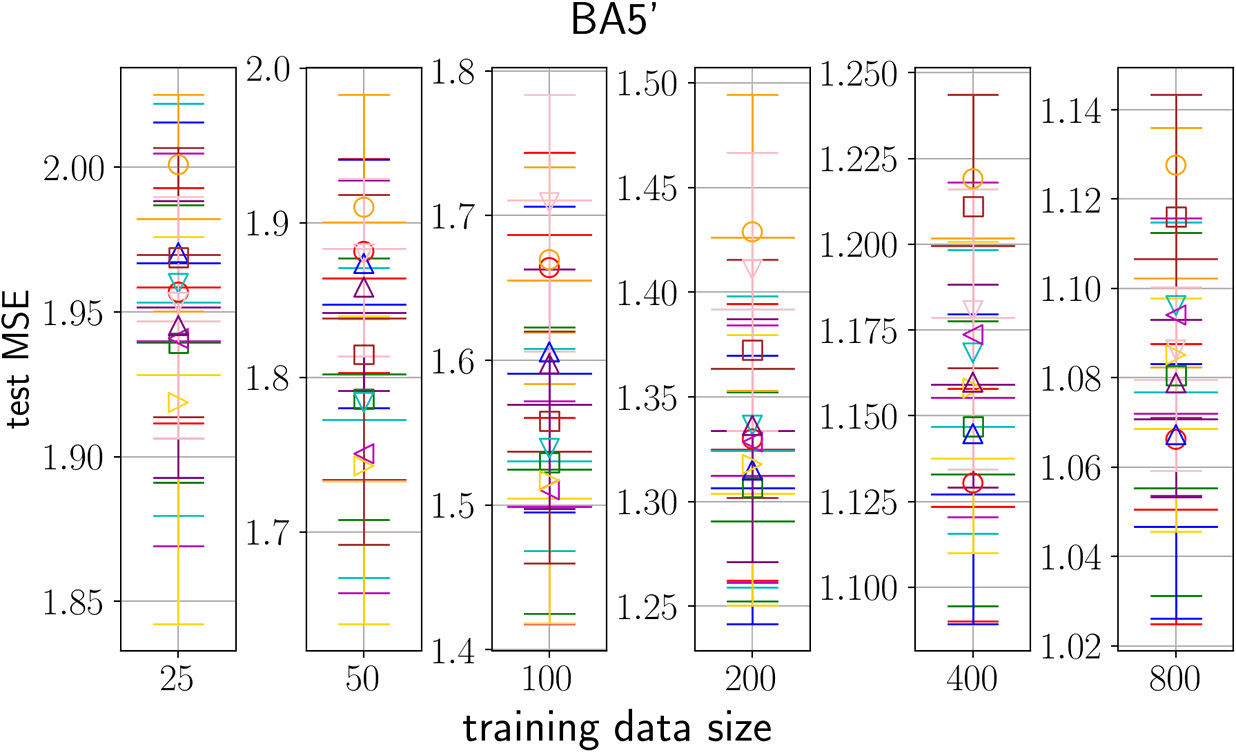}\\
\includegraphics[height=2.8cm]{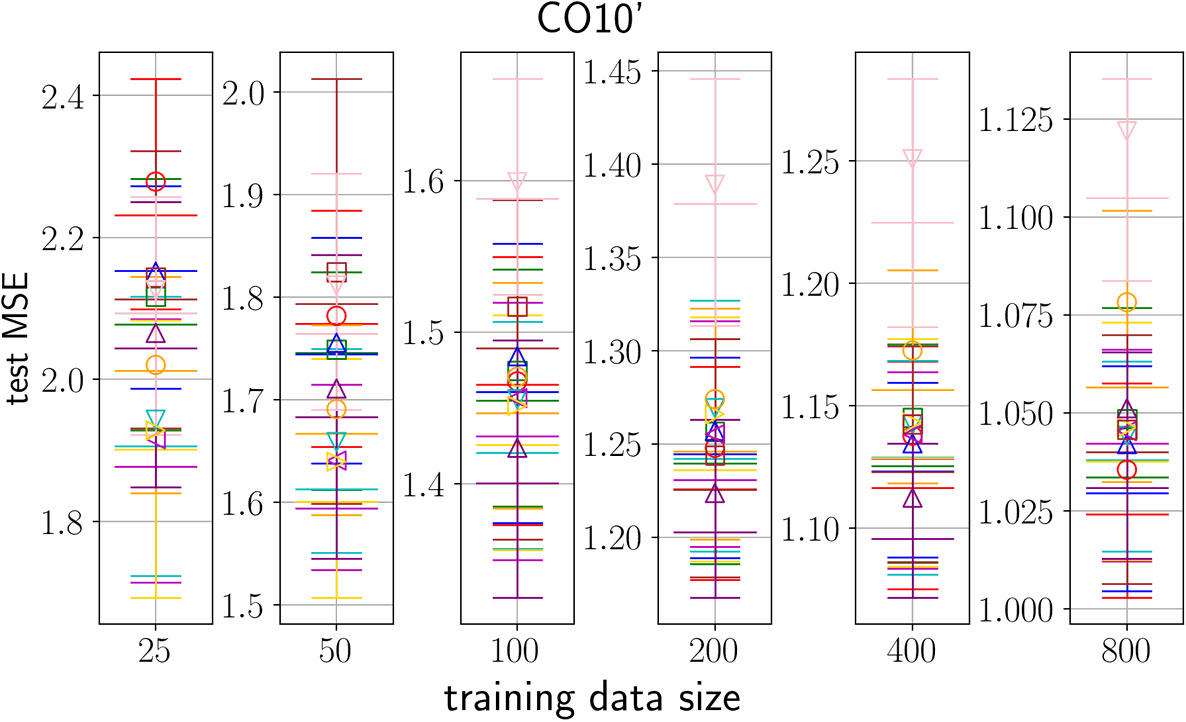}&
\includegraphics[height=2.8cm]{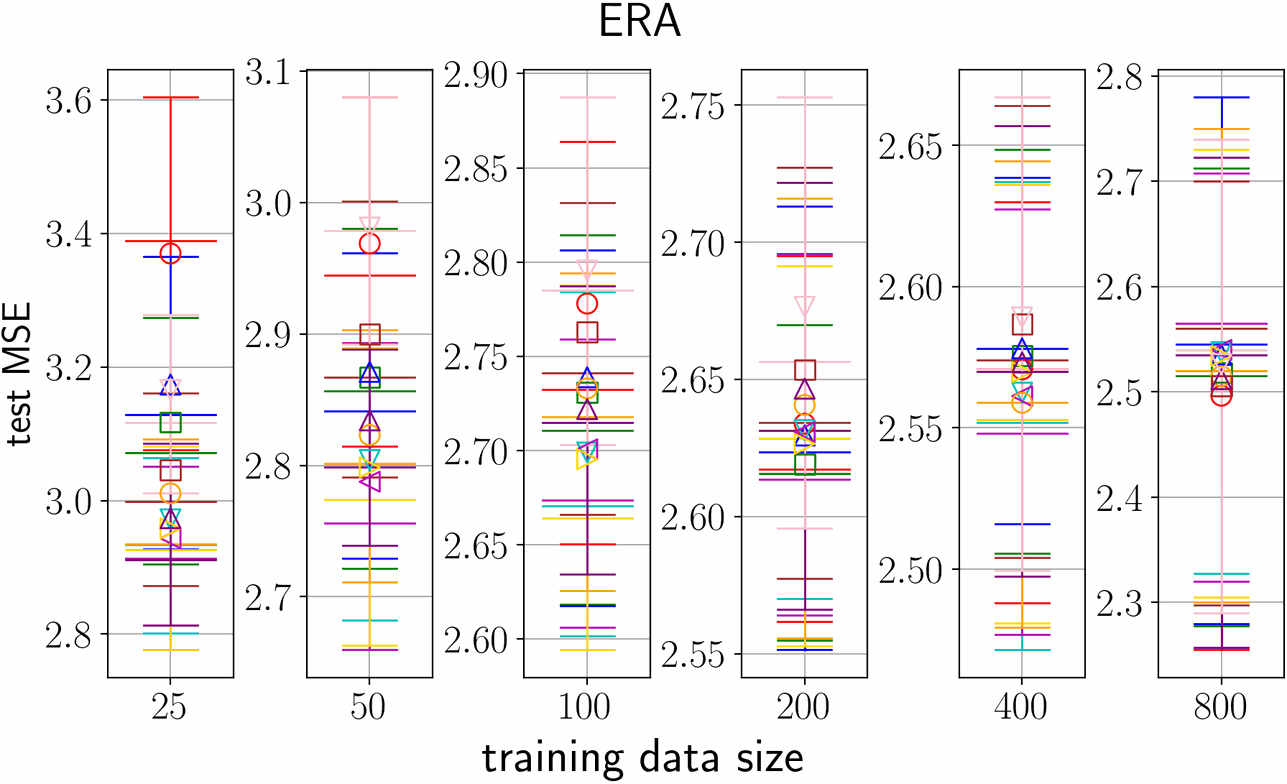}&
\includegraphics[height=2.8cm]{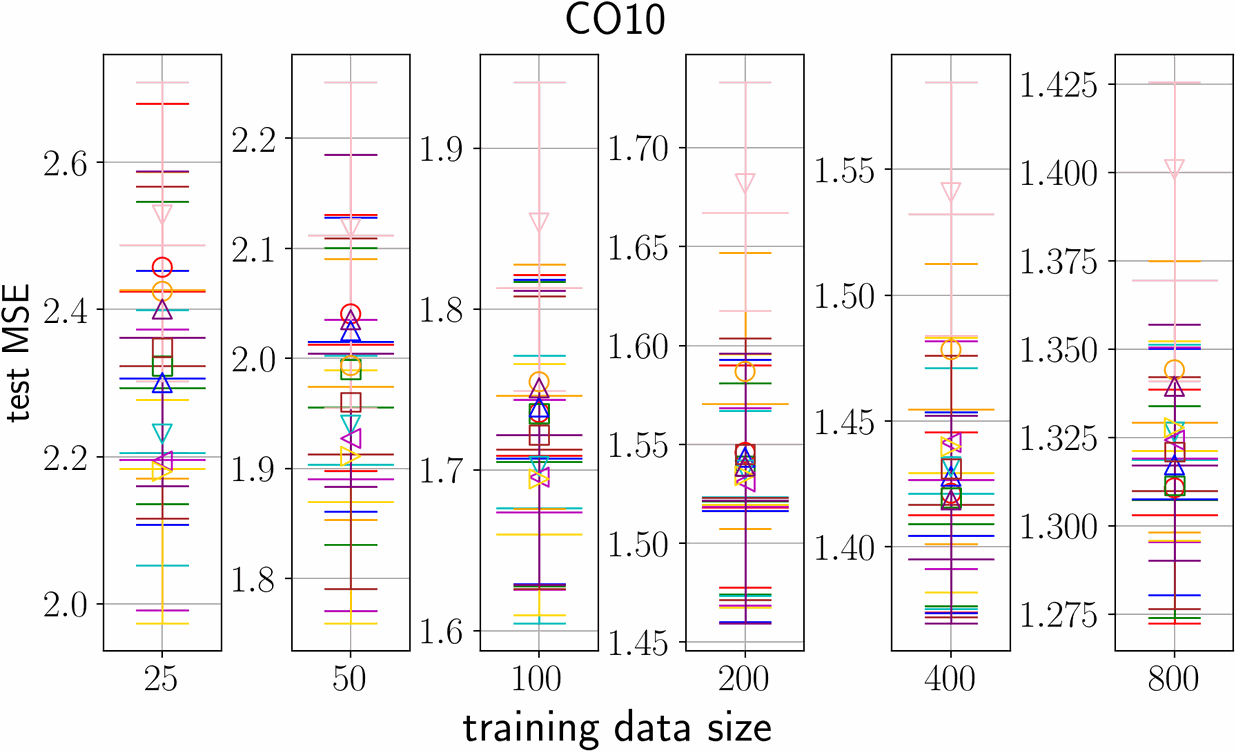}\\
\includegraphics[height=2.8cm]{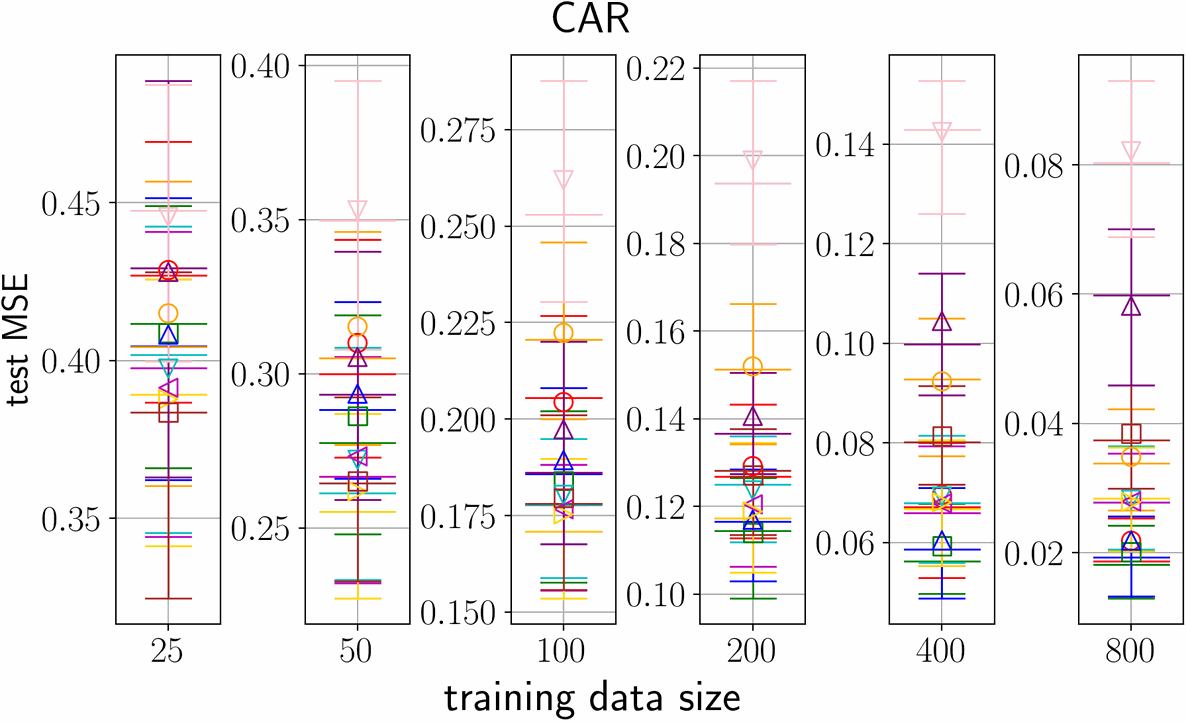}&
\includegraphics[height=2.8cm]{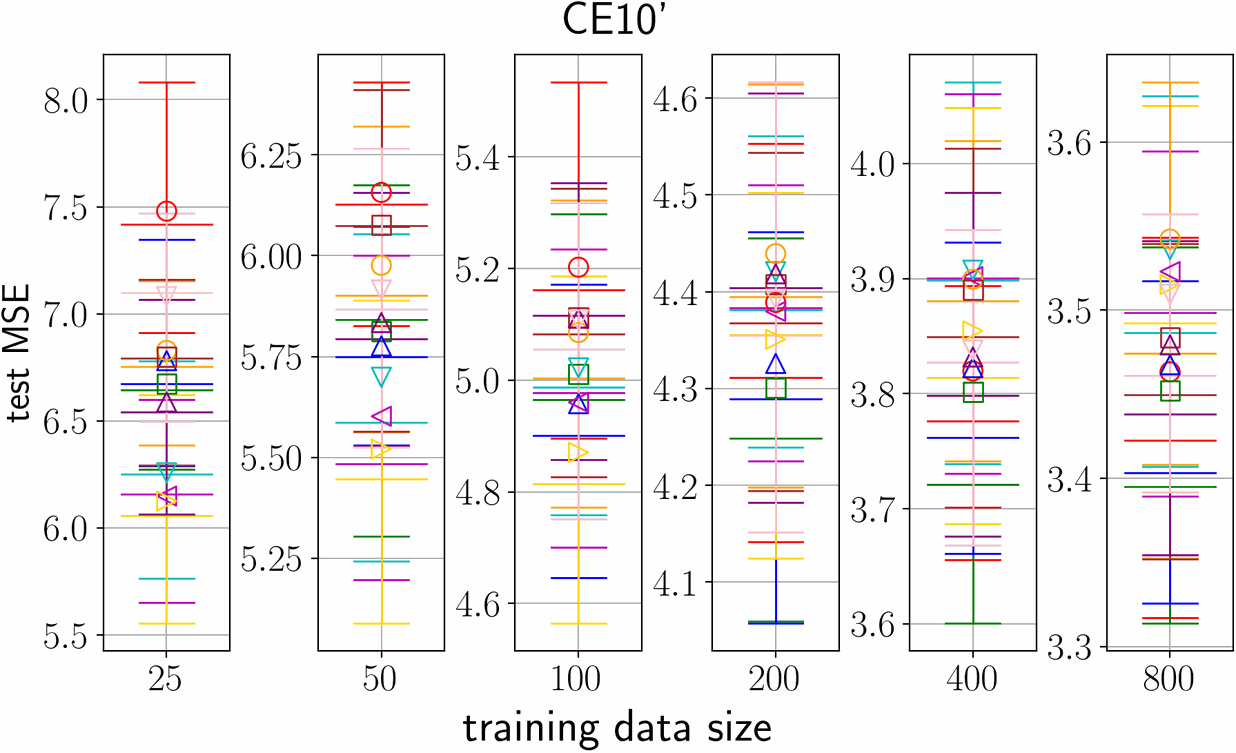}&
\includegraphics[height=2.8cm]{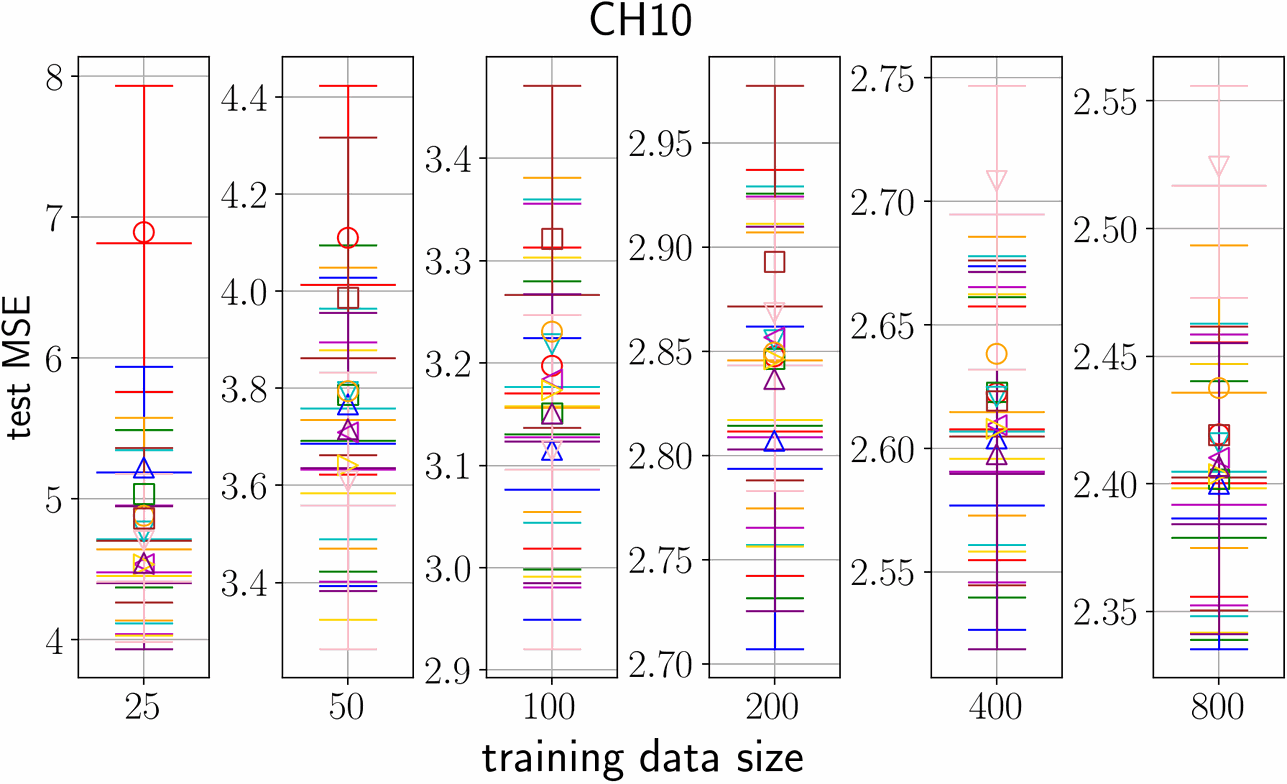}\\
\includegraphics[height=2.8cm]{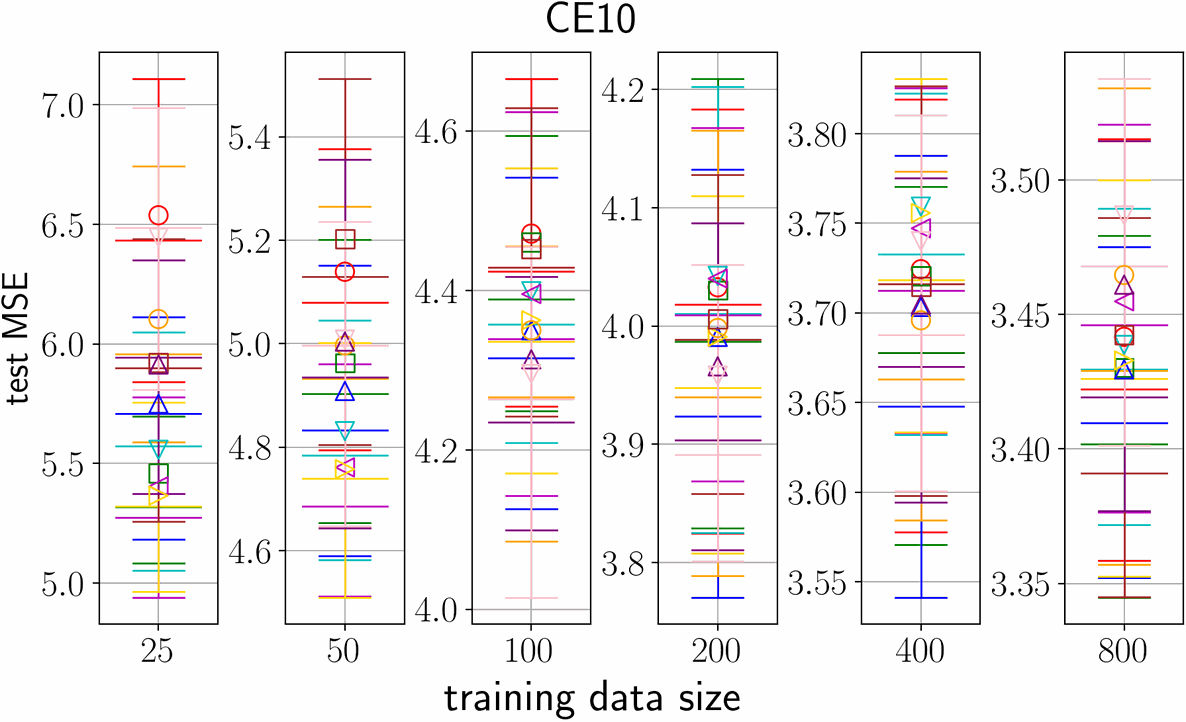}&
\includegraphics[height=2.8cm]{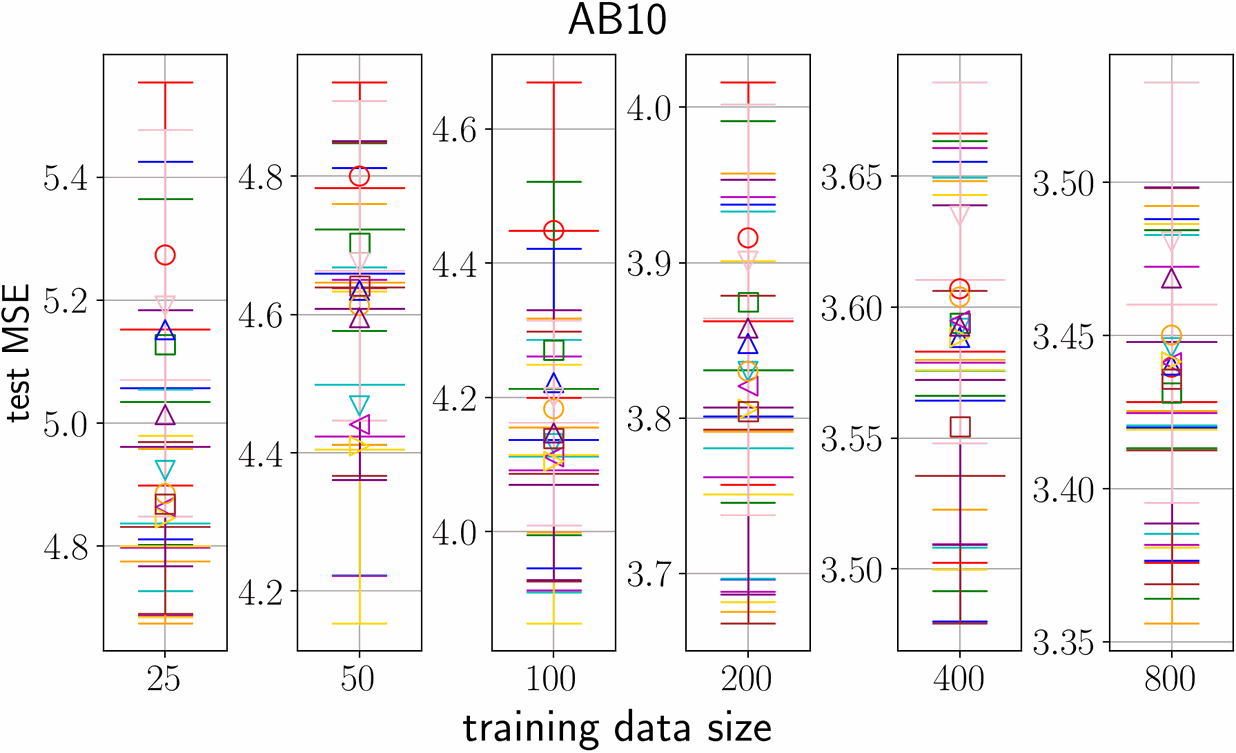}&
\includegraphics[height=2.8cm]{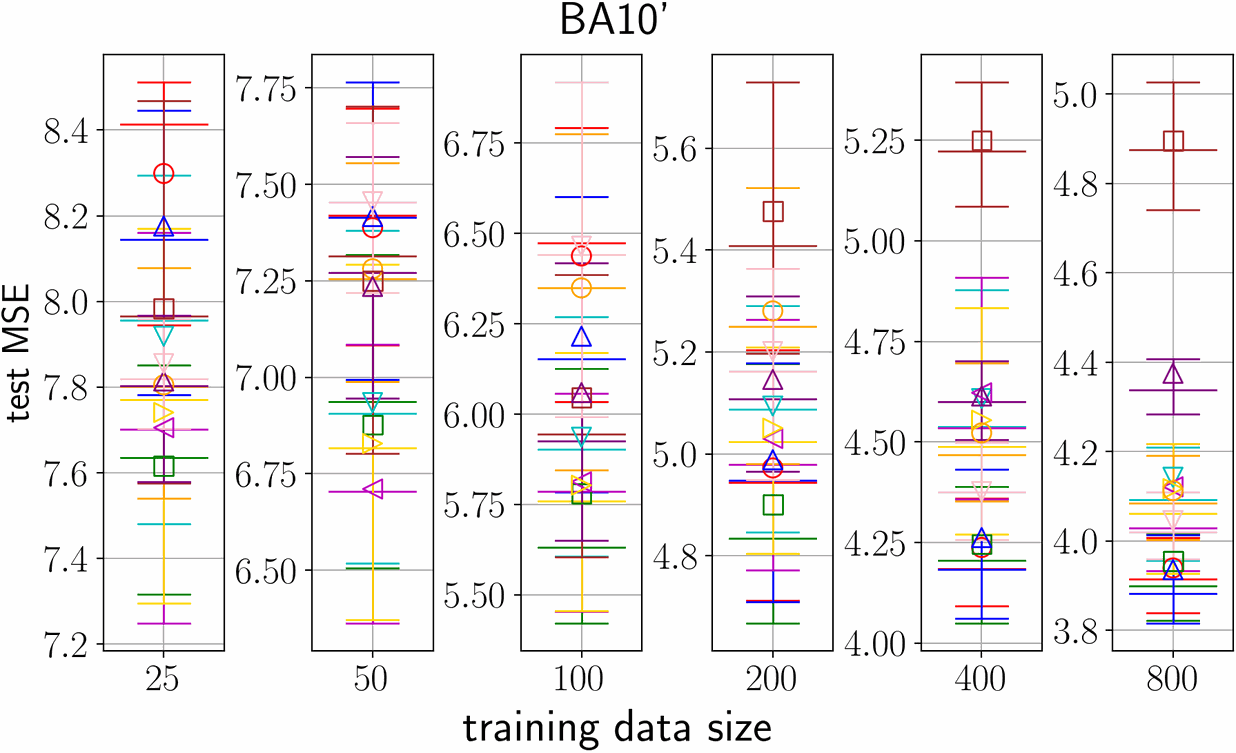}
\end{tabular}
\caption{%
Plot regarding 100-trial test MSEs
by Nonr-MLR (red solid \protect$\circ$), Prev-MLR (green solid \protect$\Box$), 
Stri-MLR (blue solid \protect$\triangle$), Nonr-AUL (cyan solid \protect$\triangledown$), 
Prev-AUL (magenta solid \protect$\triangleleft$), Stri-AUL (yellow solid \protect$\triangleright$),
Nonr-CL (orange dashed \protect$\circ$), Nonr-UL (brown dashed \protect$\Box$), 
Nonr-POCL (purple dashed \protect$\triangle$), and Nonr-POUL (pink dashed \protect$\triangledown$).
Lower short, middle long, and upper short bars and marker 
represent 0.25, 0.5, and 0.75 quantiles and mean.}
\label{fig:Performance-BestLam-MSE}
\end{figure*}

%========================================%
%==========%まだ(図は変える)
\begin{figure*}[p]
\centering%
\renewcommand{\arraystretch}{1.5}%
\renewcommand{\tabcolsep}{4pt}%
\begin{tabular}{cc}%
{\scriptsize NLL}&{\scriptsize MZE}\\
\begin{tabular}{cc}%
\begin{overpic}[width=4cm]{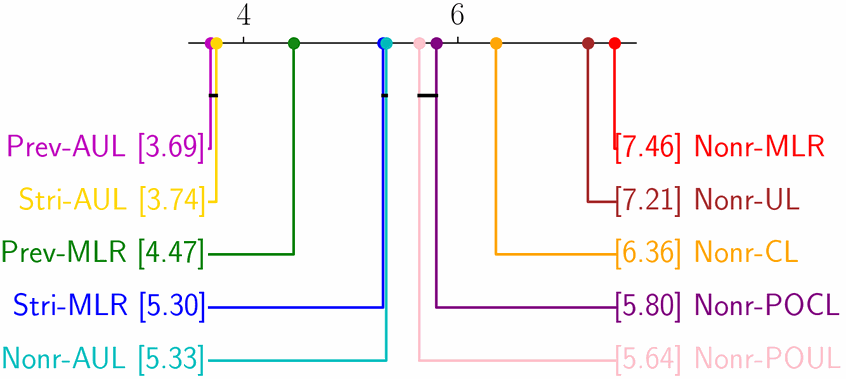}
\put(0,38){{\tiny$n_\tra\!=\!25$}}\end{overpic}&
\begin{overpic}[width=4cm]{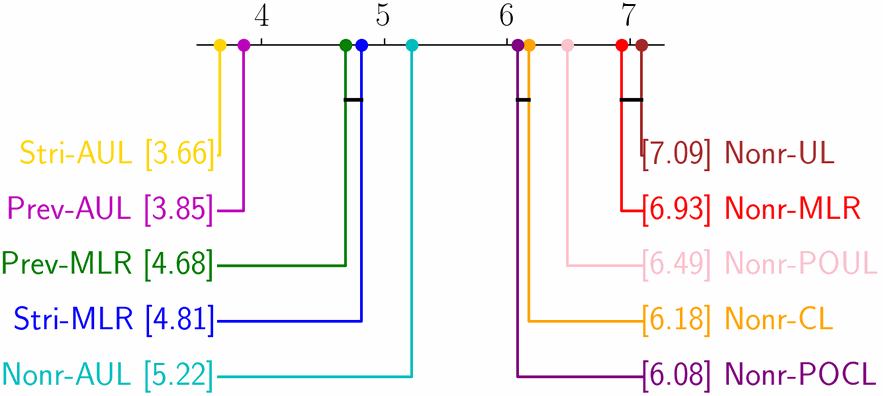}
\put(0,38){{\tiny$n_\tra\!=\!50$}}\end{overpic}\\
\begin{overpic}[width=4cm]{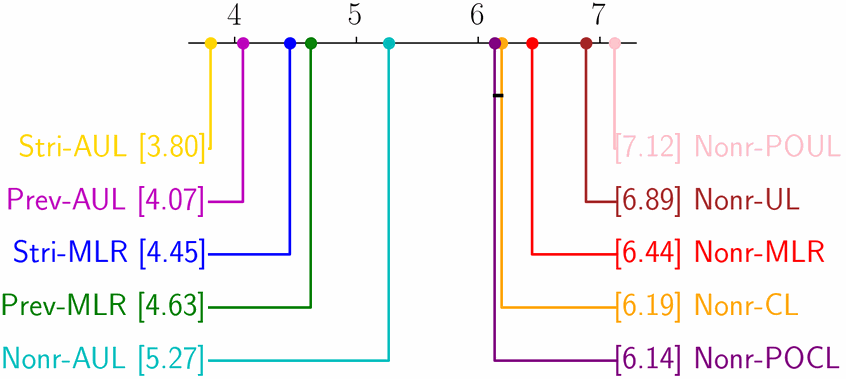}
\put(0,38){{\tiny$n_\tra\!=\!100$}}\end{overpic}&
\begin{overpic}[width=4cm]{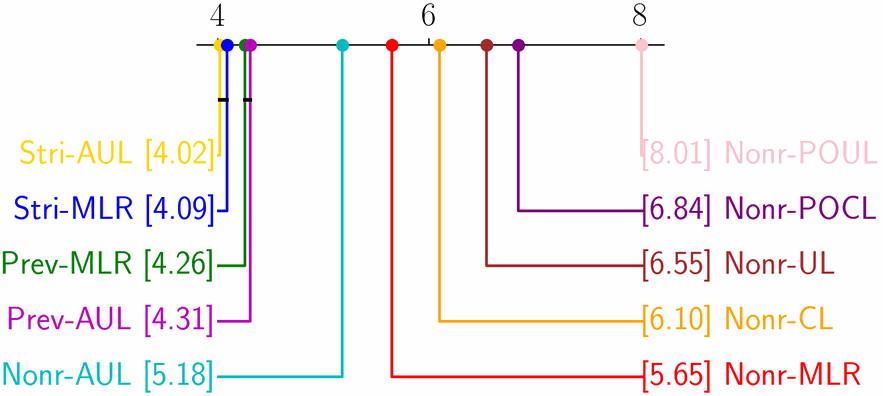}
\put(0,38){{\tiny$n_\tra\!=\!200$}}\end{overpic}\\
\begin{overpic}[width=4cm]{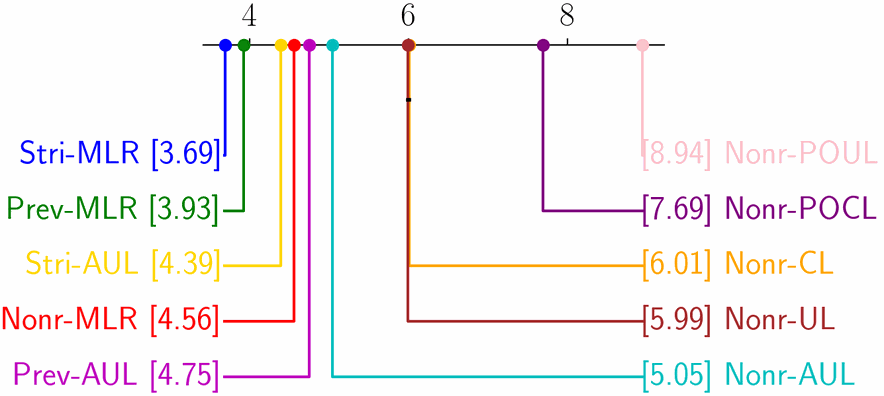}
\put(0,38){{\tiny$n_\tra\!=\!400$}}\end{overpic}&
\begin{overpic}[width=4cm]{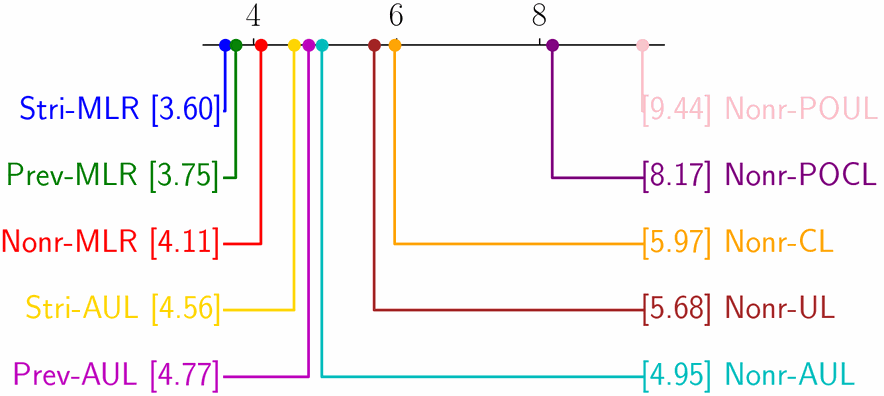}
\put(0,38){{\tiny$n_\tra\!=\!800$}}\end{overpic}\\
\end{tabular}
&
\begin{tabular}{cc}%
\begin{overpic}[width=4cm]{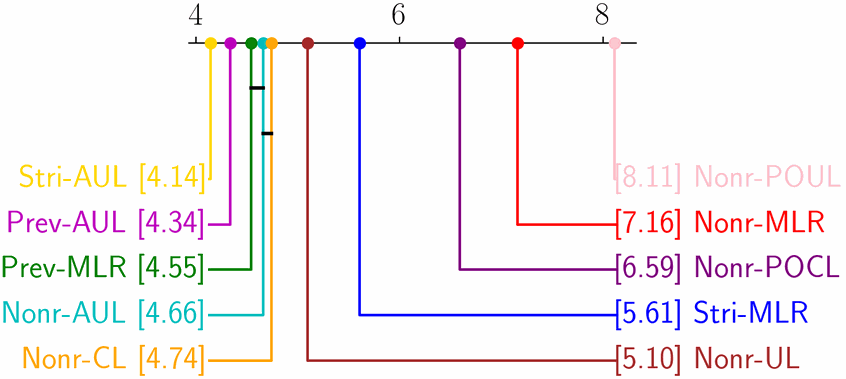}
\put(0,38){{\tiny$n_\tra\!=\!25$}}\end{overpic}&
\begin{overpic}[width=4cm]{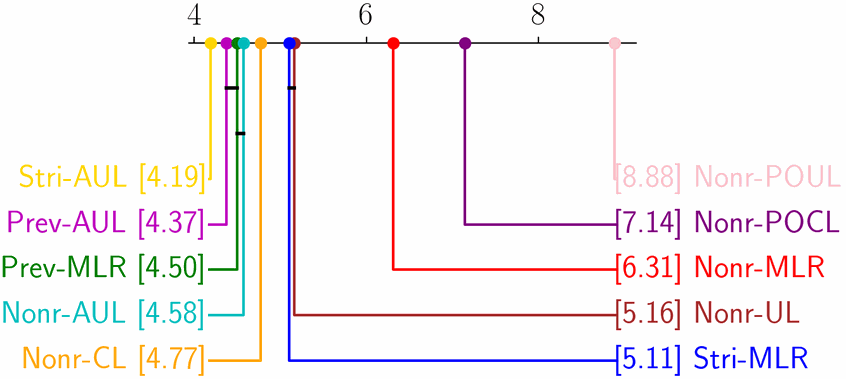}
\put(0,38){{\tiny$n_\tra\!=\!50$}}\end{overpic}\\
\begin{overpic}[width=4cm]{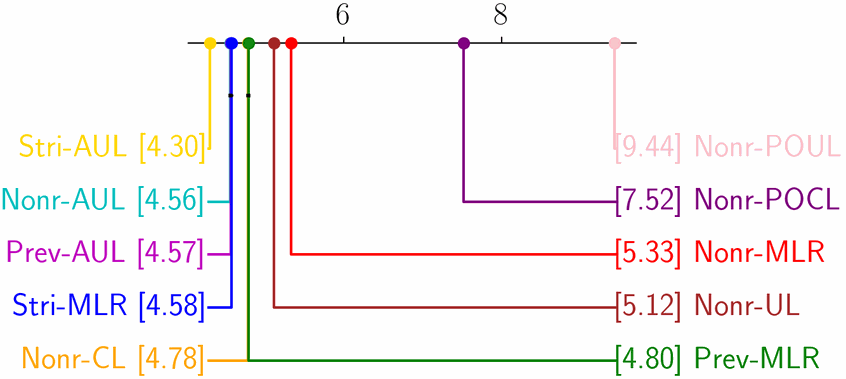}
\put(0,38){{\tiny$n_\tra\!=\!100$}}\end{overpic}&
\begin{overpic}[width=4cm]{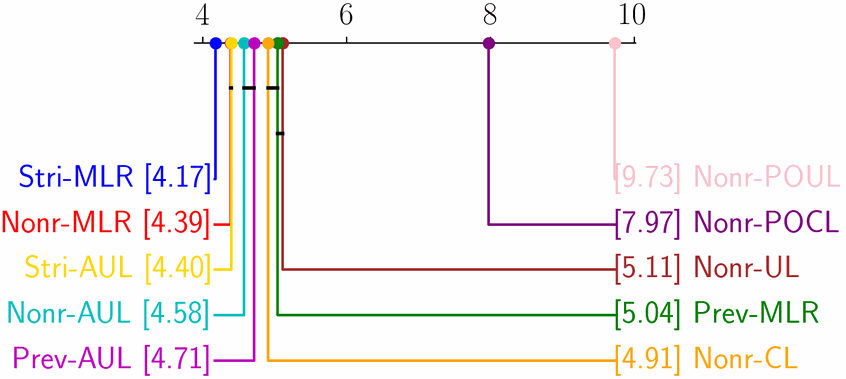}
\put(0,38){{\tiny$n_\tra\!=\!200$}}\end{overpic}\\
\begin{overpic}[width=4cm]{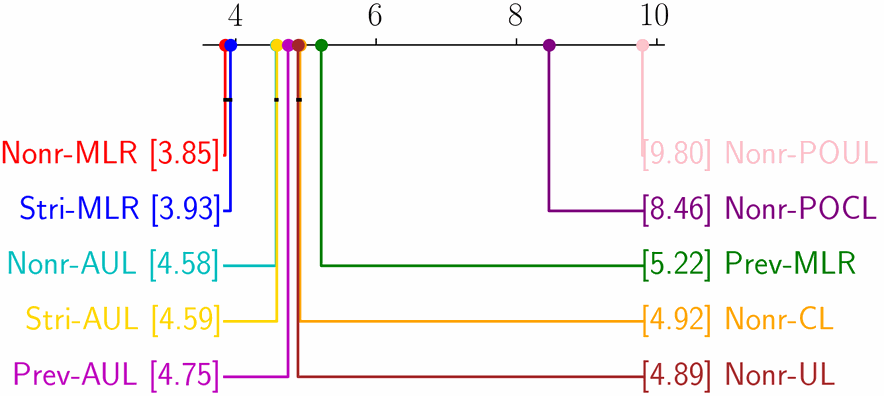}
\put(0,38){{\tiny$n_\tra\!=\!400$}}\end{overpic}&
\begin{overpic}[width=4cm]{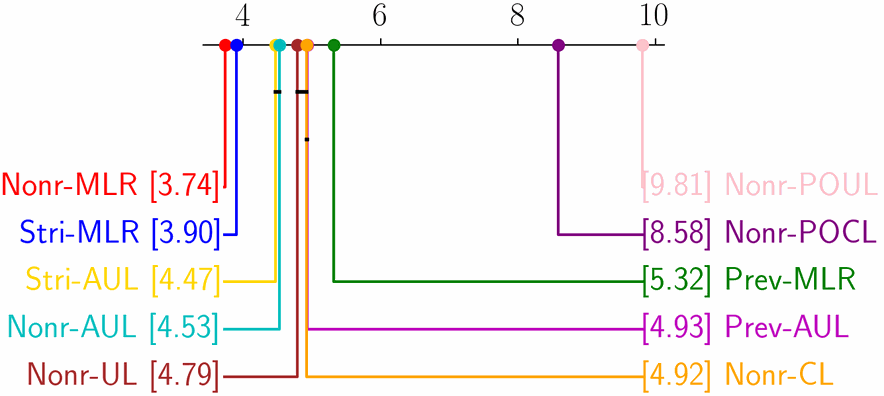}
\put(0,38){{\tiny$n_\tra\!=\!800$}}\end{overpic}\\
\end{tabular}\\
~~&~~\\
{\scriptsize MAE}&{\scriptsize MSE}\\
\begin{tabular}{cc}%
\begin{overpic}[width=4cm]{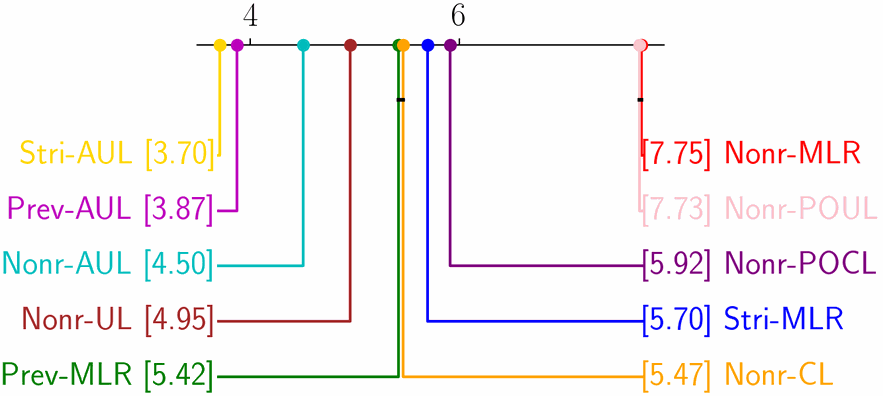}
\put(0,38){{\tiny$n_\tra\!=\!25$}}\end{overpic}&
\begin{overpic}[width=4cm]{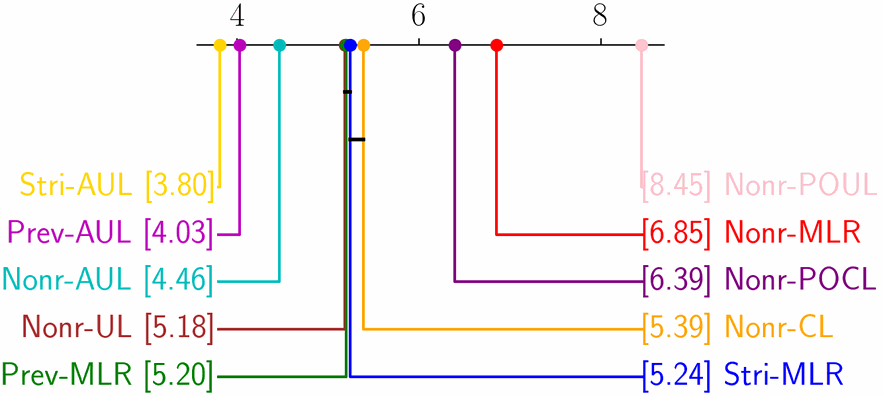}
\put(0,38){{\tiny$n_\tra\!=\!50$}}\end{overpic}\\
\begin{overpic}[width=4cm]{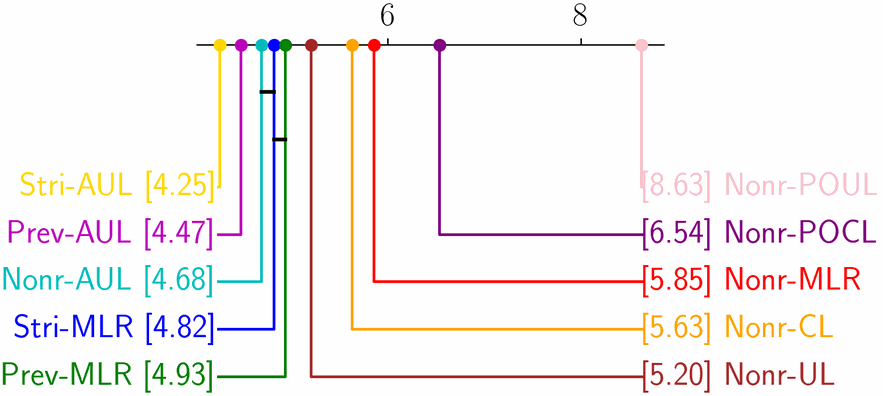}
\put(0,38){{\tiny$n_\tra\!=\!100$}}\end{overpic}&
\begin{overpic}[width=4cm]{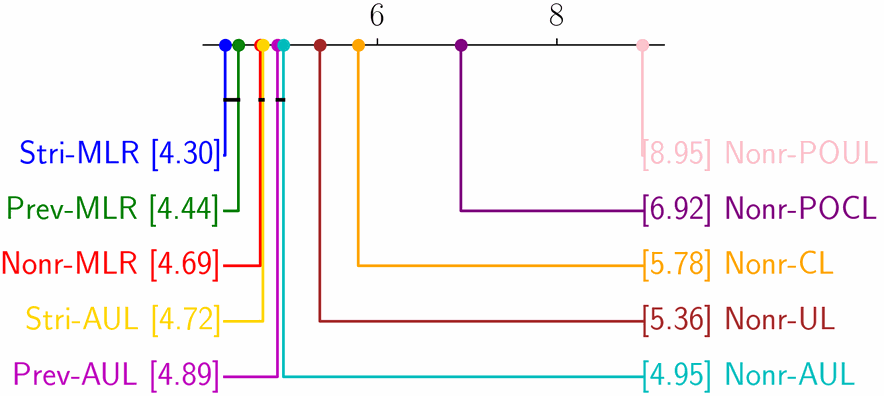}
\put(0,38){{\tiny$n_\tra\!=\!200$}}\end{overpic}\\
\begin{overpic}[width=4cm]{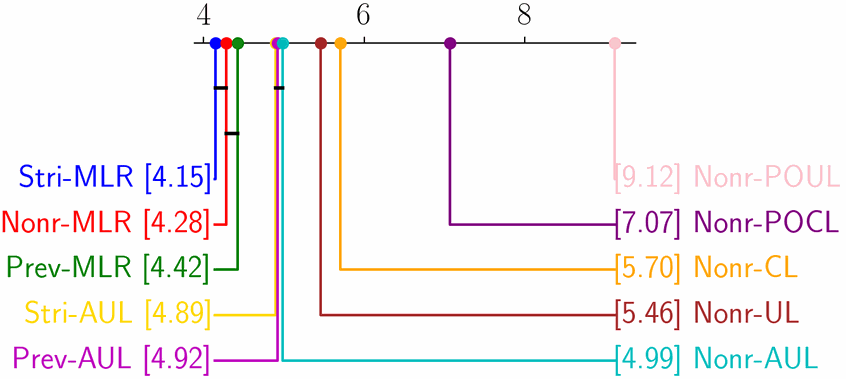}
\put(0,38){{\tiny$n_\tra\!=\!400$}}\end{overpic}&
\begin{overpic}[width=4cm]{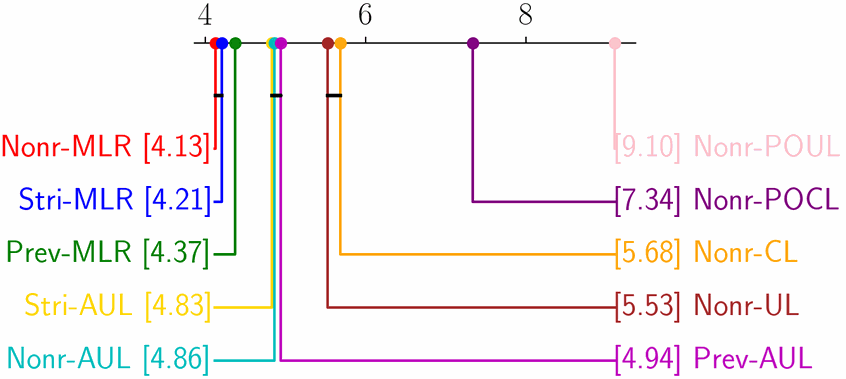}
\put(0,38){{\tiny$n_\tra\!=\!800$}}\end{overpic}\\
\end{tabular}
&
\begin{tabular}{cc}%
\begin{overpic}[width=4cm]{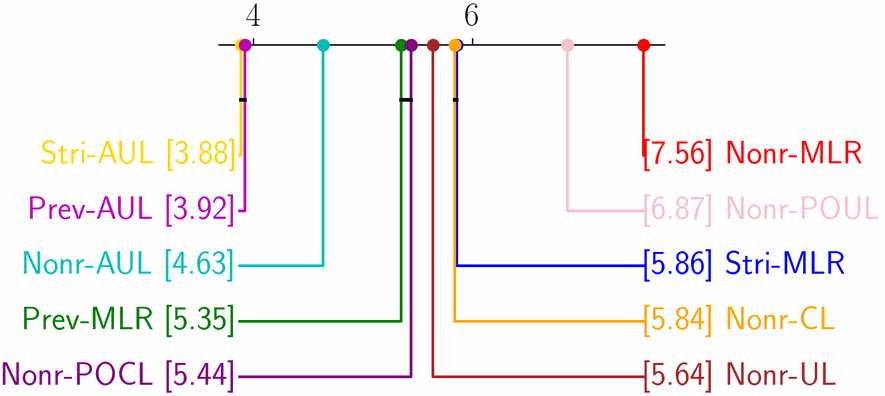}
\put(0,38){{\tiny$n_\tra\!=\!25$}}\end{overpic}&
\begin{overpic}[width=4cm]{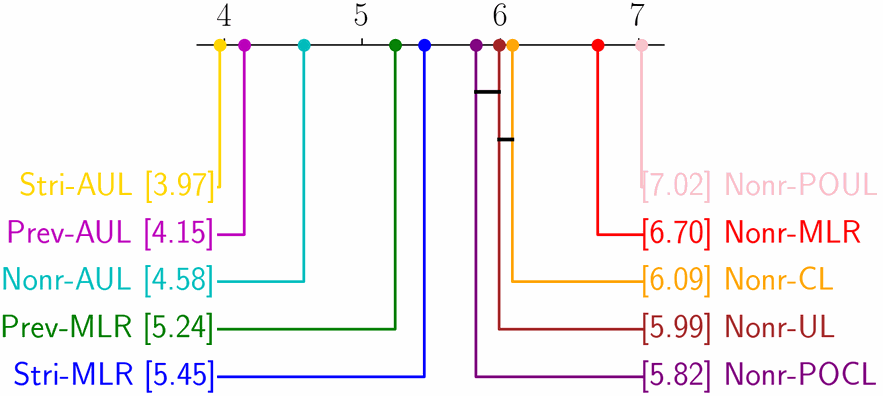}
\put(0,38){{\tiny$n_\tra\!=\!50$}}\end{overpic}\\
\begin{overpic}[width=4cm]{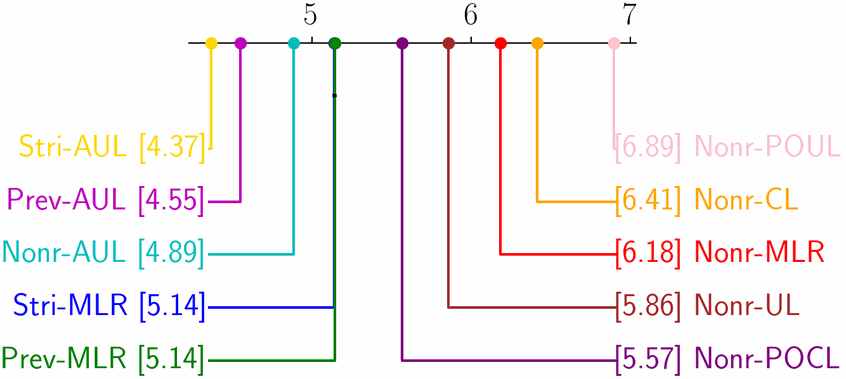}
\put(0,38){{\tiny$n_\tra\!=\!100$}}\end{overpic}&
\begin{overpic}[width=4cm]{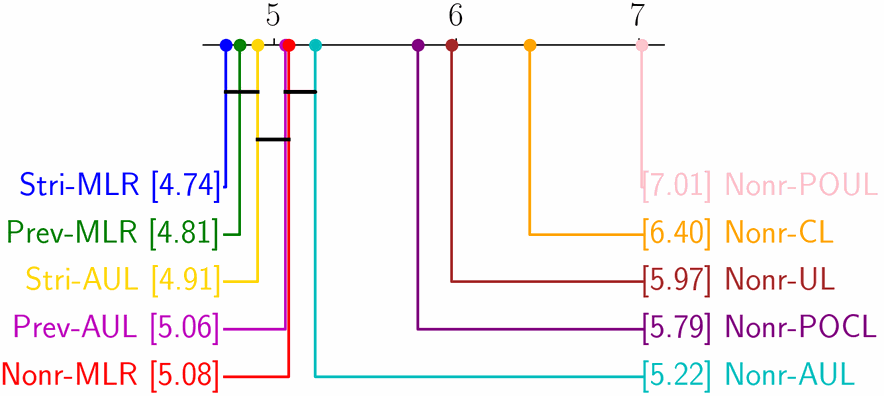}
\put(0,38){{\tiny$n_\tra\!=\!200$}}\end{overpic}\\
\begin{overpic}[width=4cm]{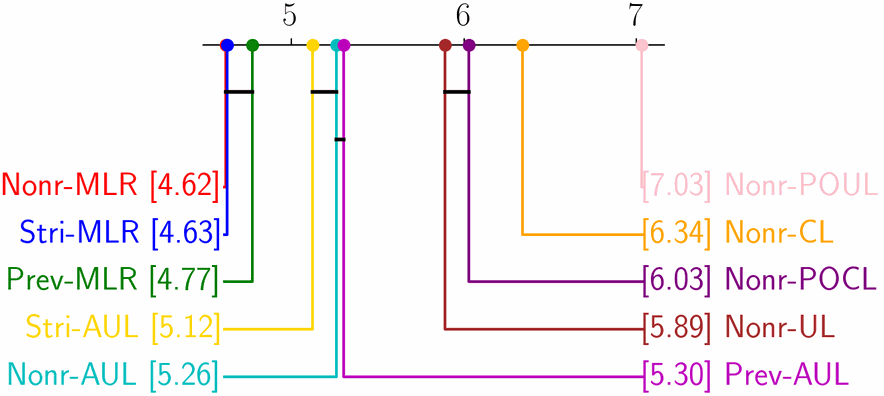}
\put(0,38){{\tiny$n_\tra\!=\!400$}}\end{overpic}&
\begin{overpic}[width=4cm]{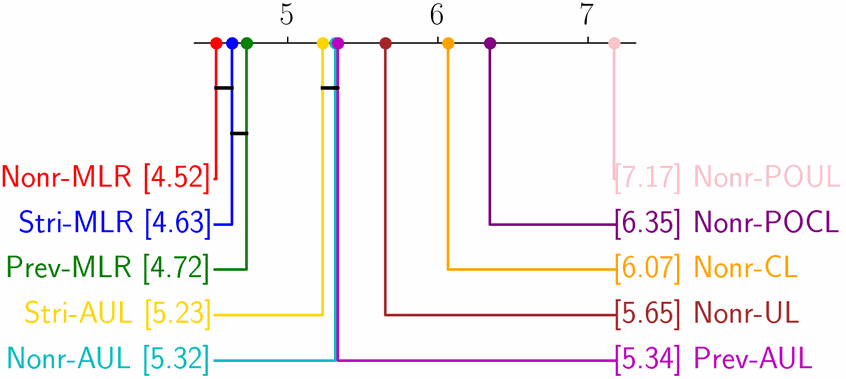}
\put(0,38){{\tiny$n_\tra\!=\!800$}}\end{overpic}\\
\end{tabular}
\end{tabular}
\caption{%
Critical difference diagram \cite{demsar06a} based on 
Conover-Friedman test \cite{conover1981rank} of 
significance level $0.05$ regarding the mean of test 
NLLs (upper left), MZEs (upper right), MAEs (lower left), 
and MSEs (lower right) with $n_\tra=25,\ldots,800$.
$A[B]$ implies that the average rank of the mean test error of the method $A$ is $B$,
and the smaller the value of $B$, the better the corresponding method $A$ is.
Black horizontal line connecting methods implies that no significant difference was detected between them.}
\label{fig:CDD-S}
\end{figure*}

%========================================%
%==========%
\begin{table*}[p]
\centering%
\renewcommand{\arraystretch}{0.75}%
\renewcommand{\tabcolsep}{0.5pt}%
\caption{%
A cell for each dataset and $n_\tra$ shows the best learning method 
(`${\rm n}$' in red if Nonr-MLR or cyan if Nonr-AUL, 
`${\rm p}$' in green if Prev-MLR or magenta if Prev-AUL, 
and `${\rm s}$' in blue if AStri-MLR or yellow if Stri-AUL) 
among methods with MLR or AUL models regarding the test 
NLLs (upper left), MZEs (upper right), MAEs (lower left), and MSEs (lower right)
(and methods that drew with the best method in the Mann-Whitney 
U-test with the significance level $0.05$ at the subscript if exists).}
\label{tab:best-Performance}
\begin{tabular}{ccc}
NLL&~~&MZE\\
\begin{tabular}{ccc|cccccc|cccccc}\toprule
\multirow{2}{*}{dataset} & \multirow{2}{*}{$K$} & \multirow{2}{*}{MS} & 
\multicolumn{6}{c|}{for MLR with $n_\tra=$}& \multicolumn{6}{c}{for AUL with $n_\tra=$}\\
&&&25&50&100&200&400&800&25&50&100&200&400&800\\
\midrule
CAR & 4 & $.0146\!\pm\!{.0095}$ &
$\cs_{\cp}$&$\cp_{\cs}$&$\cp$&$\cp_{\cs}$&$\cs_{\cp}$&$\cp_{\cs}$&$\as_{\ap}$&$\as$&$\as_{\ap\an}$&$\as_{\ap\an}$&$\as_{\ap\an}$&$\ap_{\as\an}$\\
SWD & 4 & $.3980\!\pm\!{.0211}$ &
$\cs_{\cp}$&$\cs$&$\cs$&$\cs$&$\cs_{\cp\cn}$&$\cn_{\cs\cp}$&$\as_{\ap}$&$\ap_{\as\an}$&$\ap_{\as}$&$\ap_{\as\an}$&$\ap_{\an\as}$&$\an_{\as\ap}$\\
\midrule
BA5 & 5 & $.1890\!\pm\!{.0268}$ &
$\cp$&$\cp_{\cs}$&$\cp_{\cs\cn}$&$\cs_{\cn}$&$\cp_{\cs\cn}$&$\cs_{\cn\cp}$&$\as_{\ap}$&$\as_{\ap}$&$\as_{\ap\an}$&$\as_{\an\ap}$&$\ap_{\an\as}$&$\as_{\an\ap}$\\
CO5' & 5 & $.2664\!\pm\!{.0357}$ &
$\cs$&$\cs_{\cp}$&$\cs_{\cn\cp}$&$\cs_{\cn\cp}$&$\cn_{\cp\cs}$&$\cn_{\cp\cs}$&$\as$&$\as$&$\as$&$\as_{\ap\an}$&$\as_{\an\ap}$&$\an_{\as\ap}$\\
CO5 & 5 & $.3095\!\pm\!{.0329}$ &
$\cs$&$\cs$&$\cs_{\cp\cn}$&$\cs_{\cp\cn}$&$\cs_{\cp\cn}$&$\cn_{\cs\cp}$&$\as$&$\as$&$\as$&$\as_{\an\ap}$&$\as_{\an\ap}$&$\an_{\as\ap}$\\
LEV & 5 & $.3685\!\pm\!{.0202}$ &
$\cs_{\cp}$&$\cs$&$\cs$&$\cs_{\cp}$&$\cs_{\cp\cn}$&$\cs_{\cn\cp}$&$\as_{\ap}$&$\ap_{\as\an}$&$\as_{\ap\an}$&$\ap_{\as\an}$&$\as_{\an\ap}$&$\ap_{\as\an}$\\
CH5 & 5 & $.4172\!\pm\!{.0455}$ &
$\cp$&$\cp_{\cs}$&$\cs_{\cp}$&$\cp_{\cs}$&$\cs$&$\cs_{\cp}$&$\as$&$\as_{\ap}$&$\as_{\ap}$&$\as_{\ap}$&$\as_{\ap\an}$&$\as_{\ap\an}$\\
CE5' & 5 & $.4643\!\pm\!{.0440}$ &
$\cp_{\cs}$&$\cp_{\cs}$&$\cp_{\cs}$&$\cs_{\cp\cn}$&$\cs_{\cp\cn}$&$\cn_{\cp\cs}$&$\as$&$\as$&$\as$&$\as$&$\as_{\an\ap}$&$\as_{\ap\an}$\\
CE5 & 5 & $.4760\!\pm\!{.0460}$ &
$\cp_{\cs}$&$\cp_{\cs}$&$\cs_{\cp}$&$\cs_{\cp\cn}$&$\cs_{\cn\cp}$&$\cs$&$\as$&$\as_{\ap}$&$\as_{\ap}$&$\as_{\ap\an}$&$\as_{\ap\an}$&$\as_{\ap\an}$\\
AB5 & 5 & $.4873\!\pm\!{.0324}$ &
$\cp$&$\cs_{\cp\cn}$&$\cs$&$\cs$&$\cs_{\cp}$&$\cs_{\cp}$&$\as_{\ap}$&$\as$&$\as_{\ap}$&$\as_{\ap}$&$\as_{\ap}$&$\as_{\ap\an}$\\
BA5' & 5 & $.5524\!\pm\!{.0373}$ &
$\cp$&$\cp$&$\cp$&$\cp$&$\cp_{\cs\cn}$&$\cs$&$\ap$&$\ap$&$\ap$&$\as_{\ap}$&$\as$&$\as$\\
\midrule
WQR & 6 & $.3678\!\pm\!{.0260}$ &
$\cs$&$\cs$&$\cs$&$\cs$&$\cs$&$\cs_{\cn}$&$\as_{\an\ap}$&$\as$&$\as_{\ap\an}$&$\as_{\ap\an}$&$\as_{\an\ap}$&$\an_{\as\ap}$\\
\midrule
ERA & 9 & $.7208\!\pm\!{.0198}$ &
$\cp$&$\cs_{\cp}$&$\cs$&$\cp_{\cs}$&$\cs_{\cp\cn}$&$\cs_{\cp\cn}$&$\as_{\ap}$&$\as_{\ap}$&$\as_{\ap}$&$\as_{\ap}$&$\as_{\ap\an}$&$\as_{\ap\an}$\\
\midrule
BA10 & 10 & $.3815\!\pm\!{.0287}$ &
$\cp$&$\cs_{\cp}$&$\cs_{\cp}$&$\cp_{\cs\cn}$&$\cp_{\cn\cs}$&$\cp_{\cn\cs}$&$\as_{\ap}$&$\ap_{\as}$&$\as_{\ap\an}$&$\as_{\ap\an}$&$\as_{\ap\an}$&$\ap_{\as\an}$\\
CO10' & 10 & $.4720\!\pm\!{.0369}$ &
$\cp_{\cs}$&$\cs_{\cp}$&$\cs_{\cp}$&$\cs_{\cp\cn}$&$\cp_{\cn\cs}$&$\cn_{\cp\cs}$&$\as$&$\as$&$\as$&$\as_{\ap\an}$&$\as_{\an\ap}$&$\an_{\as\ap}$\\
CO10 & 10 & $.5031\!\pm\!{.0340}$ &
$\cp_{\cs}$&$\cs_{\cp}$&$\cs_{\cp}$&$\cs_{\cp}$&$\cp_{\cn\cs}$&$\cn_{\cp\cs}$&$\as_{\ap}$&$\as$&$\as_{\ap}$&$\as_{\ap\an}$&$\ap_{\as\an}$&$\as_{\ap\an}$\\
CH10 & 10 & $.6238\!\pm\!{.0466}$ &
$\cp$&$\cp$&$\cp_{\cs}$&$\cs_{\cp}$&$\cs_{\cp}$&$\cs_{\cp}$&$\ap$&$\ap_{\as}$&$\as_{\ap}$&$\as_{\ap}$&$\as_{\ap}$&$\ap_{\as}$\\
AB10 & 10 & $.6847\!\pm\!{.0239}$ &
$\cp$&$\cp$&$\cs$&$\cs$&$\cs$&$\cs_{\cp}$&$\ap$&$\as$&$\as$&$\as_{\ap}$&$\as_{\ap}$&$\ap_{\as\an}$\\
CE10' & 10 & $.6898\!\pm\!{.0409}$ &
$\cp$&$\cp$&$\cp$&$\cp_{\cs}$&$\cp_{\cs}$&$\cp$&$\ap$&$\ap$&$\ap_{\as}$&$\as$&$\as_{\ap}$&$\as_{\ap}$\\
CE10 & 10 & $.6911\!\pm\!{.0313}$ &
$\cp$&$\cp$&$\cp_{\cs}$&$\cp_{\cs}$&$\cs_{\cp\cn}$&$\cs_{\cp}$&$\as_{\ap}$&$\ap_{\as}$&$\as_{\ap}$&$\as_{\ap}$&$\as_{\ap}$&$\as_{\ap\an}$\\
BA10' & 10 & $.7531\!\pm\!{.0182}$ &
$\cp$&$\cp$&$\cp$&$\cp$&$\cs_{\cp}$&$\cs_{\cp}$&$\ap$&$\ap$&$\ap$&$\ap$&$\as$&$\as$\\
\midrule
\multicolumn{3}{c|}{\#win of n,p,s}&
{\tiny0,\tcg{15},6}&{\tiny0,\tcg{15},6}&{\tiny0,\tcg{11},10}&{\tiny0,8,\tcb{13}}&{\tiny1,6,\tcb{14}}&{\tiny6,3,\tcb{12}}&{\tiny0,5,\tcy{16}}&{\tiny0,8,\tcy{13}}&{\tiny0,4,\tcy{17}}&{\tiny0,3,\tcy{18}}&{\tiny0,3,\tcy{18}}&{\tiny5,5,\tcy{11}}\\
\bottomrule\end{tabular}
&~~&
\begin{tabular}{ccc|cccccc|cccccc}\toprule
\multirow{2}{*}{dataset} & \multirow{2}{*}{$K$} & \multirow{2}{*}{MS} & 
\multicolumn{6}{c|}{for MLR with $n_\tra=$}& \multicolumn{6}{c}{for AUL with $n_\tra=$}\\
&&&25&50&100&200&400&800&25&50&100&200&400&800\\
\midrule
CAR & 4 & $.0146\!\pm\!{.0095}$ &
$\cs_{\cp}$&$\cp_{\cs}$&$\cp_{\cs}$&$\cp_{\cs}$&$\cs$&$\cn_{\cs\cp}$&$\as_{\ap\an}$&$\as_{\ap\an}$&$\as_{\ap\an}$&$\ap_{\as\an}$&$\ap_{\as\an}$&$\as_{\an\ap}$\\
SWD & 4 & $.3980\!\pm\!{.0211}$ &
$\cp$&$\cp$&$\cp_{\cs}$&$\cs_{\cn}$&$\cn_{\cs}$&$\cn_{\cs}$&$\ap_{\as\an}$&$\ap_{\as\an}$&$\as_{\ap\an}$&$\an_{\as}$&$\an_{\as}$&$\as_{\an}$\\
\midrule
BA5 & 5 & $.1890\!\pm\!{.0268}$ &
$\cp_{\cs}$&$\cs_{\cp}$&$\cn_{\cs}$&$\cn_{\cs}$&$\cn_{\cs}$&$\cn_{\cs}$&$\as_{\ap\an}$&$\as_{\an\ap}$&$\ap_{\as\an}$&$\an_{\ap\as}$&$\an_{\as\ap}$&$\an_{\as\ap}$\\
CO5' & 5 & $.2664\!\pm\!{.0357}$ &
$\cp_{\cs}$&$\cp_{\cs\cn}$&$\cs_{\cn\cp}$&$\cn_{\cs}$&$\cn$&$\cn$&$\as_{\ap\an}$&$\as_{\ap\an}$&$\as_{\an}$&$\as_{\an\ap}$&$\an_{\as\ap}$&$\an_{\as}$\\
CO5 & 5 & $.3095\!\pm\!{.0329}$ &
$\cp_{\cs}$&$\cn_{\cs\cp}$&$\cn_{\cs}$&$\cn_{\cs}$&$\cn_{\cs}$&$\cn_{\cs}$&$\as_{\ap\an}$&$\as_{\an\ap}$&$\as_{\an\ap}$&$\as_{\an}$&$\an_{\as}$&$\as_{\an}$\\
LEV & 5 & $.3685\!\pm\!{.0202}$ &
$\cp$&$\cp_{\cs}$&$\cs_{\cn\cp}$&$\cs_{\cn}$&$\cs_{\cn}$&$\cs_{\cn}$&$\ap_{\as\an}$&$\ap_{\as\an}$&$\as_{\an\ap}$&$\as_{\an}$&$\an_{\as}$&$\an_{\as}$\\
CH5 & 5 & $.4172\!\pm\!{.0455}$ &
$\cs_{\cp}$&$\cs$&$\cs_{\cn}$&$\cs_{\cn}$&$\cs_{\cn}$&$\cn_{\cs}$&$\ap_{\as\an}$&$\as_{\ap\an}$&$\as_{\an\ap}$&$\as_{\an\ap}$&$\ap_{\as\an}$&$\as_{\ap\an}$\\
CE5' & 5 & $.4643\!\pm\!{.0440}$ &
$\cp$&$\cp_{\cs}$&$\cs_{\cp\cn}$&$\cn_{\cs}$&$\cn_{\cs}$&$\cs_{\cn}$&$\as_{\an\ap}$&$\as_{\ap\an}$&$\an_{\as\ap}$&$\an_{\as\ap}$&$\an_{\as\ap}$&$\as_{\an\ap}$\\
CE5 & 5 & $.4760\!\pm\!{.0460}$ &
$\cp$&$\cp_{\cs}$&$\cs_{\cn}$&$\cs_{\cn}$&$\cn_{\cs}$&$\cs_{\cn}$&$\ap_{\as\an}$&$\as_{\an\ap}$&$\ap_{\as\an}$&$\an_{\as\ap}$&$\as_{\ap\an}$&$\an_{\as\ap}$\\
AB5 & 5 & $.4873\!\pm\!{.0324}$ &
$\cp_{\cs}$&$\cp_{\cs}$&$\cs_{\cp}$&$\cs$&$\cs_{\cn}$&$\cn_{\cs}$&$\as_{\an\ap}$&$\ap_{\as\an}$&$\as_{\an\ap}$&$\as_{\an\ap}$&$\as_{\an}$&$\an_{\as}$\\
BA5' & 5 & $.5524\!\pm\!{.0373}$ &
$\cp$&$\cp$&$\cp_{\cs}$&$\cs_{\cp\cn}$&$\cn$&$\cn_{\cs}$&$\ap_{\as\an}$&$\as_{\ap\an}$&$\as_{\an\ap}$&$\as_{\an\ap}$&$\as_{\an\ap}$&$\as_{\an\ap}$\\
\midrule
WQR & 6 & $.3678\!\pm\!{.0260}$ &
$\cs_{\cp}$&$\cs_{\cp}$&$\cs_{\cn}$&$\cs_{\cn}$&$\cs_{\cn}$&$\cs_{\cn}$&$\as_{\an\ap}$&$\an_{\as\ap}$&$\as_{\an\ap}$&$\as_{\an\ap}$&$\as_{\an\ap}$&$\as_{\an\ap}$\\
\midrule
ERA & 9 & $.7208\!\pm\!{.0198}$ &
$\cp$&$\cp$&$\cp$&$\cs_{\cn\cp}$&$\cp_{\cn\cs}$&$\cn_{\cs}$&$\an_{\ap\as}$&$\ap_{\as\an}$&$\as_{\ap\an}$&$\ap_{\an\as}$&$\ap_{\an\as}$&$\as_{\an\ap}$\\
\midrule
BA10 & 10 & $.3815\!\pm\!{.0287}$ &
$\cp$&$\cs_{\cp}$&$\cs$&$\cn_{\cs}$&$\cn_{\cs}$&$\cn$&$\as$&$\as_{\ap\an}$&$\as_{\an\ap}$&$\as_{\ap\an}$&$\an_{\ap\as}$&$\an_{\as\ap}$\\
CO10' & 10 & $.4720\!\pm\!{.0369}$ &
$\cp$&$\cp$&$\cs_{\cp\cn}$&$\cn$&$\cn_{\cs}$&$\cn_{\cs}$&$\as_{\ap\an}$&$\ap_{\as\an}$&$\as_{\an\ap}$&$\ap_{\as\an}$&$\an_{\as\ap}$&$\an_{\as\ap}$\\
CO10 & 10 & $.5031\!\pm\!{.0340}$ &
$\cp$&$\cp_{\cs}$&$\cp_{\cs\cn}$&$\cn_{\cs}$&$\cn_{\cs}$&$\cs_{\cn}$&$\as_{\ap\an}$&$\as_{\an\ap}$&$\as_{\an\ap}$&$\an_{\as}$&$\an_{\as\ap}$&$\an_{\as\ap}$\\
CH10 & 10 & $.6238\!\pm\!{.0466}$ &
$\cp$&$\cs_{\cp}$&$\cs$&$\cs_{\cp}$&$\cs_{\cn}$&$\cn_{\cs}$&$\ap_{\as}$&$\as_{\ap}$&$\as_{\ap\an}$&$\as_{\ap\an}$&$\ap_{\as\an}$&$\as_{\ap\an}$\\
AB10 & 10 & $.6847\!\pm\!{.0239}$ &
$\cp$&$\cp_{\cs}$&$\cp_{\cs}$&$\cs$&$\cs$&$\cn_{\cs}$&$\as_{\ap\an}$&$\ap_{\as\an}$&$\as_{\ap\an}$&$\as_{\ap}$&$\as_{\ap\an}$&$\as_{\an\ap}$\\
CE10' & 10 & $.6898\!\pm\!{.0409}$ &
$\cp$&$\cp$&$\cp_{\cs}$&$\cp_{\cs}$&$\cn_{\cs}$&$\cs_{\cn}$&$\as_{\ap\an}$&$\as_{\ap\an}$&$\as_{\an\ap}$&$\ap_{\as\an}$&$\as_{\an\ap}$&$\as_{\ap\an}$\\
CE10 & 10 & $.6911\!\pm\!{.0313}$ &
$\cp$&$\cp$&$\cs$&$\cs_{\cn}$&$\cs_{\cn}$&$\cs_{\cn}$&$\as_{\ap\an}$&$\as_{\ap\an}$&$\ap_{\an\as}$&$\as_{\ap\an}$&$\as_{\an\ap}$&$\as_{\an\ap}$\\
BA10' & 10 & $.7531\!\pm\!{.0182}$ &
$\cp$&$\cp$&$\cp$&$\cp_{\cs}$&$\cp_{\cn\cs}$&$\cn_{\cp\cs}$&$\ap_{\as\an}$&$\as$&$\as$&$\as_{\ap\an}$&$\ap_{\an\as}$&$\an_{\as\ap}$\\
\midrule
\multicolumn{3}{c|}{\#win of n,p,s}&
{\tiny0,\tcg{18},3}&{\tiny1,\tcg{15},5}&{\tiny2,8,\tcb{11}}&{\tiny7,3,\tcb{11}}&{\tiny\tcr{11},2,8}&{\tiny\tcr{14},0,7}&{\tiny1,7,\tcy{13}}&{\tiny1,6,\tcy{14}}&{\tiny1,3,\tcy{17}}&{\tiny5,4,\tcy{12}}&{\tiny\tcc{9},5,7}&{\tiny9,0,\tcy{12}}\\
\bottomrule\end{tabular}
\\
~~&~~&~~\\
MAE&~~&MSE\\
\begin{tabular}{ccc|cccccc|cccccc}\toprule
\multirow{2}{*}{dataset} & \multirow{2}{*}{$K$} & \multirow{2}{*}{MS} & 
\multicolumn{6}{c|}{for MLR with $n_\tra=$}& \multicolumn{6}{c}{for AUL with $n_\tra=$}\\
&&&25&50&100&200&400&800&25&50&100&200&400&800\\
\midrule
CAR&4&$.0146\!\pm\!{.0095}$&
$\cs_{\cp}$&$\cp_{\cs}$&$\cp_{\cs}$&$\cp_{\cs}$&$\cs_{\cp}$&$\cp_{\cs\cn}$&$\as_{\ap\an}$&$\as_{\ap\an}$&$\as_{\ap\an}$&$\ap_{\as\an}$&$\as_{\ap\an}$&$\as_{\ap\an}$\\
SWD&4&$.3980\!\pm\!{.0211}$&
$\cp$&$\cp$&$\cs_{\cp}$&$\cs_{\cn\cp}$&$\cs_{\cn}$&$\cn_{\cp\cs}$&$\ap_{\as\an}$&$\as_{\ap\an}$&$\ap_{\an\as}$&$\ap_{\as\an}$&$\ap_{\an\as}$&$\ap_{\as\an}$\\
\midrule
BA5&5&$.1890\!\pm\!{.0268}$&
$\cp_{\cs}$&$\cp_{\cs}$&$\cn_{\cs\cp}$&$\cn_{\cs\cp}$&$\cn_{\cs\cp}$&$\cn_{\cs}$&$\as_{\ap\an}$&$\as_{\an\ap}$&$\ap_{\an\as}$&$\ap_{\an\as}$&$\as_{\an\ap}$&$\an_{\as\ap}$\\
CO5'&5&$.2664\!\pm\!{.0357}$&
$\cs_{\cp}$&$\cs_{\cn\cp}$&$\cs_{\cn\cp}$&$\cn_{\cp}$&$\cn_{\cs}$&$\cn_{\cs}$&$\as_{\ap\an}$&$\as_{\ap\an}$&$\as_{\an\ap}$&$\as_{\an\ap}$&$\as_{\an\ap}$&$\an_{\as\ap}$\\
CO5&5&$.3095\!\pm\!{.0329}$&
$\cs_{\cp\cn}$&$\cs_{\cp\cn}$&$\cn_{\cs\cp}$&$\cn_{\cs\cp}$&$\cn_{\cs\cp}$&$\cn_{\cs\cp}$&$\as_{\ap\an}$&$\ap_{\as\an}$&$\as_{\an\ap}$&$\an_{\as\ap}$&$\as_{\an\ap}$&$\an_{\ap\as}$\\
LEV&5&$.3685\!\pm\!{.0202}$&
$\cp_{\cs}$&$\cs_{\cp}$&$\cs_{\cp\cn}$&$\cp_{\cs\cn}$&$\cn_{\cs\cp}$&$\cn_{\cp\cs}$&$\ap_{\as\an}$&$\as_{\an\ap}$&$\an_{\as\ap}$&$\an_{\as\ap}$&$\ap_{\an\as}$&$\an_{\ap\as}$\\
CH5&5&$.4172\!\pm\!{.0455}$&
$\cs_{\cp}$&$\cs_{\cp}$&$\cs_{\cp\cn}$&$\cs_{\cn\cp}$&$\cs_{\cp}$&$\cs_{\cp\cn}$&$\as_{\ap\an}$&$\as_{\ap\an}$&$\as_{\ap\an}$&$\an_{\as\ap}$&$\as_{\an\ap}$&$\ap_{\as\an}$\\
CE5'&5&$.4643\!\pm\!{.0440}$&
$\cp_{\cs}$&$\cs_{\cp}$&$\cs_{\cp\cn}$&$\cn_{\cp\cs}$&$\cn_{\cs}$&$\cn_{\cs\cp}$&$\as_{\ap\an}$&$\as_{\ap\an}$&$\as_{\ap\an}$&$\as_{\ap\an}$&$\as_{\ap\an}$&$\as_{\an\ap}$\\
CE5&5&$.4760\!\pm\!{.0460}$&
$\cp_{\cs}$&$\cp_{\cs}$&$\cs_{\cn}$&$\cn_{\cs\cp}$&$\cs_{\cn\cp}$&$\cs_{\cn\cp}$&$\as_{\ap}$&$\as_{\ap\an}$&$\as_{\ap\an}$&$\ap_{\as\an}$&$\as_{\ap\an}$&$\as_{\an\ap}$\\
AB5&5&$.4873\!\pm\!{.0324}$&
$\cs_{\cp}$&$\cs$&$\cs_{\cp}$&$\cs$&$\cs_{\cn\cp}$&$\cs_{\cn\cp}$&$\as_{\ap\an}$&$\as_{\ap}$&$\ap_{\as\an}$&$\an_{\as\ap}$&$\ap_{\an\as}$&$\as_{\an\ap}$\\
BA5'&5&$.5524\!\pm\!{.0373}$&
$\cp$&$\cp$&$\cp$&$\cp_{\cs}$&$\cn_{\cs\cp}$&$\cn_{\cs\cp}$&$\as_{\ap\an}$&$\as_{\ap}$&$\ap_{\as\an}$&$\as_{\ap\an}$&$\as_{\an\ap}$&$\as_{\an\ap}$\\
\midrule
WQR&6&$.3678\!\pm\!{.0260}$&
$\cp_{\cs}$&$\cs_{\cp}$&$\cs_{\cp\cn}$&$\cs_{\cn\cp}$&$\cs_{\cn\cp}$&$\cn_{\cp\cs}$&$\as_{\ap\an}$&$\ap_{\as\an}$&$\as_{\ap\an}$&$\an_{\ap\as}$&$\as_{\ap\an}$&$\as_{\an\ap}$\\
\midrule
ERA&9&$.7208\!\pm\!{.0198}$&
$\cp$&$\cp_{\cs}$&$\cp_{\cs}$&$\cp_{\cn}$&$\cn_{\cs\cp}$&$\cn_{\cs\cp}$&$\ap_{\as\an}$&$\ap_{\as\an}$&$\as_{\an\ap}$&$\as_{\ap\an}$&$\an_{\as\ap}$&$\as_{\ap\an}$\\
\midrule
BA10&10&$.3815\!\pm\!{.0287}$&
$\cp$&$\cp_{\cs}$&$\cs_{\cp}$&$\cs_{\cn\cp}$&$\cn_{\cp\cs}$&$\cn_{\cs\cp}$&$\as_{\ap}$&$\as_{\ap}$&$\as_{\ap\an}$&$\ap_{\as\an}$&$\ap_{\as\an}$&$\an_{\ap\as}$\\
CO10'&10&$.4720\!\pm\!{.0369}$&
$\cs_{\cp}$&$\cs_{\cp}$&$\cs_{\cp\cn}$&$\cn_{\cp\cs}$&$\cn_{\cp\cs}$&$\cn$&$\as_{\ap\an}$&$\an_{\ap\as}$&$\ap_{\as\an}$&$\as_{\ap\an}$&$\an_{\ap\as}$&$\an_{\ap\as}$\\
CO10&10&$.5031\!\pm\!{.0340}$&
$\cs_{\cp}$&$\cs_{\cp}$&$\cn_{\cp\cs}$&$\cn_{\cs\cp}$&$\cn_{\cs\cp}$&$\cn_{\cs\cp}$&$\as_{\ap\an}$&$\as_{\ap\an}$&$\as_{\ap\an}$&$\ap_{\an\as}$&$\ap_{\an\as}$&$\ap_{\an\as}$\\
CH10&10&$.6238\!\pm\!{.0466}$&
$\cp_{\cs}$&$\cs_{\cp}$&$\cs_{\cp}$&$\cs_{\cp\cn}$&$\cs_{\cp\cn}$&$\cp_{\cs\cn}$&$\as_{\ap}$&$\as_{\ap}$&$\as_{\ap}$&$\as_{\ap\an}$&$\ap_{\an\as}$&$\as_{\ap\an}$\\
AB10&10&$.6847\!\pm\!{.0239}$&
$\cp_{\cs}$&$\cs_{\cp}$&$\cs_{\cp}$&$\cs_{\cp}$&$\cs_{\cp\cn}$&$\cs_{\cp\cn}$&$\ap_{\as\an}$&$\as_{\ap\an}$&$\as_{\ap\an}$&$\as_{\ap\an}$&$\an_{\ap\as}$&$\as_{\an\ap}$\\
CE10'&10&$.6898\!\pm\!{.0409}$&
$\cs_{\cp}$&$\cp_{\cs}$&$\cs_{\cp}$&$\cs_{\cp\cn}$&$\cs_{\cn\cp}$&$\cs_{\cn\cp}$&$\as$&$\as_{\ap}$&$\as$&$\as_{\ap\an}$&$\as_{\ap\an}$&$\ap_{\an\as}$\\
CE10&10&$.6911\!\pm\!{.0313}$&
$\cp_{\cs}$&$\cs_{\cp}$&$\cs$&$\cs_{\cp\cn}$&$\cs_{\cp\cn}$&$\cn_{\cs\cp}$&$\as_{\ap\an}$&$\ap_{\as\an}$&$\ap_{\an\as}$&$\as_{\ap\an}$&$\ap_{\as\an}$&$\ap_{\an\as}$\\
BA10'&10&$.7531\!\pm\!{.0182}$&
$\cp$&$\cp$&$\cp$&$\cp$&$\cn_{\cp\cs}$&$\cs_{\cp\cn}$&$\as_{\ap}$&$\as_{\ap}$&$\as_{\ap}$&$\as_{\ap\an}$&$\as_{\an\ap}$&$\as_{\an\ap}$\\
\midrule
\multicolumn{3}{c|}{\#win of n,p,s}&
{\tiny0,\tcg{13},8}&{\tiny0,9,\tcb{12}}&{\tiny3,4,\tcb{14}}&{\tiny7,5,\tcb{9}}&{\tiny\tcr{11},0,10}&{\tiny\tcr{13},2,6}&{\tiny0,4,\tcy{17}}&{\tiny1,4,\tcy{16}}&{\tiny1,6,\tcy{14}}&{\tiny5,6,\tcy{10}}&{\tiny3,7,\tcy{11}}&{\tiny6,5,\tcy{10}}\\
\bottomrule\end{tabular}
&~~&
\begin{tabular}{ccc|cccccc|cccccc}\toprule
\multirow{2}{*}{dataset} & \multirow{2}{*}{$K$} & \multirow{2}{*}{MS} & 
\multicolumn{6}{c|}{for MLR with $n_\tra=$}& \multicolumn{6}{c}{for AUL with $n_\tra=$}\\
&&&25&50&100&200&400&800&25&50&100&200&400&800\\
\midrule
CAR&4&$.0146\!\pm\!{.0095}$&
$\cs_{\cp}$&$\cp_{\cs}$&$\cp_{\cs}$&$\cp_{\cs}$&$\cp_{\cs}$&$\cp_{\cn\cs}$&$\as_{\ap\an}$&$\as_{\an\ap}$&$\as_{\ap\an}$&$\as_{\ap}$&$\ap_{\as\an}$&$\ap_{\as\an}$\\
SWD&4&$.3980\!\pm\!{.0211}$&
$\cp$&$\cp$&$\cp_{\cs}$&$\cp_{\cn\cs}$&$\cn_{\cs}$&$\cs_{\cn\cp}$&$\as_{\an\ap}$&$\ap_{\an\as}$&$\an_{\ap\as}$&$\an_{\as\ap}$&$\an_{\as\ap}$&$\an_{\as\ap}$\\
\midrule
BA5&5&$.1890\!\pm\!{.0268}$&
$\cp_{\cs}$&$\cp_{\cs\cn}$&$\cn_{\cs\cp}$&$\cn_{\cs\cp}$&$\cn_{\cs\cp}$&$\cn_{\cs\cp}$&$\as_{\ap\an}$&$\as_{\ap\an}$&$\ap_{\as\an}$&$\ap_{\as\an}$&$\as_{\an\ap}$&$\ap_{\an\as}$\\
CO5'&5&$.2664\!\pm\!{.0357}$&
$\cs_{\cp\cn}$&$\cs_{\cn\cp}$&$\cn_{\cs\cp}$&$\cn_{\cp}$&$\cn_{\cp}$&$\cn_{\cp\cs}$&$\as_{\ap\an}$&$\as_{\ap\an}$&$\as_{\ap\an}$&$\ap_{\as\an}$&$\an_{\as\ap}$&$\as_{\an\ap}$\\
CO5&5&$.3095\!\pm\!{.0329}$&
$\cs_{\cp}$&$\cs_{\cp\cn}$&$\cp_{\cn\cs}$&$\cs_{\cn\cp}$&$\cn_{\cs}$&$\cs_{\cn\cp}$&$\ap_{\as\an}$&$\as_{\an\ap}$&$\as_{\an\ap}$&$\ap_{\as\an}$&$\as_{\an\ap}$&$\an_{\ap\as}$\\
LEV&5&$.3685\!\pm\!{.0202}$&
$\cp_{\cs}$&$\cs_{\cp\cn}$&$\cs_{\cp\cn}$&$\cs_{\cp\cn}$&$\cn_{\cs\cp}$&$\cn_{\cs\cp}$&$\ap_{\as\an}$&$\ap_{\as\an}$&$\as_{\ap\an}$&$\as_{\ap\an}$&$\an_{\ap\as}$&$\ap_{\an\as}$\\
CH5&5&$.4172\!\pm\!{.0455}$&
$\cp_{\cs}$&$\cp_{\cs}$&$\cs_{\cn\cp}$&$\cp_{\cs\cn}$&$\cs_{\cp\cn}$&$\cs_{\cp\cn}$&$\as_{\ap\an}$&$\as_{\ap\an}$&$\as_{\ap\an}$&$\as_{\an\ap}$&$\an_{\as\ap}$&$\as_{\an\ap}$\\
CE5'&5&$.4643\!\pm\!{.0440}$&
$\cs_{\cp}$&$\cp_{\cs}$&$\cp$&$\cp_{\cn\cs}$&$\cn_{\cs\cp}$&$\cs_{\cn\cp}$&$\ap_{\as\an}$&$\as_{\ap\an}$&$\as_{\ap}$&$\as_{\ap\an}$&$\as_{\ap\an}$&$\as_{\an\ap}$\\
CE5&5&$.4760\!\pm\!{.0460}$&
$\cp_{\cs}$&$\cp_{\cs}$&$\cs_{\cp\cn}$&$\cs_{\cn\cp}$&$\cn_{\cs\cp}$&$\cn_{\cs\cp}$&$\ap_{\as\an}$&$\an_{\ap\as}$&$\as_{\ap\an}$&$\as_{\an\ap}$&$\an_{\as\ap}$&$\as_{\an\ap}$\\
AB5&5&$.4873\!\pm\!{.0324}$&
$\cs_{\cp}$&$\cs_{\cp}$&$\cs_{\cp}$&$\cs_{\cn\cp}$&$\cn_{\cs\cp}$&$\cs_{\cn\cp}$&$\ap_{\an\as}$&$\as_{\an\ap}$&$\as_{\ap\an}$&$\ap_{\an\as}$&$\as_{\ap\an}$&$\as_{\ap\an}$\\
BA5'&5&$.5524\!\pm\!{.0373}$&
$\cp_{\cn}$&$\cp$&$\cp$&$\cp_{\cs}$&$\cn_{\cs\cp}$&$\cs_{\cn\cp}$&$\as_{\ap}$&$\as_{\ap}$&$\ap_{\as}$&$\as_{\ap\an}$&$\as_{\an\ap}$&$\as_{\ap}$\\
\midrule
WQR&6&$.3678\!\pm\!{.0260}$&
$\cp_{\cs}$&$\cs_{\cp}$&$\cs_{\cp}$&$\cs_{\cn\cp}$&$\cn_{\cs\cp}$&$\cn_{\cs}$&$\ap_{\as\an}$&$\ap_{\an\as}$&$\as_{\an\ap}$&$\an_{\as\ap}$&$\as_{\an\ap}$&$\an_{\as\ap}$\\
\midrule
ERA&9&$.7208\!\pm\!{.0198}$&
$\cp_{\cs}$&$\cp_{\cs}$&$\cp_{\cs}$&$\cp_{\cs\cn}$&$\cn_{\cp\cs}$&$\cn_{\cp\cs}$&$\ap_{\as\an}$&$\ap_{\as\an}$&$\as_{\ap\an}$&$\as_{\ap\an}$&$\ap_{\an\as}$&$\as_{\ap\an}$\\
\midrule
BA10&10&$.3815\!\pm\!{.0287}$&
$\cp$&$\cp_{\cs}$&$\cs_{\cp}$&$\cs_{\cn\cp}$&$\cn_{\cp\cs}$&$\cn_{\cp\cs}$&$\as_{\ap}$&$\as_{\ap}$&$\as_{\ap\an}$&$\an_{\as\ap}$&$\as_{\ap\an}$&$\ap_{\an\as}$\\
CO10'&10&$.4720\!\pm\!{.0369}$&
$\cp_{\cs}$&$\cp_{\cs\cn}$&$\cp_{\cn\cs}$&$\cn_{\cp\cs}$&$\cn_{\cs\cp}$&$\cn_{\cs\cp}$&$\ap_{\as\an}$&$\as_{\ap\an}$&$\as_{\an\ap}$&$\ap_{\as\an}$&$\ap_{\as\an}$&$\an_{\as\ap}$\\
CO10&10&$.5031\!\pm\!{.0340}$&
$\cs_{\cp}$&$\cp_{\cs\cn}$&$\cp_{\cs\cn}$&$\cs_{\cp\cn}$&$\cp_{\cs\cn}$&$\cn_{\cp\cs}$&$\as_{\ap\an}$&$\as_{\ap\an}$&$\as_{\ap\an}$&$\ap_{\as\an}$&$\an_{\as\ap}$&$\ap_{\an\as}$\\
CH10&10&$.6238\!\pm\!{.0466}$&
$\cp_{\cs}$&$\cs_{\cp}$&$\cs_{\cp}$&$\cs$&$\cs_{\cp\cn}$&$\cp_{\cs\cn}$&$\as_{\ap}$&$\as_{\ap}$&$\as_{\ap\an}$&$\as_{\an\ap}$&$\ap_{\as\an}$&$\as_{\ap\an}$\\
AB10&10&$.6847\!\pm\!{.0239}$&
$\cp_{\cs}$&$\cs$&$\cs_{\cp}$&$\cs_{\cp}$&$\cs_{\cp\cn}$&$\cp_{\cs\cn}$&$\as_{\ap\an}$&$\as_{\ap\an}$&$\as_{\ap\an}$&$\as_{\ap\an}$&$\as_{\an\ap}$&$\ap_{\as\an}$\\
CE10'&10&$.6898\!\pm\!{.0409}$&
$\cp_{\cs}$&$\cs_{\cp}$&$\cs_{\cp}$&$\cp_{\cs}$&$\cp_{\cs\cn}$&$\cp_{\cs\cn}$&$\as_{\ap\an}$&$\as_{\ap}$&$\as$&$\as_{\ap}$&$\as_{\ap}$&$\as_{\ap\an}$\\
CE10&10&$.6911\!\pm\!{.0313}$&
$\cp$&$\cs_{\cp}$&$\cs$&$\cs_{\cp\cn}$&$\cs_{\cp\cn}$&$\cp_{\cs\cn}$&$\as_{\ap}$&$\ap_{\as\an}$&$\as_{\ap\an}$&$\as_{\an}$&$\ap_{\as\an}$&$\as_{\an\ap}$\\
BA10'&10&$.7531\!\pm\!{.0182}$&
$\cp$&$\cp$&$\cp$&$\cp_{\cn\cs}$&$\cp_{\cn\cs}$&$\cs_{\cp\cn}$&$\ap_{\as}$&$\ap_{\as}$&$\as_{\ap}$&$\ap_{\as\an}$&$\as_{\an\ap}$&$\as_{\ap\an}$\\
\midrule
\multicolumn{3}{c|}{\#win of n,p,s}&
{\tiny0,\tcg{15},6}&{\tiny0,\tcg{12},9}&{\tiny2,9,\tcb{10}}&{\tiny3,8,\tcb{10}}&{\tiny\tcr{13},4,4}&{\tiny\tcr{9},5,7}&{\tiny0,9,\tcy{12}}&{\tiny1,6,\tcy{14}}&{\tiny1,2,\tcy{18}}&{\tiny3,7,\tcy{11}}&{\tiny6,5,\tcy{10}}&{\tiny4,6,\tcy{11}}\\
\bottomrule\end{tabular}
\end{tabular}
\end{table*}

% %========================================%
% \section*{Acknowledgment}
% %==========%
% Anonymized

%========================================%
\bibliographystyle{IEEEtran}
\bibliography{age_estimation, hetero, LabelSmoothing, machine_learning, multitask_learning, ordinal_regression, robust, sparse, dataset}

\end{document}